\documentclass{article}
\usepackage{iclr2026_conference,times}

\usepackage{amsmath,amsfonts,bm}

\def\eqref#1{equation~\ref{#1}}

\def\1{\bm{1}}

\DeclareMathAlphabet{\mathsfit}{\encodingdefault}{\sfdefault}{m}{sl}
\SetMathAlphabet{\mathsfit}{bold}{\encodingdefault}{\sfdefault}{bx}{n}

\usepackage{hyperref}
\usepackage{url}

\usepackage{graphicx}
\usepackage{booktabs}

\usepackage[accsupp]{axessibility}  %

\usepackage{hyperref}

\usepackage{orcidlink}

\usepackage{svg}
\usepackage{url}
\usepackage{hyperref}       %
\usepackage{url}            %
\usepackage{booktabs}       %
\usepackage{amsfonts}       %
\usepackage{nicefrac}       %
\usepackage{microtype}      %
\usepackage{xcolor}         %
\usepackage{algorithm}
\usepackage{algpseudocode}
\usepackage{pgfplots}
\usepackage{amsmath}
\usepackage{bm}
\usepackage{amsthm}
\usepgfplotslibrary{groupplots}
\pgfplotsset{compat=1.17}
\usepackage{array}
\usepackage{makecell}
\usepackage[table]{xcolor}
\usepackage{colortbl}
\usepackage{booktabs}
\usepackage{pifont}
\usepackage{multirow}
\usepackage{siunitx}
\usepackage{caption}
\usepackage{amssymb}
\usepackage{enumitem}
\usepackage[most]{tcolorbox} %

\usepackage{rotating}

\usepackage{pgfplotstable}

\usepackage{pgfplots}
\pgfplotsset{compat=1.18}
\usepackage{pgfplotstable}
\usepackage{xstring} %
\usepackage{etoolbox} %

\usepackage{pgfplotstable} %

\usepackage{subcaption}

\usepackage{booktabs}
\usepackage{makecell}
\usepackage{multirow}
\usepackage{array}
\usepackage[table]{xcolor} %
\usepackage{caption}

\usepackage{tabularx}

\usepackage{tikz}

\usetikzlibrary{positioning,arrows.meta,shapes.geometric,fit,calc}
\usetikzlibrary{decorations.markings}

\definecolor{method1color}{RGB}{31, 119, 180} %
\definecolor{method2color}{RGB}{255, 127, 14} %
\definecolor{method3color}{RGB}{44, 160, 44} %
\definecolor{method4color}{RGB}{214, 39, 40} %
\definecolor{method5color}{RGB}{148, 103, 189} %
\definecolor{method6color}{RGB}{140, 86, 75}   %

\definecolor{methodcolor1}{RGB}{0, 101, 189}
\definecolor{method2color}{RGB}{159, 186, 54}  %

\definecolor{method3color}{RGB}{247, 177, 30}  %
\definecolor{method4color}{RGB}{217, 81, 23}  %

\definecolor{pb}{HTML}{4C72B0}    %
\definecolor{pg}{HTML}{55A868}    %
\definecolor{pr}{HTML}{C44E52}    %
\definecolor{pp}{HTML}{8172B2}    %

\definecolor{cloudblue}{HTML}{1f77b4}      %
\definecolor{gaussorange}{HTML}{ff7f0e}    %
\definecolor{sobolgreen}{HTML}{2ca02c}     %

\usepackage{hyperref}

\newcolumntype{L}{>{\raggedright\arraybackslash}p{3.6cm}}
\newcolumntype{C}{>{\centering\arraybackslash}p{1.5cm}}

\usepackage[table,xcdraw]{xcolor}
\usetikzlibrary{shapes.geometric, arrows.meta, positioning, fit}

\tikzset{
  >=Latex, 
  block/.style = {
    draw=black, fill={rgb,1:red,0.7;green,0.7;blue,0.7}, 
    thick, rounded corners, 
    minimum width=2.5cm, minimum height=1cm, 
    align=center
  },
  arrow/.style = {
    -{Stealth[scale=1.2]}, thick, black
  },
  phase/.style = {
    draw=black, rounded corners=3mm, inner sep=8pt
  },
  title/.style = {
    font=\large\bfseries, black
  },
  small image node/.style={
    inner sep=0pt,
    outer sep=0pt,
    yshift=0.1cm 
  },
}
\tikzset{
    lock/.pic={
        \node[inner sep=0pt] at (0,0) {\includegraphics[width=0.22cm]{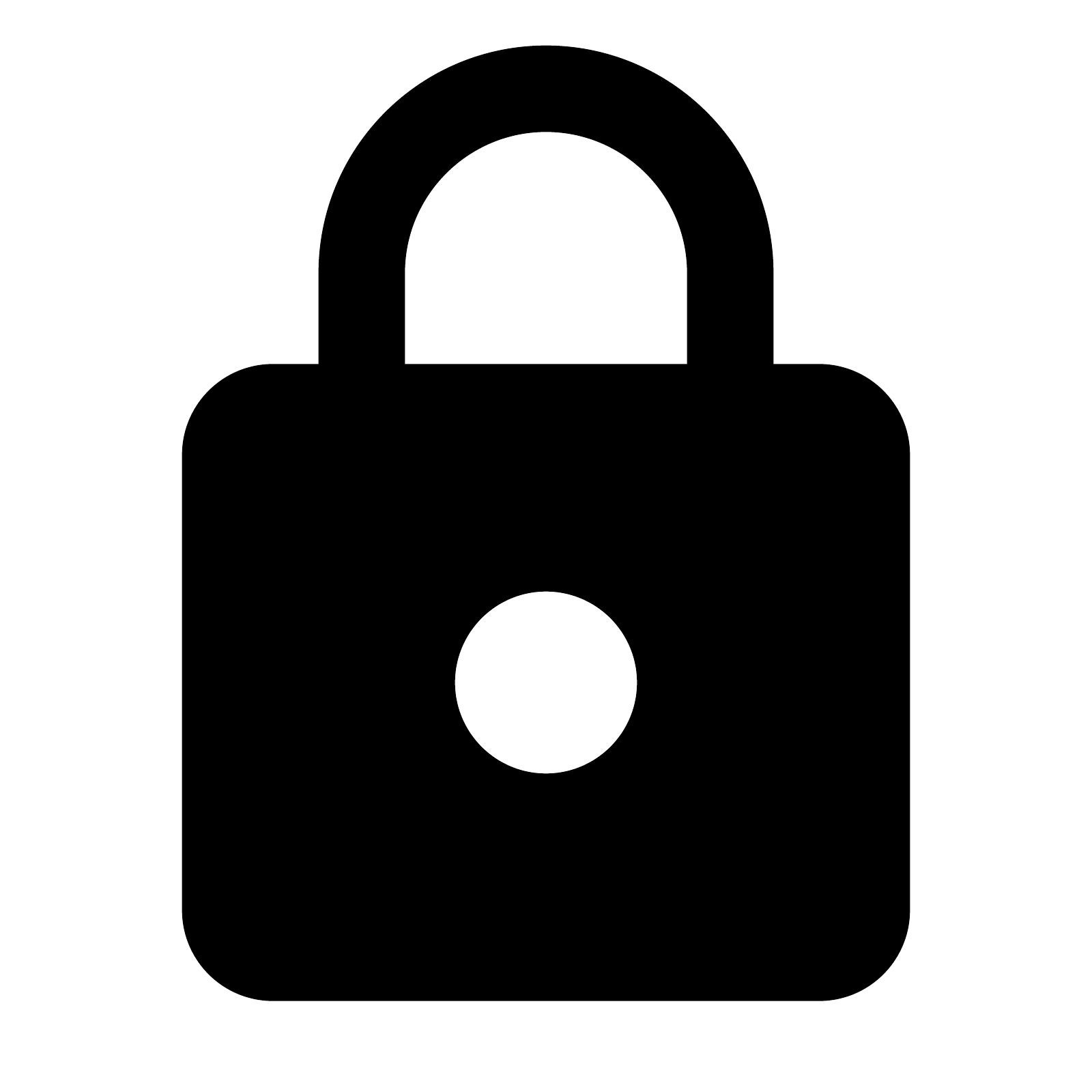}};
    }
}

\newcommand{\cmark}{\textbf{\ding{52}}} 
\newcommand{\xmark}{\ding{56}}

\usepackage{tcolorbox}
\usepackage{tcolorbox}
\usepackage{hyperref}
\usepackage{lmodern}

\definecolor{iclrblue}{RGB}{60,90,160}

\definecolor{headergray}{gray}{0.35}
\definecolor{boxgray}{gray}{0.92}

\newtcolorbox{iclrbox}[1]{
  colback=boxgray,
  colframe=black,
  title=#1,
  fonttitle=\bfseries\large,
  coltitle=white,
  colbacktitle=headergray,
  boxed title style={
    arc=6pt,
    outer arc=6pt,
    boxrule=0pt,
  }
}

\usepackage{hyperref}

\definecolor{darkblue}{rgb}{0, 0, 0.5}
\definecolor{navyblue}{RGB}{0, 31, 101}
\definecolor{light_grey}{gray}{0.50} 
\definecolor{dark_grey}{gray}{0.35}
\hypersetup{
    colorlinks=true,    %
    linkcolor=light_grey, %
    citecolor=light_grey, %
    urlcolor=light_grey,  %
}

\makeatletter
\renewcommand\paragraph{\@startsection{paragraph}{4}{\z@}%
  {1.5ex \@plus 0.5ex \@minus .2ex}%
  {-1em}%
  {\normalfont\normalsize\bfseries}}  %
\makeatother

\title{Trust-Region Noise Search for Black-Box Alignment of Diffusion and Flow Models}

\usepackage[affil-it]{authblk}

\author[1,2]{Niklas Schweiger}
\author[1,2,$\dagger$]{Daniel Cremers}
\author[1,2,$\dagger$]{Karnik Ram}

\affil[1]{Technical University of Munich, Germany}
\affil[2]{Munich Center for Machine Learning, Germany}

\affil[ ]{\texttt{niklas.schweiger@tum.de}}
\begingroup
  \renewcommand\thefootnote{}
  \footnotetext{Accepted to ECCV 2026. Shorter version at ReALM-GEN Workshop, ICLR 2026.}
\endgroup
\begingroup
  \renewcommand\thefootnote{}
  \footnotetext{$^\dagger$Equal supervision.}
\endgroup

\iclrfinalcopy 

\makeatletter
\renewcommand{\headrulewidth}{0pt} %
\makeatother

\usepackage[capitalize]{cleveref}

\begin{document}

\maketitle

\fancyhead{}
\renewcommand{\headrulewidth}{0pt}

\begin{abstract}
Optimizing the noise samples of diffusion and flow models is an increasingly popular approach to align these models to target rewards at inference time. However, we observe that these approaches are usually restricted to differentiable or cheap reward models, the formulation of the underlying pre-trained generative model, or are memory/compute inefficient. We instead propose a simple trust-region based search algorithm (TRS) which treats the pre-trained generative and reward models as a black-box and only optimizes the source noise. Our approach achieves a good balance between global exploration and local exploitation, and is versatile and easily adaptable to various generative settings and reward models with minimal hyperparameter tuning. We evaluate TRS across text-to-image, molecule and protein design tasks, and obtain significantly improved output samples over the base generative models and other inference-time alignment approaches which optimize the source noise sample, or even the entire reverse-time sampling noise trajectories in the case of diffusion models. Our source code is publicly available%
\renewcommand{\thefootnote}{\fnsymbol{footnote}}%
\footnote[1]{Project page: \url{https://niklasschweiger.github.io/trust-region-noise-search/}}%
\renewcommand{\thefootnote}{\arabic{footnote}}%
.
\end{abstract}

\newcommand{\imgwithreward}[3]{%
\begin{tikzpicture}
\node[inner sep=0pt] (img) {%
  \includegraphics[
    width=#1,
    trim=40 10 40 10, %
    clip
  ]{#2}
};
\node[
    anchor=north east,
    font=\scriptsize\bfseries,
    text=white,
    fill=black,
    fill opacity=0.6,
    text opacity=1,
    inner sep=1.5pt
] at (img.north east) {#3};
\end{tikzpicture}
}

\newcommand{\structureswithreward}[3]{%
\begin{tikzpicture}
\node[
    inner sep=0pt,
    minimum width=#1,
    minimum height=2.2cm  %
] (img) {%
  \includegraphics[
    width=#1,
    trim=8 8 8 8,
    clip
  ]{#2}
};
\node[
    anchor=north east,
    font=\scriptsize\bfseries,
    text=white,
    fill=black,
    fill opacity=0.6,
    text opacity=1,
    inner sep=1.5pt
] at (img.north east) {#3};
\end{tikzpicture}
}

\begin{figure}[h]
\centering
\setlength{\tabcolsep}{3pt}
\renewcommand{\arraystretch}{1.0}

\begin{tabular}{@{}c@{}c@{}} %

\begin{minipage}[t]{1.0\textwidth}
\centering

\begin{tikzpicture}
    \draw[thick] (-2,0) -- (1.0,0); %
    \node[font=\small\bfseries] at (3.0,0) {Aesthetic Score ($\uparrow$)}; %
    \draw[->, thick] (5,0) -- (8,0); %
\end{tikzpicture}

\setlength{\tabcolsep}{1pt}
\begin{tabular}{*{6}{c}}
\imgwithreward{0.145000\textwidth}{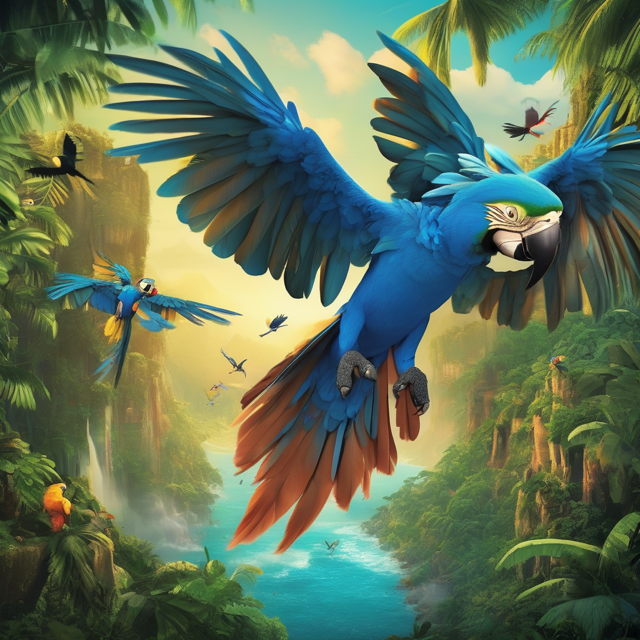}{6.9} &
\imgwithreward{0.145000\textwidth}{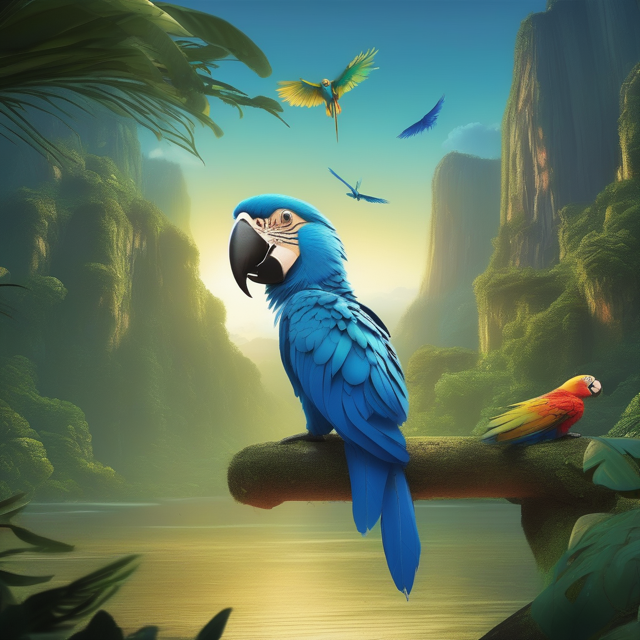}{7.5} &
\imgwithreward{0.145000\textwidth}{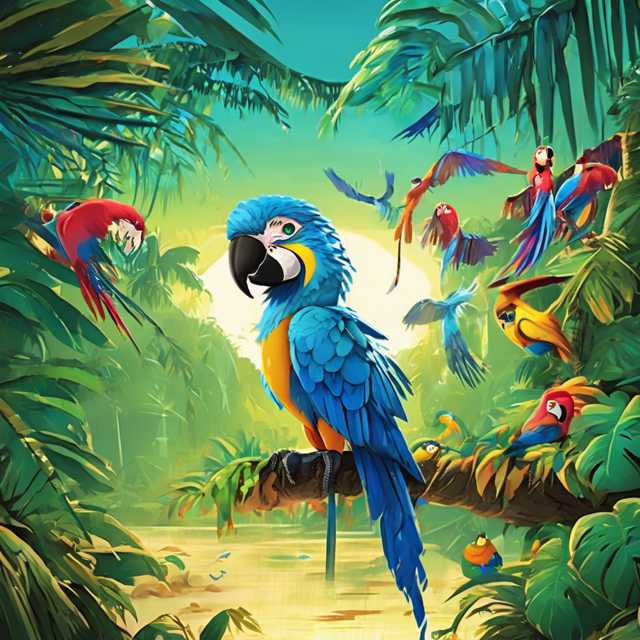}{7.9} &
\imgwithreward{0.145000\textwidth}{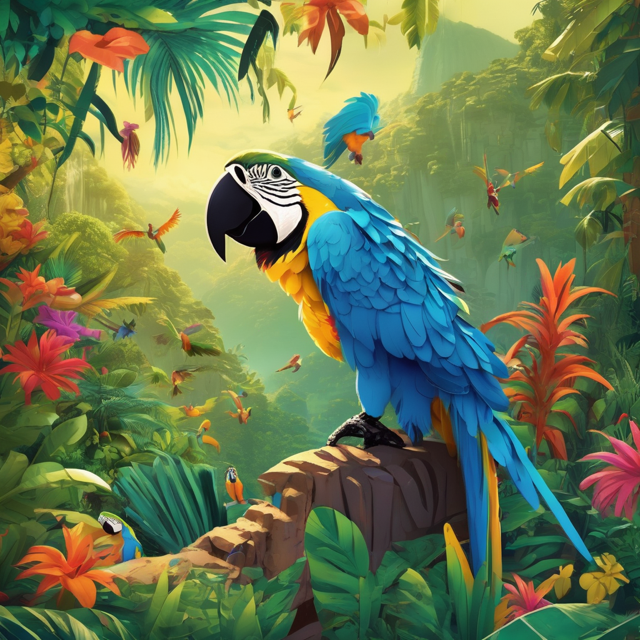}{7.7} &
\imgwithreward{0.145000\textwidth}{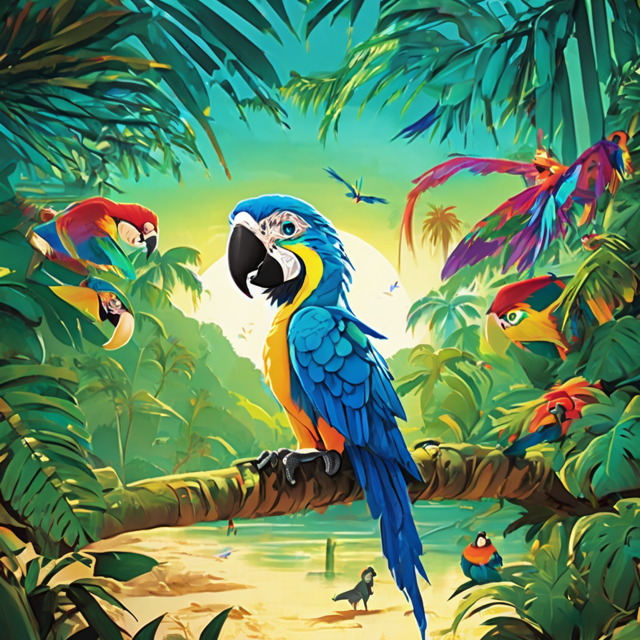}{8.1} &
\imgwithreward{0.145000\textwidth}{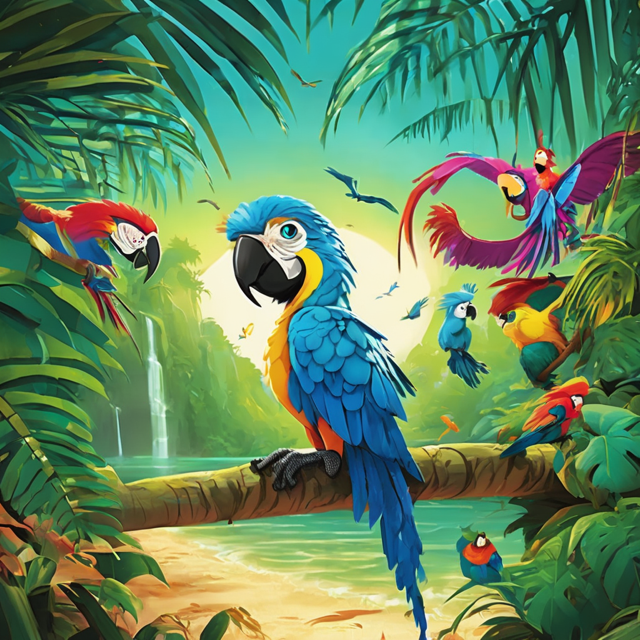}{8.2} \\
\end{tabular}

\vspace{0.5em}

\begin{tikzpicture}
    \draw[thick] (-2,0) -- (1.0,0); %
    \node[font=\small\bfseries] at (3.0,0) {Property Alignment ($\downarrow$)}; %
    \draw[->, thick] (5,0) -- (8,0); %
\end{tikzpicture}%

\setlength{\tabcolsep}{1pt}
\begin{tabular}{*{6}{c}}

\structureswithreward{0.145000\textwidth}{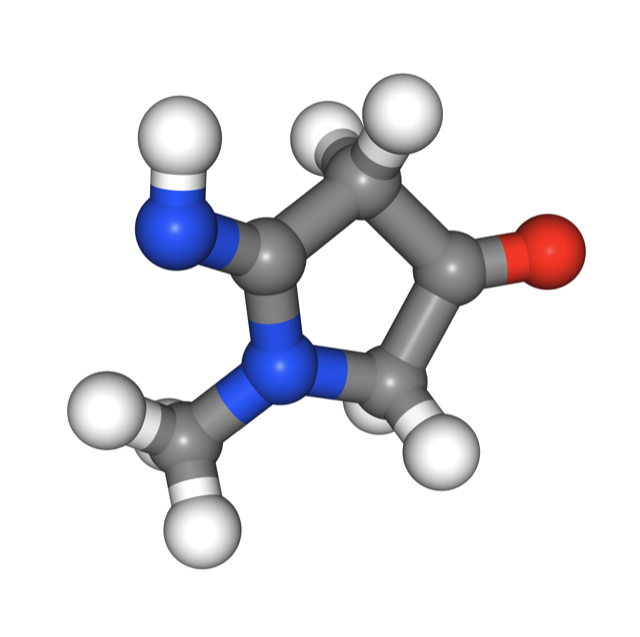}{0.75} &
\structureswithreward{0.145000\textwidth}{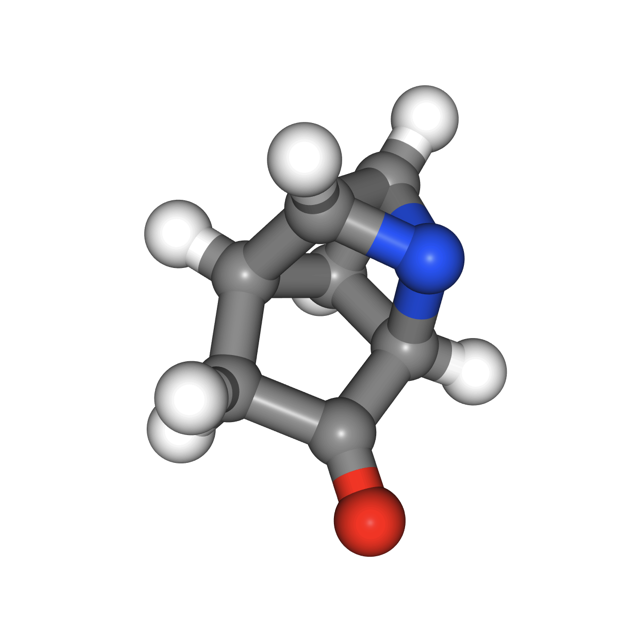}{0.65} &
\structureswithreward{0.145000\textwidth}{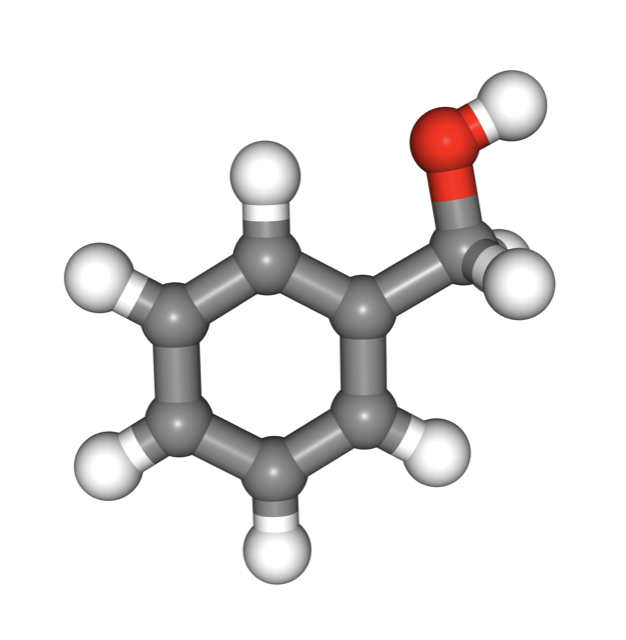}{0.63} &
\structureswithreward{0.145000\textwidth}{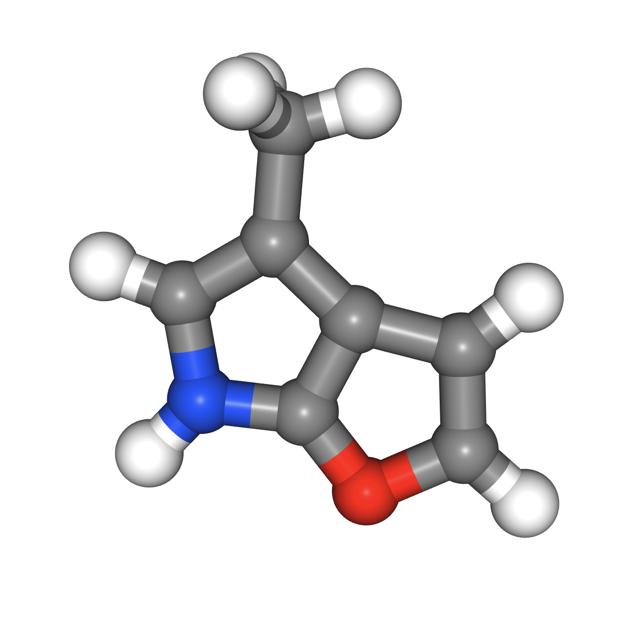}{0.43} &
\structureswithreward{0.145000\textwidth}{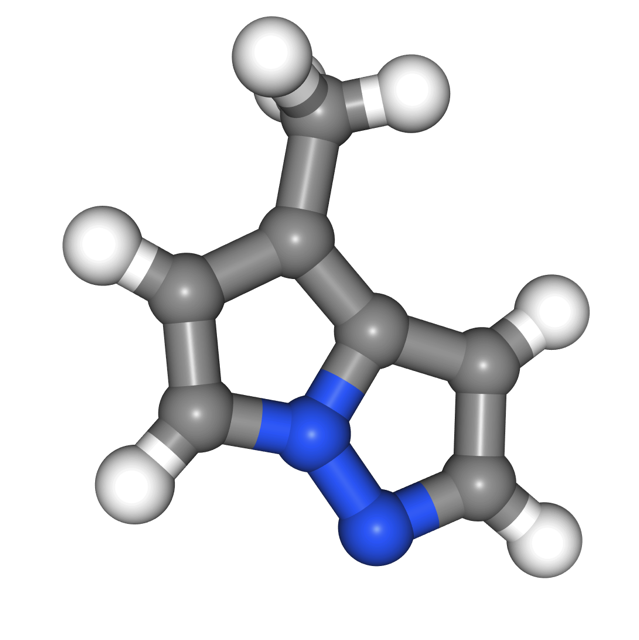}{0.41} &
\structureswithreward{0.145000\textwidth}{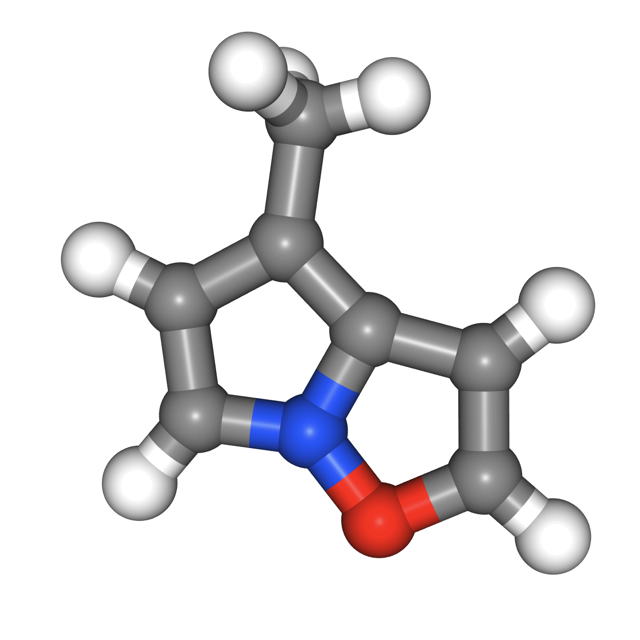}{0.35} \\

\end{tabular}
\setlength{\tabcolsep}{1pt}

\vspace{0.0em}

\begin{tikzpicture}
    \draw[thick] (-2,0) -- (1.0,0); %
    \node[font=\small\bfseries] at (3.0,0) {Designability ($\uparrow$)}; %
    \draw[->, thick] (5,0) -- (8,0); %
\end{tikzpicture}%

\setlength{\tabcolsep}{1pt}
\begin{tabular}{*{6}{c}}

\structureswithreward{0.145000\textwidth}{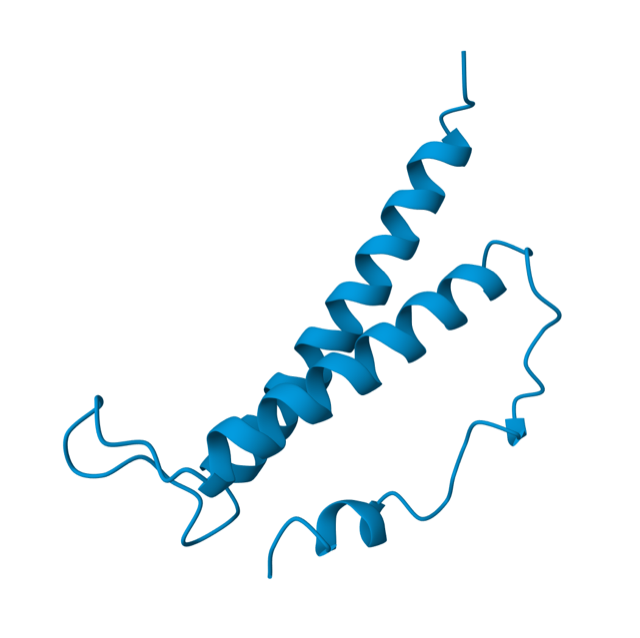}{0.041} &

\structureswithreward{0.145000\textwidth}{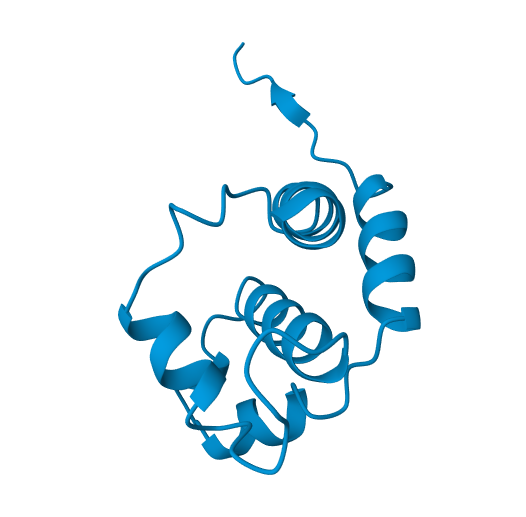}{0.057} &

\structureswithreward{0.145000\textwidth}{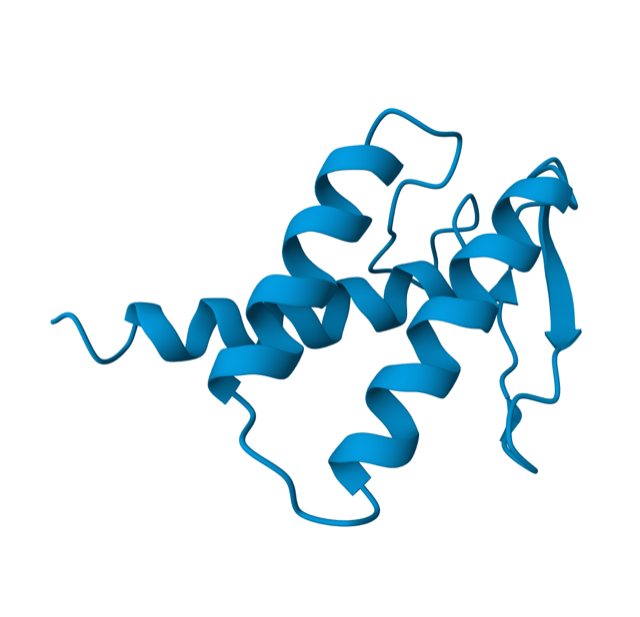}{0.059} &

\structureswithreward{0.145000\textwidth}{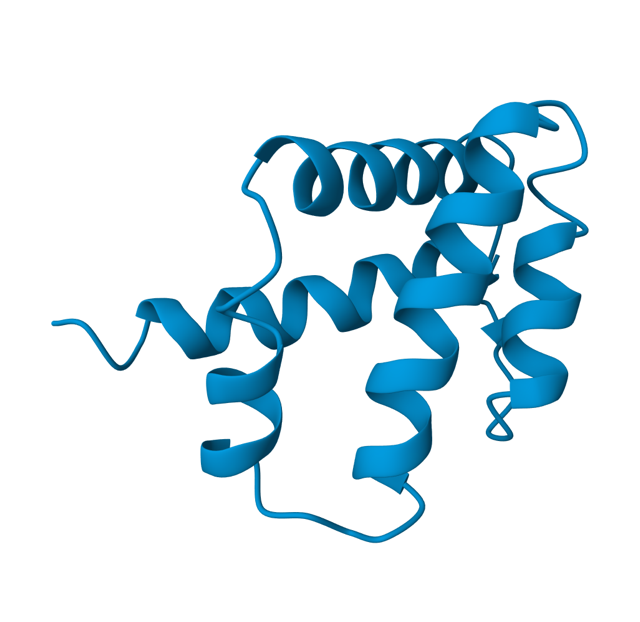}{0.201} &

\structureswithreward{0.145000\textwidth}{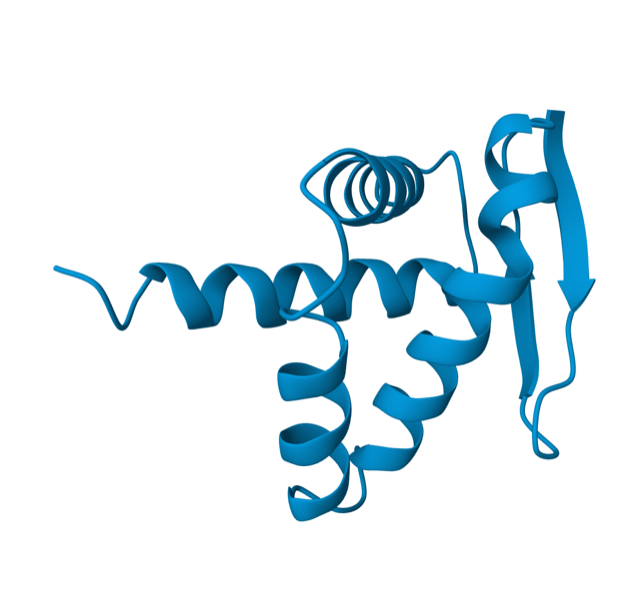}{0.346} &

\structureswithreward{0.145000\textwidth}{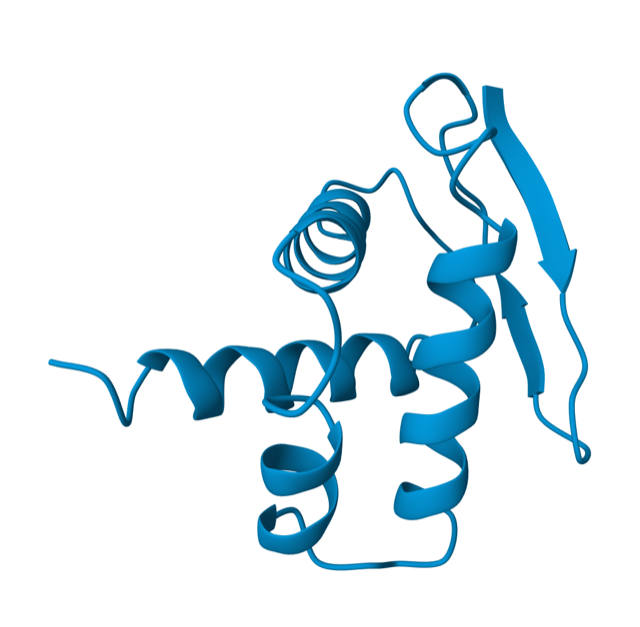}{0.506}

\\
\end{tabular}
\\[0.2em]
\setlength{\tabcolsep}{1pt}

\setlength{\tabcolsep}{1pt}

\end{minipage}
\end{tabular}
\captionsetup{font=small}

\caption{
\textbf{Progression of optimized samples.} Output samples produced by our algorithm (TRS) across iterations $i \in \{0, \dots, 15\}$ for text-to-image, molecules, and proteins. \textit{Top}: Aesthetic reward alignment for the prompt ``Animated movie poster...''. \textit{Middle}: Molecule property alignment. \textit{Bottom}: Protein designability. The horizontal arrow indicates the direction of optimization, while the arrows $\uparrow / \downarrow$ indicate whether the objective is maximized or minimized. 
}
\label{fig:optimization_progression}
\end{figure}

\section{Introduction}
\label{introduction}

Generative models such as diffusion and flow-based models have revolutionized diverse domains, including high-fidelity image and video synthesis, molecule generation, and language modeling, by learning complex data distributions from large-scale pretraining data \citep{ho2020denoising, lipman2023flowmatching}. 
While scaling model capacity and training compute has significantly improved sample quality, pretrained models often fall short of specific, fine-grained requirements. This includes generating molecules with precise binding affinities \citep{guan2023d} or images that strictly align with complex, multi-attribute prompts. 

This gap has motivated the emergence of inference-time alignment for diffusion and flow-based models, an exciting new paradigm where the quality of generated samples is optimized post-training using feedback from target reward models~\citep{bansal2024universal, eyring2024reno}. Unlike fine-tuning, this approach requires no additional training data and instead invests additional compute at inference time to steer samples toward desired properties.

Several approaches for inference-time alignment have been proposed. \textit{Gradient-based} methods back-propagate through the entire (often ODE-based) iterative process of the model to adjust the initial noise sample~\citep{ben2024d, wang2024training}. But these are prone to high GPU-memory costs and risk drifting off the training-data manifold. \textit{Sequence-based} methods, typically used for SDE samplers, range from filtering approaches that iteratively reweigh and resample~\citep{singhal2025general, kim2025test} to tree-search based methods~\citep{jain2025diffusion, li2025dynamic}. But these often require a high number of expensive reward calls or rely on accurate value estimates for terminal rewards, which are not always available or reliable. In contrast, \textit{black-box search} methods which are relative underexplored, treat the generator and reward models as a black box and apply search heuristics for the source noise sample~\citep{ma2025inference, tan2025fast}. While versatile and applicable to any generative architecture and reward model, we notice that existing approaches often struggle to find a good balance between global exploration and local exploration, tending toward one extreme or the other.

Inspired by Bayesian optimization algorithms~\citep{eriksson2019scalable}, which also optimize expensive black-box functions, we propose a simple and effective \textit{trust region search} (TRS) approach that balances between global exploration and local exploitation in a structured manner. By controlling only the source noise, TRS is readily applicable to a wide range of generative models and reward functions without internal modifications. TRS begins by exploring multiple seed noise samples, which are then pruned and iteratively refined with local perturbations. Crucially, these perturbations are adaptively controlled in both magnitude and direction based on observed reward values, ensuring the search remains within the data manifold to produce stable and high-quality samples.

The contributions of our work are as follows:

\begin{enumerate}
    \item We introduce \textit{trust region search} (TRS), a simple approach for inference-time reward alignment of black-box diffusion and flow models via adaptive source noise control. 
    \item We provide an extensive evaluation on text-to-image generation, demonstrating that TRS yields significantly more aligned and higher-quality samples compared to existing search heuristics and even full-noise sequence search baselines under identical compute budgets.
    \item We demonstrate the versatility of our approach through an extended evaluation on small molecule and protein design tasks, where TRS proves effective even with expensive reward functions and requires minimal hyperparameter tuning.
\end{enumerate}

\section{Background and Related Work}
\label{sec:preliminaries}

\subsection{Diffusion and Flow-based Models}
\label{sec:prel:diffusion_and_flow}

Diffusion and flow-based generative models aim to transport a simple noise distribution $p_{\text{noise}} = p_0$ to a complex data distribution $p_{\text{data}} = p_1$. A sample $\mathbf{x}_0 \sim p_0$ is gradually transformed into a data sample $\mathbf{x}_1 \sim p_1$ through a continuous-time process $\{\mathbf{x}_t\}_{t\in[0,1]}$, typically discretized into $T$ steps with $t_k = k/T$. In continuous data regimes, $p_0$ is usually white Gaussian noise $\mathcal{N}(\mathbf{0},\mathbf{I})$.\\

\noindent
\textbf{Diffusion models} are characterized by a forward noising process that gradually transforms data into noise. This process is governed by a stochastic differential equation (SDE):
\[
    d\mathbf{x}_t = \mathbf{f}(t)\mathbf{x}_t dt + g(t)d\mathbf{w}_t,
\]
where $\mathbf{f}(t)$ and $g(t)$ are the drift and diffusion coefficients that determine the noising schedule \citep{song2020score}. To generate new samples, this process must be reversed. The generative (reverse) process relies on the score function $\bm{\epsilon}(\mathbf{x}_t, t) \triangleq \nabla_{\mathbf{x}} \log p_t(\mathbf{x_t})$, which represents the gradient of the log-density of the noisy sample. Once the score is estimated via a neural network, sampling can be performed using either the deterministic or stochastic formulations shown in  \Cref{tab:ode_sde}.\\

\noindent
\textbf{Flow matching models} instead learn a continuous vector field $\mathbf{v}_\theta(\mathbf{x}_t, t) = d\mathbf{x}_t/dt$ \citep{lipman2023flowmatching} that defines the velocity of the samples. Unlike diffusion, the paths are not necessarily tied to a corruption process; a common choice is the optimal transport (OT) path \citep{liu2023flow}, defined by linear interpolation
\[
\mathbf{x}_t = (1-t)\mathbf{x}_0 + t \mathbf{x}_1,
\]
which yields a constant target velocity $\mathbf{v}_t = \mathbf{x}_1 - \mathbf{x}_0$. Standard flow matching integrates the learned ODE for sampling, while stochastic variants \citep{bose2024sestochastic} can introduce Brownian motion and optional score corrections as detailed in  \Cref{tab:ode_sde}.\\

\noindent
\textbf{Unified view.} Both paradigms can be expressed in a general form $d\mathbf{x}_t = \mathbf{f}(\mathbf{x}_t, t) dt + g(t) d\mathbf{w}_t$. As summarized in  \Cref{tab:ode_sde}, the two frameworks differ primarily in how the drift $\mathbf{f}(\mathbf{x}_t, t)$ is parameterized, either through the score function $\bm{\epsilon}(\mathbf{x}_t, t)$ to reverse a corruption process, or through a velocity field $\mathbf{v}_\theta(\mathbf{x}_t, t)$ to match a probability flow.

\begin{table}[h]
\small
\centering
\caption{\textbf{Generative sampling formulations} ($t=0$ as noise, $t=1$ as data). Note that for diffusion, $\bm{\epsilon}(\mathbf{x}_t, t)$ denotes the score function $\nabla_{\mathbf{x}} \log p_t(\mathbf{x})$.}
\label{tab:ode_sde}
\setlength{\tabcolsep}{9pt} %
\begin{tabular}{l l l}
\toprule
\textbf{Formulation} & \textbf{ODE} & \textbf{SDE} \\
\midrule
Diffusion & 
\begin{tabular}[t]{@{}l@{}} 
    $d\mathbf{x}_t = [ \mathbf{f}(t)\mathbf{x}_t$ \\ 
    $\qquad \hspace{0.3cm} - \frac{1}{2}g^2(t)\bm{\epsilon}(\mathbf{x}_t, t) ] dt$ 
\end{tabular} & 
\begin{tabular}[t]{@{}l@{}} 
    $d\mathbf{x}_t = [ \mathbf{f}(t)\mathbf{x}_t$ \\ 
    $\qquad \hspace{0.3cm} - g^2(t)\bm{\epsilon}(\mathbf{x}_t, t) ] dt + g(t)d\mathbf{w}_t$ 
\end{tabular} \\
\addlinespace[4.5pt] 
Flow Matching & $d\mathbf{x}_t = \mathbf{v}_\theta(\mathbf{x}_t, t) dt$ & $d\mathbf{x}_t = \mathbf{v}_\theta(\mathbf{x}_t, t) dt + g(t)d\mathbf{w}_t$ \\
\bottomrule
\end{tabular}
\end{table}

\subsection{Reward alignment of diffusion and flow models.}\label{sec:prel:alignment}

While pre-trained diffusion and flow models generate high-quality samples, they often fail to meet specific downstream criteria, such as aesthetics~\cite{wu2023human} in the case of image generation, or specific properties such as stability or binding affinity in the case of molecule \citep{guan2023d, xu2022geodiff} and protein design \citep{illuminating, watson2023denovo}. This alignment can be achieved by fine-tuning the model's parameters~\citep{clark2024directly, black2024training} but this requires collecting new samples for every new target property, and is incompatible with new properties that are presented only at inference-time. A promising alternative approach is \textit{reward alignment} where the noise samples during generation are aligned to meet a target reward objective. Conceptually, these aligned or optimized noise samples create a new target distribution that balances two competing objectives: staying true to the structural distribution learned during pretraining while shifting the probability mass towards target samples that maximize the reward. We categorize these noise optimization or search methods as follows.

\paragraph{Gradient-based guidance.}
One approach to noise optimization involves back-propagating the gradients from a reward function through the entire iterative sampling process to update the source (or intermediate) noise samples \citep{ben2024d, wang2024training, guo2024initno, tang2025inference}.
While effective, these methods require differentiable reward functions and incur substantial GPU memory and computational overhead, as they necessitate storing or recomputing the full diffusion trajectory during inference. This limitation becomes particularly severe in high-dimensional settings, such as image or 3D generation, or when many solver steps are required. Furthermore, pure gradient-based refinement often shifts the generation off the natural data manifold, necessitating additional regularization or alignment terms to preserve sample quality \citep{wang2024training}.
An exception is direct noise optimization (DNO) \citep{tang2025inference}, which additionally supports gradient approximations for non-differentiable rewards and an SDE-based mode that can refine intermediate noises along the trajectory. Despite these extensions, the same disadvantages of high runtime and GPU memory consumption remain.

\paragraph{Noise sequence search.} 
Another popular approach is to guide the generation process throughout the whole or parts of the sampling trajectory $\{\mathbf{x}_0, \dots, \mathbf{x}_1\}$. This includes sequential Monte Carlo (SMC) approaches using resampling such as FK-Steering (FKS)~\citep{singhal2025general} and DAS~\citep{kim2025test}, tree search methods such as DSearch and DTS~\citep{li2025dynamic,jain2025diffusion}, and Fast Direct~\citep{tan2025fast} which uses Gaussian process-based black-box optimization over the full noise sequence~\citep{tan2025fast}. While these methods can be effective for stochastic sampling, they typically rely on intermediate reward approximations~\citep{li2024derivative} or are difficult to utilize with batched evaluations. Moreover, SMC-based methods such as FKS and DAS require SDE-based sampling and substantially more reward evaluations, which restricts them to cheap reward models in practice.

\paragraph{Black-box search.}
Most versatile are source noise optimization algorithms which treat the generative model and the reward model as a black-box. They can be applied to any generative model, that maps a source noise distribution to some data distribution, and any reward model which can be differentiable or non-differentiable. They are conceptually much simpler and easy to implement, while also leading to surprisingly good results. This category of approaches is relatively underexplored but includes some recent work by Ma \textit{et al.}~\cite{ma2025inference} where they apply random search and zero-order search for the source noise. Some additional recent (concurrent) work is by Jajal and Eliopoulos \textit{et al.}~\cite{jajal2025inference} where they apply evolutionary and genetic search algorithms for the source noise.
Our work belongs to this category and we apply trust-region search, inspired by Bayesian optimization algorithms~\cite{eriksson2019scalable}, and show strong results across data modalities and reward models. We summarize the conceptual differences between these noise optimization methods in \Cref{tab:comparison}.

\begin{table}[t]
\centering
\small 
\caption{\textbf{Feature comparison across noise optimization methods}. We compare methods by their ability to support gradient-free objectives, black-box models, parallelized batch efficiency, and global or local optimization.}
\label{tab:comparison}
\setlength{\tabcolsep}{5.5pt}
\begin{tabular}{lccccc}
\toprule
\textbf{Method} & \textbf{Grad-free} & \textbf{Black-box} & \textbf{Batch Eff.} & \textbf{Global} & \textbf{Local} \\ \midrule
OC-Flow~\citep{wang2024training}      & \xmark & \xmark & \xmark & \xmark & \cmark \\
Fast Direct \citep{tan2025fast}          & \cmark & \cmark & \cmark & \cmark & \xmark \\
DTS* \citep{jain2025diffusion} & \cmark & \xmark & \xmark & \cmark & \xmark \\ 
Random Search \citep{ma2025inference} & \cmark & \cmark & \cmark & \cmark & \xmark \\ 
Zero-Order  \citep{ma2025inference}  & \cmark & \cmark & \cmark & \xmark & \cmark \\ \midrule
\textbf{TRS (Ours)} & \cmark & \cmark & \cmark & \cmark & \cmark \\ \bottomrule
\end{tabular}
\end{table}

\clearpage

\section{Methodology}
\label{sec:methodology}

\subsection{Problem Statement}
\label{sec:methodology:problem_statement}

\begin{figure}[htbp]
  \centering
  \begin{minipage}[t]{0.33\textwidth}
   \vspace{0pt} %
  Let \(\mathcal{F}:\mathbb{R}^M\to\mathbb{R}^D\) be a pretrained generative diffusion or flow-based model which maps source noise $\mathbf{x}_0$ to a data sample $\mathbf{x}_1$ and \(R:\mathbb{R}^D\to\mathbb{R}\) a reward model processing $\mathbf{x}_1$ and returning a scalar reward $r$. Note that generative models typically operate in a

  \end{minipage}
  \hfill
  \begin{minipage}[t]{0.64\textwidth}
    \vspace{0pt}
    \centering
    \resizebox{\textwidth}{!}{
      \input{figures/pipeline.tex}
    }
    \captionsetup{font=small}
    \caption{\textbf{The optimization black-box.} It is defined by the generative model $\mathcal{F}$ and the reward function $R(\mathbf{x}_1)$, which connect noise samples $\mathbf{x}_0$ with scalar rewards $r$.}
    \label{fig:pipeline_black_box}
  \end{minipage}
  \vspace{-0.2cm}
\end{figure}
\vspace{-0.2cm}
\noindent
 compressed latent space \cite{rombach2022high}, leading to $M<D$.
We do not intervene with the model in any other way, treating it as  
a black-box, as illustrated in  \Cref{fig:pipeline_black_box}. The resulting objective is:
\vspace{-0.1cm}
\begin{equation}
\mathbf{x}_{0}^\star \;=\; \arg\max_{\mathbf{x}_0 \in \mathbb{R}^M}\; R\big(\mathcal{F}(\mathbf{x}_0)\big).
\label{eq:objective}
\end{equation}

\noindent
We treat the entire mapping $R(\mathcal{F}(\mathbf{x}_0))$ as a computationally expensive black-box, as illustrated in \Cref{fig:pipeline_black_box}, either due to the
generative model, the reward model, or the use of large batch sizes.
Consequently, we focus on methods that can effectively steer generation under a strict evaluation budget.

\input{figures/trs_concept}

\subsection{Trust-Region Search (TRS)}
\label{sec:methodology:trs}

TRS is a trust-region search algorithm with structured noise sampling designed for inference-time alignment with expensive black-box evaluations.
Our approach is inspired by TuRBO~\citep{eriksson2019scalable} for Bayesian optimization, but introduces several important modifications for inference-time steering of large generative models.
In particular, unlike surrogate-based Bayesian optimization, TRS relies purely on structured sampling, since we find that surrogates do not contribute meaningfully to performance, due to the highly non-linear noise space, which is difficult to predict under constrained budgets. Additionally, we use a different center selection scheme, by always choosing the top-$k$ observed noises, which is crucial for the performance.
An overview of the method is given in \Cref{alg:trs_compact}.

\begin{algorithm}[H]
\caption{Trust-Region Search (TRS)}
\label{alg:trs_compact}
\begin{algorithmic}[1]
\footnotesize
\Require Total/Warmup Budget $N_{\text{total}}/ N_{\text{warm}}$, Batch size $B$, Regions $k$, Model $\mathcal{F}$, Reward function $R$

\State \textbf{Warm-up:} Sample $\{\mathbf{x}_{0,i}\}_{i=1}^{N_{\text{warm}}} \sim p_0$; evaluate $r_i = R(\mathcal{F}(\mathbf{x}_{0,i}))$ in batches of size $B$.

\State \textbf{Init:} Set centers $\{\mathbf{x}_{0,j}^{\mathrm{c}}\}_{j=1}^k$ to top $k$; set lengths $\ell_j \gets \ell_{\text{init}}$; set remaining budget $N \gets N_{\text{total}} - N_{\text{warm}}$
\While{budget $N \geq B$}
    \State $\mathcal{S}_{\text{batch}} \gets \emptyset$
    \For{region $\mathcal{T}_j$ and sample $b \in \{1..B/k\}$}
        \State $p_{j,b} \sim \text{U}(p_{\min}, p_{\max})$; \ $\mathbf{m}_{j,b}\sim \text{Ber}(p_{j,b})^M$ \ 
        \State $\tilde{\mathbf{x}}_{0,j,b} \sim \text{Perturb}(\ell_j)$ \Comment{Gauss or Sobol}
        \State $\mathbf{x}_{0,j,b} \gets \mathbf{x}_{0,j}^{\mathrm{c}} + (\tilde{\mathbf{x}}_{0,j,b} \odot \mathbf{m}_{j,b})$ \ 
        \State $\mathcal{S}_{\text{batch}} \gets \mathcal{S}_{\text{batch}} \cup \{\mathbf{x}_{0,j,b}\}$
    \EndFor
    \State \textbf{Evaluate:} $\{r_{j,b}\} \gets R(\mathcal{F}(\mathcal{S}_{\text{batch}}))$ 
    \State \textbf{Adapt:} Update lengths $\{\ell_j\}$
    \State \textbf{Shift:} Re-center $\{\mathbf{x}_{0,j}^{\mathrm{c}}\}$ to the global top $k$.
    \State $N \gets N - B$
\EndWhile
\State \Return best $\{\mathcal{F}(\mathbf{x}_0), r\}$
\end{algorithmic}
\end{algorithm}

\paragraph{Warm-up.}
We begin with a short warm-up phase for bootstrapping the search.
We sample $N_{\text{warm}}$ initial noise samples from the model prior $p_0$,
\begin{equation}
\mathbf{x}_{0,i} \sim p_0, \qquad
\mathbf{x}_{1,i} = \mathcal{F}(\mathbf{x}_{0,i}), \qquad
r_i = R(\mathbf{x}_{1,i}),
\end{equation}
with $i = 1, \dots, N_{\text{warm}}$,
and select the top-performing $k$ points as initial trust-region centers
$\{\mathbf{x}_{0,j}^{\mathrm{c}}\}_{j=1}^k$.
All trust regions are initialized with the same side length
$\ell_j = \ell_{\text{init}}$.
In practice, this warm-up is implemented using standard iterations with batch size $B$. We find allocating approximately $20\%$ of the total evaluation budget $N_{\text{total}}$ to this phase to be a robust heuristic across tasks and budgets; an ablation is provided in Appendix~\ref{sec:app:ablations:warm-up}.

\paragraph{Trust-region iterations.}
After warm-up, the algorithm maintains $k$ hypercubic trust regions
$\mathcal{T}^j \subset \mathbb{R}^M$, each defined by a center
$\mathbf{x}_{0,j}^{\mathrm{c}}$ and side length $\ell_j$. In each iteration, we perform the following steps, where a global batch set $\mathcal{S}_{\text{batch}}$ of size $B$ across all regions is maintained.

\begin{enumerate}[leftmargin=0.7cm, labelsep=0.5em]
  \item \textbf{Propose:} For each region $j$, generate $B/k$ candidate noise vectors $\mathbf{x}_{0,j,b}$ by perturbing the center $\mathbf{x}_{0,j}^{\mathrm{c}}$. Following \citep{eriksson2019scalable}, perturbations $\tilde{\mathbf{x}}_{0,j,b}$ are generated within an axis-aligned hypercube with side length $\ell_j$. Additionally, we combine these perturbations with a stochastic coordinate mask. This mask is built by drawing a perturbation probability $p_{j,b} \sim \text{Uniform}(p_{\min}, p_{\max})$ and applying $\mathbf{m}_{j,b} \sim \mathrm{Bernoulli}(p_{j,b})^M$. The resulting candidates are:
\begin{equation}
    \mathbf{x}_{0,j,b} = \mathbf{x}_{0,j}^{\mathrm{c}} + \big(\tilde{\mathbf{x}}_{0,j,b}\big) \odot \mathbf{m}_{j,b}.
\end{equation}
  \item \textbf{Evaluate:} All $B$ candidates across regions are aggregated into $\mathcal{S}_{\text{batch}}$ and evaluated in parallel to obtain the rewards $\{r_{j,b}\} = R(\mathcal{F}(\mathcal{S}_{\text{batch}}))$.
  \item \textbf{Update:} After batch evaluation, we adapt trust-region side lengths $\{\ell_j\}$ using success-based rules per region. A key distinction from vanilla TuRBO~\citep{eriksson2019scalable} is that trust regions are not treated independently.
While exploring multiple regions is beneficial early on, we observe that only a subset typically remains promising.
Accordingly, after each batch iteration we re-center all trust regions $\{\mathbf{x}_{0,j}^{\mathrm{c}}\}$ at the globally best $k$ points observed so far.
This mechanism naturally shifts computation from exploration toward exploitation, reallocating evaluation budget to promising regions over time.
An ablation study on different re-centering strategies is provided in Appendix~\ref{sec:app:ablations:center_selection}.
\end{enumerate}

\paragraph{Perturbations.} For generating the perturbations in the proposal step we use two different schemes. Deterministic low-discrepancy samplers such as Sobol~\citep{sobol1967distribution} are efficient to fill search spaces, but can only be computed only up to 21k dimensions \citep{doi:10.1137/070709359}. For higher dimensional search spaces, like in our experiment with SDXL-Lightning in \Cref{sec:exp:text-to-image} with noise dimension 65,536, we use a Gaussian perturbation scheme, that is designed to follow the trust-region hypercube exploration closely. 

For Sobol-based proposals, a point $\mathbf{u}_{j,b} \in [0,1]^M$ is sampled and mapped affinely into the trust region.
Let
$\mathbf{a}_j = -\tfrac{1}{2}\ell_j \mathbf{1}_M$
and
$\mathbf{b}_j = \tfrac{1}{2}\ell_j \mathbf{1}_M$,
where $\mathbf{1}_M$ denotes a $M$-dimensional vector of ones. 
The resulting proposal is

\begin{equation}
\tilde{\mathbf{x}}_{0,j,b} = \mathbf{a}_j + (\mathbf{b}_j - \mathbf{a}_j) \odot \mathbf{u}_{j,b}.
\end{equation}

For Gaussian perturbations, we sample
\begin{equation}
\tilde{\mathbf{x}}_{0,j,b} \sim \mathcal{N}(\mathbf{0}, \sigma_j^2 \mathbf{I}), 
\qquad 
\sigma_j = \ell_j / \sqrt{12},
\end{equation}
where $\mathbf{I} \in \mathbb{R}^{M \times M}$ is the identity matrix. The standard deviation $\sigma_j = \ell_j / \sqrt{12}$ is chosen to match the variance of a uniform distribution $\mathcal{U}[-\ell_j/2, \ell_j/2]$, ensuring the Gaussian proposals cover the trust-region hypercube with equivalent spread.

\paragraph{Trust-region adaptation.} The update rules for the trust-region side length largely follow the strategy introduced in TuRBO~\citep{eriksson2019scalable}. For each region, we maintain success and failure counters based on whether newly evaluated candidates improve upon the best value observed within that region. If a candidate yields an improvement, the success counter is incremented and the failure counter is reset.  Otherwise, the failure counter is incremented and the success counter is reset. Two thresholds, $c_{\mathrm{succ}}$ and $c_{\mathrm{fail}}$, govern the adaptation of the trust-region side length $\ell_j$.
When the success counter reaches $c_{\mathrm{succ}}$, the trust region is expanded according to
\vspace{-0.1cm}
\begin{equation}
\ell_j^{\text{new}} = \min(\ell_j \cdot \alpha_{\ell}, \ell_j^{\max}),
\end{equation}

where $\alpha_{\ell}$ is an expansion factor for the length (we set it to $1.5$ by default).
Conversely, when the failure counter reaches $c_{\mathrm{fail}}$, the trust region is contracted as
\vspace{-0.1cm}
\begin{equation}
\ell_j^{\text{new}} =  \max(\ell_j / \alpha_{\ell}, \ell_j^{\min}).
\end{equation}
This mechanism enables adaptive control over the exploration scale based on recent optimization progress. We further illustrate the core steps of TRS in \Cref{fig:trs_flow_final} and provide additional details in the appendix.

\section{Experiments}
\label{sec:benchmarks}

We evaluate our trust-region search (TRS) method across three diverse generative settings: text-to-image diffusion models in \Cref{sec:exp:text-to-image}, ODE-based molecular flow matching in \Cref{sec:exp:molecule_generation}, and protein backbone design in \Cref{sec:exp:protein_design}. These experiments span different modalities, reward models, and sampling procedures, allowing us to assess both the effectiveness and robustness of TRS under varying optimization budgets and reward costs. Across all settings, we compare against representative gradient-based, noise-sequence, and black-box search baselines under matched compute constraints. Overall, the results demonstrate that TRS consistently achieves stronger alignment with target objectives while maintaining sample quality and requiring minimal task-specific tuning.

\begin{figure}[htbp]
    \centering
    \definecolor{highlight}{rgb}{0.94, 0.96, 1.0}
    
    \setlength{\tabcolsep}{0pt} 
    \setlength{\aboverulesep}{2pt} 
    \setlength{\belowrulesep}{2pt}

    \newlength{\hsep}
    \newlength{\vsep}
    \setlength{\hsep}{4pt} 
    \setlength{\vsep}{28pt}
    
    \renewcommand{\arraystretch}{0} 

    \begin{tabular}{
        >{\raggedright\arraybackslash\tiny\itshape}m{1.5cm} %
        @{\hspace{\hsep}} >{\columncolor{highlight}}c %
        @{\hspace{\hsep}}c@{\hspace{\hsep}}c@{\hspace{\hsep}}c@{\hspace{\hsep}}c@{\hspace{\hsep}}c %
    }
        \toprule
        \rule{0pt}{3ex} \textbf{\scriptsize Prompt} & \textbf{\scriptsize TRS (Ours)} & \textbf{\scriptsize RS} & \textbf{\scriptsize ZO} & \textbf{\scriptsize DTS} & \textbf{\scriptsize FD} & \textbf{\scriptsize OC-Flow} \\[1.5ex]
        \midrule
        
        Three cats and two dogs sitting on the grass.& 
        $\vcenter{\hbox{\includegraphics[width=0.138\textwidth]{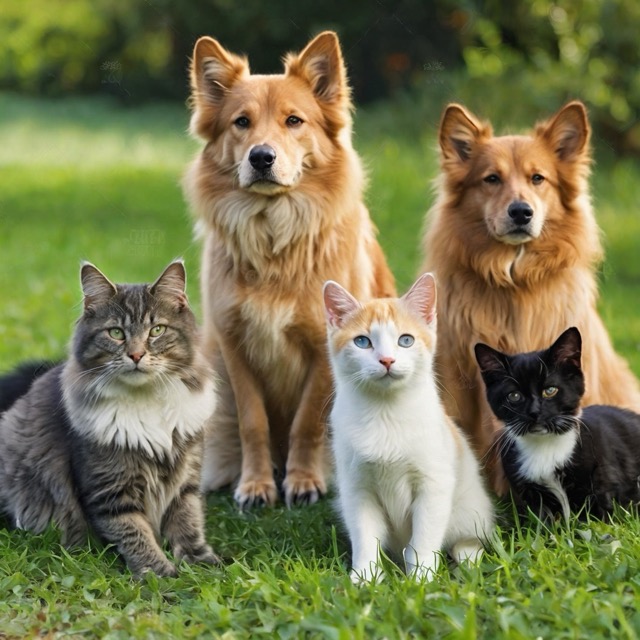}}}$ & 
        $\vcenter{\hbox{\includegraphics[width=0.138\textwidth]{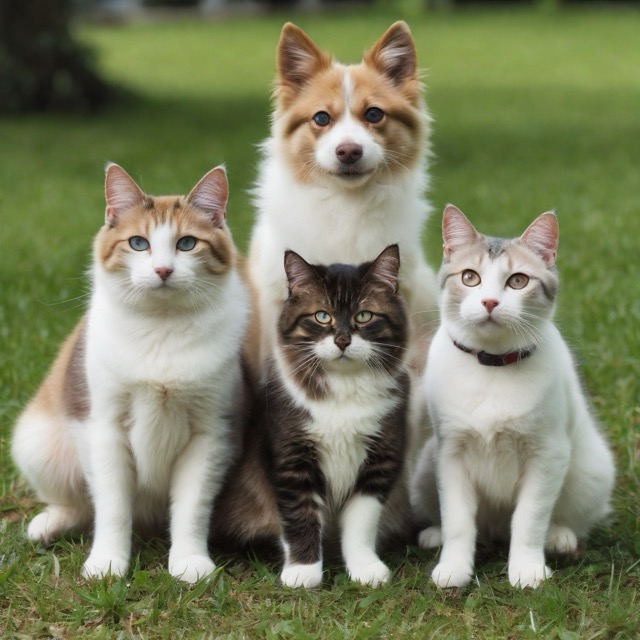}}}$ & 
        $\vcenter{\hbox{\includegraphics[width=0.138\textwidth]{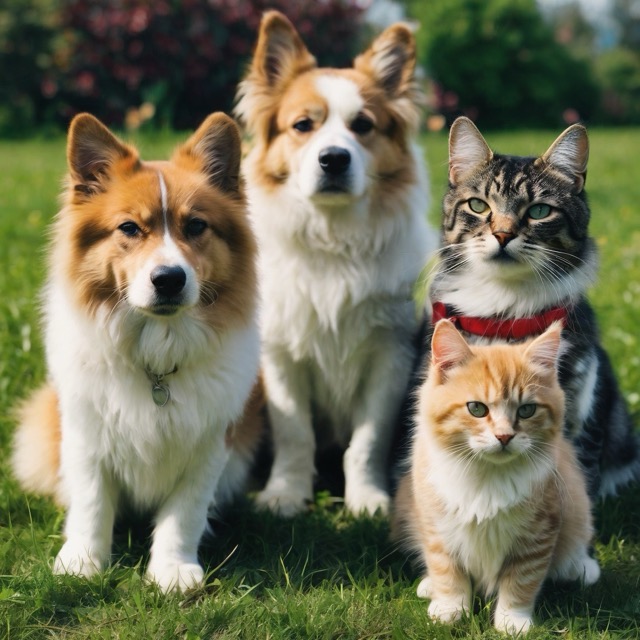}}}$ & 
        $\vcenter{\hbox{\includegraphics[width=0.138\textwidth]{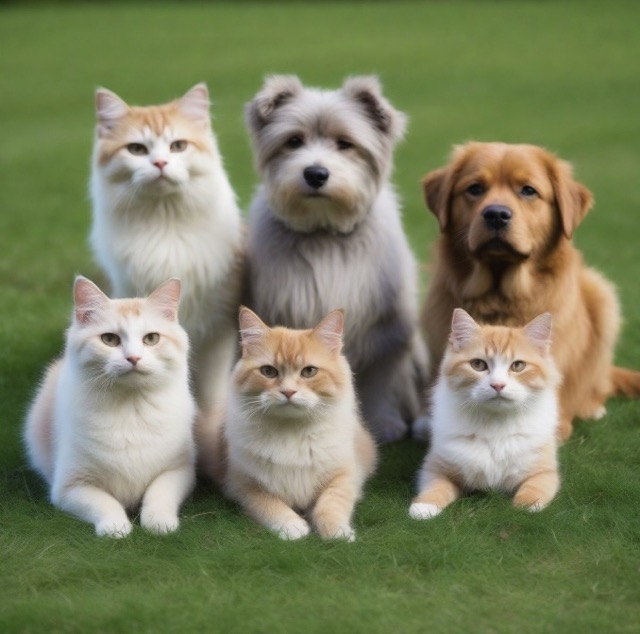}}}$ & 
        $\vcenter{\hbox{\includegraphics[width=0.138\textwidth]{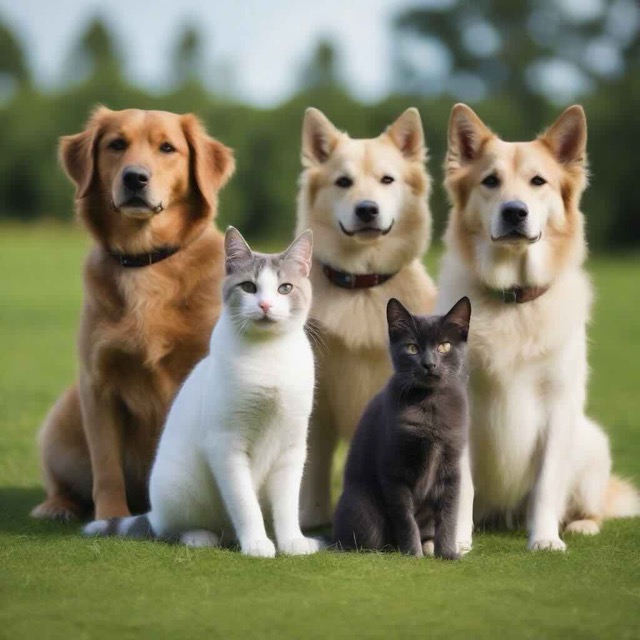}}}$ &
        $\vcenter{\hbox{\includegraphics[width=0.138\textwidth]{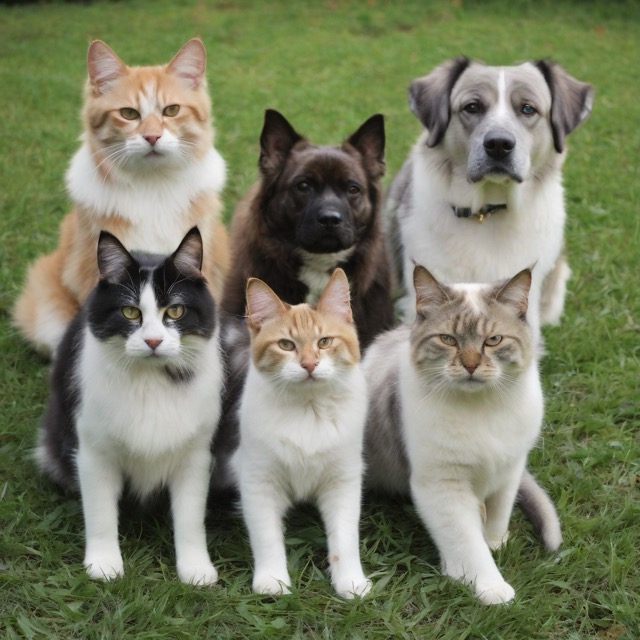}}}$ 
        \\[\vsep] %

        A storefront with 'Google Brain Toronto' written on it.
 & 
        $\vcenter{\hbox{\includegraphics[width=0.138\textwidth]{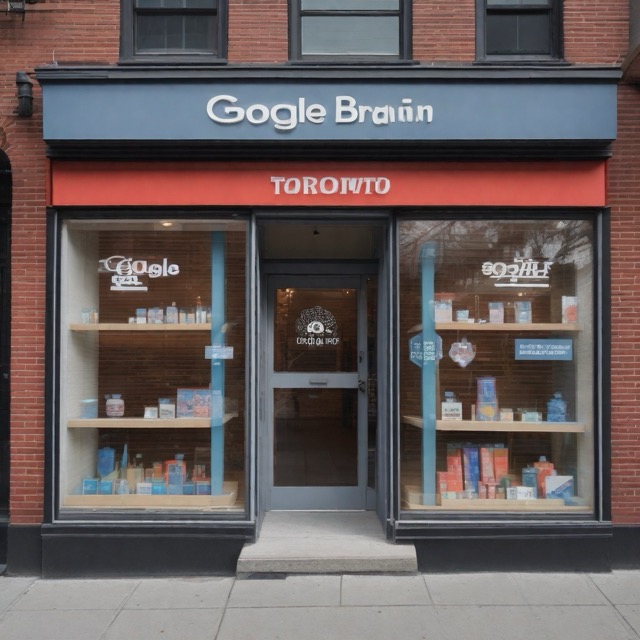}}}$ & 
        $\vcenter{\hbox{\includegraphics[width=0.138\textwidth]{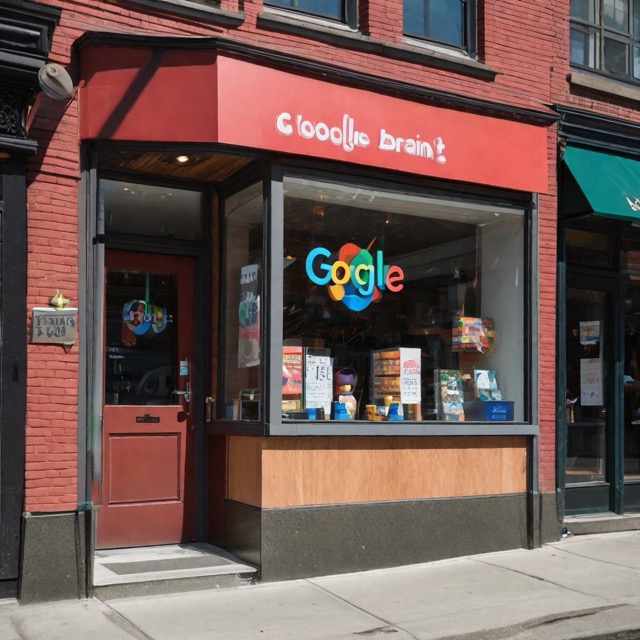}}}$ & 
        $\vcenter{\hbox{\includegraphics[width=0.138\textwidth]{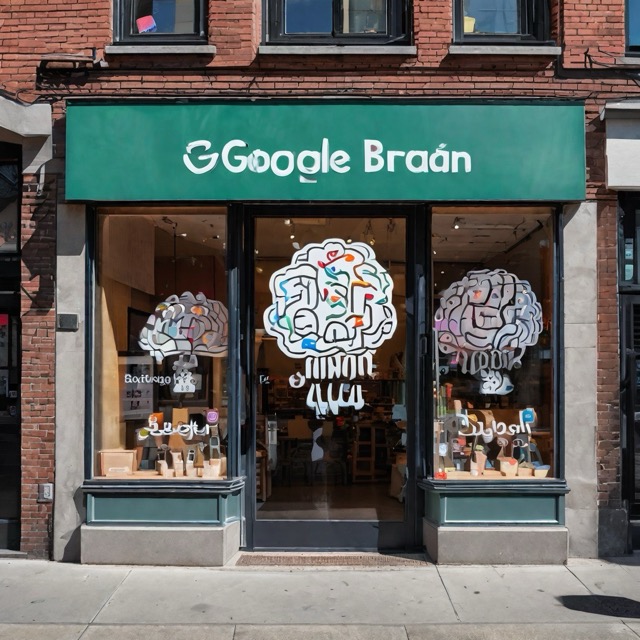}}}$ & 
        $\vcenter{\hbox{\includegraphics[width=0.138\textwidth]{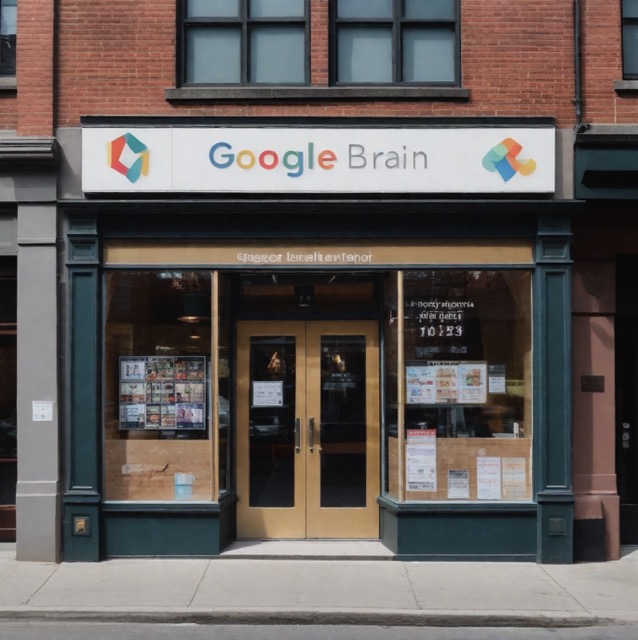}}}$ & 
        $\vcenter{\hbox{\includegraphics[width=0.138\textwidth]{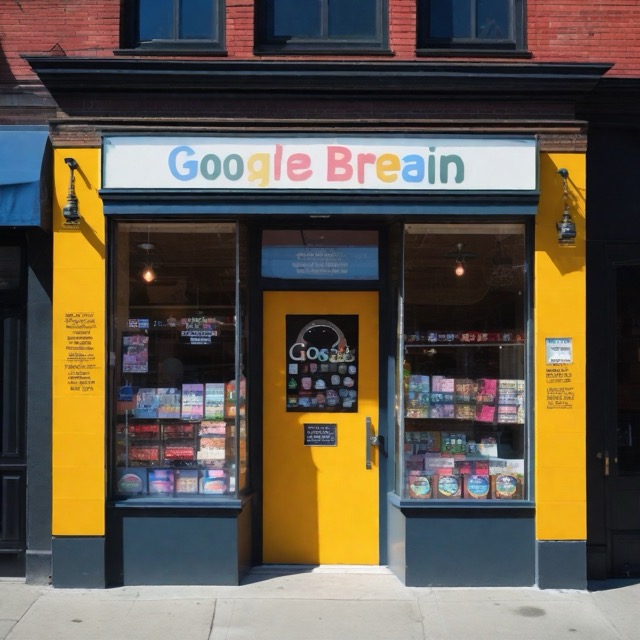}}}$ &
        $\vcenter{\hbox{\includegraphics[width=0.138\textwidth]{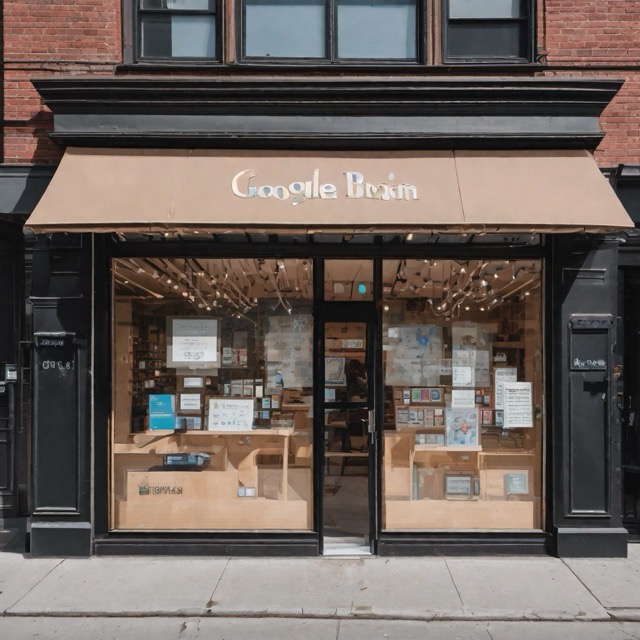}}}$ \\[\vsep] %
        
        A zebra underneath a broccoli.
 & 
        $\vcenter{\hbox{\includegraphics[width=0.138\textwidth]{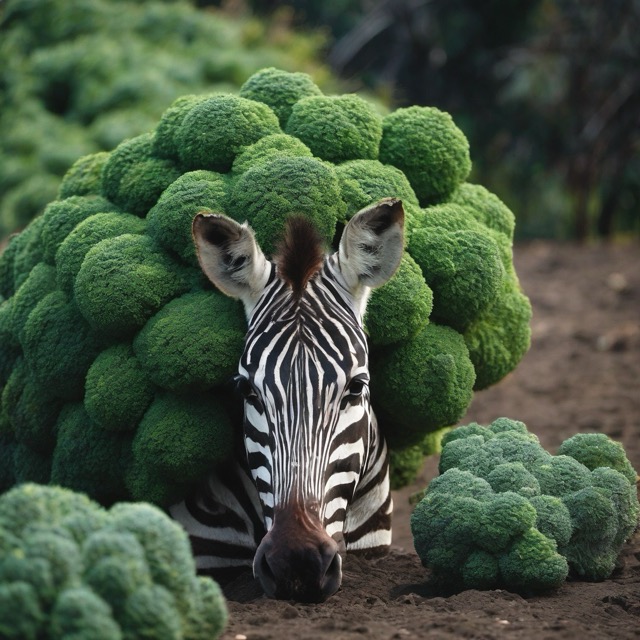}}}$ & 
        $\vcenter{\hbox{\includegraphics[width=0.138\textwidth]{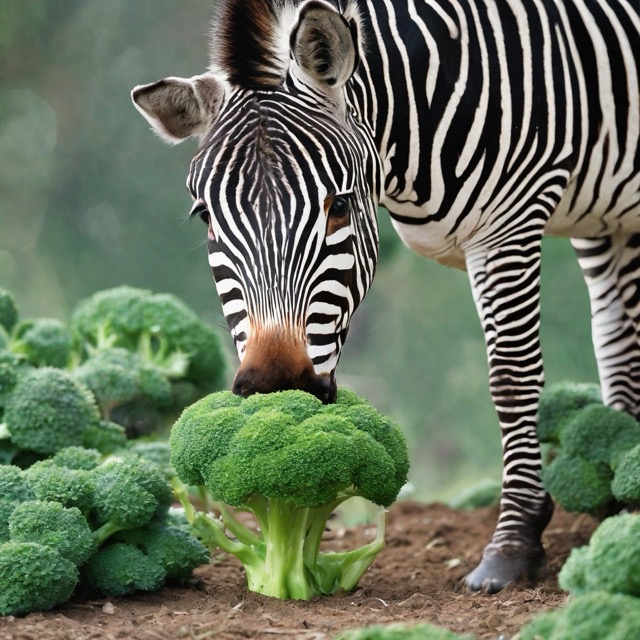}}}$ & 
        $\vcenter{\hbox{\includegraphics[width=0.138\textwidth]{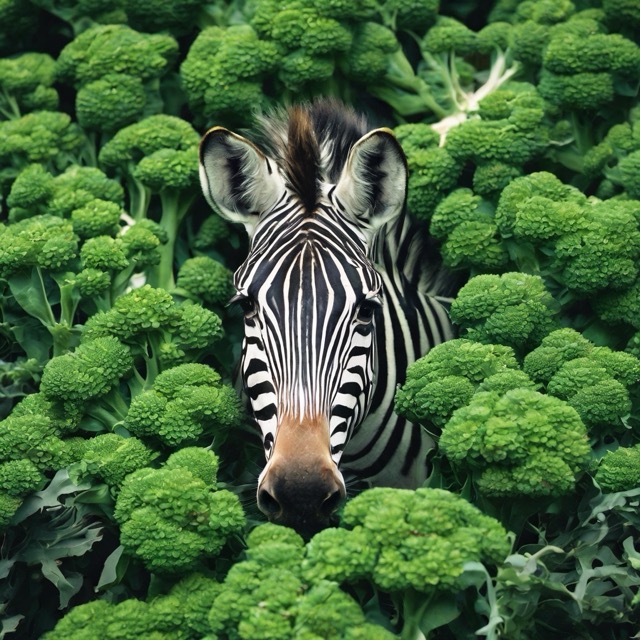}}}$ & 
        $\vcenter{\hbox{\includegraphics[width=0.138\textwidth]{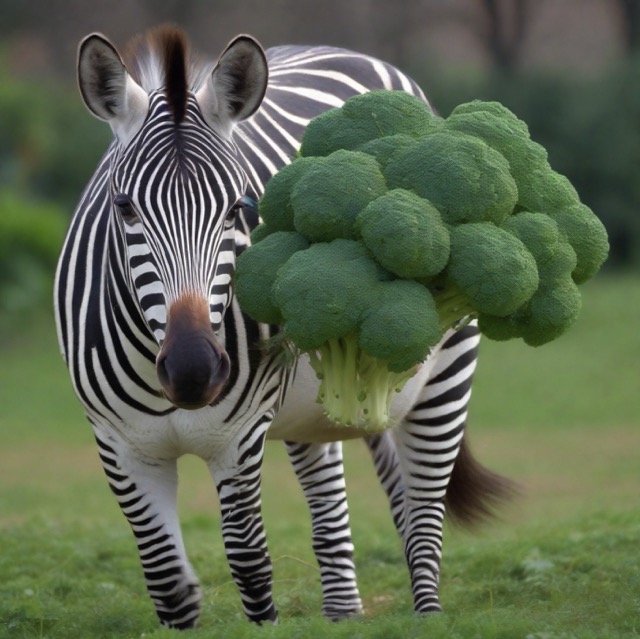}}}$ & 
        $\vcenter{\hbox{\includegraphics[width=0.138\textwidth]{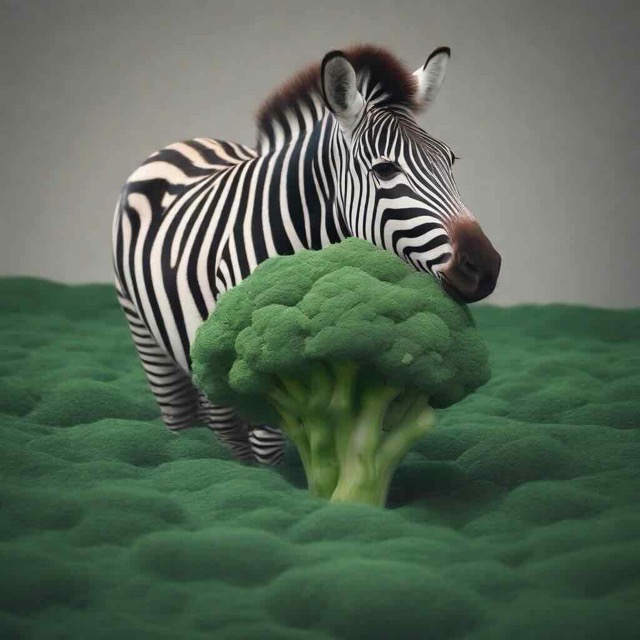}}}$ &
        $\vcenter{\hbox{\includegraphics[width=0.138\textwidth]{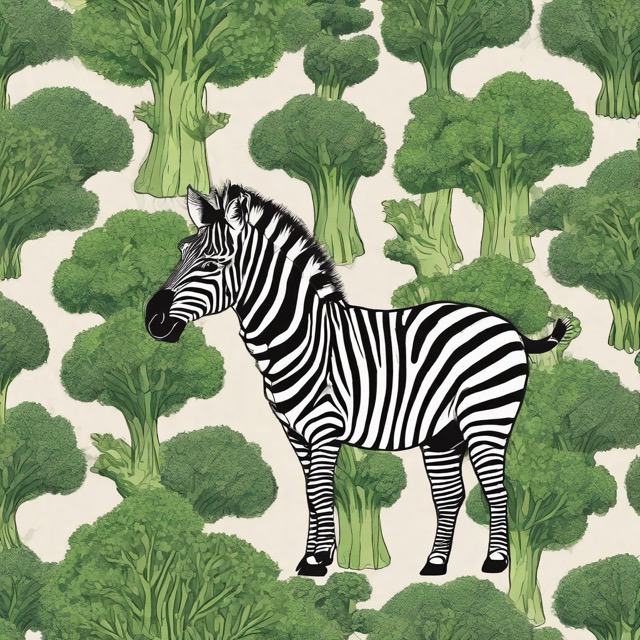}}}$ \\
        
        \bottomrule
    \end{tabular}
    \raggedright
    \caption{\textbf{Optimized text-to-image samples.} Examples from different algorithms for all methods in Section~\ref{sec:exp:text-to-image}, with SDXL-Lightning \citep{lin2024sdxllightning}. The first row is optimized with ImageReward \citep{xu2023imagereward} and the lower two rows with HPSv2 \citep{wu2023human}. All prompts are from DrawBench \citep{saharia2022photorealistic}. Outputs from TRS adhere to the prompt more closely in terms of specified animal count, text, and relative positions. Further examples (incl. randomized) are given in the appendix.}
    \label{fig:exp:img:qualitative}
\end{figure}

\subsection{Text-to-Image}
\label{sec:exp:text-to-image}

\paragraph{Setup.} 
We evaluate on the DrawBench \citep{saharia2022photorealistic} benchmark which comprises 200 prompts across diverse categories (e.g., counting, composition). We use two generative models: SD1.5 \citep{rombach2022high} (50 steps) and SDXL-Lightning \citep{lin2024sdxllightning} (8 steps), with ImageReward \citep{xu2023imagereward} and HPSv2 \citep{wu2023human} as the reward models. TRS is compared against gradient-based OC-Flow \citep{wang2024training}, state-of-the-art Diffusion Tree Sampling (DTS)* \citep{jain2025diffusion}, Fast Direct (FD) \citep{tan2025fast}, and black-box methods including random and zero-order search (RS, ZO) \citep{ma2025inference}. Additionally, we compare to diagonal CMA-ES, which is close to concurrent work about GA/ES approaches for noise search \citep{jajal2025inference}, and a variant of our TRS algorithm without adaptive region lengths. We use diffusion backbones for T2I, but note that TRS treats the generator as a deterministic map from source noise to sample and is therefore agnostic to whether sampling is ODE- or SDE-based; our molecule and protein experiments already employ flow-matching models (EquiFM, Proteina), supporting the generality of our approach. The number of function evaluations (NFE) is fixed across methods. Further implementation details for the baselines are in Appendix~\ref{sec:app:baselines}. 

\paragraph{Metrics.}
We report the mean of the best per-prompt reward \Cref{tab:draw_bench_comprehensive} and average best rewards across varying NFE budgets in \Cref{fig:exp:text_to_image_scale}. 
Additionally, we report the number of reward oracle evaluations induced by the fixed NFE budget (\Cref{tab:draw_bench_comprehensive}). Direct NFE comparison for backpropagation-based OC-Flow is provided via runtime analysis in Appendix~\ref{sec:app:runtime}. Note that the ImageReward outputs are designed to roughly follow a normal distribution, so outputs beyond the range of $[-2,2]$ are rare. HPSv2 shows differences in a smaller scale where typical high-quality images fall within the range of $[0.25, 0.35]$.

\paragraph{Results.}  \Cref{tab:draw_bench_comprehensive} demonstrates that TRS is consistently better than all baselines across generative models and reward functions. On both models, TRS exceeds the state-of-the-art DTS* performance with up to $4\times$ reduction in wall-clock time and fewer reward evaluations. OC-Flow and Fast Direct underperform relative to random search, highlighting the difficulty of surrogate modeling and gradient-based optimization in high-dimensional noise spaces, while TRS shows the best results. Diagonal CMA-ES (with zero initialization, as full CMA-ES is infeasible due to its $\mathcal{O}(d^2)$ memory) performs comparably to TRS on the image task, yet TRS remains best overall. The SMC-based FK-Steering and DAS are likewise outperformed at an equal NFE budget, while requiring SDE-based sampling and substantially more reward calls. Finally, TRS with matched warm-up and top-$k$ selection but a fixed radius (i.e., without trust-region adaptation) is also inferior to full TRS (\Cref{tab:draw_bench_comprehensive}), isolating trust-region adaptation as an important component.
\Cref{fig:exp:img:qualitative} shows some qualitative results and \Cref{fig:exp:text_to_image_scale} further illustrates the scaling trends from small to high budgets, where TRS generally scales best, whereas the other algorithms saturate earlier.
Full hyperparameter configurations of all methods are displayed in Appendix~\ref{sec:app:hyper_params}.

\begin{table}[t]
\centering
\fontsize{8}{9}\selectfont
\captionsetup{font=small}
\caption{\textbf{Comparison on DrawBench.} We report the mean best rewards (IR, HPSv2) for SD1.5 and SDXL-Lightning. Reward calls, memory (GB), and wall-clock time (s per prompt) are reported for SD1.5+IR where available. $\uparrow / \downarrow$ indicate the direction of improvement. Best values are in \textbf{bold}, second best are \underline{underlined}. Unlike the other entries, CMA-ES~\citep{hansen2001completely} is a general-purpose optimizer that we adapt to source-noise search; the concurrent ES/GA line of work is~\citep{jajal2025inference}.}
\label{tab:draw_bench_comprehensive}

\renewcommand{\arraystretch}{1.2}
\setlength{\tabcolsep}{6pt}

\resizebox{\textwidth}{!}{%
\begin{tabular}{@{}l cc cc ccc@{}}
\toprule
& \multicolumn{2}{c}{\textbf{SD1.5}} & \multicolumn{2}{c}{\textbf{SDXL-L}} & & & \\
\cmidrule(lr){2-3} \cmidrule(lr){4-5}
\textbf{Algorithm} & {IR $\uparrow$} & {HPS $\uparrow$} & {IR $\uparrow$} & {HPS $\uparrow$} & \textbf{Calls $\downarrow$} & \textbf{Mem. $\downarrow$} & \textbf{Time $\downarrow$} \\
\midrule

Base & -0.16 & 0.246 & 0.50 & 0.262 & 1 & 4.32 & 2.5 \\

\rowcolor{black!5}[0pt][0pt]\multicolumn{8}{l}{\textbf{Gradient-based guidance}} \\
OC-Flow ~\citep{wang2024training} & 0.42 & 0.277 & 0.85 & 0.287 & -- & 25.45 & 335 \\

\rowcolor{black!5}[0pt][0pt]\multicolumn{8}{l}{\textbf{Noise sequence search}} \\
FK-Steering~\citep{singhal2025general} & 1.48 & 0.301 & 1.55 & 0.322 & 2400 & 16.20 & 234.0 \\
DAS~\citep{kim2025test} & 1.51 & 0.306 & 1.61 & 0.327 & 2400 & 76.78 & 378.9 \\
Fast Direct ~\citep{tan2025fast} & 1.35 & 0.296 & 1.50 & 0.314 & 420 & 16.32 & 231.6 \\
DTS* ~\citep{jain2025diffusion} & 1.59 & 0.303 & 1.62 & 0.327 & 503 & 11.9 & 924.9 \\

\rowcolor{black!5}[0pt][0pt]\multicolumn{8}{l}{\textbf{Black-box search}} \\
Random ~\citep{ma2025inference} & 1.44 & 0.302 & 1.54 & 0.324 & 400 & 15.62 & 208.3 \\
Zero-order ~\citep{ma2025inference} & 1.50 & 0.313 & 1.59 & 0.332 & 400 & 15.64 & 215.3 \\
CMA-ES (diag)~\citep{hansen2001completely} & 1.56 & \underline{0.319} & \textbf{1.67} & \textbf{0.344} & 400 & 15.68 & 226.6 \\
\midrule
TRS w/o TR adapt. & \underline{1.60} & 0.318 & 1.64 & 0.333 & 400 & 15.64 & 226.5 \\

\textbf{TRS (Ours)} & {\textbf{1.62}} & {\textbf{0.322}} & \underline{1.66} & \underline{0.340} & 400 & 15.64 & 226.5 \\
\bottomrule
\end{tabular}%
}
\end{table}

\begin{figure}[h!]
\centering
\begin{tikzpicture}
\begin{groupplot}[
    group style={
        group size=4 by 1,
        horizontal sep=0.8cm, %
    },
    width=4.1cm, 
    height=4cm,
    grid=both,
    grid style={gray!30},
    tick label style={font=\scriptsize},
    label style={font=\small},
    title style={font=\scriptsize},
    scaled x ticks=false,
    scaled y ticks=false,
    yticklabel={\pgfmathprintnumber[fixed,precision=2]{\tick}},
    xtick={5000,10000,15000,20000},
    xticklabels={5k,10k,15k,20k},
    every axis plot/.append style={
        line width=1.1pt,
        mark size=1.2
    }
]

\nextgroupplot[
    title={SD1.5 (IR) ($\uparrow$)},
    ymin=1.28, ymax=1.66,
    ytick={1.30,1.40,1.50,1.60},
    ylabel={Mean best reward},
    xlabel={NFE},
    legend to name=sharedlegend4,
    legend columns=4,
    legend style={font=\small, draw=none}
]
\addplot[color=method1color, mark=*]
    coordinates {(5000,1.3166) (10000,1.527) (15000,1.60036) (20000,1.6217)};
\addlegendentry{TRS (Ours)}
\addplot[color=method2color, mark=*]
    coordinates {(5000,1.30711) (10000,1.3877) (15000,1.411) (20000,1.442)};
\addlegendentry{Random}
\addplot[color=method3color, mark=*]
    coordinates {(5000,1.3877) (10000,1.4153) (15000,1.496) (20000,1.508)};
\addlegendentry{Zero-order}
\addplot[color=method4color, mark=*]
    coordinates {(5000,1.35) (10000,1.48) (15000,1.54) (20000,1.5775)};
\addlegendentry{DTS*}

\nextgroupplot[
    title={SD1.5 (HPS) ($\uparrow$)},
    ymin=0.293, ymax=0.325,
    xlabel={NFE},
    ytick={0.30,0.31,0.32}
]
\addplot[color=method1color, mark=*] coordinates {(5000,0.304) (10000,0.3137) (15000,0.3184) (20000,0.3223)};
\addplot[color=method2color, mark=*] coordinates {(5000,0.2942) (10000,0.2982) (15000,0.3001) (20000,0.3018)};
\addplot[color=method3color, mark=*] coordinates {(5000,0.3019) (10000,0.3079) (15000,0.3110) (20000,0.3126)}; 
\addplot[color=method4color, mark=*] coordinates {(5000,0.299) (10000,0.303) (15000,0.304) (20000,0.305)};

\nextgroupplot[
    title={SDXL (IR) ($\uparrow$)},
    ymin=1.44, ymax=1.68,
    ytick={1.45,1.55,1.65},
    xtick={800,1600,2400,3200},
    xlabel={NFE},
    xticklabels={0.8k,1.6k,2.4k,3.2k}
]
\addplot[color=method1color, mark=*] coordinates {(800,1.456) (1600,1.570) (2400,1.614) (3200,1.66)};
\addplot[color=method2color, mark=*] coordinates {(800,1.470) (1600,1.522) (2400,1.540) (3200,1.551)};
\addplot[color=method3color, mark=*] coordinates {(800,1.505) (1600,1.552) (2400,1.576) (3200,1.59)};
\addplot[color=method4color, mark=*] coordinates {(800,1.504) (1600,1.568) (2400,1.597) (3200,1.619)};

\nextgroupplot[
    title={SDXL (HPS) ($\uparrow$)},
    ymin=0.315, ymax=0.345,
    ytick={0.32,0.33,0.34},
    xtick={800,1600,2400,3200},
    xlabel={NFE},
    xticklabels={0.8k,1.6k,2.4k,3.2k}
]
\addplot[color=method1color, mark=*] coordinates {(800,0.323) (1600,0.330) (2400,0.335) (3200,0.339)};
\addplot[color=method2color, mark=*] coordinates {(800,0.318) (1600,0.321) (2400,0.323) (3200,0.324)};
\addplot[color=method3color, mark=*] coordinates {(800,0.325) (1600,0.329) (2400,0.330) (3200,0.331)};
\addplot[color=method4color, mark=*] coordinates {(800,0.316) (1600,0.321) (2400,0.325) (3200,0.327)};

\end{groupplot}

\node[yshift=-1.4cm] at ($(group c1r1.south)!0.5!(group c4r1.south)$) {\pgfplotslegendfromname{sharedlegend4}};

\end{tikzpicture}
\captionsetup{font=small}
\caption{\textbf{Inference-time scaling for text-to-image.} Here we plot the mean best rewards for SD1.5/SDXL optimizing for HPSv2 and ImageReward across different NFE budgets. TRS shows the best scaling performance among all methods.}
\label{fig:exp:text_to_image_scale}
\end{figure}

\setlength{\tabcolsep}{0pt}
\setlength{\aboverulesep}{2pt}
\setlength{\belowrulesep}{2pt}
\newlength{\structcellwidth}

\setlength{\structcellwidth}{\dimexpr(\linewidth - 6\hsep - 34pt)/7\relax}
\newlength{\molblockwidth}
\newlength{\protblockwidth}
\setlength{\molblockwidth}{4\structcellwidth}
\setlength{\protblockwidth}{3\structcellwidth}
\renewcommand{\arraystretch}{0}

\begin{figure}[htbp]
\raggedright
\small

\begin{subfigure}[t]{\molblockwidth}
  \centering
  \subcaption{\textbf{Molecules}}\label{fig:exp:mols_and_prots:a}\par\vspace{0.5ex}
  \begin{tabular}{@{}c@{\hspace{\hsep}}c@{\hspace{\hsep}}c@{\hspace{\hsep}}c@{}}
    \toprule
    \rule{0pt}{2.5ex}
    \textbf{\scriptsize OC-Flow} & \textbf{\scriptsize Random} & \textbf{\scriptsize Zero-order} & \textbf{\scriptsize TRS}
    \\[0.6ex]
    \midrule
    \rule{0pt}{\structcellwidth}
    \imgwithreward{\structcellwidth}{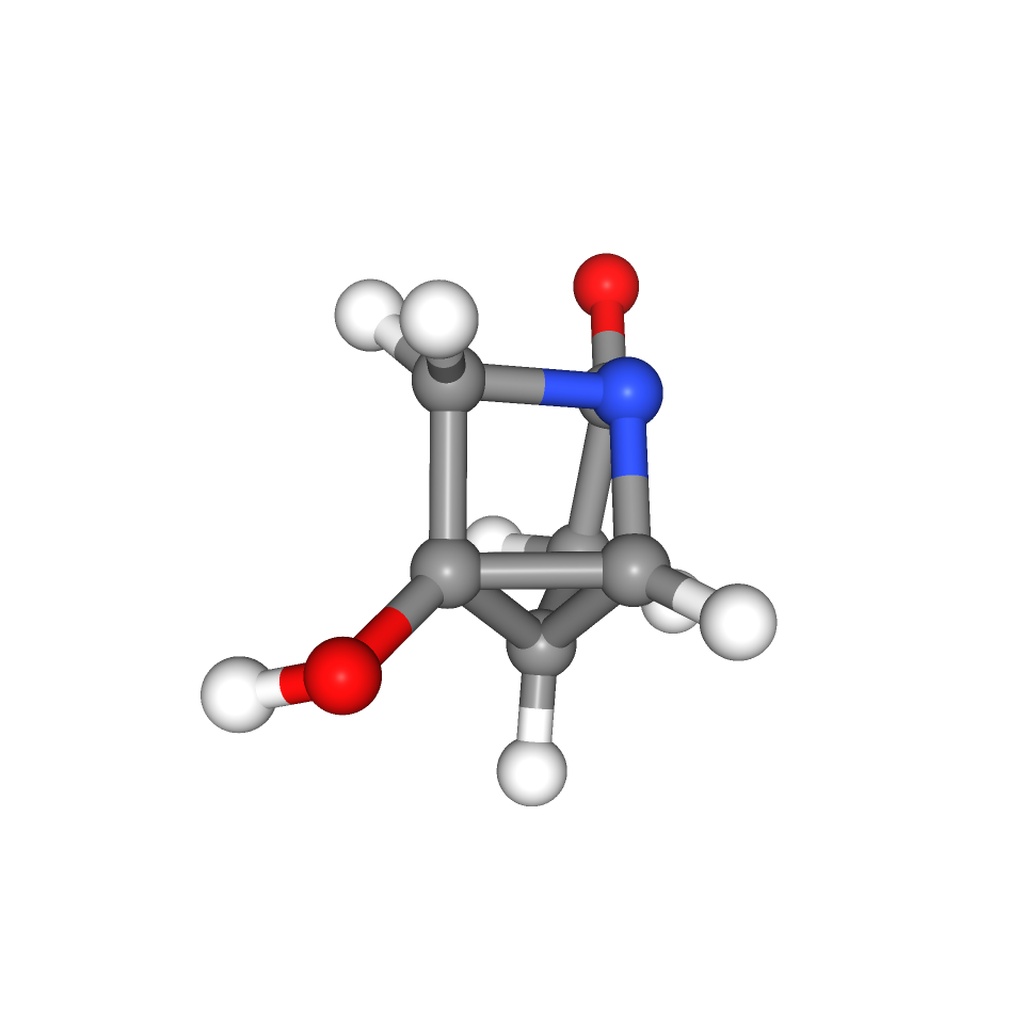}{1.04}
    & \imgwithreward{\structcellwidth}{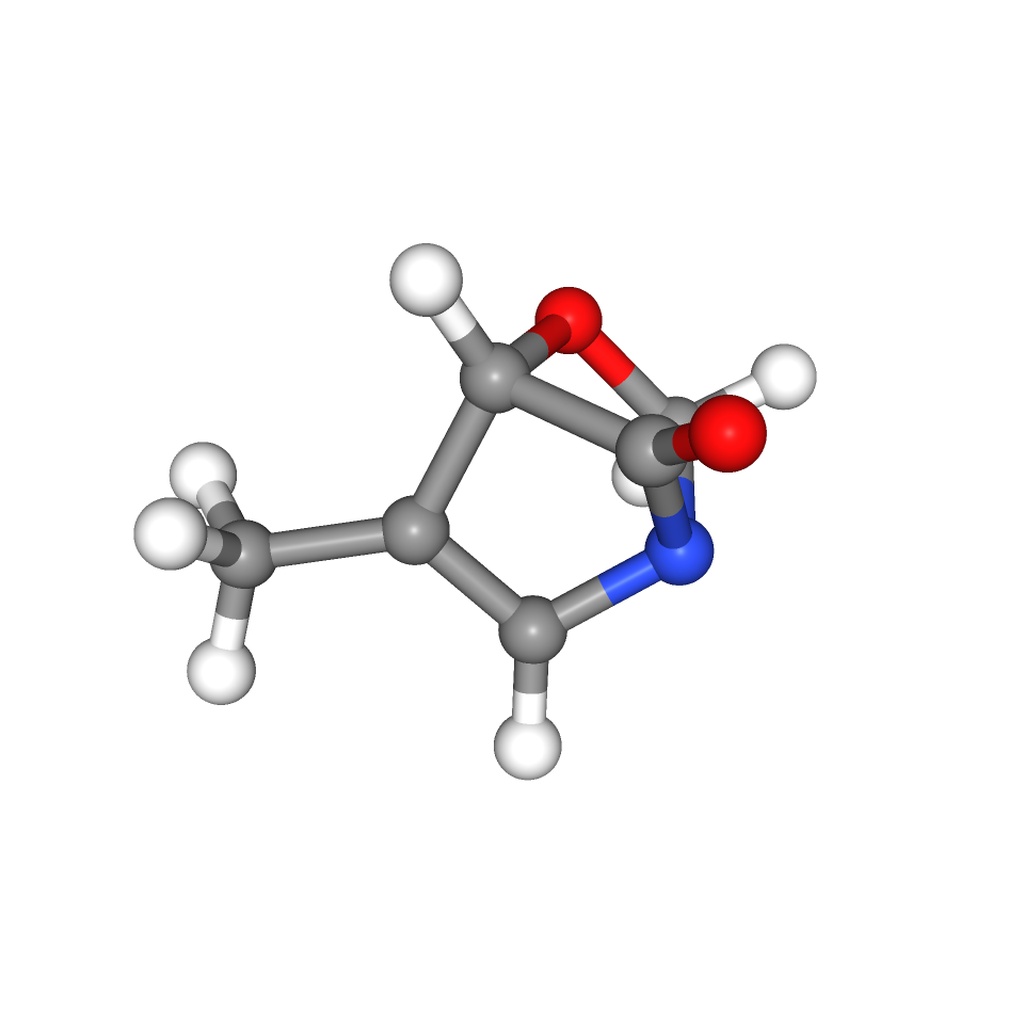}{0.63}
    & \imgwithreward{\structcellwidth}{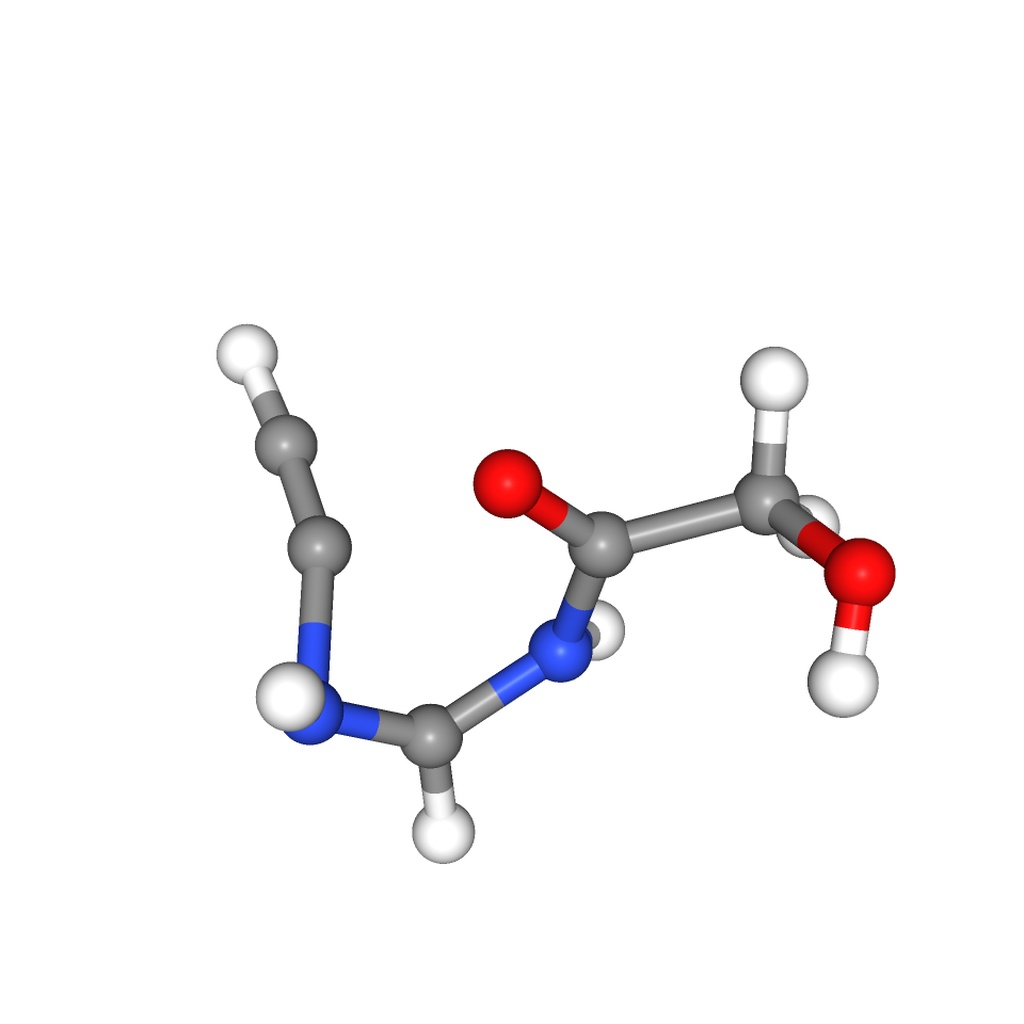}{0.60}
    & \imgwithreward{\structcellwidth}{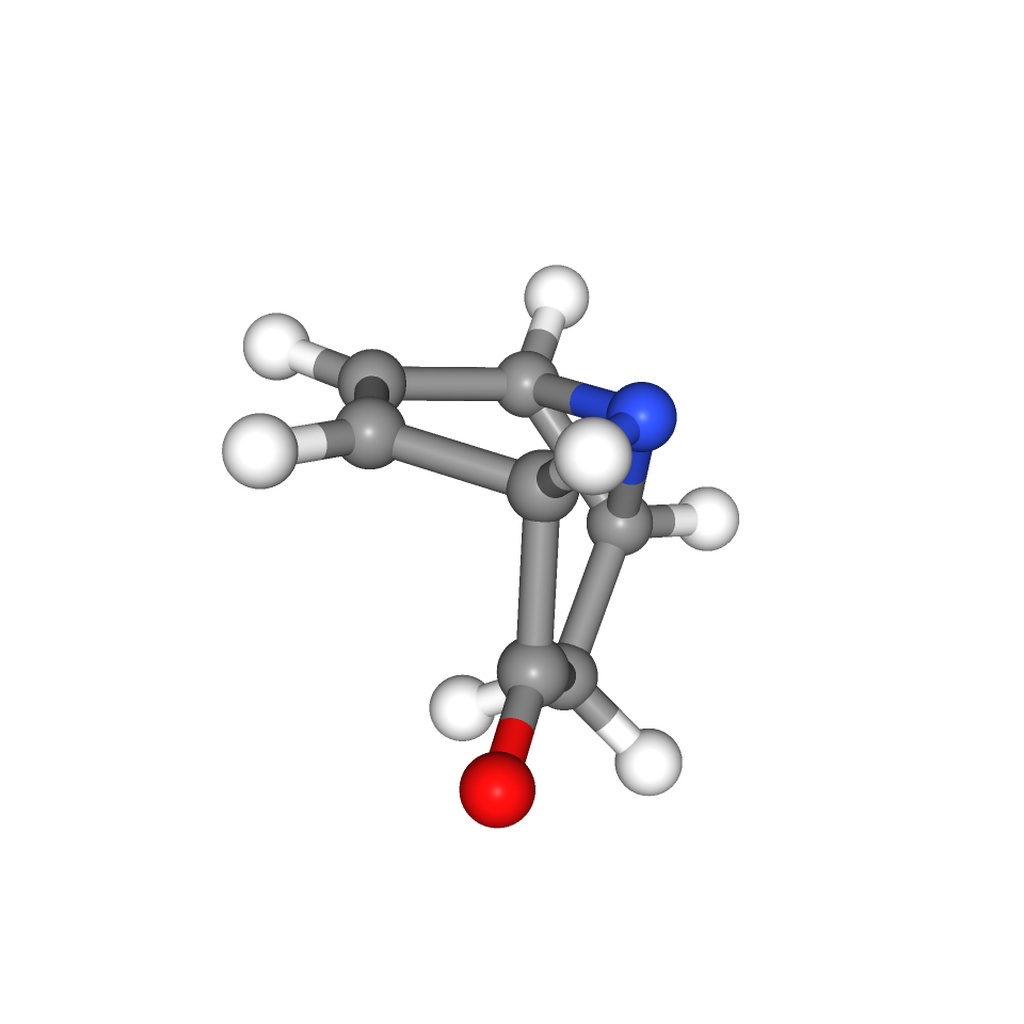}{0.37}
    \\[\vsep]
    \rule{0pt}{\structcellwidth}
    \imgwithreward{\structcellwidth}{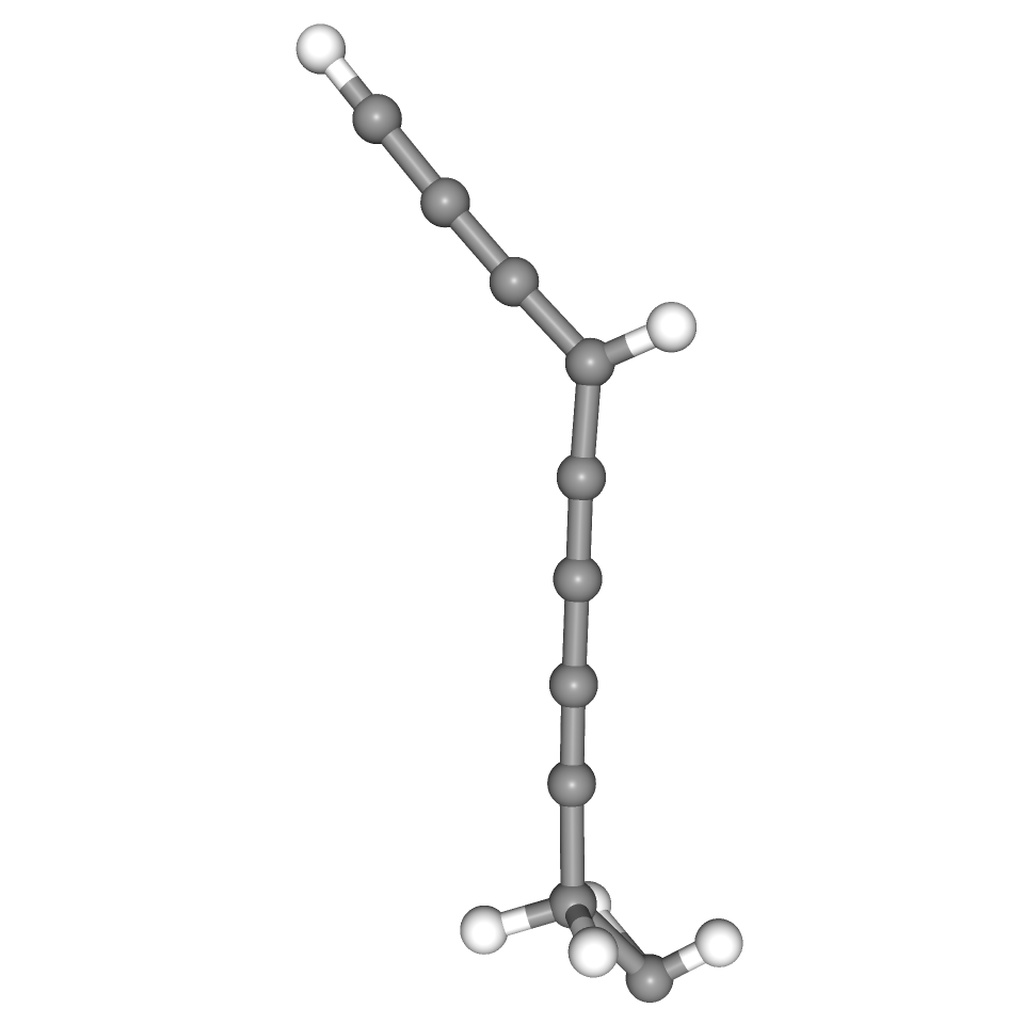}{0.93}
    & \imgwithreward{\structcellwidth}{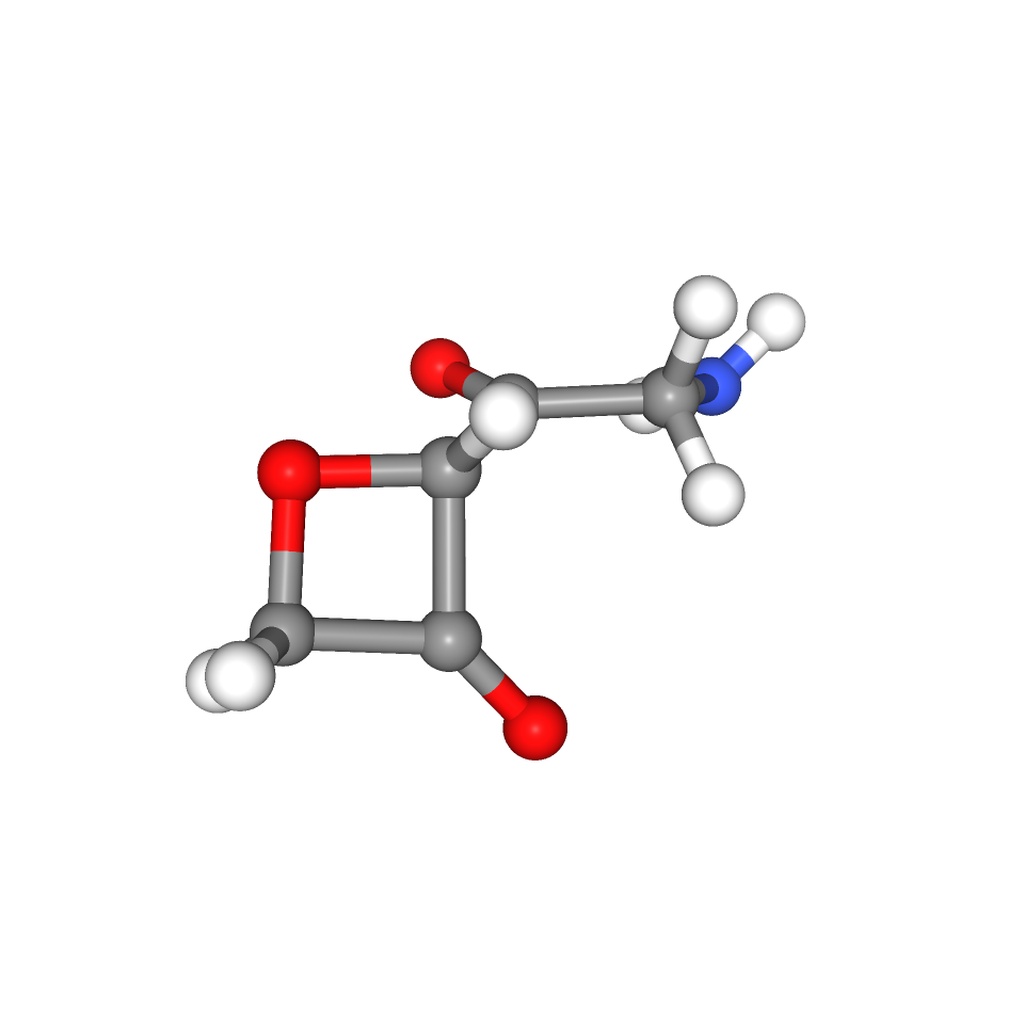}{0.62}
    & \imgwithreward{\structcellwidth}{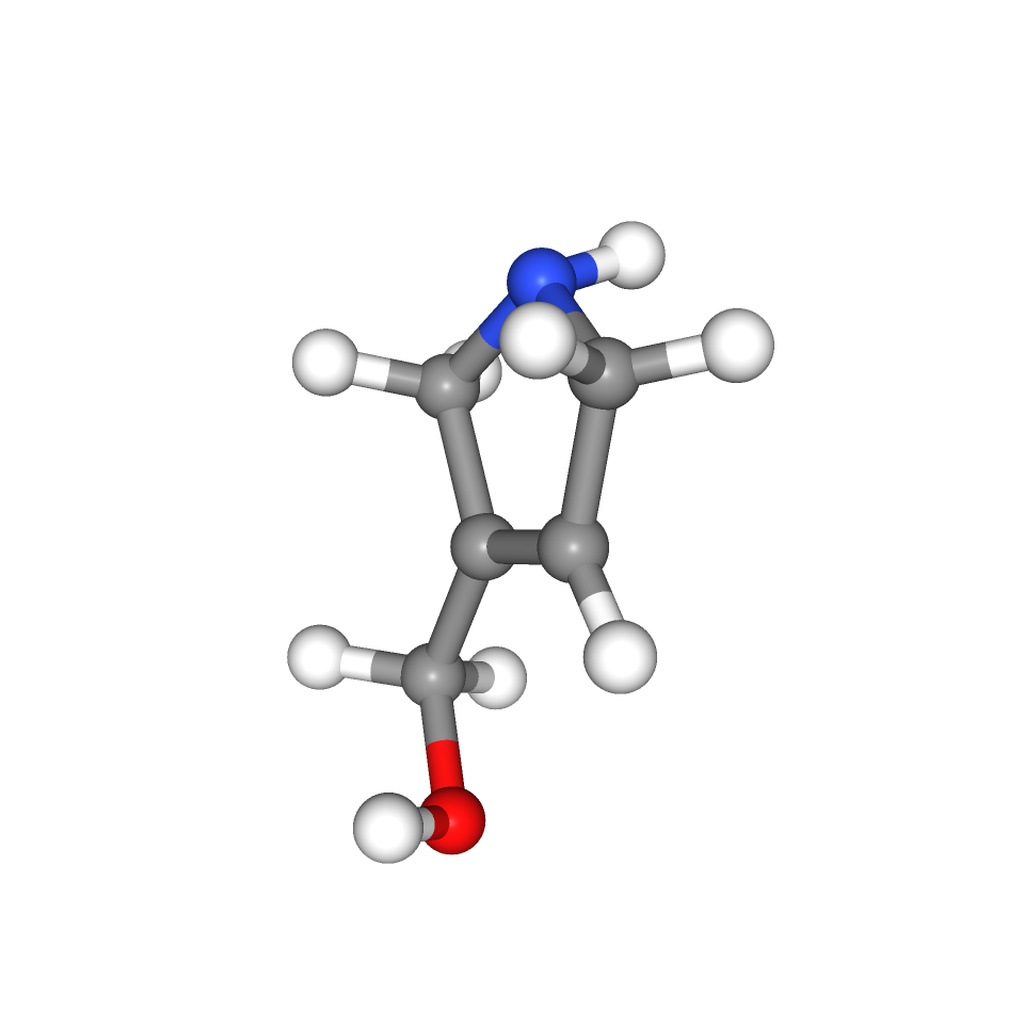}{0.59}
    & \imgwithreward{\structcellwidth}{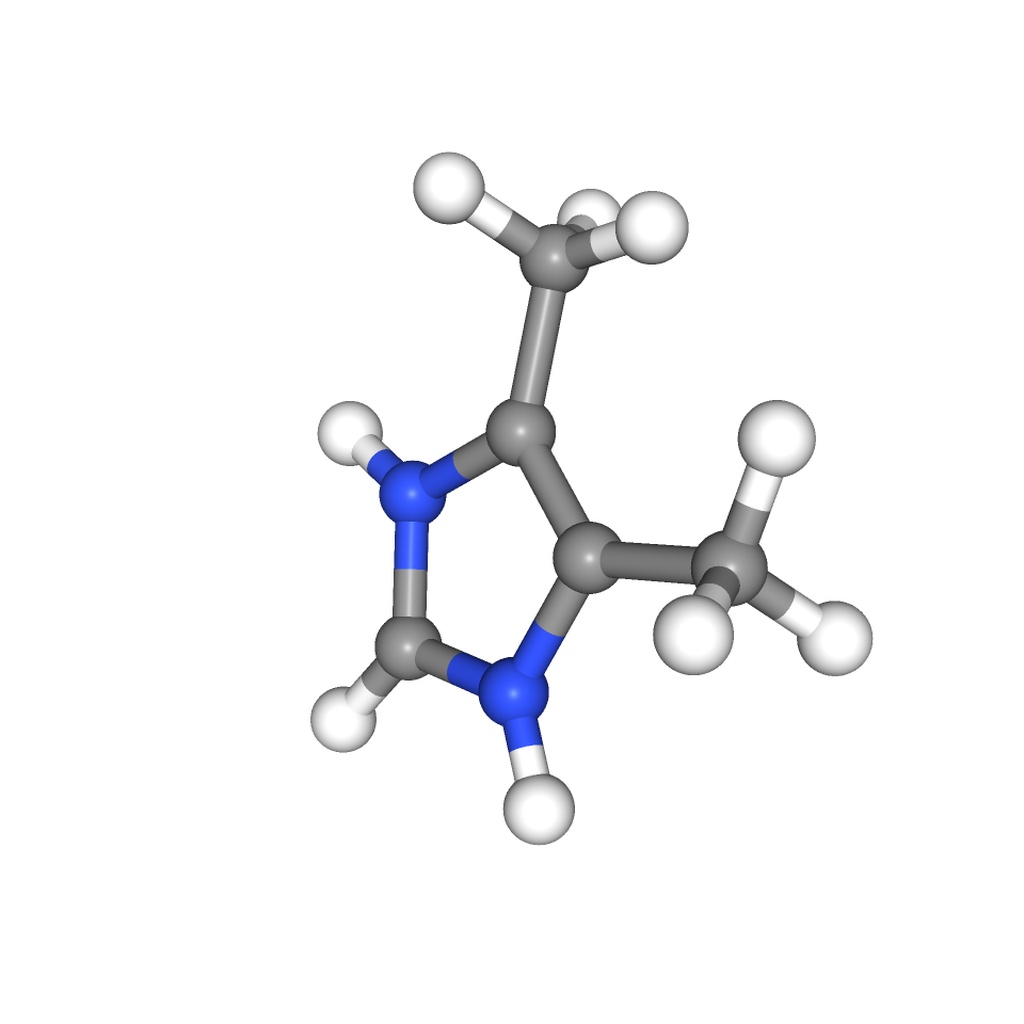}{0.55}
    \\
    \bottomrule
  \end{tabular}
\end{subfigure}\hspace{2em}%
\begin{subfigure}[t]{\protblockwidth}
  \centering
\subcaption{\textbf{Proteins}}\label{fig:exp:mols_and_prots:b}\par\vspace{0.5ex}
  \begin{tabular}{@{}c@{\hspace{\hsep}}c@{\hspace{\hsep}}c@{}}
    \toprule
    \rule{0pt}{2.5ex}
    \textbf{\scriptsize Random} & \textbf{\scriptsize Zero-order} & \textbf{\scriptsize TRS}
    \\[0.6ex]
    \midrule
    \rule{0pt}{\structcellwidth}
    \imgwithreward{\structcellwidth}{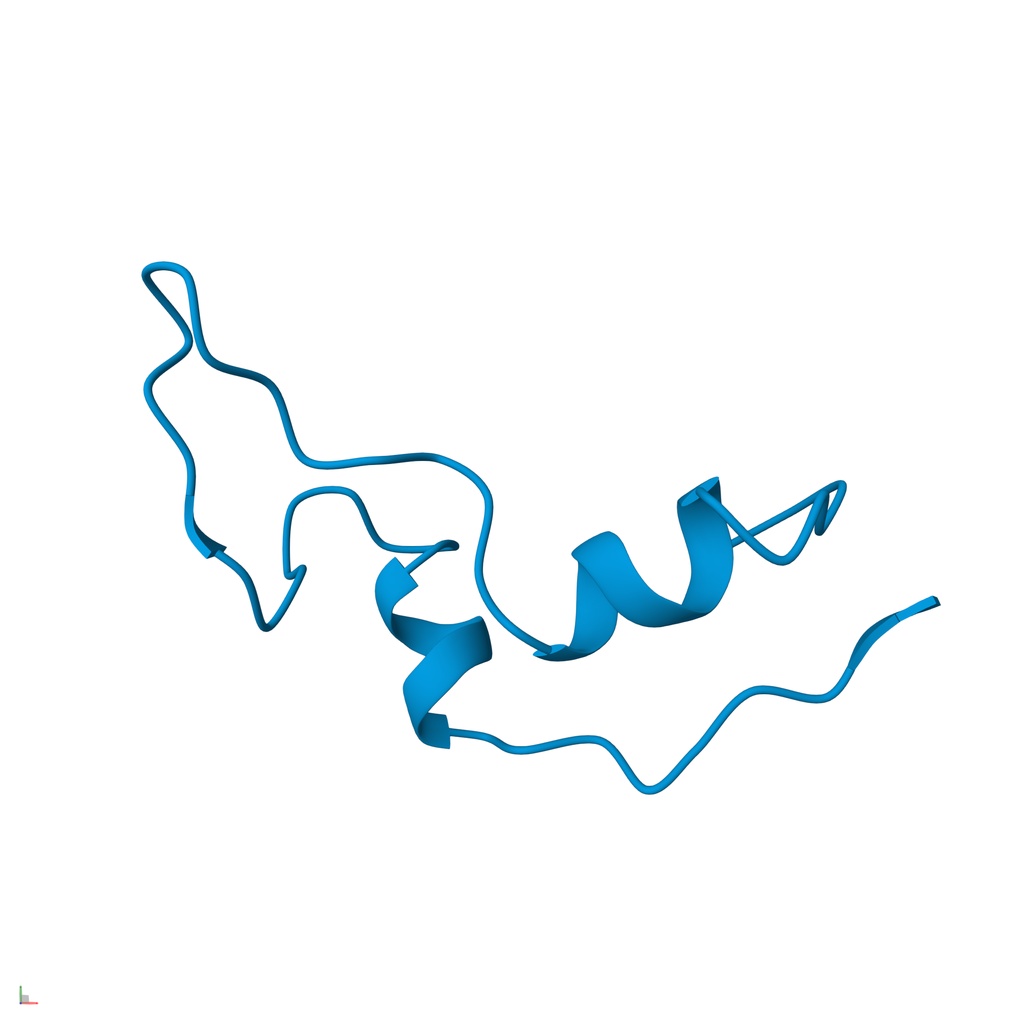}{0.53}
    & \imgwithreward{\structcellwidth}{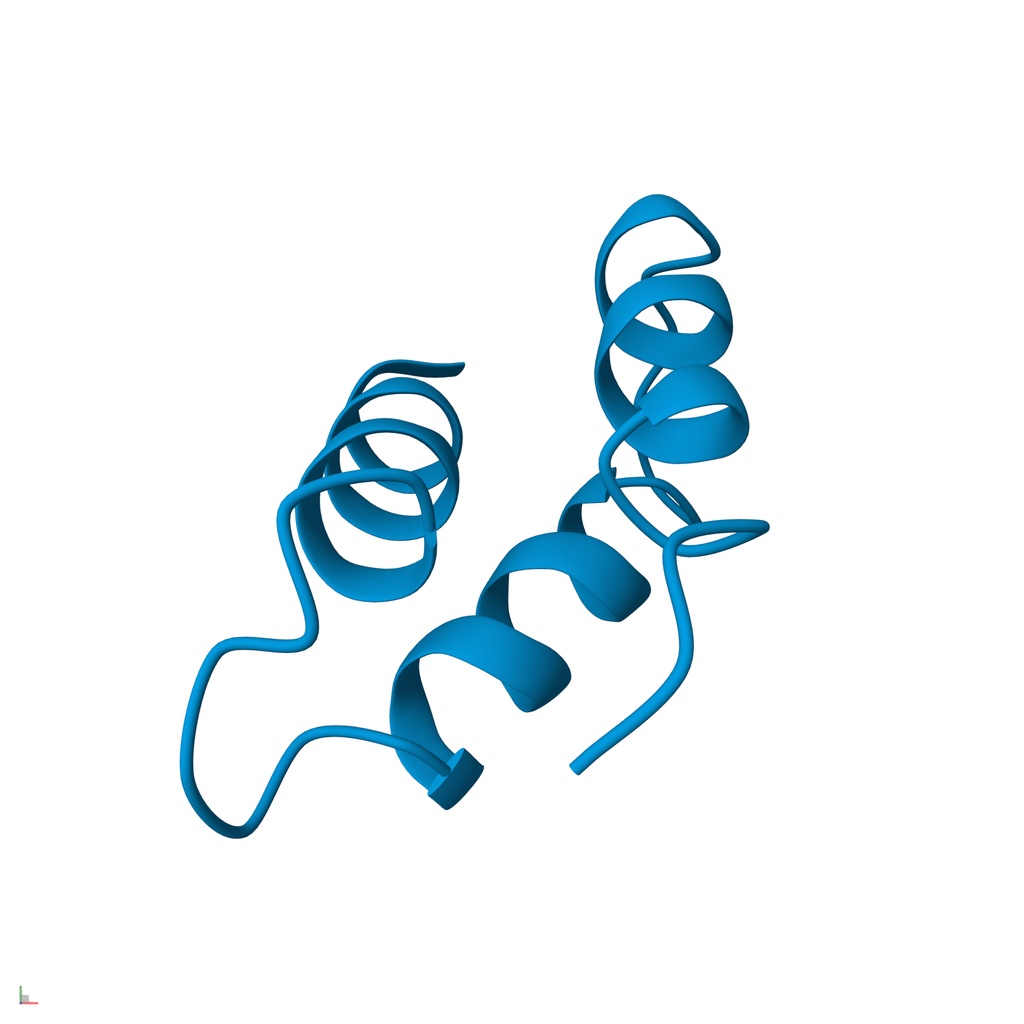}{0.59}
    & \imgwithreward{\structcellwidth}{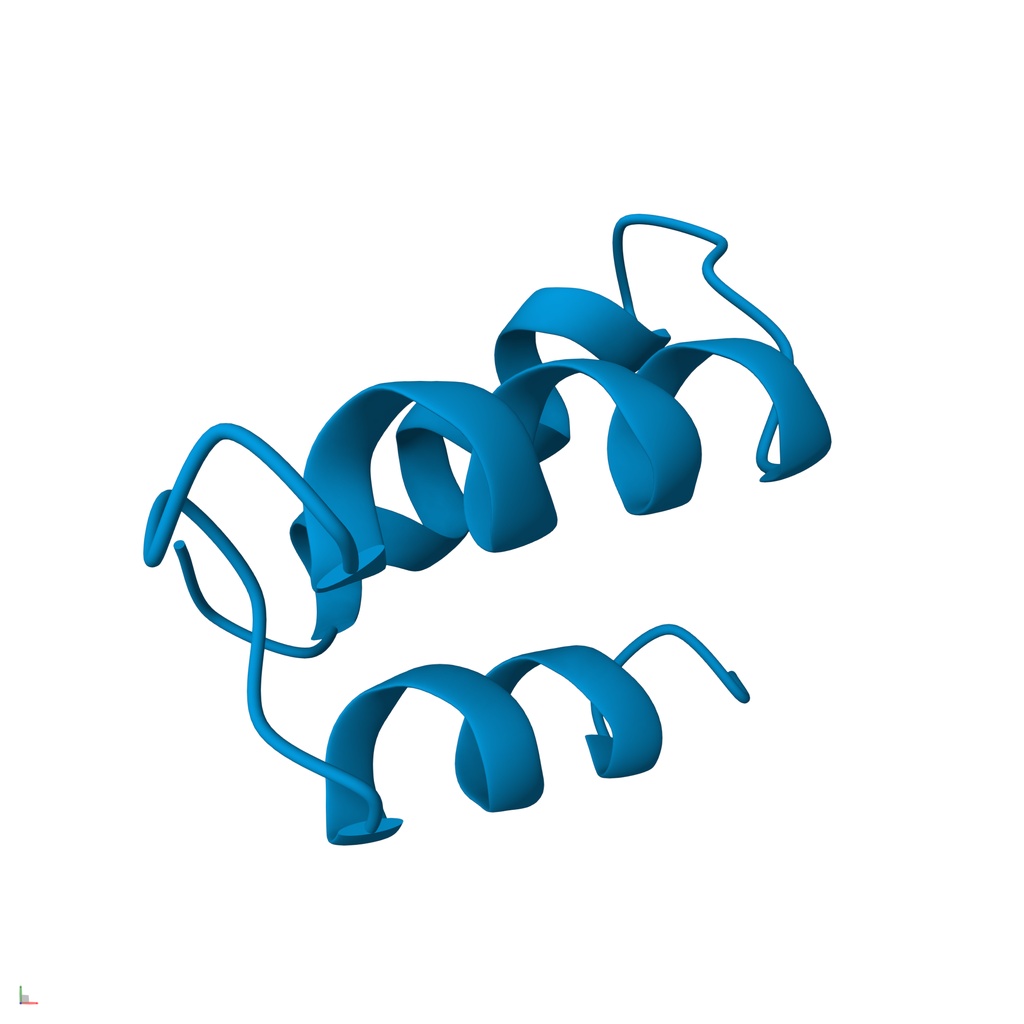}{0.65}
    \\[\vsep]
    \rule{0pt}{\structcellwidth}
    \imgwithreward{\structcellwidth}{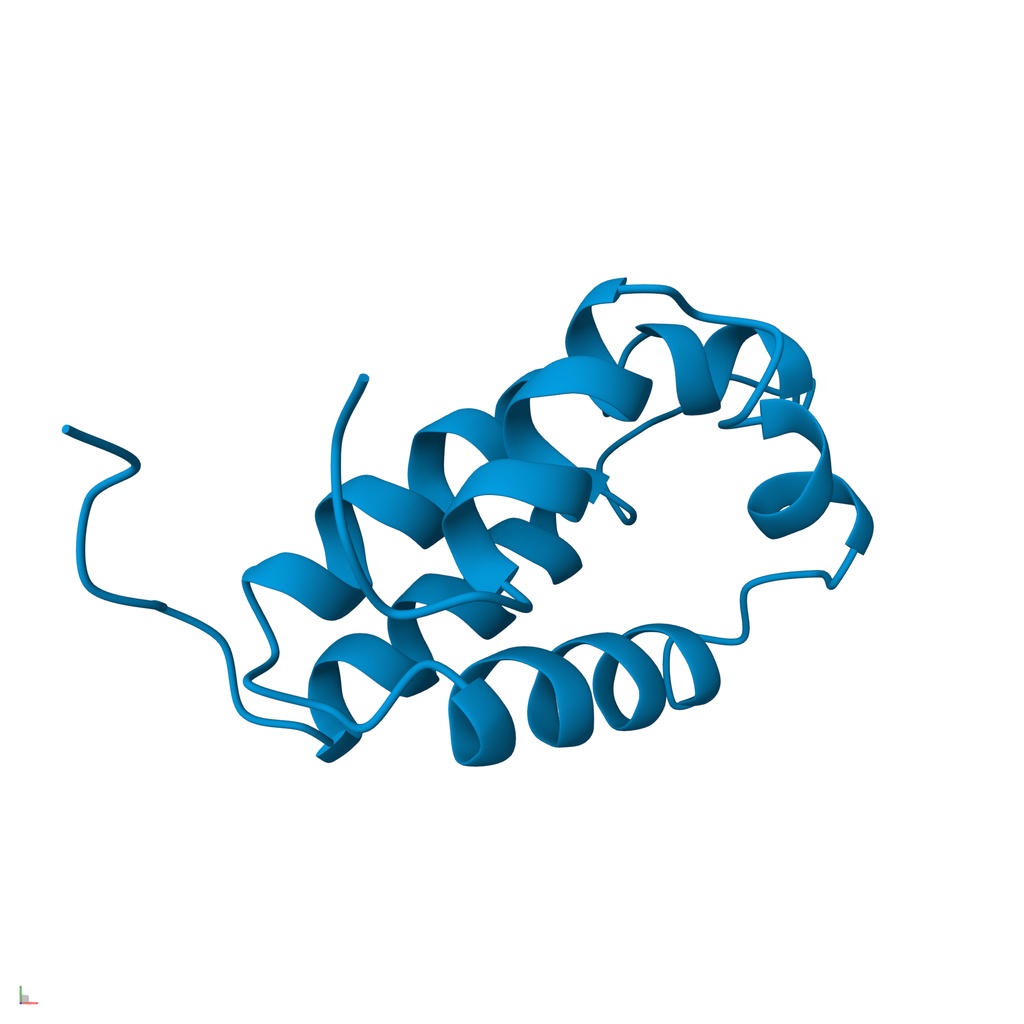}{0.33}
    & \imgwithreward{\structcellwidth}{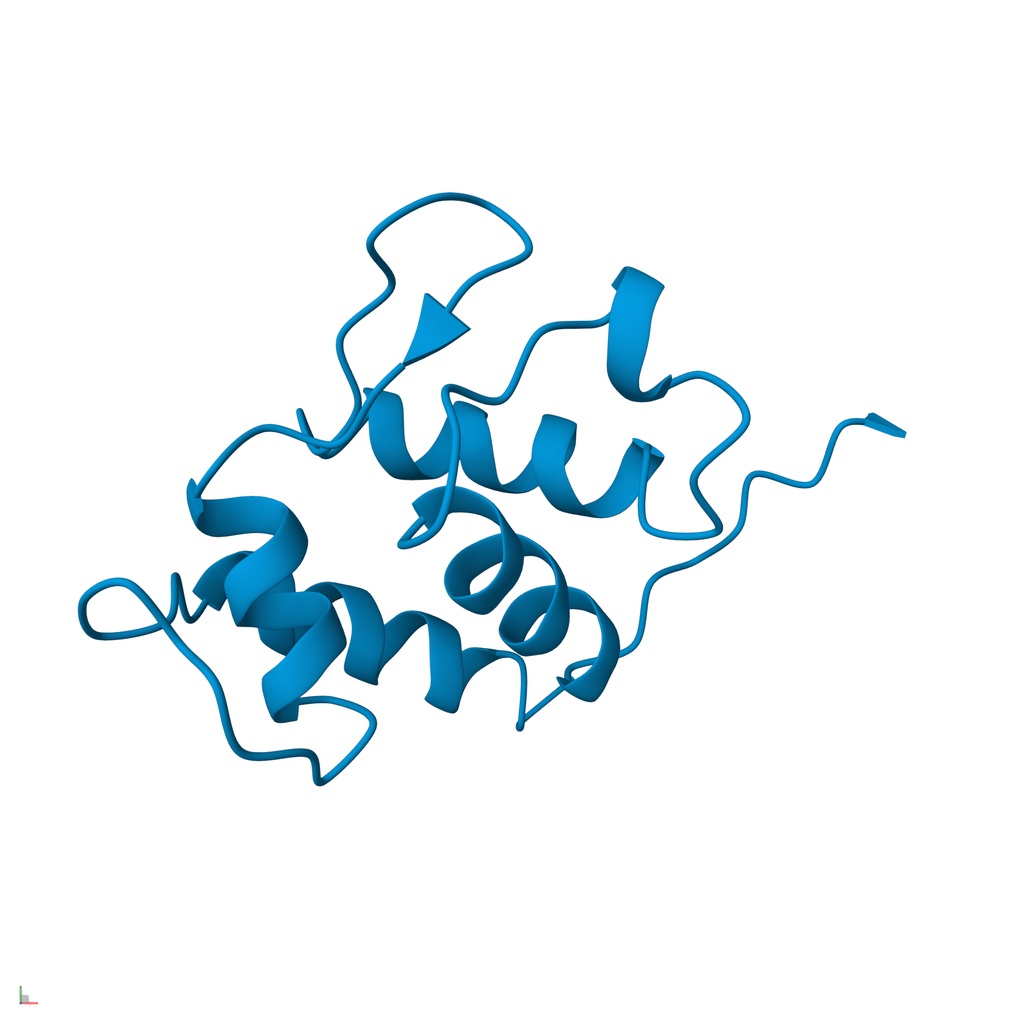}{0.39}
    & \imgwithreward{\structcellwidth}{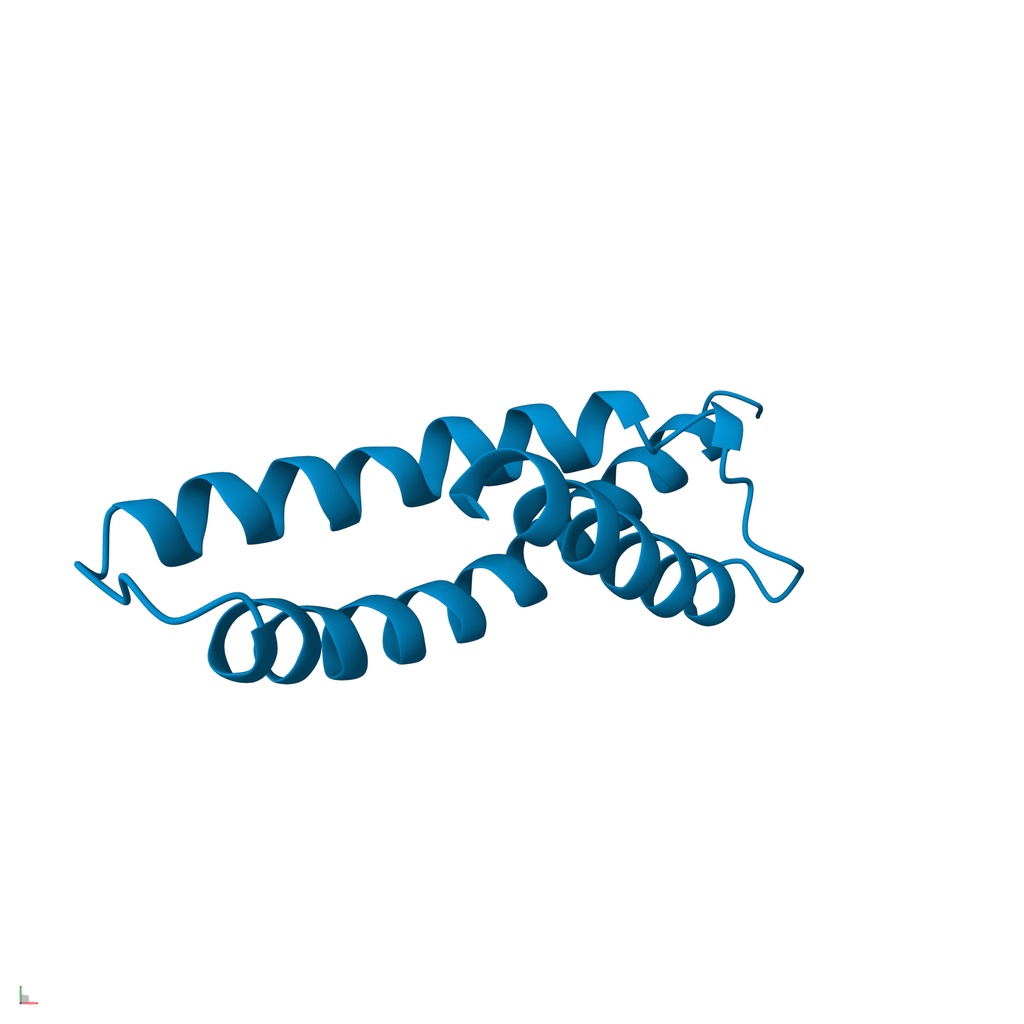}{0.45}
    \\
    \bottomrule
  \end{tabular}
\end{subfigure}

\vspace{0.4em}
\caption{\textbf{Optimized molecule and protein samples.} We visualize randomly selected samples produced by different optimization methods in the molecule and protein experiments. 
(a) In the top row optimization is conducted with three target properties ($L_3$), while the bottom row with the six ($L_6$). Lower is better. 
(b) In the top row $n_{\text{res}}{=}50$ and in the bottom row $n_{\text{res}}{=}100$. The proteins are optimized for the designability reward.}
\label{fig:exp:mols_and_prots}
\end{figure}

\subsection{Molecule Generation}
\label{sec:exp:molecule_generation}

\paragraph{Setup.}
In this experiment, we move to ODE-based flow matching and consider a different data modality.
Our goal is to generate small molecules with specified target values of chemical properties~\citep{ramakrishnan2014quantum}.
We use EquiFM~\citep{song2023equivariant} with 50 integration steps, operating on a joint continuous Gaussian coordinate- and encoded feature space ($291$ total dimensions), with pretrained chemical property prediction models. This setting is a standard benchmark for evaluating classifier-free guidance \citep{hoogeboom2022equivariant} and was also used by OC-Flow~\citep{wang2024training} for noise optimization toward single-property targets. However, we observe that random search based methods already achieve very strong performance in this regime and therefore extend the benchmark to multi-property target matching.
Specifically, we define a distance-based loss as the mean absolute deviation between predicted and target values across multiple chemical properties. Since our framework is formulated as reward maximization, we maximize the negative of this loss.
We consider two reward functions, $R_3$ and $R_6$, involving 3 and 6 chemical properties, respectively and additional property details are provided in Appendix~\ref{sec:app:experiment_details:molecule_generation}.
As baselines, we exclude noise-sequence search methods, as they are incompatible with ODE-based flow matching, and compare against OC-Flow, random search, zero-order search and diagonal CMA-ES, which is concurrent work \citep{jajal2025inference}.

\paragraph{Metrics.}
As our primary metric, we report the mean best losses $L_3(x) = -R_3(x)$ and $L_6(x) = -R_6(x)$ (mean distance to the target properties) over 200 molecules, with the target property values sampled randomly. In addition, we track other relevant metrics such as the molecule stability percentage (MSP), and valid and unique percentage (VUP).

\paragraph{Results.}
The quantitative results are summarized in \Cref{fig:exp:molecule_main}. Additionally, we visualize some randomly chosen optimized samples from different solvers in \Cref{fig:exp:mols_and_prots}.
We observe that TRS consistently achieves the lowest losses, indicating the highest alignment with the specified multi-property targets.
Importantly, this improvement does not come at the expense of other quality metrics: molecule stability and novelty remain comparable to those of the base model.
In contrast, the gradient-based OC-Flow method exhibits degraded stability and novelty, even with regularization. This suggests that gradient-based optimization tends to drift off the data manifold in this setting. We find that sampling-based approaches are better suited for this task; notably, even random search outperforms OC-Flow. This performance gap likely stems from conflicting gradients between different property classifiers, which makes local optimization difficult. As illustrated in the bottom row of \Cref{fig:exp:mols_and_prots}, OC-Flow often fails to escape unfavorable initial samples, whereas sampling-based methods explore the space more effectively. Here, diagonal CMA-ES only matches the best-of-N random search baseline, likely due to tuning sensitivity, whereas TRS achieves the best alignment with the same hyperparameters used across all experiments.

\begin{figure}[ht]
\centering
\captionsetup{font=small}

\begin{subfigure}[b]{0.43\columnwidth}
    \centering
    \vbox to 4.2cm{ %
        \vfill
        \renewcommand{\arraystretch}{1.3}
        \setlength{\tabcolsep}{3pt} 
        \small %
        \vspace{-0.4cm}
        
        \resizebox{\linewidth}{!}{%
        \begin{tabular}{@{} l c c c c @{}}
        \toprule
        \textbf{Algorithm}
        & {$L_3$(\(\downarrow\))}
        & {$L_6$(\(\downarrow\))}
        & {\textbf{MSP}}
        & {\textbf{VUP}} \\
        \midrule
        Base        & 1.21           & 1.15         & 88.3  & 88.3 \\
        OC-Flow     & 0.83           & 0.93         & 68.0  & 68.0  \\
        Random      & 0.47           & 0.62         & 84.2  & 84.1  \\
        Zero-order  & 0.43 & \underline{0.59} & 84.4  & 84.3  \\
        CMA-ES (diag) & 0.47 & 0.61 & 84.5 & 84.0 \\
        \midrule
        TRS w/o TR adapt. & \underline{0.42} & \textbf{0.55} & 86.5 & 86.5 \\
        \textbf{TRS (Ours)} & \textbf{0.39} & \textbf{0.55} & 85.7 & 85.6 \\
        \bottomrule
        \end{tabular}%
        }
        \vfill
    }
    \caption{}
    \label{subfig:table}
\end{subfigure}
\hfill
\begin{subfigure}[b]{0.54\columnwidth}
    \centering
    \begin{tikzpicture}
    \begin{groupplot}[
        group style={
            group size=2 by 1,
            horizontal sep=1.0cm, %
        },
        width=4.2cm,  
        height=4cm, 
        grid=both,
        grid style={gray!30},
        tick label style={font=\small}, 
        label style={font=\small},
        title style={font=\small},
        scaled x ticks=false,
        scaled y ticks=false,
        xtick={25000,50000,75000,100000},
        xticklabels={25k,50k,75k,100k},
        every axis plot/.append style={
            line width=1.1pt,
            mark size=1.2
        }
    ]
    
    \nextgroupplot[
        title={$L_3$},
        ymin=0.38, ymax=0.62,
        ytick={0.4, 0.5, 0.6},
        ylabel={Loss (Property dist.)},
        xlabel={NFE},
        legend to name=sharedlegend_mol_final,
        legend columns=-1,
        legend style={font=\small, draw=none, fill=none}
    ]
    \addplot[color=method1color, mark=*] coordinates {(25000,0.5269) (50000,0.4627) (75000,0.42147) (100000,0.394)};
    \addlegendentry{TRS (Ours)~~}
    \addplot[color=method2color, mark=*] coordinates {(25000,0.542) (50000,0.5129) (75000,0.481) (100000,0.473)};
    \addlegendentry{Random~~}
    \addplot[color=method3color, mark=*] coordinates {(25000,0.5112) (50000,0.473) (75000,0.452) (100000,0.432)};
    \addlegendentry{ZO}
    
    \nextgroupplot[
        title={$L_6$},
        ymin=0.48, ymax=0.72,
        ytick={0.5, 0.6, 0.7}, 
        xlabel={NFE},
    ]
    \addplot[color=method1color, mark=*] coordinates {(25000,0.640) (50000,0.591) (75000,0.572) (100000,0.548)};
    \addplot[color=method2color, mark=*] coordinates {(25000,0.6787) (50000,0.65336) (75000,0.627) (100000,0.617)};
    \addplot[color=method3color, mark=*] coordinates {(25000,0.661) (50000,0.623) (75000,0.606) (100000,0.591)};
    
    \end{groupplot}
    \node[yshift=-1.3cm] at ($(group c1r1.south)!0.5!(group c2r1.south)$) {\pgfplotslegendfromname{sharedlegend_mol_final}};
    \end{tikzpicture}
    
    \caption{}
    \label{subfig:plots}
\end{subfigure}

\vspace{0.5em}
\caption{\textbf{Molecule optimization results.} (a) Distance to the combined targets of 3 and 6 chemical property values, together with MSP stability and validity VUP metrics which we do not explicitly optimize. The best alignment values are in \textbf{bold}, second best are \underline{underlined}. (b) Scaling trends over the NFE budget.}
\label{fig:exp:molecule_main}
\end{figure}

\subsection{Protein Design}
\label{sec:exp:protein_design}

\paragraph{Setup.}
Protein design is another challenging data modality with expensive reward models and is relatively underexplored by inferenece-time alignment methods. State of the art 3D protein generation models, like Proteina~\citep{geffner2025proteina}, are flow matching models that are typically trained via ODE integration, but sampling is often performed using an SDE scheduler with noise reduction to improve designability \citep{bose2024sestochastic, lin2024manyonedesigningscaffolding}. However, this procedure alters the target distribution and no longer samples from the full flow-matching distribution, which can negatively impact other metrics such as diversity and novelty. In this experiment, we therefore focus on steering ODE-based sampling while preserving the full distribution. For completeness, we provide a comparison to SDE-based sampling in Appendix~\ref{sec:app:proteina_discussion}. We further note that SDE-based noise-sequence optimization methods are not applicable in this setting: they either rely on value estimation functions that are unavailable for 3D protein design, or are computationally infeasible due to the high cost of reward evaluation.
We design the experiment by fixing the number of residues $n_{\text{res}} = 50$ and $n_{\text{res}} = 100$. For the reward function, we use the computationally expensive designability reward based on large structure extraction and protein folding models (see Appendix~\ref{sec:app:experiment_details} for details).
We compare against purely black-box baselines, including random search, zero-order search \citep{ma2025inference} and diagonal CMA-ES, which is concurrent work \citep{jajal2025inference}.

\paragraph{Metrics.} 
As alignment metric we report the mean best rewards for each run. Additionally we show the cluster diversity and the pairwise TM-score and a novelty metric, which compares the proteins to the entire PDB dataset~\citep{berman2000protein}. Further details of these are given in Appendix~\ref{sec:app:metrics:proteins}.

\begin{figure}[ht]
\centering
\captionsetup{font=small}

\begin{subfigure}[t]{0.44\columnwidth}
    \centering
    \vspace{0pt} %
    \renewcommand{\arraystretch}{1.25}
    \setlength{\tabcolsep}{2pt} 
    \small 
    
    \resizebox{\linewidth}{!}{%
    \begin{tabular}{@{} l l c c c c @{}}
    \toprule
    $n_\text{res}$ & \textbf{Method} & {Des.$\uparrow$} & {Clu.$\uparrow$} & {TM$\downarrow$} & {Nov.$\downarrow$} \\
    \midrule
    \multirow{6}{*}{50}
        & Base       & 0.05 & 0.71 & 0.50 & 0.70 \\
        & Random         & 0.53   & 0.27 & 0.68 & 0.85 \\
        & Zero-order         & 0.59 & 0.39 & 0.62 & 0.85 \\
        & CMA-ES (diag) & 0.62 & 0.19 & 0.76 & 0.92 \\
        & TRS w/o TR adapt. & \underline{0.63} & 0.30 & 0.67 & 0.89 \\
        & \textbf{TRS(Ours)} & \textbf{0.65} & 0.30 & 0.67 & 0.89 \\
    \midrule
    \multirow{6}{*}{100}
        & Base       & 0.02 & 0.71 & 0.56 & 0.86 \\
        & Random         & 0.33 & 0.49 & 0.55 & 0.84 \\
        & Zero-order         & 0.39 & 0.59 & 0.54 & 0.84 \\
        & CMA-ES (diag) & \underline{0.44} & 0.28 & 0.63 & 0.90 \\
        & TRS w/o TR adapt. & 0.42 & 0.45 & 0.56 & 0.85 \\
        & \textbf{TRS(Ours)} & \textbf{0.45} & 0.44 & 0.57 & 0.86 \\
    \bottomrule
    \end{tabular}%
    }
    \vspace{0.5cm}
    \caption{}
    \label{subfig:protein_table}
\end{subfigure}
\hfill
\begin{subfigure}[t]{0.54\columnwidth}
\centering
\vspace{0pt} %
\begin{tikzpicture}
\begin{groupplot}[
    group style={
        group size=2 by 1,
        horizontal sep=1.1cm, %
    },
    width=4.1cm,  
    height=4.0cm, %
    grid=both,
    grid style={gray!30},
    tick label style={font=\small},
    label style={font=\small},
    title style={font=\small},
    scaled x ticks=false,
    scaled y ticks=false,
    xtick={16000, 32000, 48000, 64000},
    xticklabels={16k, 32k, 48k, 64k},
    every axis plot/.append style={
        line width=1.1pt,
        mark size=1.2
    }
]

\nextgroupplot[
    title={$n_{\text{res}} =50$},
    ymin=0.38, ymax=0.65,
    ytick={0.4,0.5,0.6},
    ylabel={Mean Best Reward},
    xlabel={NFE},
    legend to name=sharedlegend_prot_final,
    legend columns=-1,
    legend style={font=\small, draw=none, fill=none}
]
\addplot[color=method1color, mark=*] coordinates {(16000,0.425) (32000,0.525) (48000,0.601) (64000,0.633)};
\addlegendentry{TRS (Ours)~~}
\addplot[color=method2color, mark=*] coordinates {(16000,0.413) (32000,0.471) (48000,0.505) (64000,0.533)};
\addlegendentry{RS~~}
\addplot[color=method3color, mark=*] coordinates {(16000,0.447) (32000,0.542) (48000,0.574) (64000,0.588)};
\addlegendentry{ZO}

\nextgroupplot[
    title={$n_{\text{res}} =100$},
    ymin=0.18, ymax=0.45,
    ytick={0.2,0.3,0.4},
    xlabel={NFE},
]
\addplot[color=method1color, mark=*] coordinates {(16000,0.242) (32000,0.344) (48000,0.404) (64000,0.437)};
\addplot[color=method2color, mark=*] coordinates {(16000,0.227) (32000,0.286) (48000,0.312) (64000,0.333)};
\addplot[color=method3color, mark=*] coordinates {(16000,0.245) (32000,0.332) (48000,0.362) (64000,0.386)};
\end{groupplot}

\node[yshift=-1.3cm] at ($(group c1r1.south)!0.5!(group c2r1.south)$) {\pgfplotslegendfromname{sharedlegend_prot_final}};
\end{tikzpicture}

\caption{}
\label{subfig:protein_plots}
\end{subfigure}

\vspace{0.5em}
\caption{\textbf{Protein optimization results.} (a) Optimized mean designability rewards at 64k NFE, together with diversity and novelty metrics which we do not explicitly optimize. The best values are in \textbf{bold}, second best are \underline{underlined}. (b) Reward improvement across NFE budgets.}
\label{fig:exp:protein_combined}
\end{figure}

\paragraph{Results.}
In terms of the designability steering, we see in \Cref{fig:exp:protein_combined}, that we significantly improve over other search algorithms in both settings.
We see that generally the rewards are higher when we optimize proteins with 50 residues, which is expected, but also their diversity and novelty metrics decrease compared to the base model. We assume that, since the solution space is rather small, it is likely that well designable proteins share similar features. However, this is not comparable to the mode collapse, which we observe for SDE noise reduction (see  Appendix~\ref{sec:app:proteina_discussion}).
Naturally, due to a larger solution space, optimizing proteins with 100 residues leads to better diversity and novelty, while still gaining major improvements over the base model. This counts for all optimization methods, but TRS achieves the best reward alignment again. Diagonal CMA-ES is competitive in this setting, consistent with the image experiments, yet TRS still attains the highest designability in both residue-length regimes.

\section{Conclusion}
\label{sec:conclusion}

In this work we investigate inference-time scaling and preference alignment, where our simple trust-region source noise search achieves state-of-the-art performance across text-to-image, molecule, and protein design tasks. Our approach is model and reward agnostic, making it particularly suited for real-world settings where reward functions are often expensive or unknown. It also offers good balance between exploration and exploitation by searching multiple noise regions early and refining promising ones, and does not drift off the data manifold and remains stable as we observe in  \Cref{sec:exp:molecule_generation}. While all methods are limited by the accuracy of the reward models (Appendix~\ref{sec:app:limitations}), scaling improvements \citep{wu2025rewarddance} suggest this limitation will diminish and our efficient source noise optimization is particularly well-suited for this development. Future work includes exploring the geometry of the source noise space further and developing improved perturbation schemes that adhere more to this geometry.

\subsubsection*{Contribution Statement}

NS proposed and developed the trust-region search idea, implemented all the experiments and wrote the first draft. KR initiated the project with the idea to apply black-box Bayesian optimization for source noise search, helped with writing and supervised the project. DC advised on the project, and secured funding and the compute necessary for the project.

\subsubsection*{Acknowledgments}

We thank Weirong Chen and Christian Koke for their valuable feedback. This work was supported by the European Research Council (ERC) Advanced Grant SIMULACRON and by the GNI Project ``AI4Twinning''.

\newpage
\appendix

\crefname{section}{Appendix}{Appendices}
\Crefname{section}{Appendix}{Appendices}
\crefname{subsection}{Appendix}{Appendices}
\Crefname{subsection}{Appendix}{Appendices}

\section{Appendix}
\label{sec:Appendix}

This appendix provides supplementary material to support the results presented in the main text. We begin by detailing the hyperparameters and implementation of TRS in \Cref{sec:app:hyper_params}, followed by expanded experimental settings in \Cref{sec:app:experiment_details}. We then present additional ablation studies (\Cref{sec:app:ablations}) and discuss the comparative impact of ODE versus SDE sampling (\Cref{sec:app:proteina_discussion}). Furthermore, we provide a comprehensive overview of all baselines (\Cref{sec:app:baselines}), offer deeper insights into the evaluation metrics (\Cref{sec:app:metrics}), and report the runtime for all methods (\Cref{sec:app:runtime}). Finally, we address the limitations of our approach in \Cref{sec:app:limitations} and showcase random samples from the image experiments in \Cref{sec:app:optimized_samples_examples}.

\section{Algorithm Details and Hyperparameters}
\label{sec:app:hyper_params}

In this section we explain the remaining details about our method of Section~\ref{sec:methodology:trs} and show all hyperparameter details from the experiments in Section~\ref{sec:benchmarks} for TRS and across all baselines. 

\paragraph{Restart logic.}
When the failure tolerance threshold $c_{\text{fail}}$ is triggered while the trust-region side length is already at its minimum value $\ell_j^{\min}$, a region restart is triggered.
In this case, the side length is reset, and the region is subsequently re-centered at one of the globally best points observed so far.
This restart mechanism allows the algorithm to escape from locally saturated regions and to continue allocating evaluations toward more promising areas of the search space.

\paragraph{Length-dependent perturbation constraints.}
To maintain stable behavior across different trust-region scales, we apply simple constraints that couple the side length $\ell_j$ with the perturbation probability $p_{j,b}$ used for coordinate masking $\mathbf{m}_{j,b}$, by rejection sampling.
These constraints limit the number of perturbed dimensions when the trust region is large, which is particularly important in high-dimensional noise spaces.
\begin{itemize}
    \item If $\ell_j \ge 2.0 \rightarrow p_j \le 0.2$.
    \item If $\ell_j \ge 1.6  \rightarrow p_j \le 0.5$.
    \item If $\ell_j \ge 1.2  \rightarrow  p_j \le 0.7$.
\end{itemize}
These constraints were found to improve robustness without introducing additional tuning complexity. We provide an intuitive visualization for those rules in \Cref{fig:trs_comparison_grid}, showing under which combinations of mask and region length the perturbations fail.

\paragraph{Hyperparameters.} While TRS contains several hyper-parameters, we find them to be robust across all the experiments we conduct in Section~\ref{sec:benchmarks}. We show this in \Cref{tab:hyperparameters}, where we can see that the number of regions is the only one we change, which is due to the different batch sizes we use for efficiency. But even this shows good robustness, which we show in an ablation in \Cref{fig:ablations_combined}. 

\begin{table}[H]
\small
\centering
\caption{\textbf{Hyperparameters.} For all our experiments in Section~\ref{sec:benchmarks}. When entries are separated with a slash / , we refer to the difference between SD1.5 / SDXL. The asterix * excludes OC-Flow and DTS. IR abbreviates ImageReward and HPS refers to the HPSv2 reward.}
\label{tab:hyperparams}

\renewcommand{\arraystretch}{1.15}
\setlength{\tabcolsep}{9pt}

\begin{tabular}{l c c c}
\toprule
\textbf{Hyperparameter} & {\makecell{\textbf{T2I}}} 
                & {\makecell{\textbf{Moleculs}}} 
                & {\makecell{\textbf{Protein}}} \\
\midrule

\rowcolor{black!5}\multicolumn{4}{l}{\textit{All algorithms}} \\
Noise Dimension $D$ & 16384 / 65536 & 291 & $3 \times n_{\text{res}}$ \\
Batchsize $B$* & 20 & 100 & 8 \\
Inference steps & 50/8 &  50 & 400 \\
\addlinespace[3pt]

\rowcolor{black!5}\multicolumn{4}{l}{\textbf{OC-Flow} ~\citep{wang2024training}} \\

Batch size $B$ & 1 & 1 & \text{--} \\
Step size $\eta$ & 2.5 &\text{--} & \text{--} \\
Weight decay & 0.995 & 1.0 & \text{--} \\
Weight constraint & 0.4 & 0.01 & \\
Optimization steps & 15 & 5 & \text{--} \\
Optimizer & SGD & L-BFGS & \text{--} \\
Learning rate $\alpha$  &   1.0    &   1.0   &    \text{--}   \\

\rowcolor{black!5}\multicolumn{4}{l}{\textbf{DTS*} ~\citep{jain2025diffusion}} \\


Expansion steps     & \makecell[c]{{\scriptsize [50, 40, 30, 20, 10]} \\ {\scriptsize / [8, 6, 4, 2]}} & -- & -- \\

Exploration Constant $\lambda$ &  0.1 \text{(IR)},  0.01 \text{(HPS)}  & \text{--} & \text{--} \\
Progressive Width Constant $C$ & 2.0  & \text{--}& \text{--} \\
Progressive Width $\alpha$  & 0.4 & \text{--}& \text{--} \\
Exploration type   & UCB & \text{--}& \text{--} \\

\rowcolor{black!5}\multicolumn{4}{l}{\textbf{Zero-Order Search} ~\citep{ma2025inference}} \\
Added noise $\epsilon$ & 0.1 & 0.1 & 0.1 \\

\rowcolor{black!5}\multicolumn{4}{l}{\textbf{Fast Direct} ~\citep{tan2025fast}} \\
Step size  $\alpha$ & 80 & \text{--} & \text{--} \\
Total steps $T$ & 6& \text{--} & \text{--} \\
Noise sigma (GP) $\sigma_n$ & 0.1 & \text{--} & \text{--} \\

\rowcolor{black!5}\multicolumn{4}{l}{\textbf{CMA-ES (diag)} ~\citep{hansen2001completely}} \\
Initial step size $\sigma_0$ & 0.2 & 0.2 & 0.2 \\
Covariance & diagonal & diagonal & diagonal \\
Population $\lambda$ ($\mu = \lfloor\lambda/2\rfloor$) & 20 & 100 & 8 \\
Initialization & zero & zero & zero \\

\rowcolor{black!5}\multicolumn{4}{l}{\textbf{TRS (Ours)}} \\
Perturbations & Sobol / Gaussian & Sobol  & Sobol \\
\# Trust regions $k$ & 15 & 20 & 5 \\
Initial Trust Region length $l_{init}$ & 0.8 & 0.8 & 0.8 \\ 
Minimal Length $l_{\text{min}}$ & 0.05 & 0.05 & 0.05 \\ 
Maximal Lenght $l_{\text{max}}$ & 2.4 & 2.4 & 2.4 \\
Length Update factor $\alpha_{\ell}$ &  1.5  &  1.5  & 1.5  \\
Success counter threshold $c_{succ}$ & 3 & 3 & 3 \\ 
Failure counter threshold $c_{fail}$ & 3 & 3 & 3 \\
Fraction warm-up iterations & 20\% & 20\%  & 20\% \\
Min. prob. for masks $p_{\text{min}}$ &  0.1  &  0.1 &  0.1  \\
Max. prob. for masks $p_{\text{max}}$ &  0.9  &  0.9 &  0.9  \\
\addlinespace[3pt]

\bottomrule
\end{tabular}
\label{tab:hyperparameters}
\end{table}

\section{Experiment Details}
\label{sec:app:experiment_details}

In this section, we provide additional details of the experiments in Section~\ref{sec:benchmarks} with a specific focus on the generative models and the reward functions that are applied. We provide the links to all external resources in \Cref{tab:master_resources}.

\begin{table}[ht]
\centering
\small
\renewcommand{\arraystretch}{1.2} 
\setlength{\tabcolsep}{6pt} 

\caption{\textbf{Generative models and reward functions.} Used for the experiments in Section~\ref{sec:benchmarks}.}
\label{tab:master_resources}

\newcolumntype{C}{>{\centering\arraybackslash}p{2cm}}
\newcolumntype{P}{>{\centering\arraybackslash}p{1.8cm}}

\begin{tabularx}{\textwidth}{@{} C l P >{\raggedright\arraybackslash}X @{}}
\toprule
\textbf{Domain} & \textbf{Model / Tool} & \textbf{Params} & \textbf{Source / Repository} \\ \midrule

\textbf{T2I} & SD v1.5 & $\sim$860M & \href{https://huggingface.co/stable-diffusion-v1-5/stable-diffusion-v1-5}{HF: SD-v1.5} \\
 & SDXL-Lightning & $\sim$2.6B & \href{https://huggingface.co/ByteDance/SDXL-Lightning}{HF: SDXL-Lightning} \\
 & ImageReward & $\sim$446M & \href{https://github.com/THUDM/ImageReward}{GitHub: ImageReward} \\
 & HPSv2 & $\sim$986M & \href{https://github.com/tgxs002/HPSv2}{GitHub: HPSv2} \\
 & Aesthetic Pred. & $\sim$428M & \href{https://github.com/LAION-AI/aesthetic-predictor}{GitHub: Aesthetic} \\ 
 \addlinespace[4pt]

\textbf{Molecules} & EquiFM & $\sim$22M & \href{https://github.com/WangLuran/Guided-Flow-Matching-with-Optimal-Control}{GitHub: EquiFM / OC-Flow} \\
 \addlinespace[4pt]

\textbf{Proteins} & Proteina ($\mathcal{M}^{\text{small}}_{\text{FS}}$) & $\sim$60M & \href{https://github.com/NVIDIA-Digital-Bio/proteina}{GitHub: Proteina} \\
 & ProteinMPNN & $\sim$1.7M & \href{https://github.com/dauparas/ProteinMPNN}{GitHub: ProteinMPNN} \\
 & ESMFold-v1 & $\sim$3B & \href{https://github.com/facebookresearch/esm}{GitHub: ESMFold} \\ \bottomrule
\end{tabularx}
\end{table}

\subsection{Text-to-Image}
\label{sec:app:experiment_details:text-to-image}

\paragraph{Generative models.}
We use Stable Diffusion \citep{rombach2022high} v1.5($\sim$859.5M parameters),
which generates $3 \times 512 \times 512$ images from a noise space of
$4 \times 64 \times 64$, and a distilled version of the larger SDXL \cite{lin2024sdxllightning}
($\sim$2.57B parameters), which generates $3 \times 1024 \times 1024$ images
from a noise space of $4 \times 128 \times 128$.
Both models encode text prompts using CLIP-based text encoders
\citep{radford2021learning}.
Classifier-free guidance \citep{dhariwal2021diffusion} is applied during sampling
by linearly combining conditional and unconditional model predictions with
conditioning signal $c$.
For image sampling, we use the DDIM scheduler.
We set $\eta = 0$ (deterministic sampling) for black-box source noise optimization
methods and $\eta = 1.0$ (stochastic sampling) for noise sequence optimization methods.
The number of inference steps is set to $50$ for SD1.5 and $8$ for SDXL, and all
experiments are conducted in \texttt{float16} precision.

\paragraph{Reward functions.}
We use three reward functions for the text-to-image experiments, each defined by
the scalar output of a pretrained model.
ImageReward \citep{xu2023imagereward} and HPSv2 \citep{wu2023human} evaluate image–prompt alignment and are defined
as $R(\mathbf{x}_1, c)$, where $\mathbf{x}_1$ denotes the generated image and $c$ the text prompt.
In contrast, the Aesthetic predictor \citep{schuhmann2022laion} evaluates only the generated image and
is defined as $R(\mathbf{x}_1)$.

\subsection{Molecule Generation}
\label{sec:app:experiment_details:molecule_generation}

\paragraph{Generative model.}
We use EquiFM \citep{song2023equivariant} ($\sim$22M parameters) to generate small
QM9 molecules and base our implementation on the code of OC-Flow \citep{wang2024training}. EquiFM is an ODE-based flow matching model that operates on a
combined continuous Gaussian noise representation for both atomic coordinates
and atom types, with total dimensionality $M = 291$.
The number of atoms per molecule ranges from 3 to 29.
To accommodate variable-sized molecules while keeping a fixed input dimension,
unused atom entries are masked to zero.
In the experiments described in Section~\ref{sec:exp:molecule_generation},
the number of atoms is sampled according to the empirical distribution in the
QM9 training set, which most frequently yields molecules with 15 to 20 atoms.

\paragraph{Reward functions.}
The reward is defined as the negative sum of absolute deviations between target
property values and the corresponding predictions of pretrained regression
models.
For the $R_6$ reward, we consider the properties
$\{\alpha, \mu, \Delta\varepsilon, \varepsilon_{\text{HOMO}},
\varepsilon_{\text{LUMO}}, c_v\}$, corresponding to isotropic polarizability,
dipole moment, HOMO--LUMO gap, HOMO energy, LUMO energy, and heat capacity.
For the $R_3$ reward, we restrict the objective to the subset
$\{\alpha, \mu, \varepsilon_{\text{LUMO}}\}$.

\subsection{Protein Design}
\label{sec:app:experiment_details:protein_design}

\paragraph{Generative model.}
For protein design, we use the $\mathcal{M}^{\text{small}}_{\text{FS}}$ variant of
Proteina \citep{geffner2025proteina} with 400 integration steps, which consists of approximately 60M
transformer parameters and does not include triangle layers.
The source noise is drawn from a continuous Gaussian distribution
$\mathcal{N}(\mathbf{0}, \mathbf{I})$ with dimensionality $n_{\text{res}} \times 3$, corresponding
to the 3D coordinates of $n_{\text{res}}$ residues.

Once the model has learned the ODE dynamics, sampling can be performed either via
deterministic ODE integration,
$
\mathrm{d}\mathbf{x}_t = \mathbf{v}_t^\theta(\mathbf{x}_t, \tilde{c}) \, \mathrm{d}t,
$
or using stochastic SDE sampling,
$
\mathrm{d}\mathbf{x}_t =
\big[\mathbf{v}_t^\theta(\mathbf{x}_t, \tilde{c})
- \gamma g(t) \mathbf{s}_t^\theta(\mathbf{x}_t, \tilde{c})\big] \, \mathrm{d}t
+ \sqrt{2 \gamma g(t)} \, \mathrm{d}\mathbf{w}_t.
$
We discuss the implications of these sampling choices in more detail in
\Cref{sec:app:proteina_discussion}.
In practice, setting $\gamma < 1$ improves designability at the expense of
diversity and novelty.

\paragraph{Reward functions.}
Following \citep{geffner2025proteina}, we optimize the designability of generated
3D protein backbones.
Given a generated structure, a pretrained inverse folding model,
ProteinMPNN \citep{dauparas2022robust} ($\sim$1.7M parameters), is used to design a
compatible amino acid sequence.
This sequence is subsequently processed by a folding model, for which we use the
transformer-based ESMFold-v1 \citep{lin2023evolutionary} ($\sim$3B parameters), to
predict its 3D structure.
We compare the predicted structure to the original generated backbone by
computing the self-consistency root mean square deviation (scRMSD).
The scalar reward is then defined as
$
R(\mathbf{x}_1) = \exp(-\mathrm{scRMSD(\mathbf{x}_1)}),
$
which maps the score to the range $(0, 1]$, with values closer to 1 indicating
higher designability. Note that usually $8$ sequences are extracted and folded and the best of those is chosen, but for computational efficiency we only extract $1$ per protein backbone.

\section{Ablations}
\label{sec:app:ablations}

Here we perform further ablation studies to justify the design choices in Section~\ref{sec:methodology:trs} and hyperparameters in \Cref{sec:app:hyper_params}. Most importantly, \Cref{sec:app:ablations:center_selection} emphasises the importance of our top-$k$ center selection instead of keeping them strictly apart. \Cref{sec:app:ablation:lengths_and_masks} provides visual and quantitative insights into the relationship of the trust-region lengths $\ell_j$ and the mask probabilities $p_j$, while \Cref{sec:app:ablations:warm-up} shows that $20\%$ is a good heuristic for the fraction of warm-up budgets across multiple total budgets. Finally, \Cref{sec:app:ablations:n_regions} shows that even the number of regions $k$, is a robust hyperparameter, which is the only one that we change across our experiments in Section~\ref{sec:benchmarks}.

\subsection{Center Selection}
\label{sec:app:ablations:center_selection}

\paragraph{Setup.} 
We investigate four center selection strategies for text-to-image generation using Stable Diffusion 1.5 (SD1.5) and ImageReward as reward function. \textit{LocalBest} and \textit{LocalLastIter} enforce strict region separation, updating centers using either the historical best or the most recent best sample within each region $\mathcal{T}_j$, respectively. In contrast, \textit{GlobalTopk} and \textit{GlobalLastIter} allow interaction by re-centering regions globally based on the overall top-$k$ samples observed so far or in the latest iteration. All experiments use a batch size $B=20$ and 20 iterations ($20k$ NFE), consistent with the configuration in Section~\ref{sec:exp:text-to-image}.

\paragraph{Metrics.} 
Performance is evaluated on the full DrawBench benchmark. We use the mean best ImageReward as the primary metric to rank the generated noise samples and determine center updates. Random search is included as baseline as a reference to compare the relative improvement of each selection strategy.

\paragraph{Results.} 
As shown in \Cref{fig:center_selection}, \textit{GlobalTopk} consistently outperforms all other strategies. The results indicate that enforcing strict separation between regions (as in the original TuRBO algorithm \citep{eriksson2019scalable}) limits performance in this setting. Conversely, global re-centering enables a more effective allocation of the evaluation budget toward the most promising areas of the search space, significantly improving generation quality.

\begin{figure}[h]
\centering
\begin{tikzpicture}
\begin{axis}[
    title={Center Selection Strategies (SD1.5 + IR)},
    width=0.6\linewidth,
    height=0.35\linewidth,
    xbar,
    bar width=10pt,
    xmin=1.40,
    xmax=1.64,
    grid=both,
    grid style={gray!30},
    tick label style={font=\small},
    label style={font=\small},
    title style={font=\small\bfseries},
    xlabel={Mean best reward},
    symbolic y coords={
        RandomSearch,
        LocalLastIter,
        LocalBest,
        GlobalLastIter,
        GlobalTopk
    },
    ytick=data,
    axis y line*=left,
]

\addplot[
    fill=method1color,
    draw=method1color
] coordinates {
    (1.43,RandomSearch)
    (1.455,LocalLastIter)
    (1.494,LocalBest)
    (1.553,GlobalLastIter)
    (1.616,GlobalTopk)
};

\end{axis}
\end{tikzpicture}
\caption{\textbf{Effect of center-selection strategy.} Evaluated on DrawBench using SD1.5 and ImageReward. We compare the mean best rewards across the prompts in the benchmark.}
\label{fig:center_selection}
\end{figure}

\subsection{Trust Region Dynamics: Interaction of Length and Masks}
\label{sec:app:ablation:lengths_and_masks}

\paragraph{Setup.} 
We perform a grid ablation across varying initial trust region lengths $\ell_{\text{init}}$ and masking sparsity ranges defined by $p_{\text{min}}$ and $p_{\text{max}}$. 
First, we do this quantitatively by investigating different combinations of initial trust-region lengths $\ell_{\text{init}}$ and set the mask probabilities in different ranges, with different min and max probabilities $p_{\text{min}}$, $p_{\text{max}}$. 
This experiment is conducted using SD1.5 on a representative subset of 55 prompts from DrawBench, covering all categories of the original benchmark to ensure broad coverage of the latent space. Additionally, we visualize the relationship of the trust-region length $\ell_j$ and sampled mask probability $p_j$ in \Cref{fig:trs_comparison_grid}. 

\paragraph{Metrics.} 
In \Cref{fig:trs_comparison_grid} we aim to track the point at which the samples begin to exhibit structural degradation and visualize these noising effects, while we show the mean best ImageReward for the quantitative experiment in \Cref{tab:tab:trs_robustness}. 

\paragraph{Results.} 
\Cref{fig:trs_comparison_grid} reveals a clear stability threshold at approximately $\ell_j \approx 1.6$, beyond which images show visible noise. However, we find that low-probability masks (e.g., $p_j=0.05$) act as a regularizer and they allow for significantly higher exploration lengths, up to $\ell_j=6.4$, while maintaining valid, coherent structures. 
\Cref{tab:tab:trs_robustness} reveals that the exact masking range is not particalarely important, it works well for different ranges. The initial shows also good robustness, even when changing it to very high or low numbers, due to the adaptive nature of the algorithm.

\begin{table}[ht]
\centering
\renewcommand{\arraystretch}{1} %
\setlength{\tabcolsep}{6pt}      %

\caption{\textbf{Sensitivity to trust-region init length.} Mean best rewards across varying trust-region init lengths ($\ell_{\text{init}}$) and [$p_{\text{min}}$, $p_{\text{max}}$] settings for SD1.5 and ImageReward.}
\label{tab:tab:trs_robustness}
\begin{tabular}{lcccc}
\toprule
\textbf{TR init length} & \multicolumn{4}{c}{\textbf{Mask probabilities [$p_{\text{min}}$, $p_{\text{max}}$]}} \\
\cmidrule(lr){2-5}
($\ell_{\text{init}}$) & [0.05, 0.50] & [0.10, 0.90] & [0.30, 0.70] & [0.50, 1.00] \\
\midrule
0.2 (Low)    & 1.6223 & 1.6430 & 1.6353 & 1.6564 \\
0.8 (Medium)  & 1.6928 & 1.6761 & 1.6820 & 1.6543 \\
2.0 (Large) & 1.6404 & 1.6197 & 1.6306 & 1.6185 \\
\bottomrule
\end{tabular}
\end{table}

\begin{figure}[ht]
    \centering
    \small
    \setlength{\tabcolsep}{1.2pt}
    \renewcommand{\arraystretch}{1.0}

    \begin{tabular}{c cccccc @{\hspace{12pt}} cccccc}
        & \multicolumn{6}{c}{\textbf{(a) SD1.5 ($D = 16384$)}} & \multicolumn{6}{c}{\textbf{(b) SDXL ($D = 65536$)}} \\[4pt]

        & \textbf{0.05} & \textbf{0.2} & \textbf{0.4} & \textbf{0.6} & \textbf{0.8} & \textbf{1.0} 
        & \textbf{0.05} & \textbf{0.2} & \textbf{0.4} & \textbf{0.6} & \textbf{0.8} & \textbf{1.0} \\

        \textbf{0.4} & 
        \includegraphics[width=0.07\textwidth]{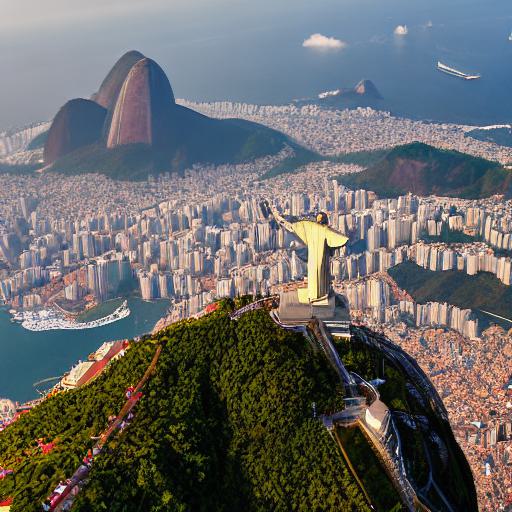} &
        \includegraphics[width=0.07\textwidth]{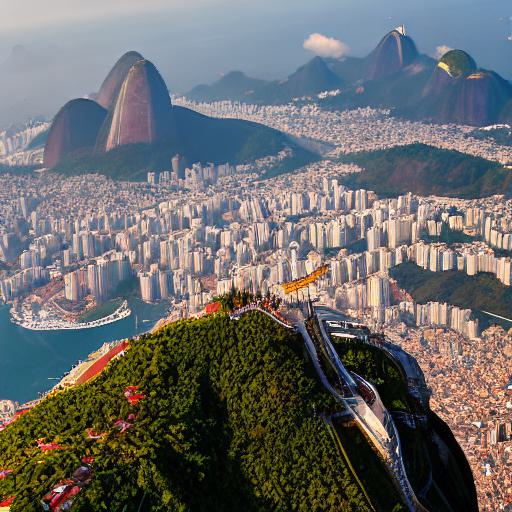} &
        \includegraphics[width=0.07\textwidth]{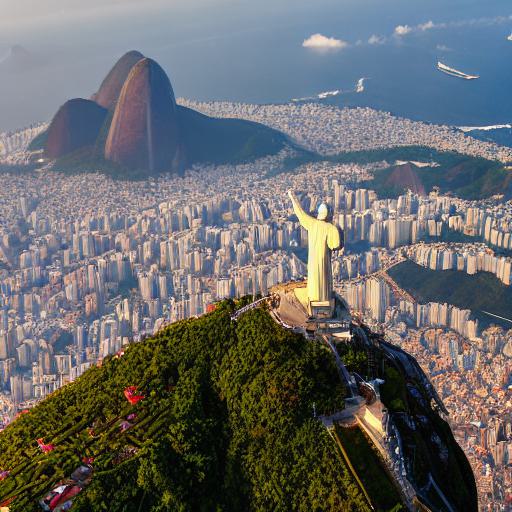} &
        \includegraphics[width=0.07\textwidth]{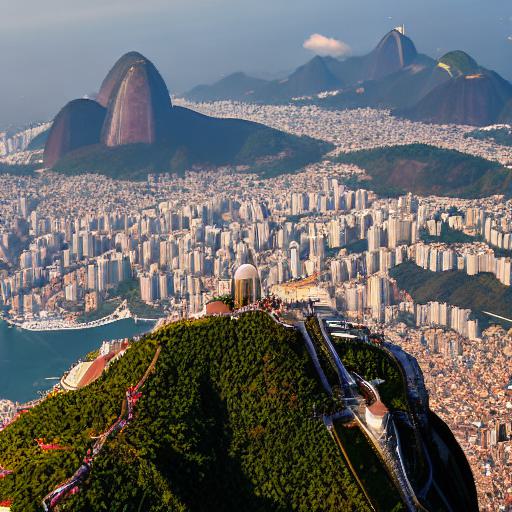} &
        \includegraphics[width=0.07\textwidth]{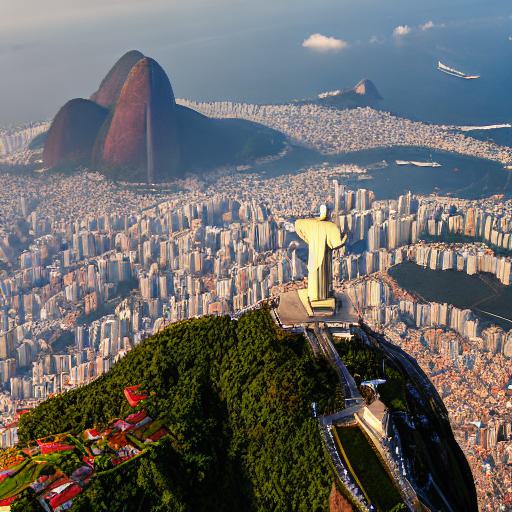} &
        \includegraphics[width=0.07\textwidth]{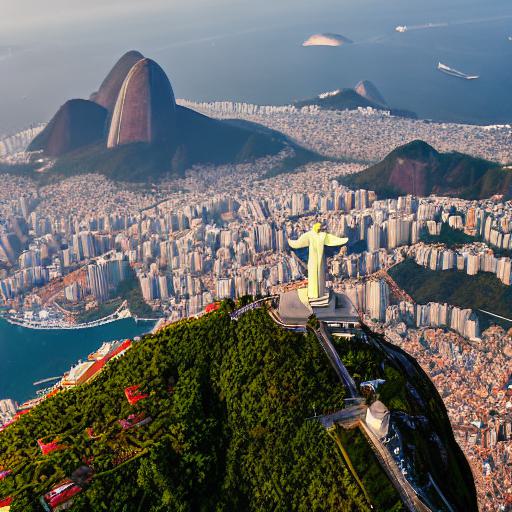} &
        \includegraphics[width=0.07\textwidth]{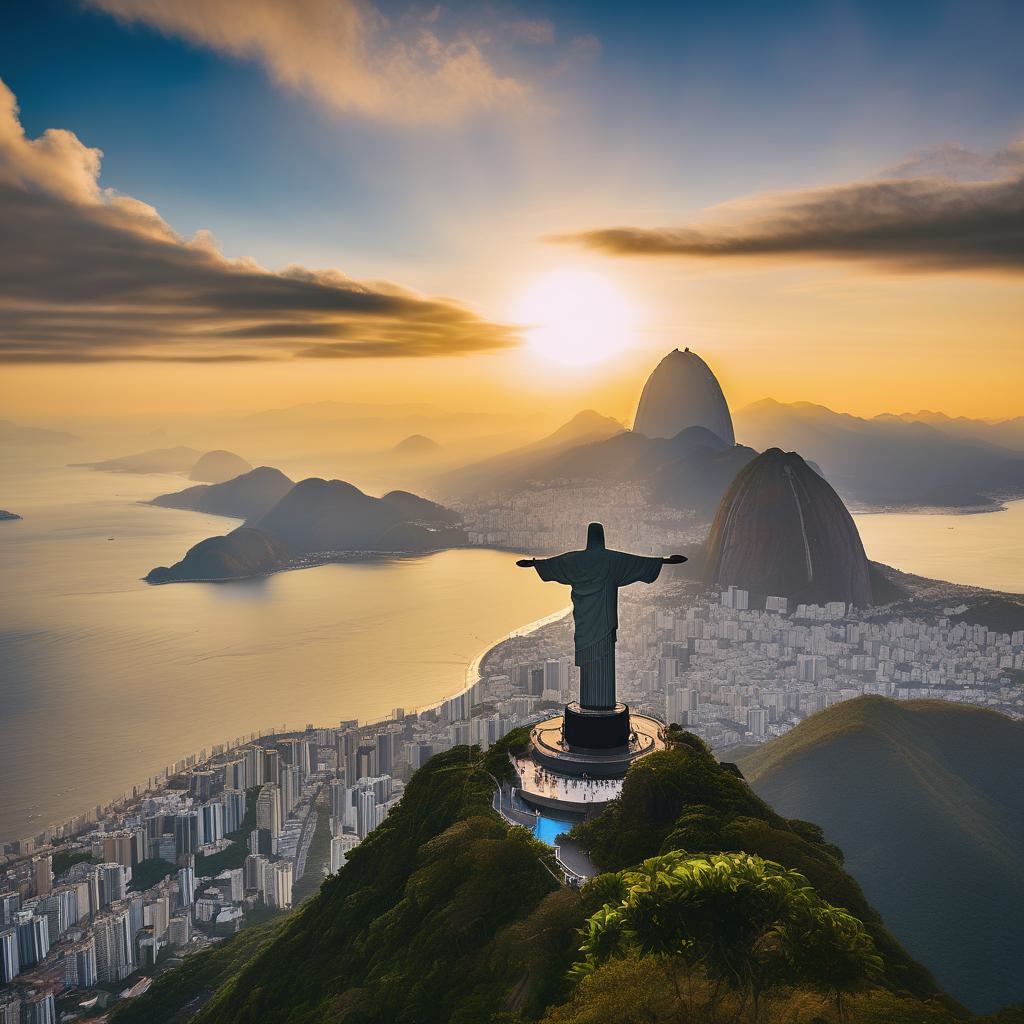} &
        \includegraphics[width=0.07\textwidth]{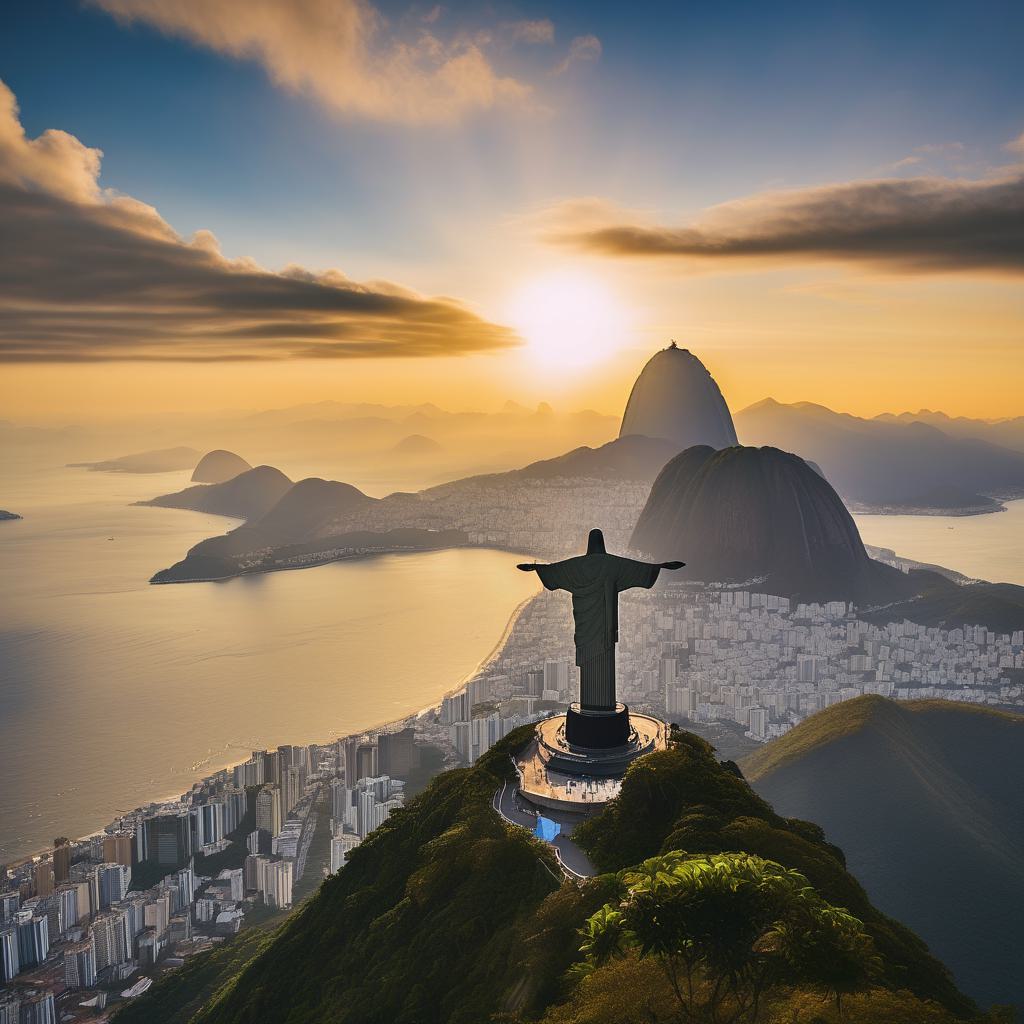} &
        \includegraphics[width=0.07\textwidth]{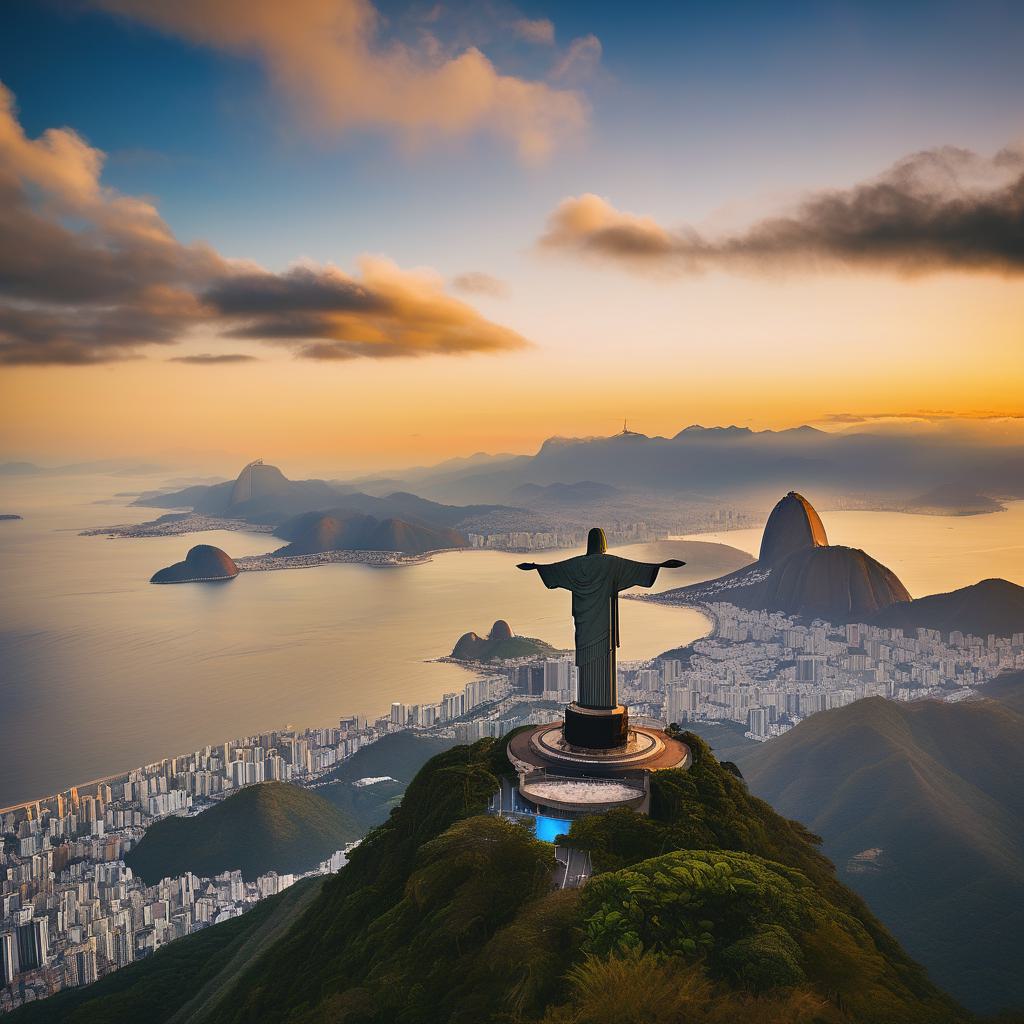} &
        \includegraphics[width=0.07\textwidth]{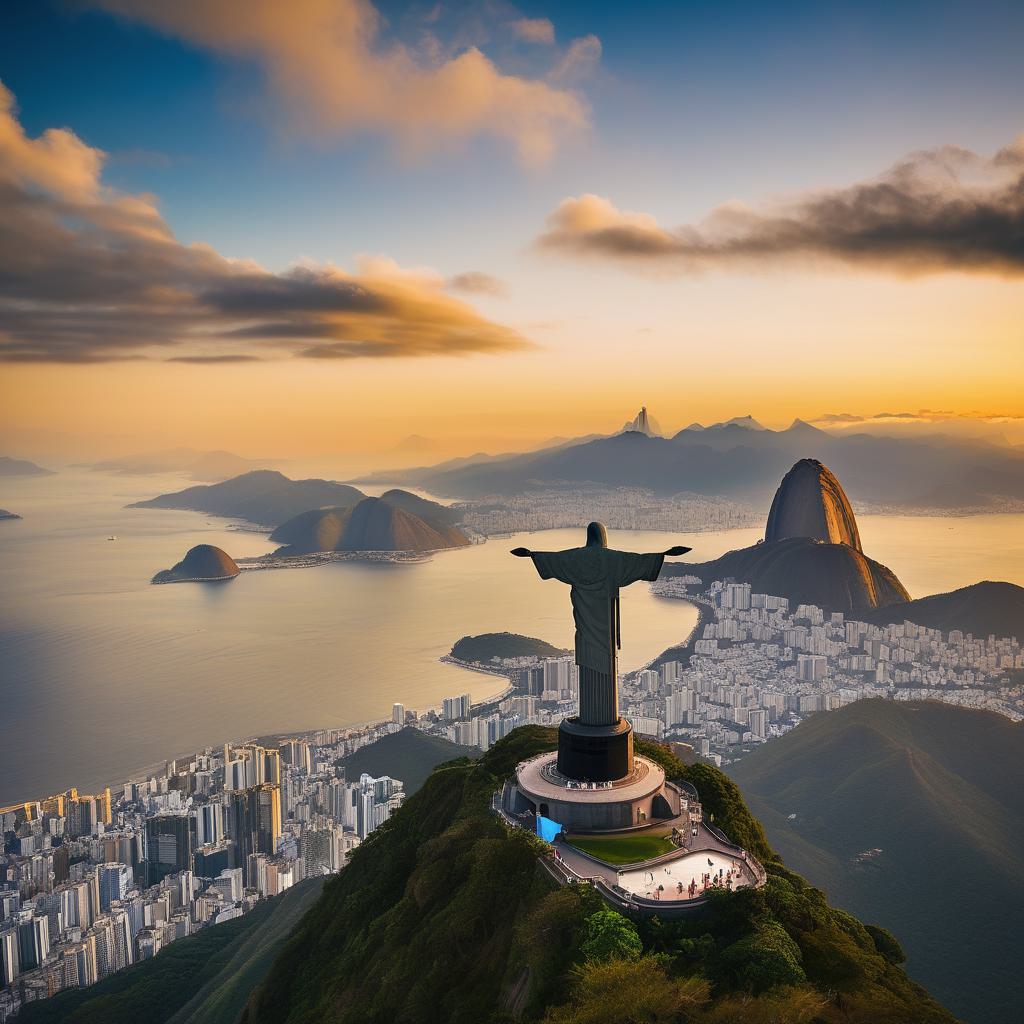} &
        \includegraphics[width=0.07\textwidth]{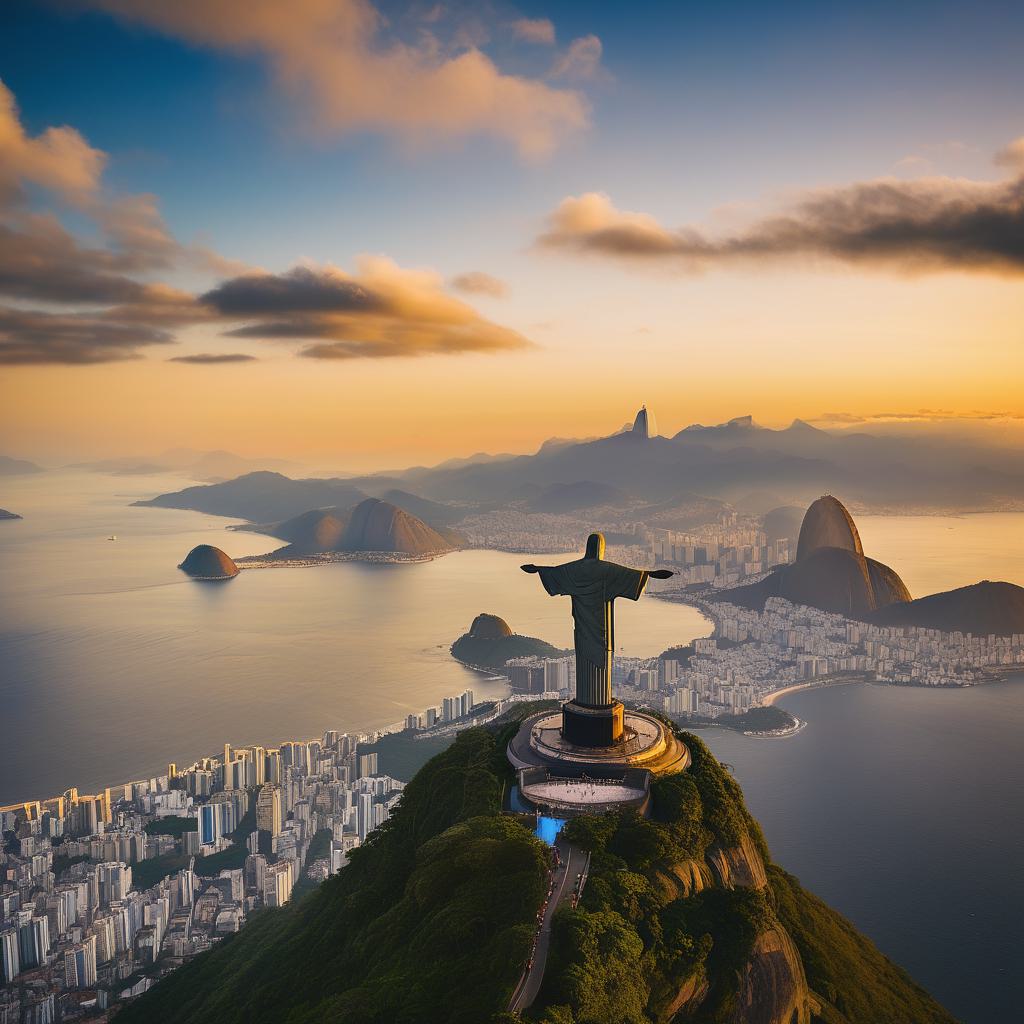} &
        \includegraphics[width=0.07\textwidth]{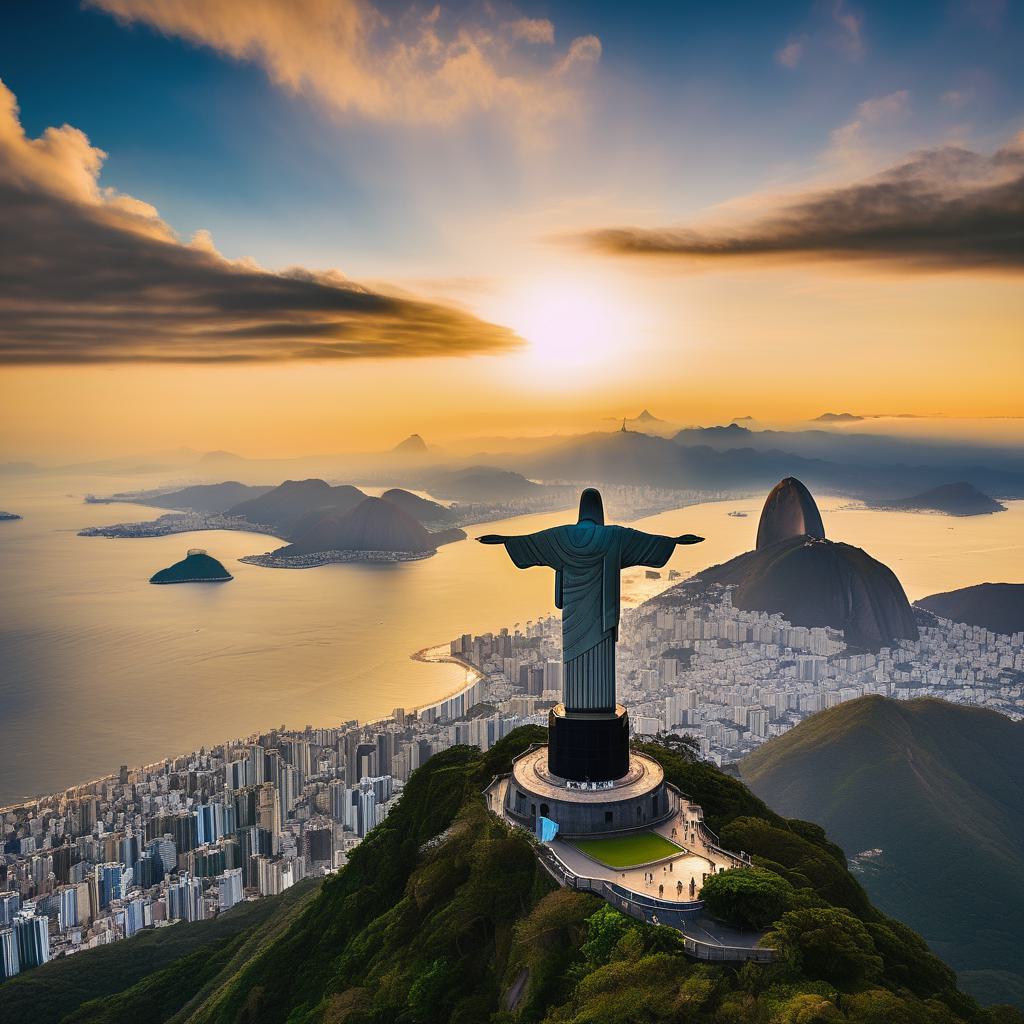} \\

        \textbf{0.8} & 
        \includegraphics[width=0.07\textwidth]{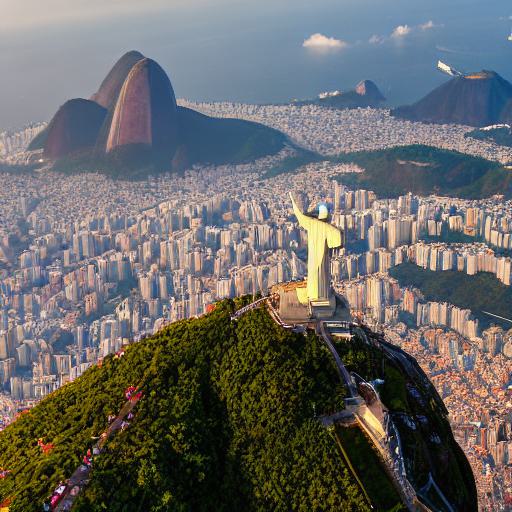} &
        \includegraphics[width=0.07\textwidth]{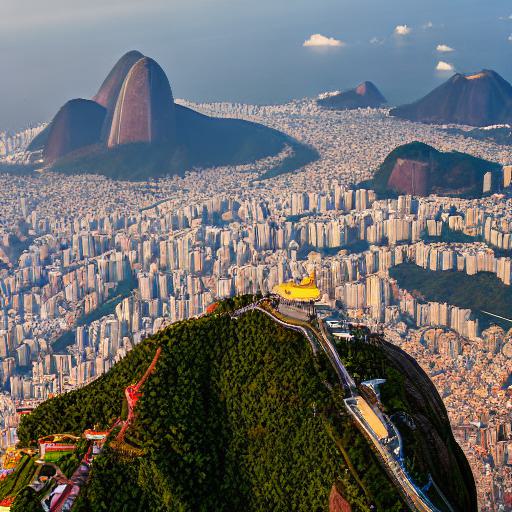} &
        \includegraphics[width=0.07\textwidth]{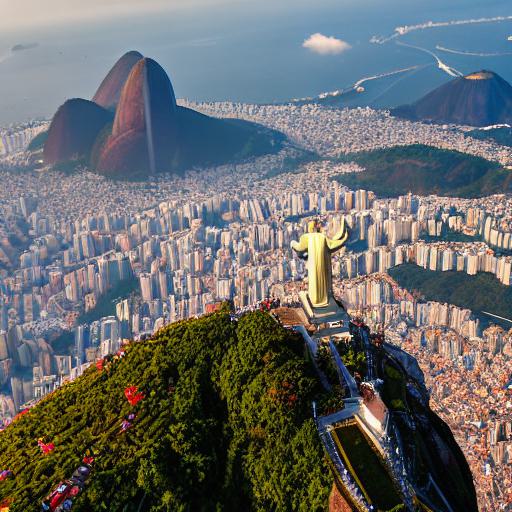} &
        \includegraphics[width=0.07\textwidth]{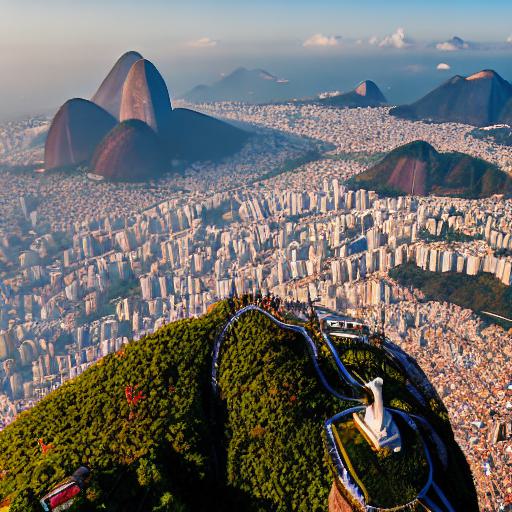} &
        \includegraphics[width=0.07\textwidth]{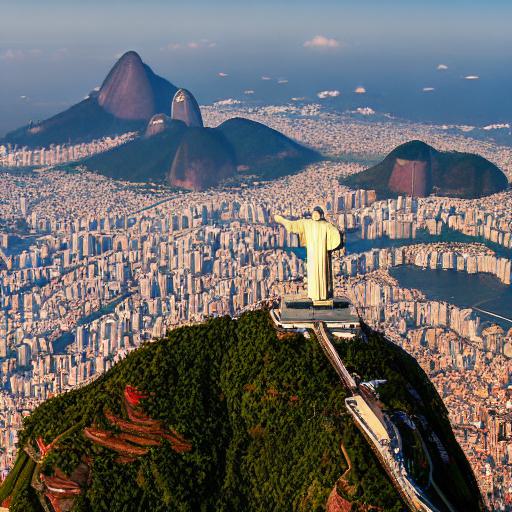} &
        \includegraphics[width=0.07\textwidth]{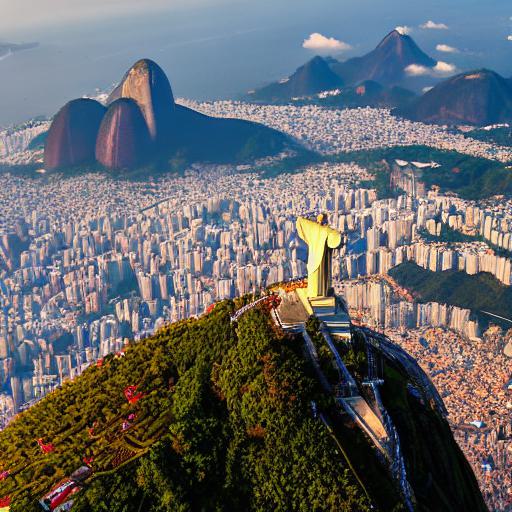} &
        \includegraphics[width=0.07\textwidth]{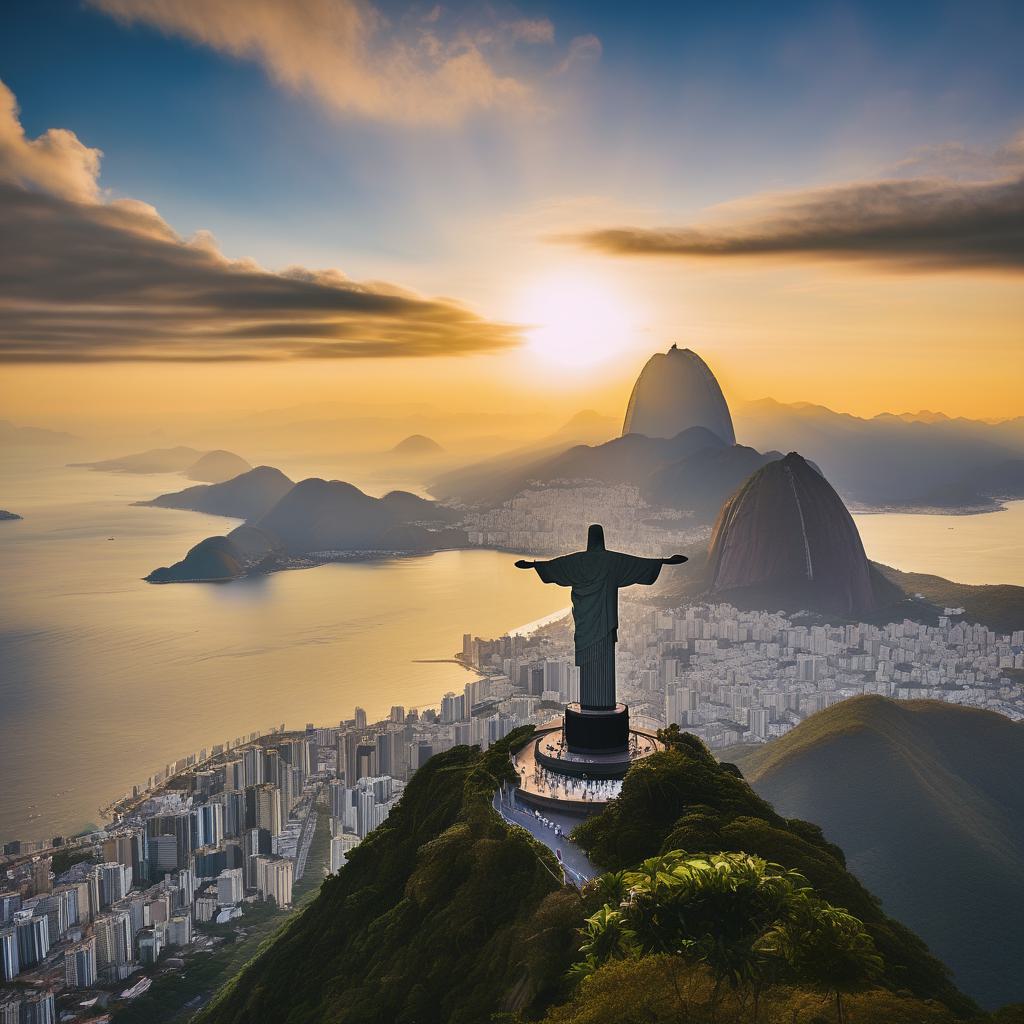} &
        \includegraphics[width=0.07\textwidth]{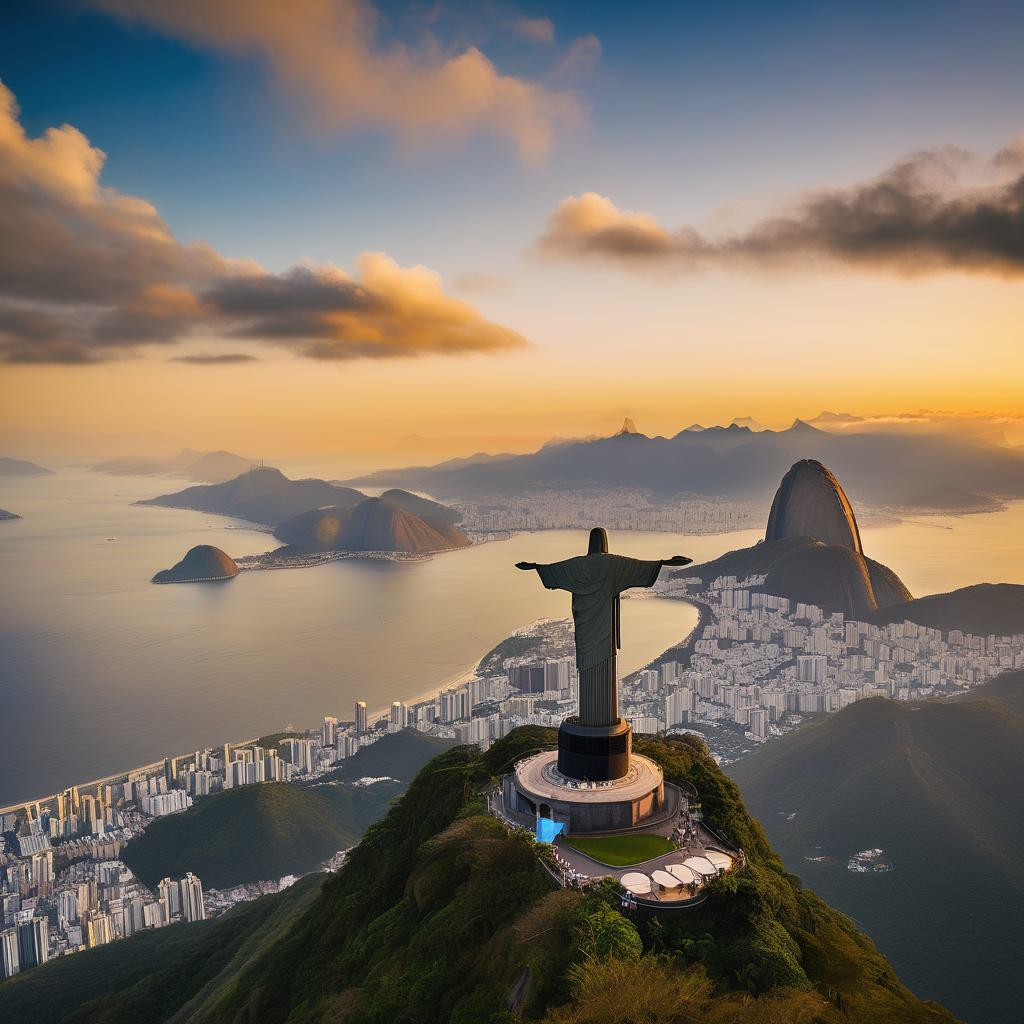} &
        \includegraphics[width=0.07\textwidth]{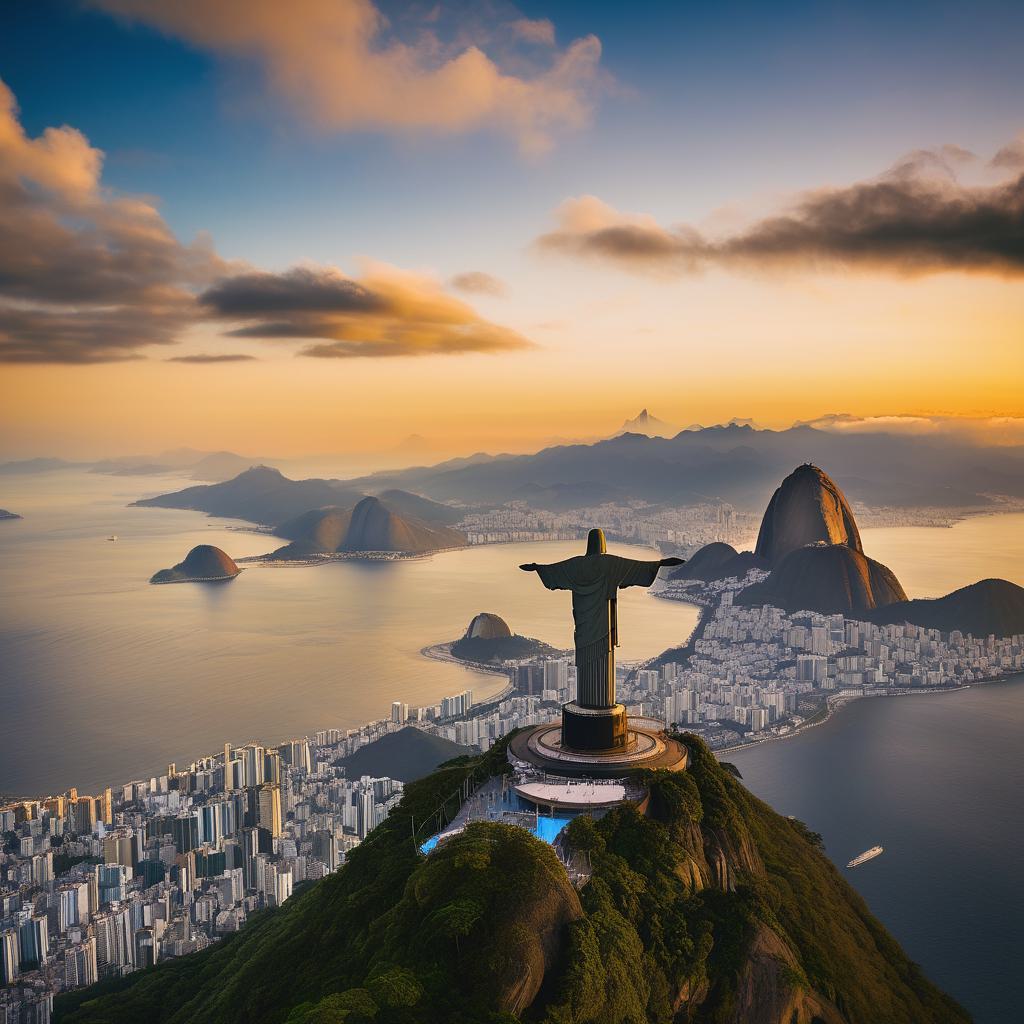} &
        \includegraphics[width=0.07\textwidth]{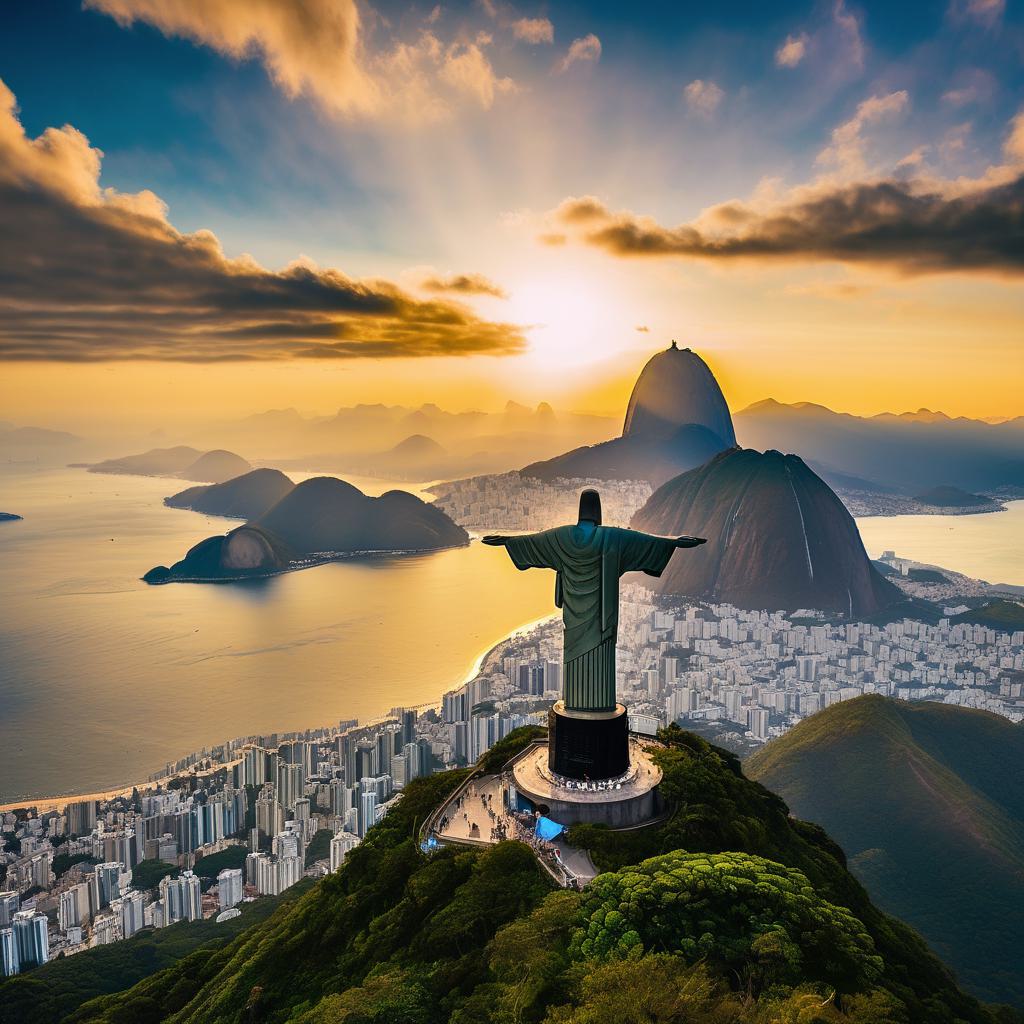} &
        \includegraphics[width=0.07\textwidth]{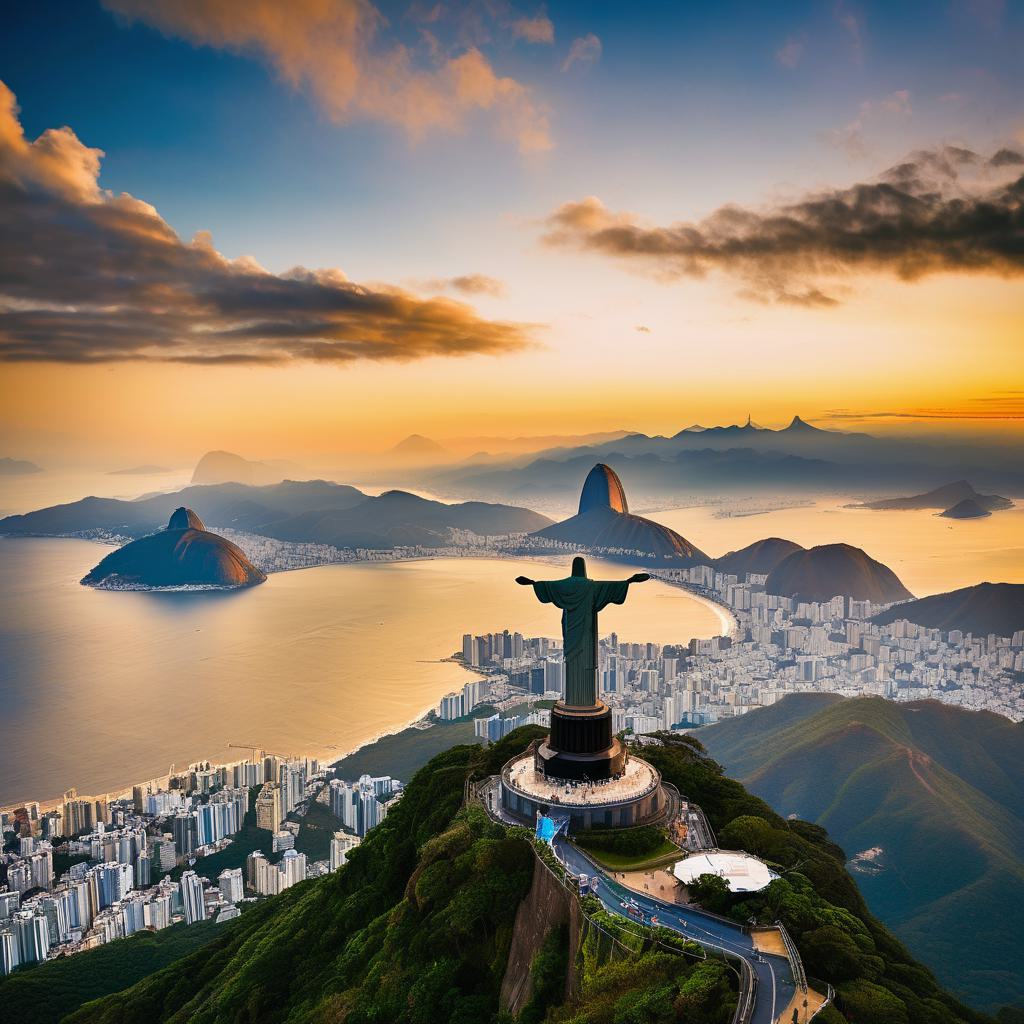} &
        \includegraphics[width=0.07\textwidth]{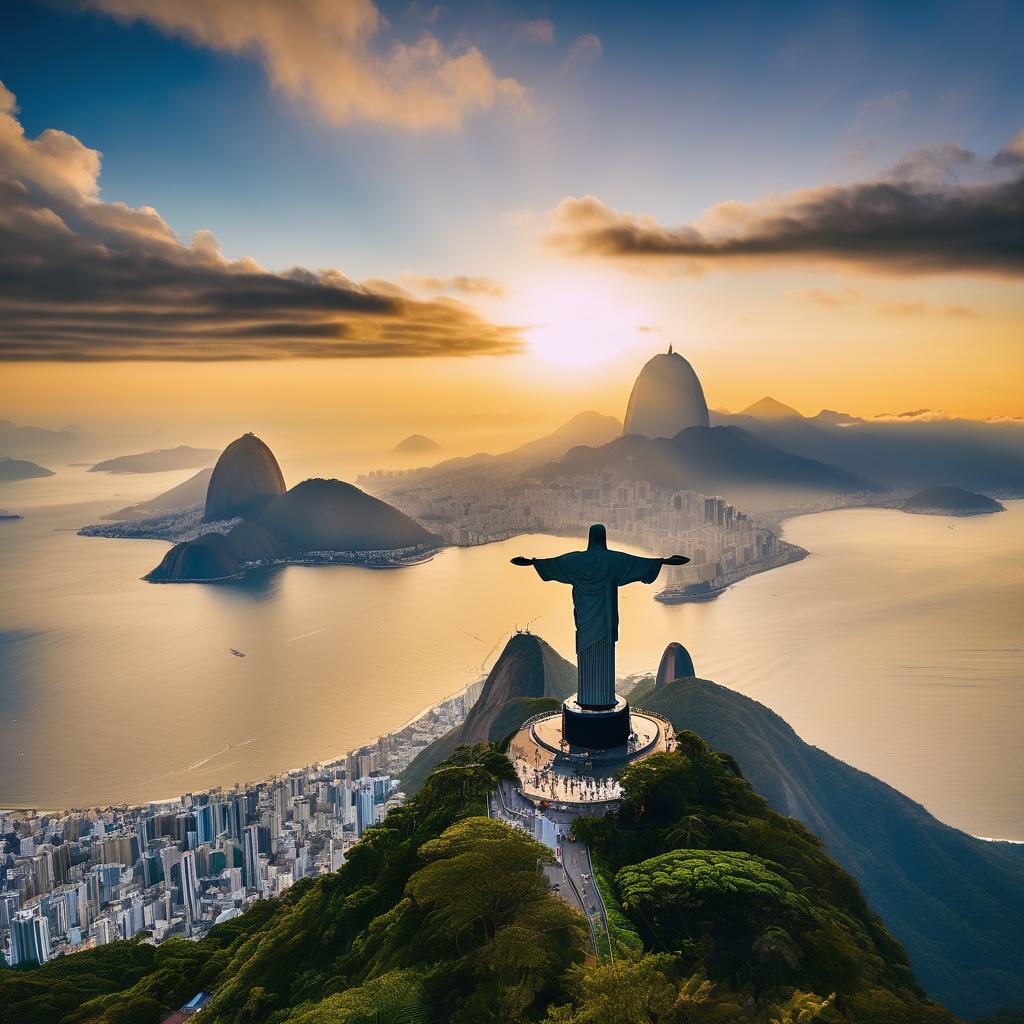} \\

        \textbf{1.6} & 
        \includegraphics[width=0.07\textwidth]{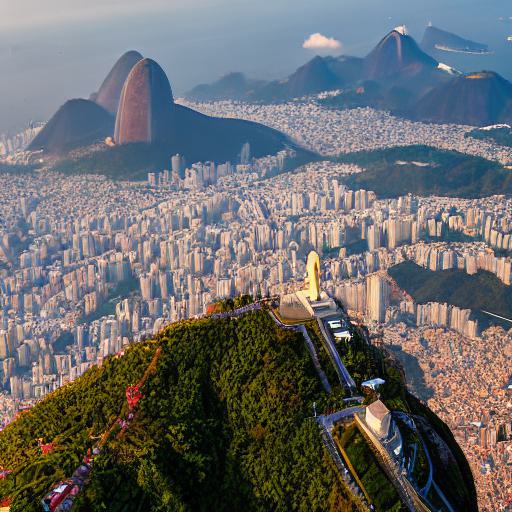} &
        \includegraphics[width=0.07\textwidth]{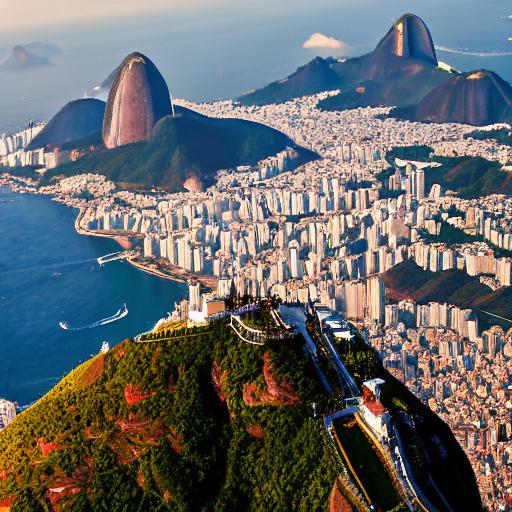} &
        \includegraphics[width=0.07\textwidth]{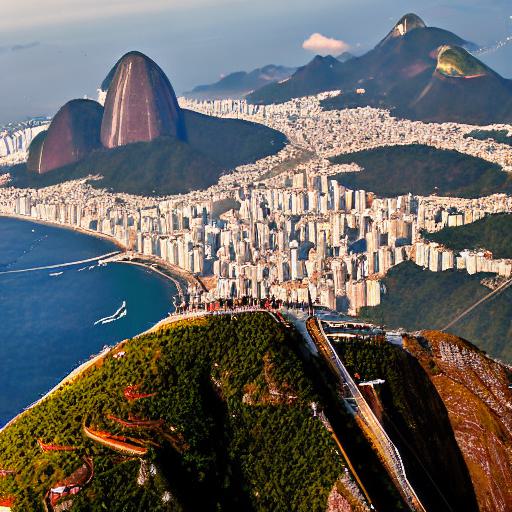} &
        \includegraphics[width=0.07\textwidth]{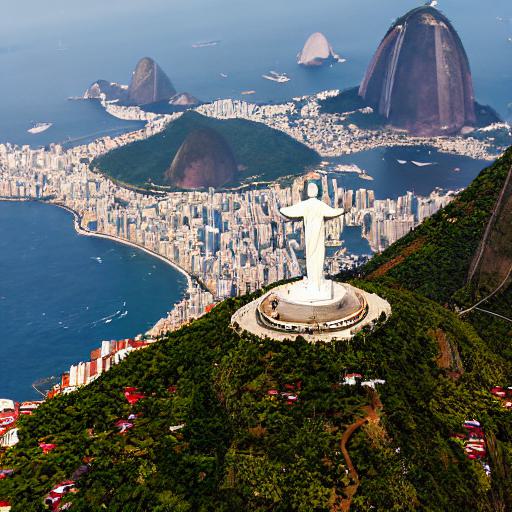} &
        \includegraphics[width=0.07\textwidth]{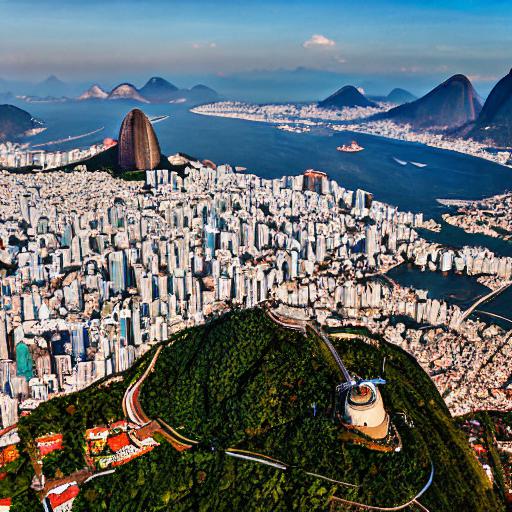} &
        \includegraphics[width=0.07\textwidth]{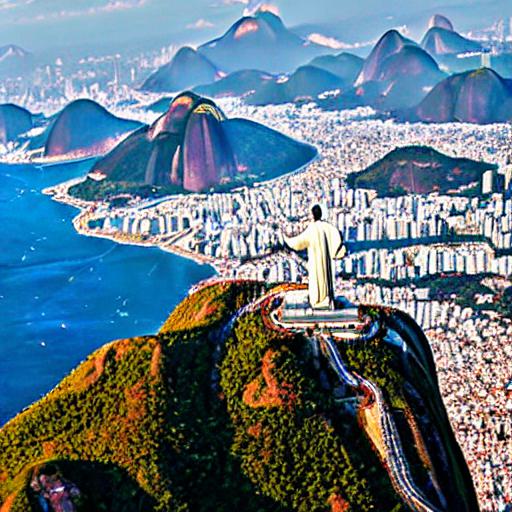} &
        \includegraphics[width=0.07\textwidth]{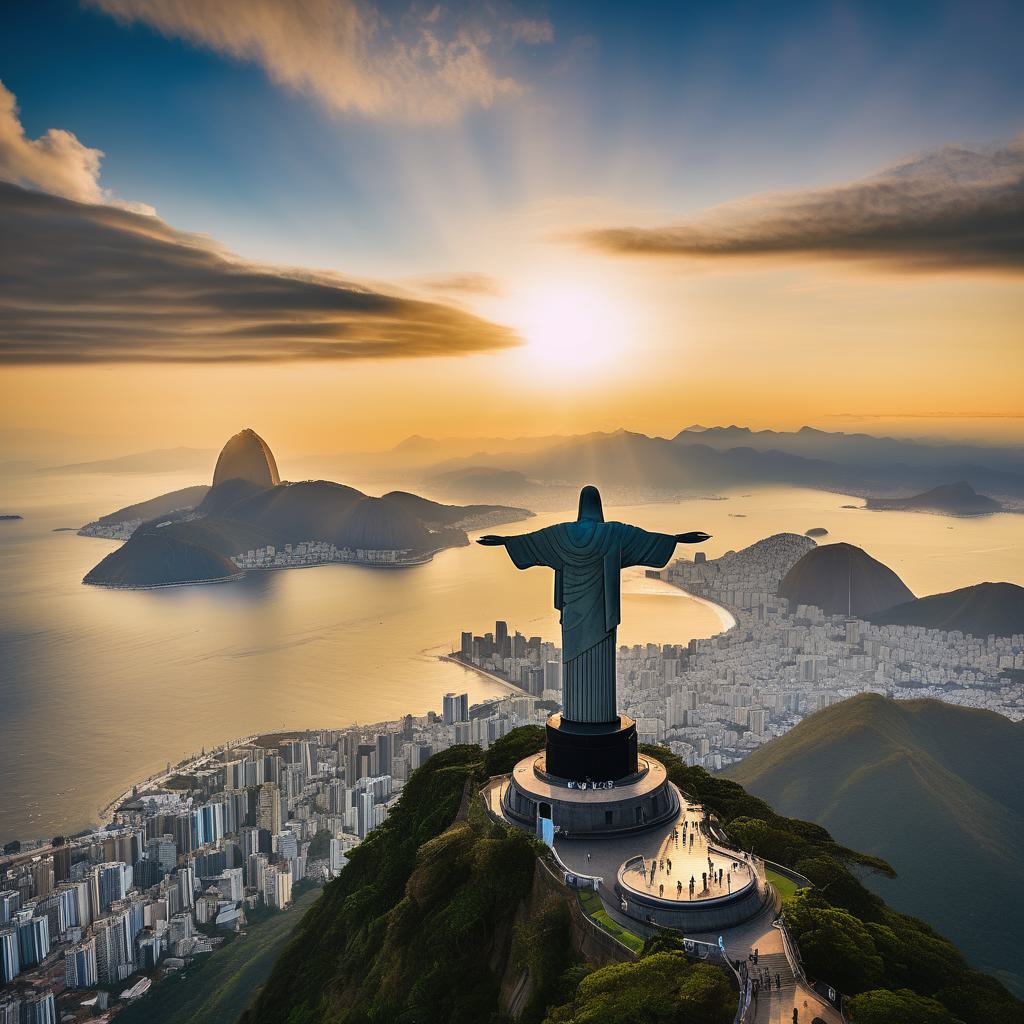} &
        \includegraphics[width=0.07\textwidth]{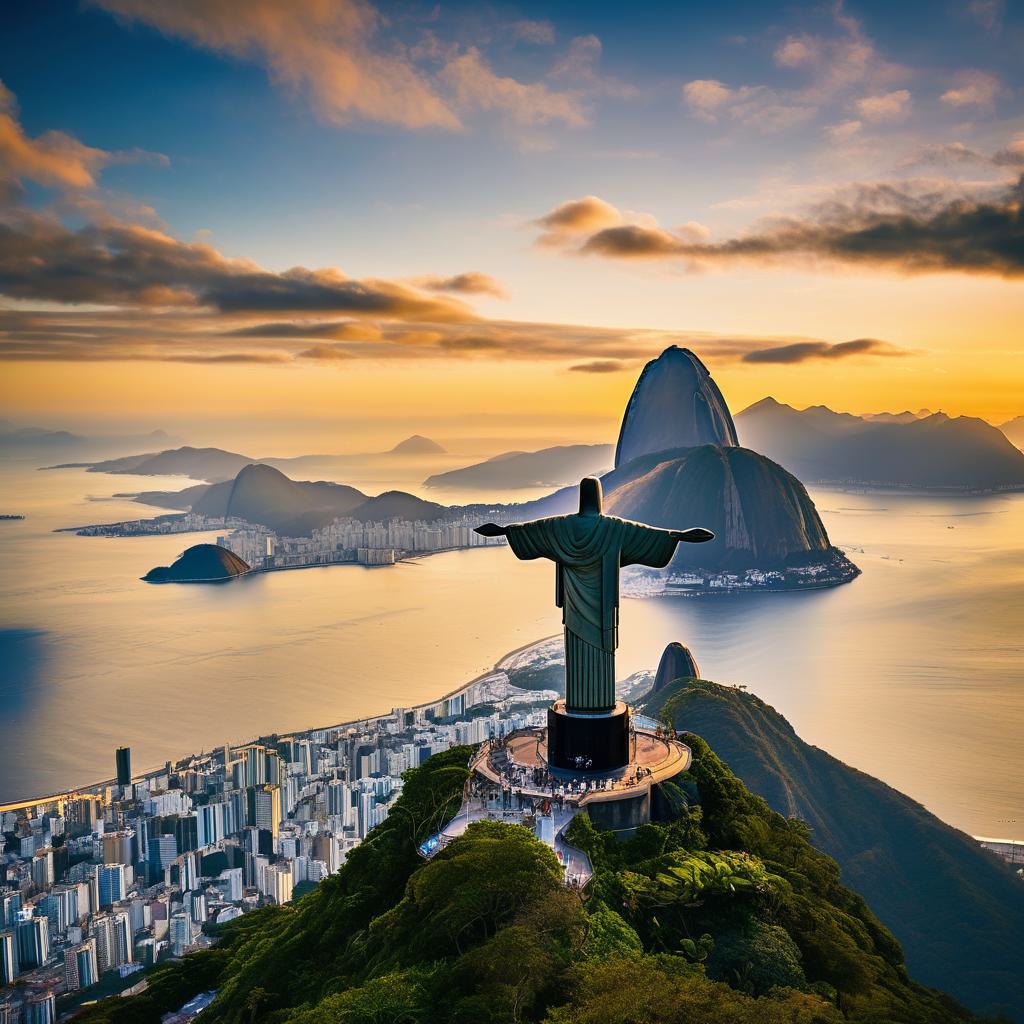} &
        \includegraphics[width=0.07\textwidth]{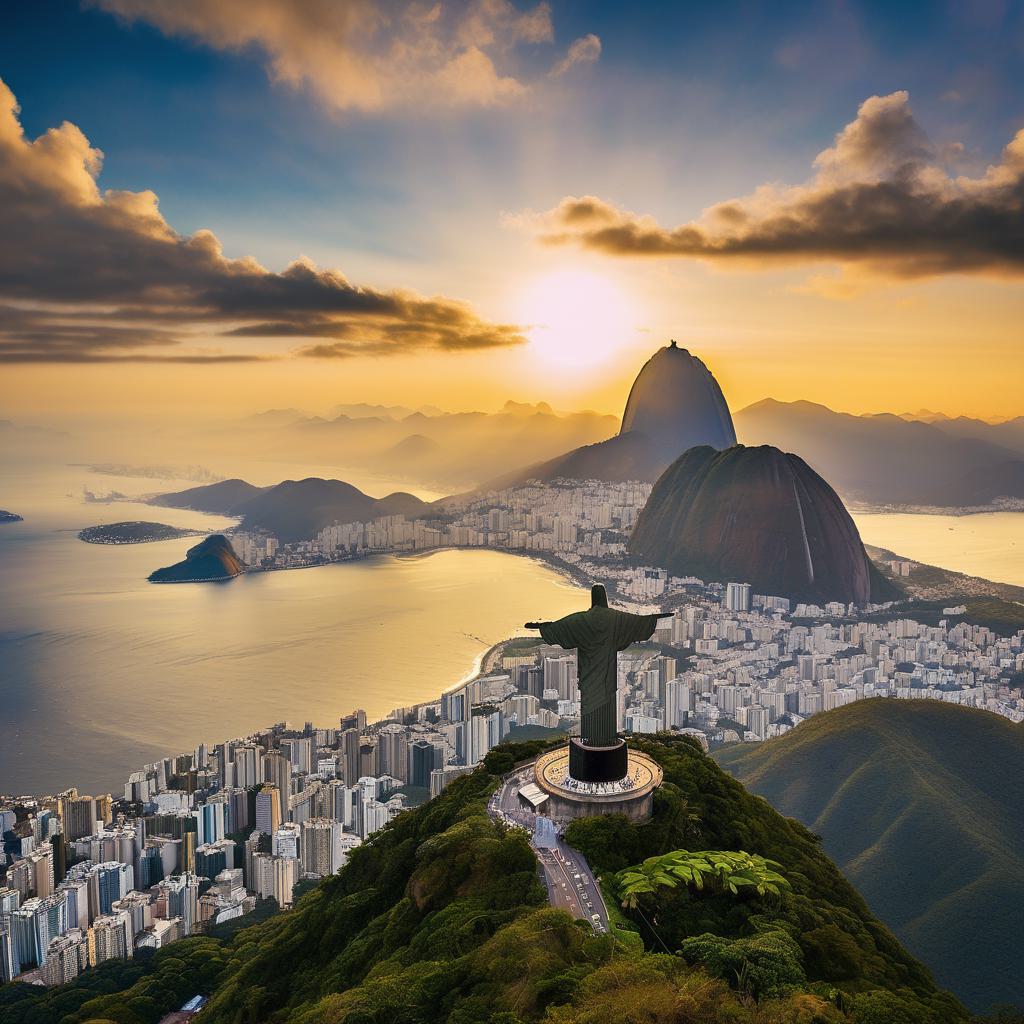} &
        \includegraphics[width=0.07\textwidth]{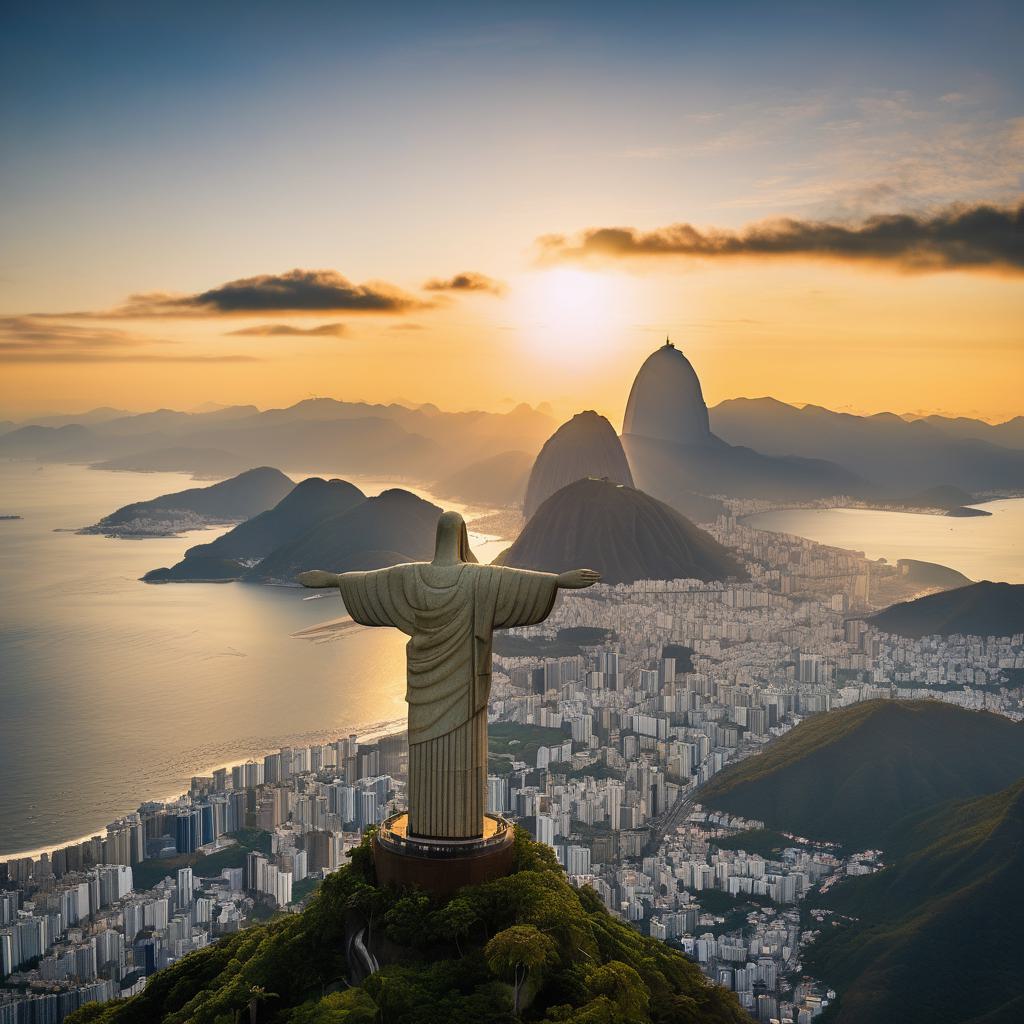} &
        \includegraphics[width=0.07\textwidth]{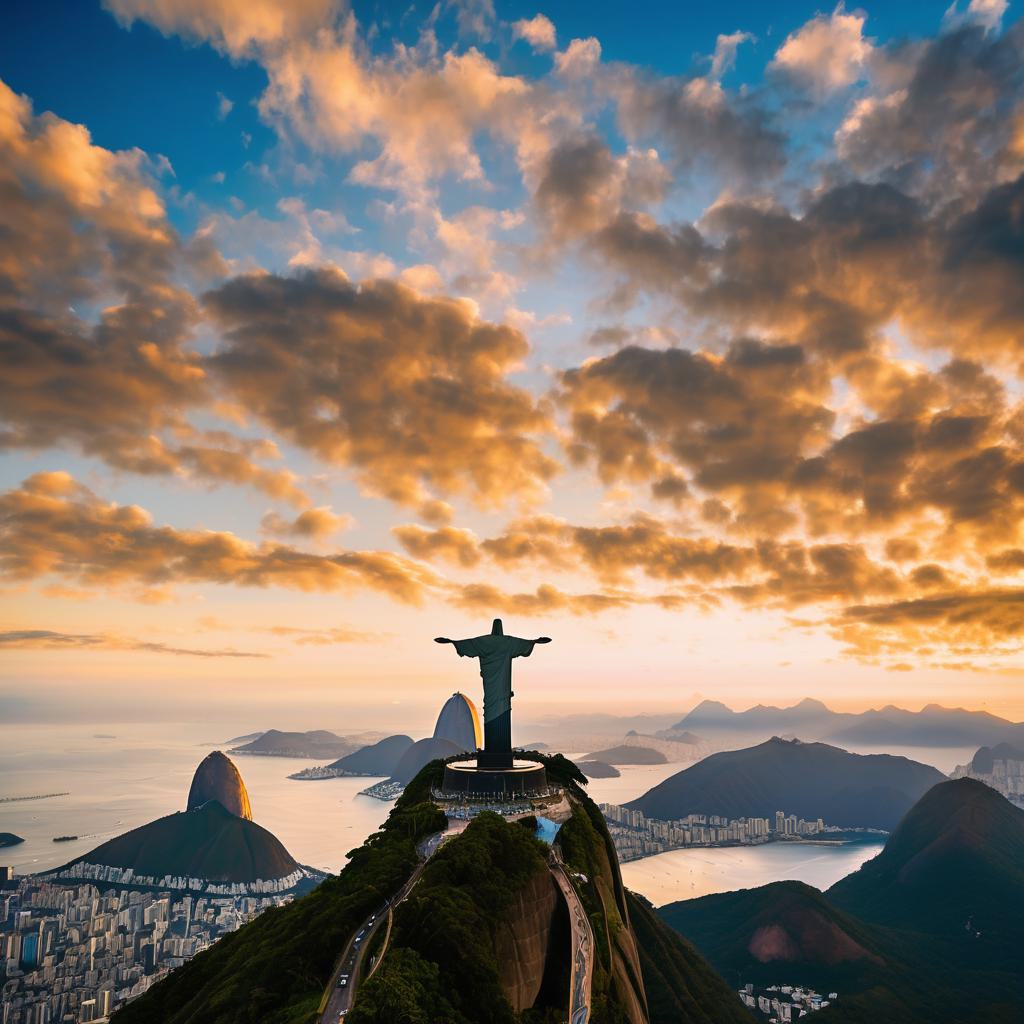} &
        \includegraphics[width=0.07\textwidth]{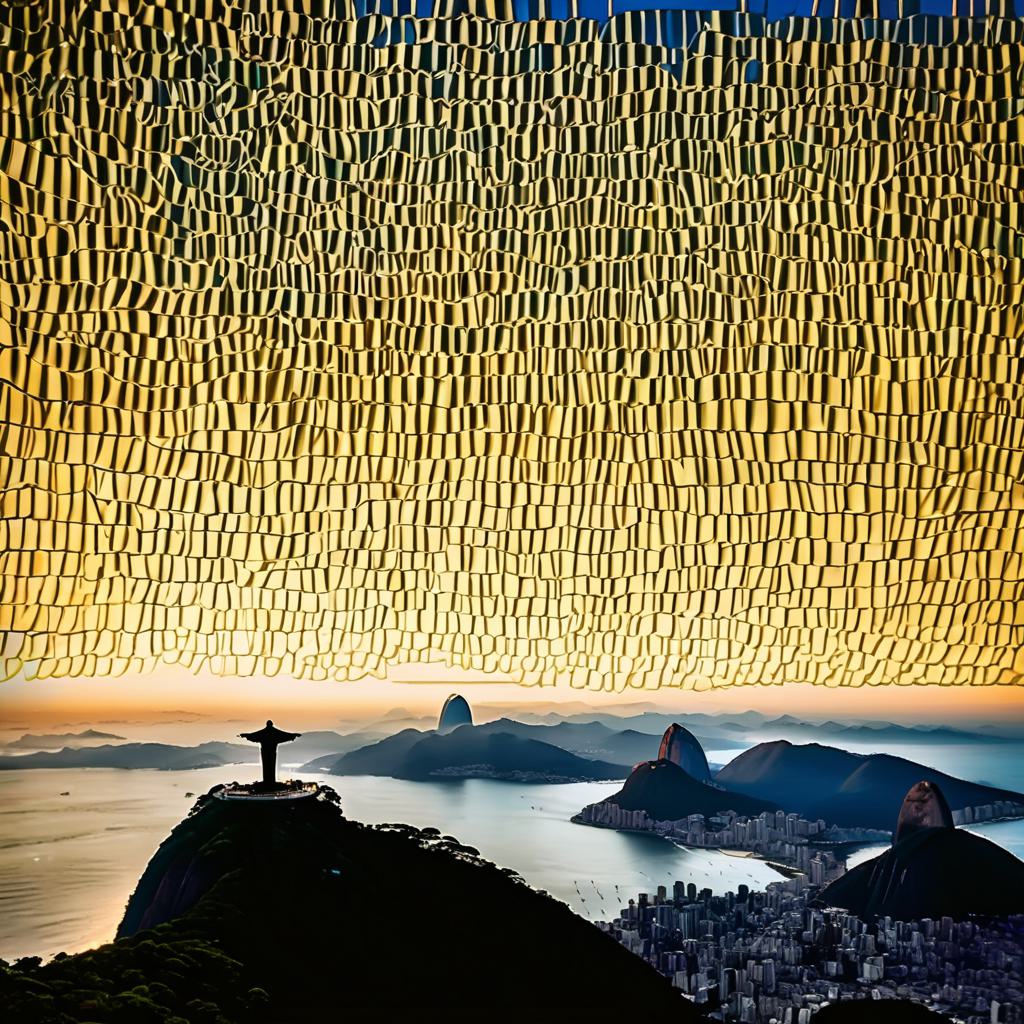} \\

        \textbf{3.2} & 
        \includegraphics[width=0.07\textwidth]{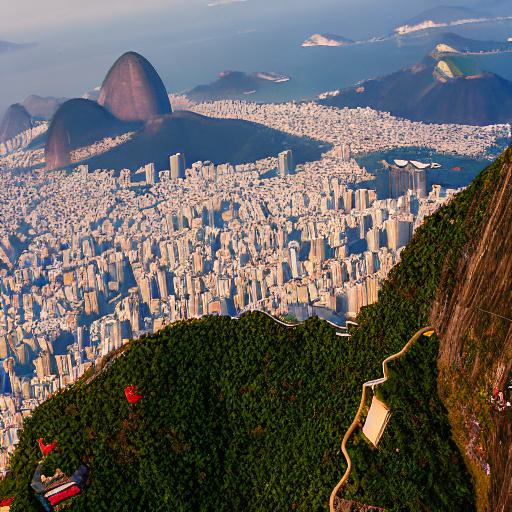} &
        \includegraphics[width=0.07\textwidth]{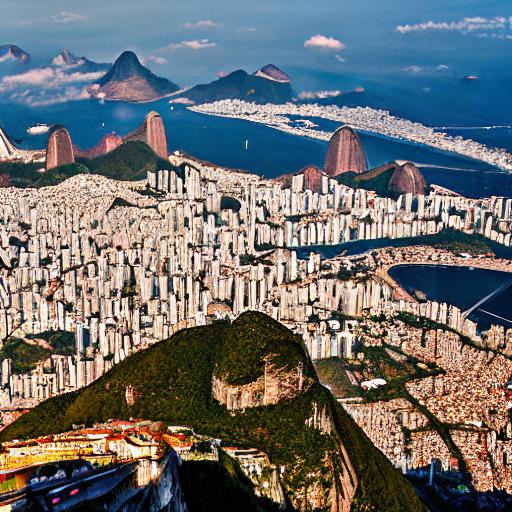} &
        \includegraphics[width=0.07\textwidth]{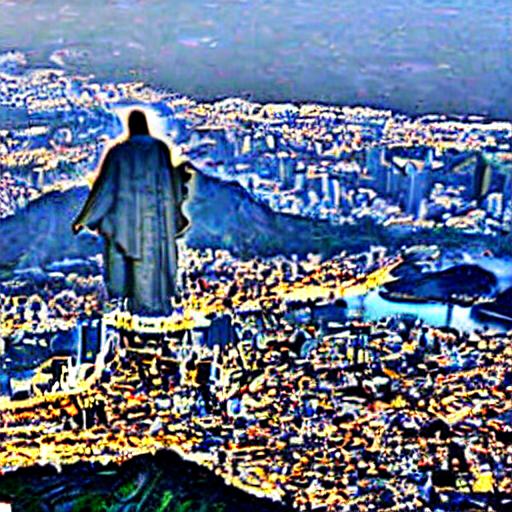} &
        \includegraphics[width=0.07\textwidth]{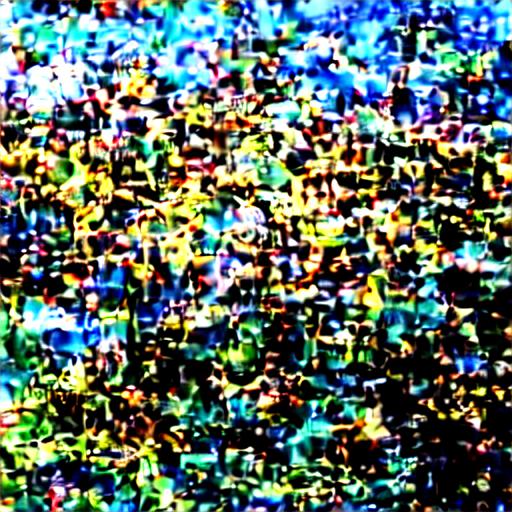} &
        \includegraphics[width=0.07\textwidth]{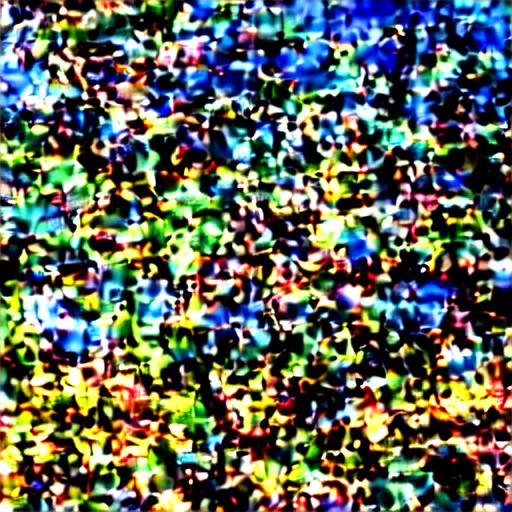} &
        \includegraphics[width=0.07\textwidth]{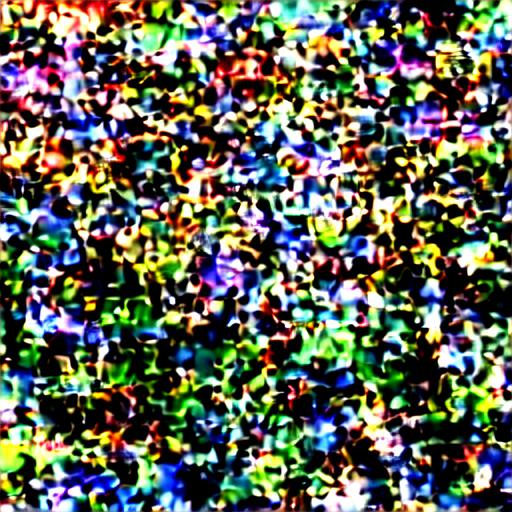} &
        \includegraphics[width=0.07\textwidth]{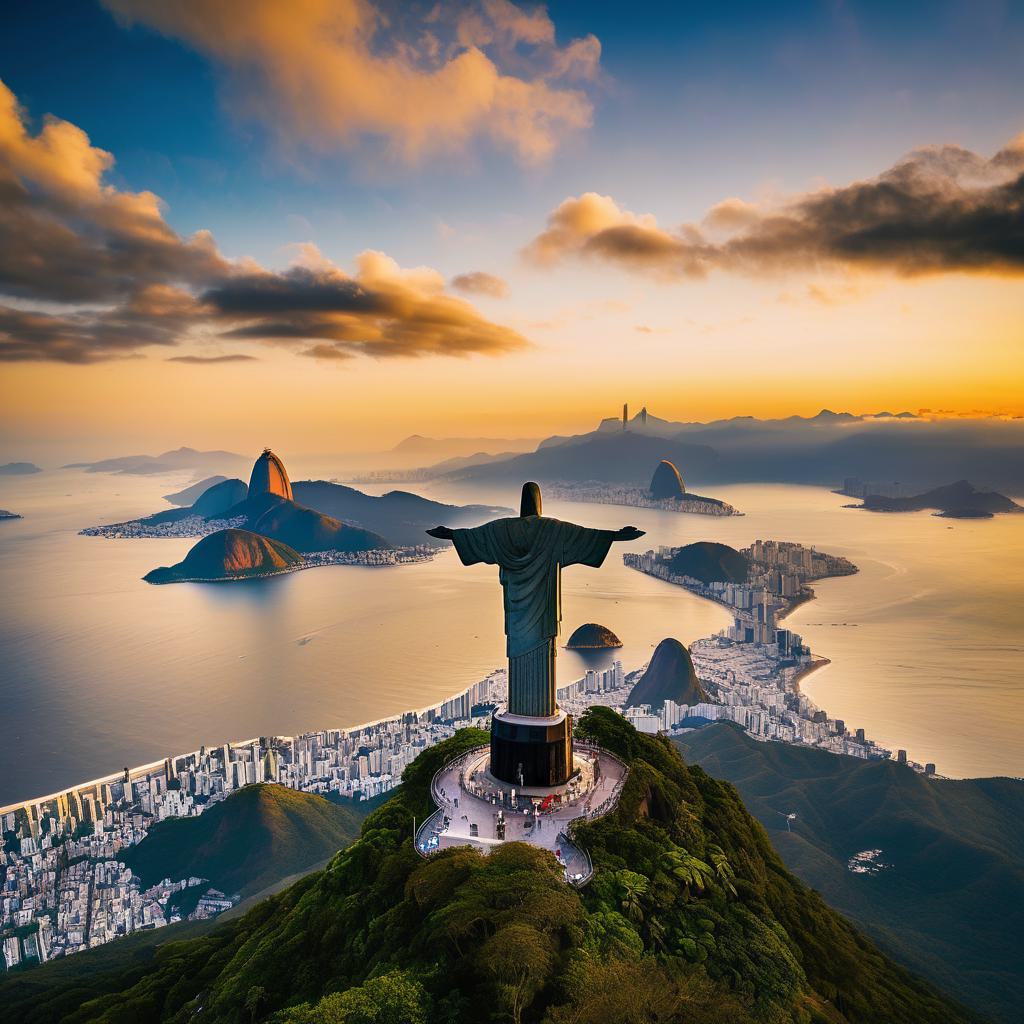} &
        \includegraphics[width=0.07\textwidth]{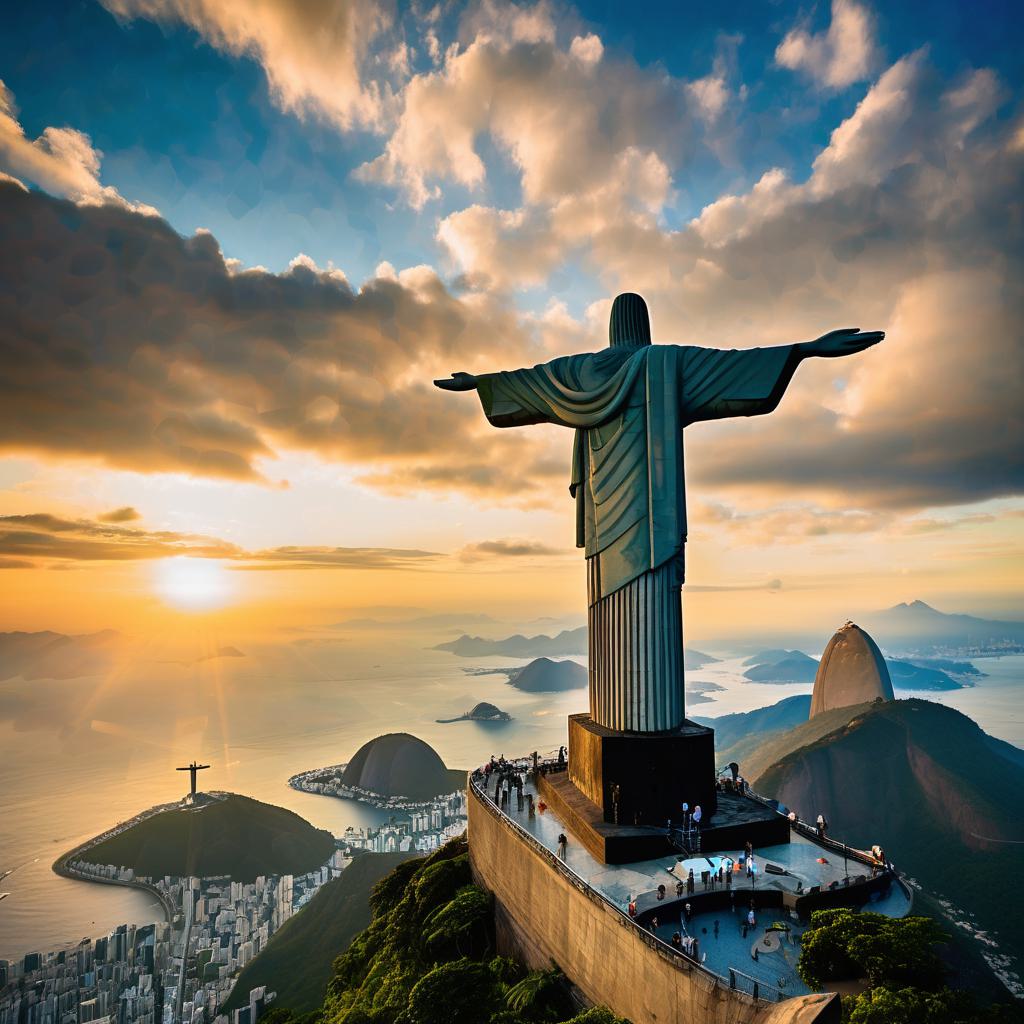} &
        \includegraphics[width=0.07\textwidth]{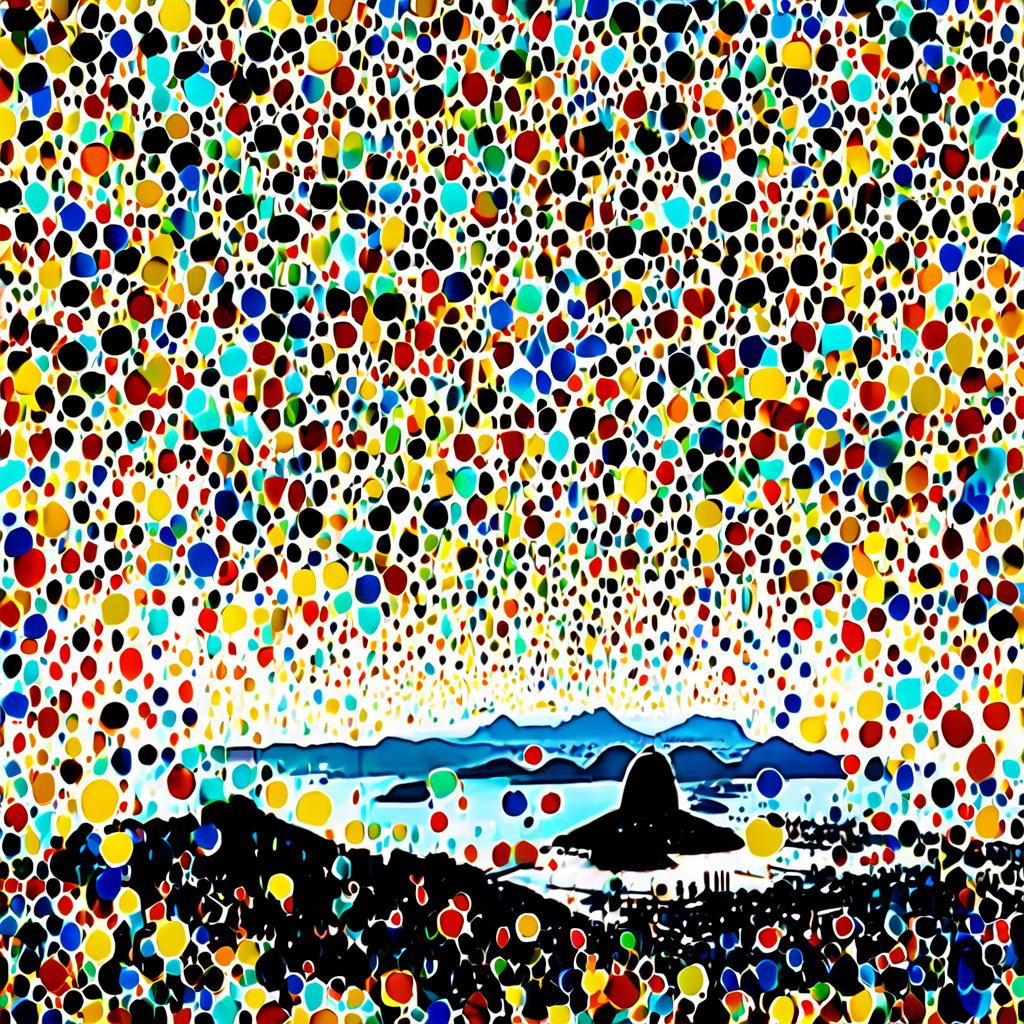} &
        \includegraphics[width=0.07\textwidth]{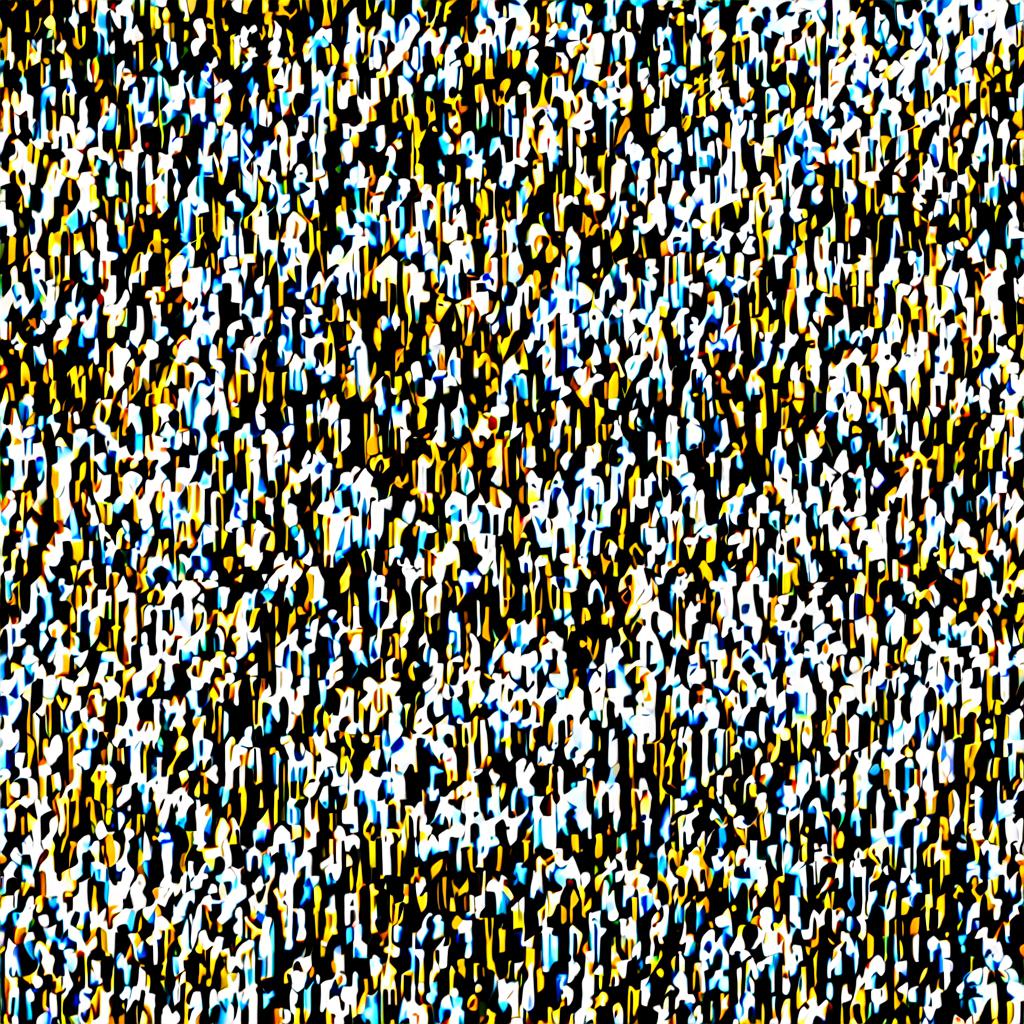} &
        \includegraphics[width=0.07\textwidth]{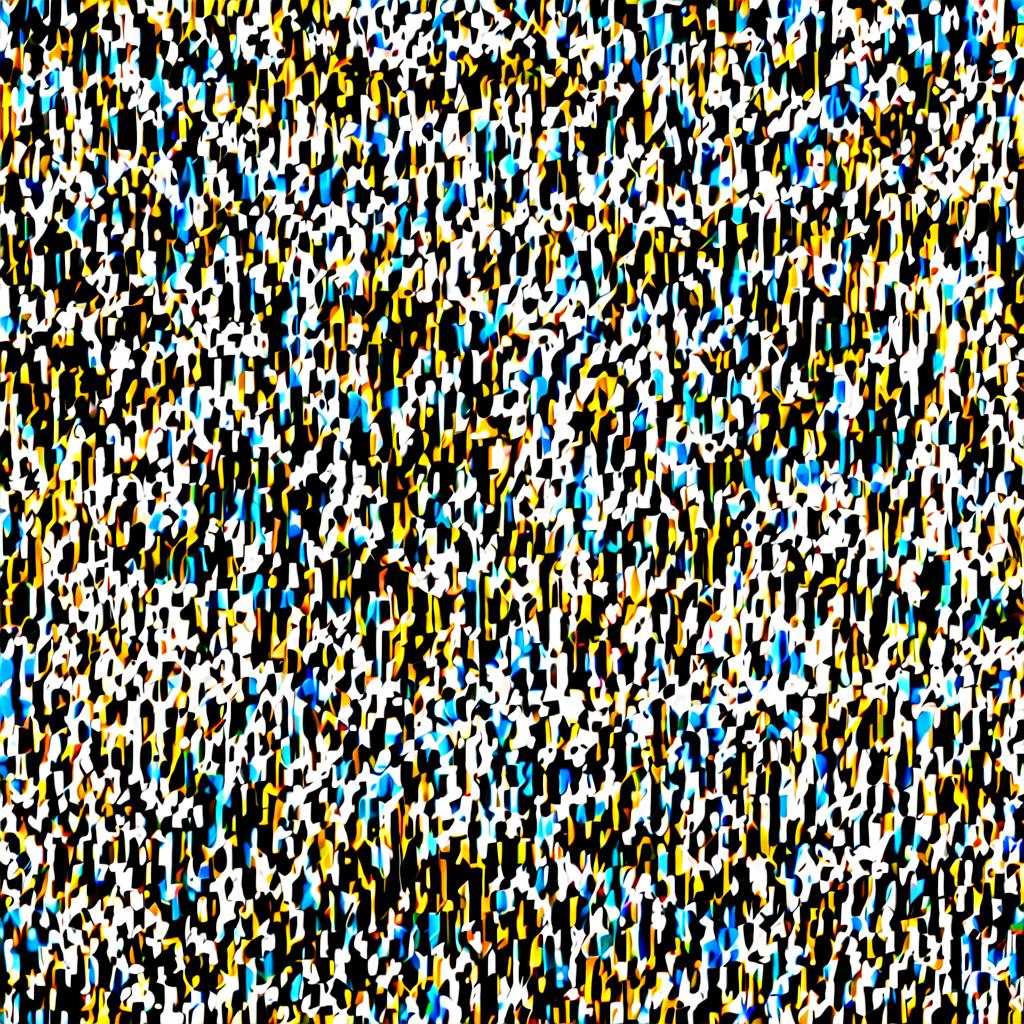} &
        \includegraphics[width=0.07\textwidth]{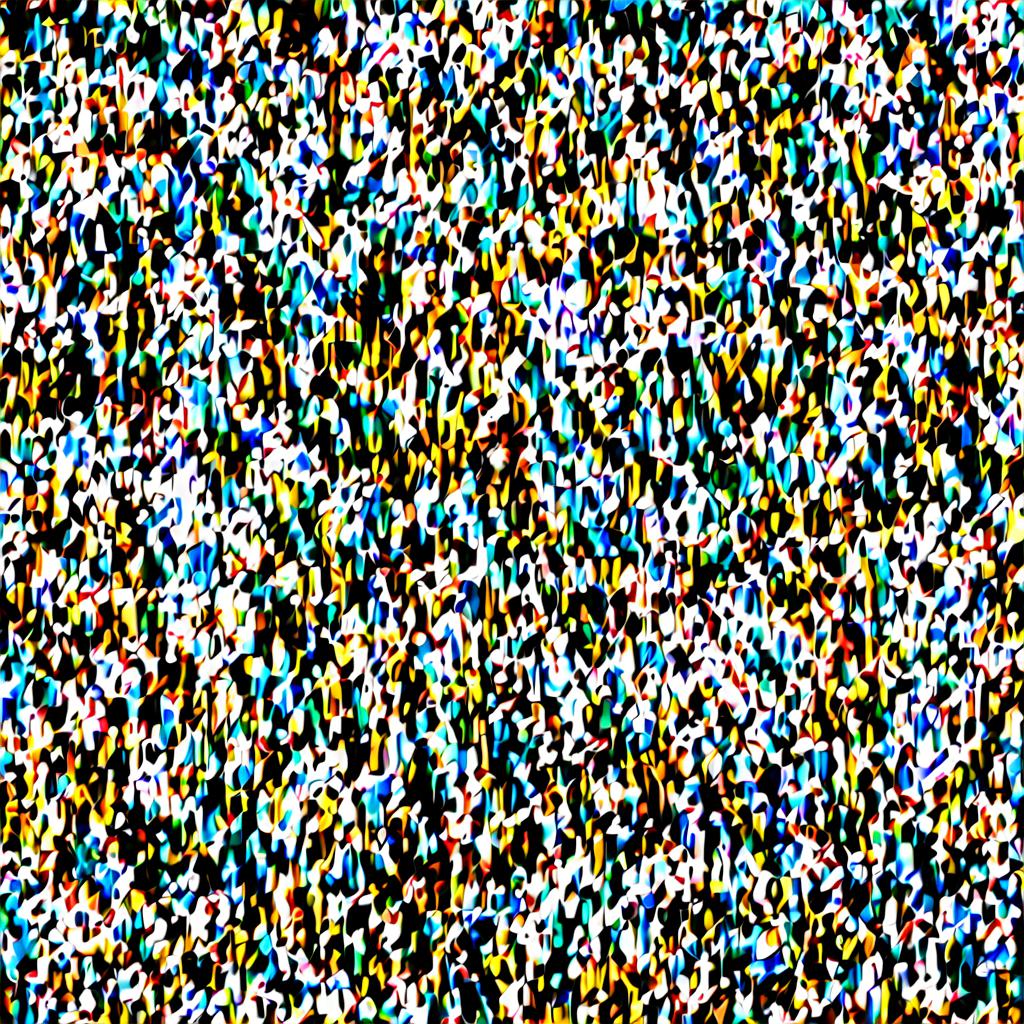} \\

        \textbf{6.4} & 
        \includegraphics[width=0.07\textwidth]{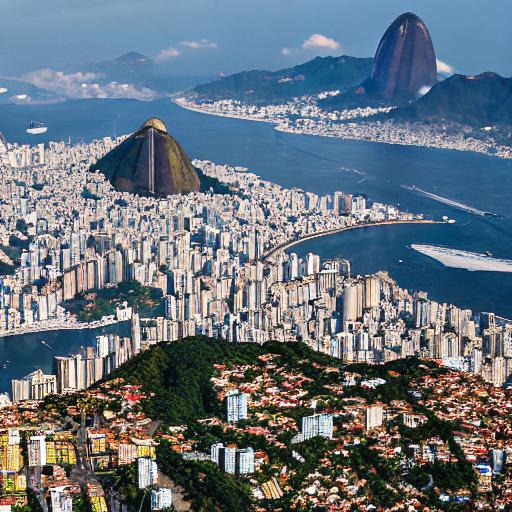} &
        \includegraphics[width=0.07\textwidth]{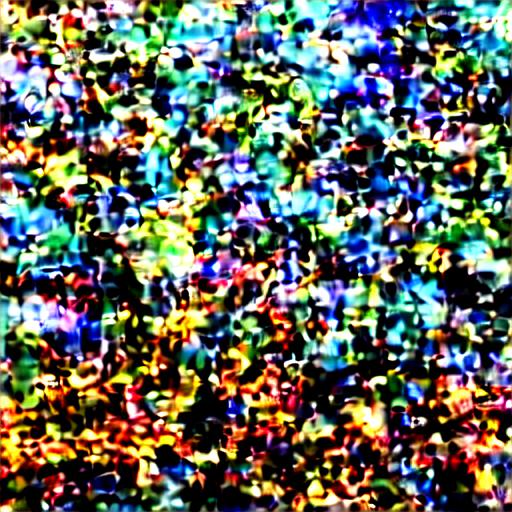} &
        \includegraphics[width=0.07\textwidth]{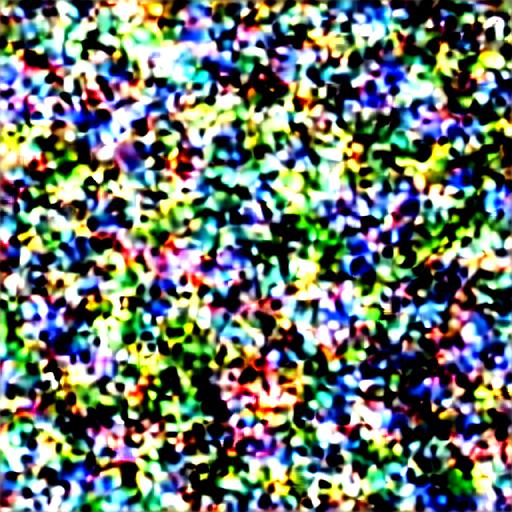} &
        \includegraphics[width=0.07\textwidth]{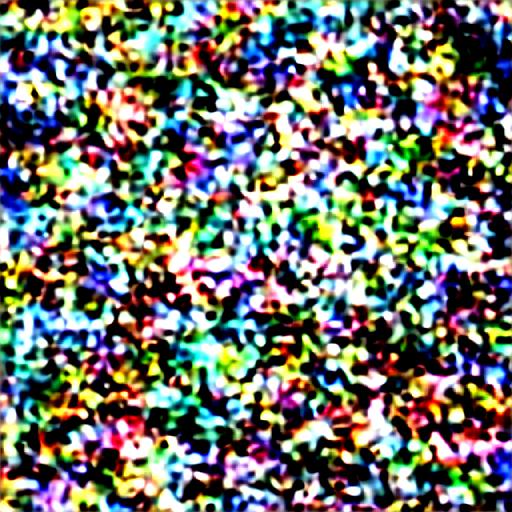} &
        \includegraphics[width=0.07\textwidth]{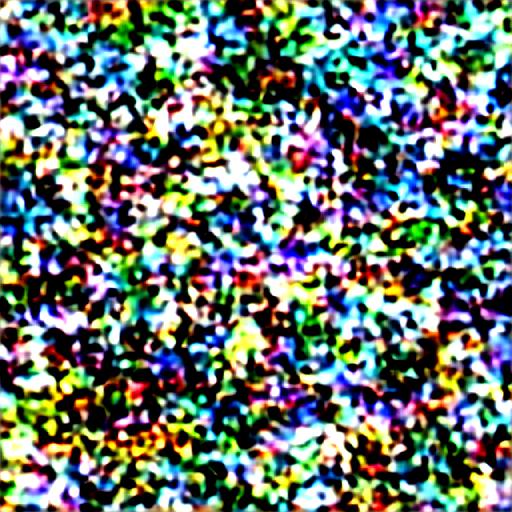} &
        \includegraphics[width=0.07\textwidth]{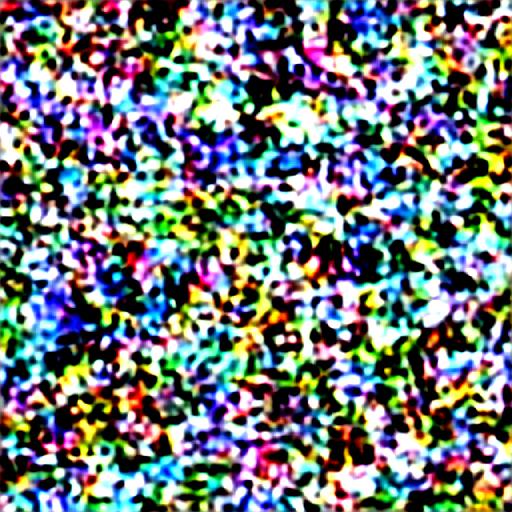} &
        \includegraphics[width=0.07\textwidth]{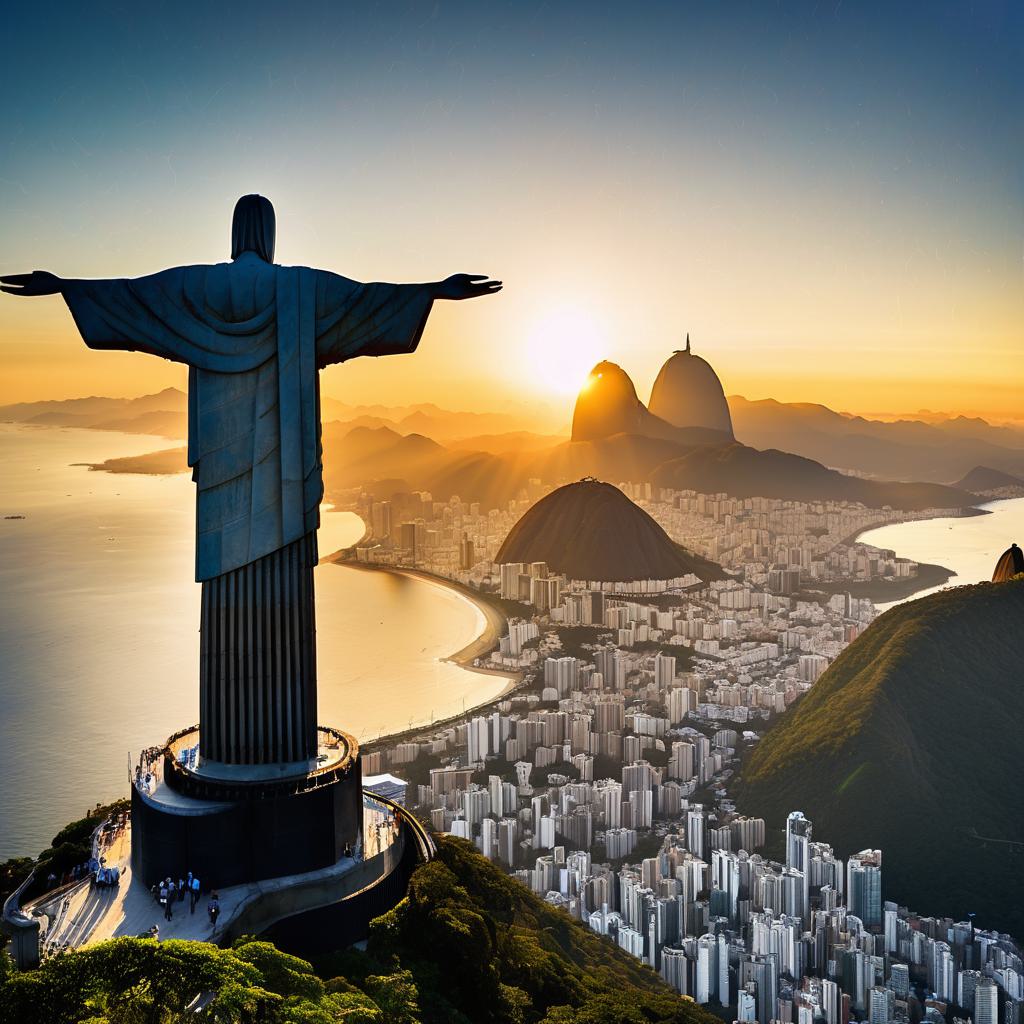} &
        \includegraphics[width=0.07\textwidth]{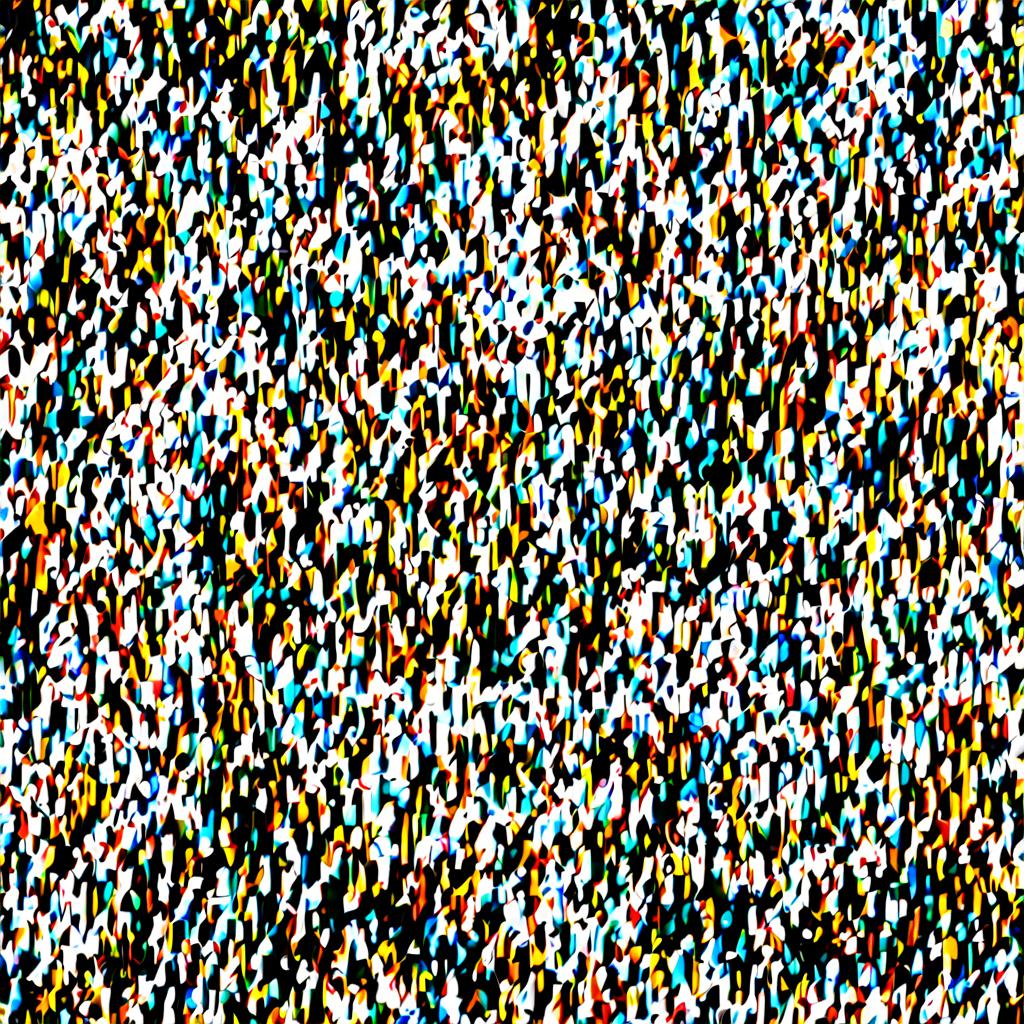} &
        \includegraphics[width=0.07\textwidth]{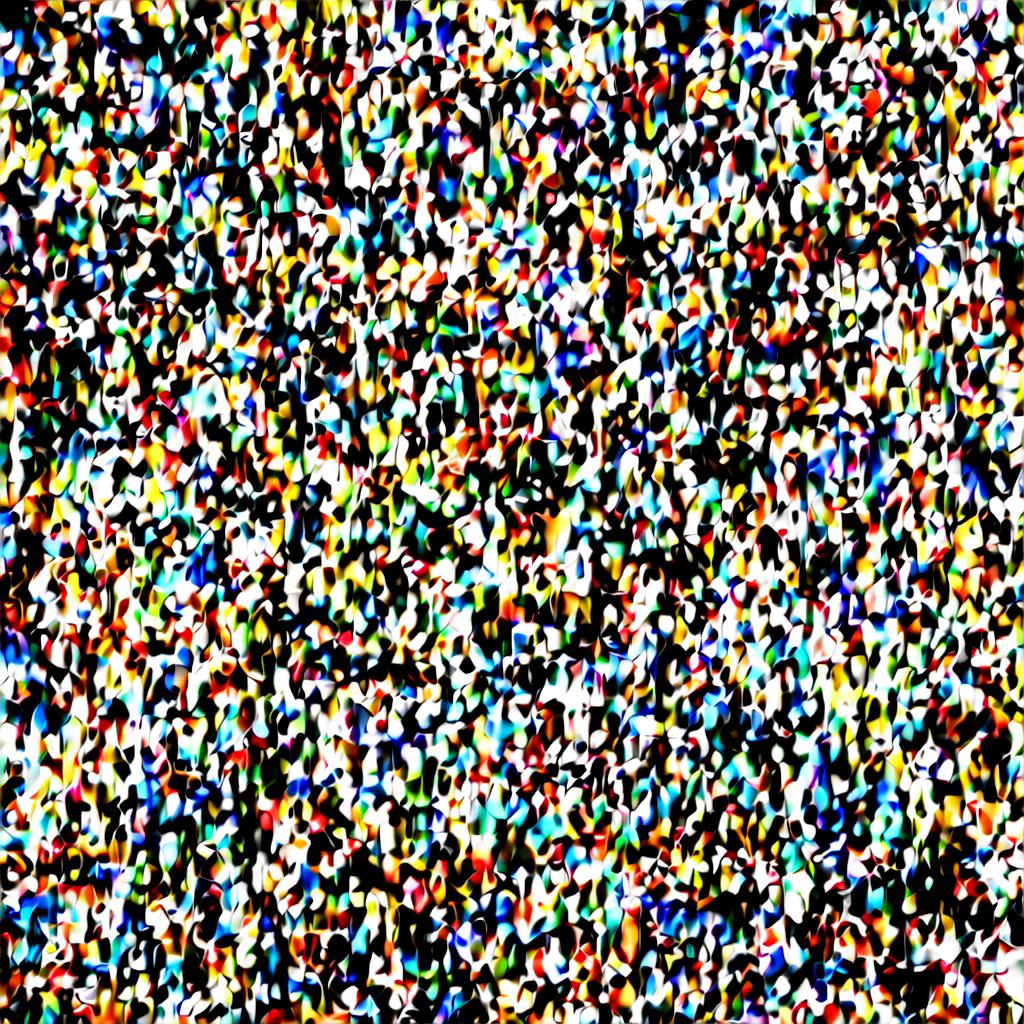} &
        \includegraphics[width=0.07\textwidth]{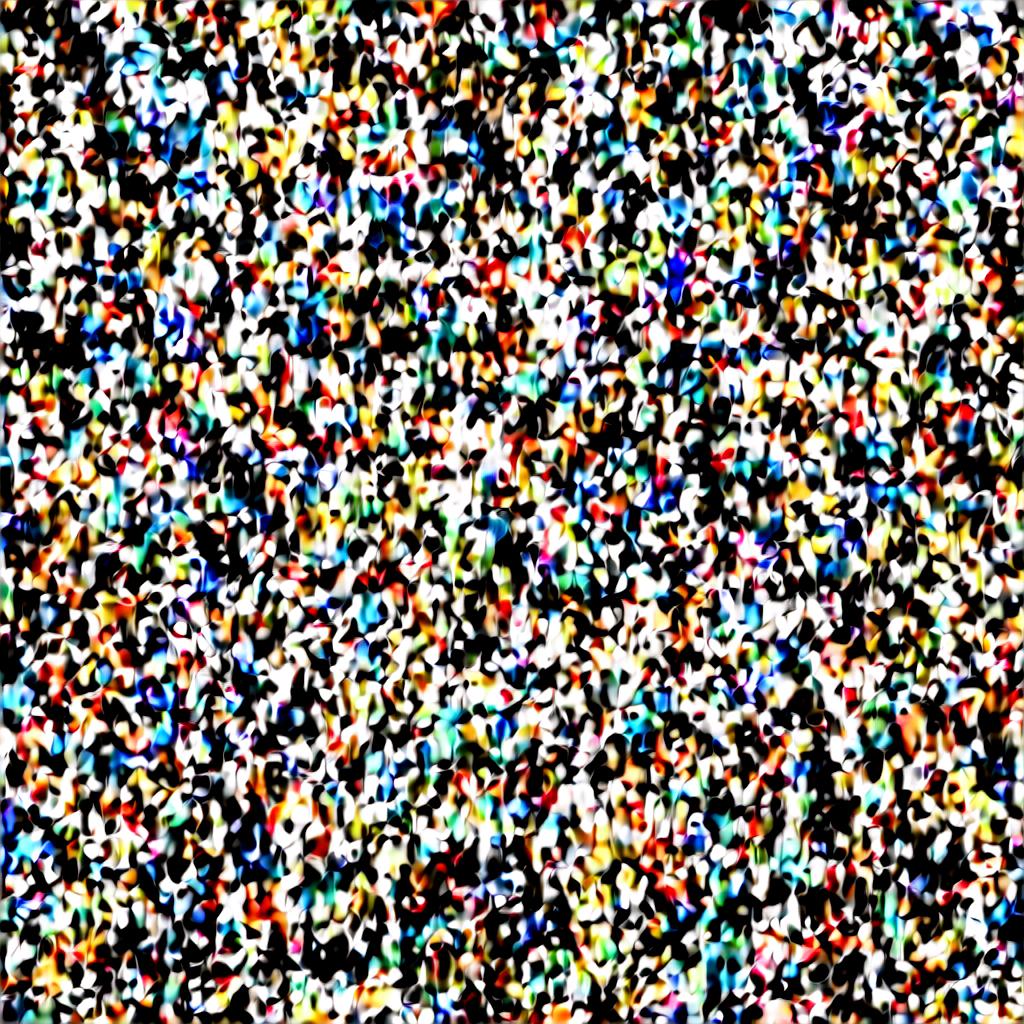} &
        \includegraphics[width=0.07\textwidth]{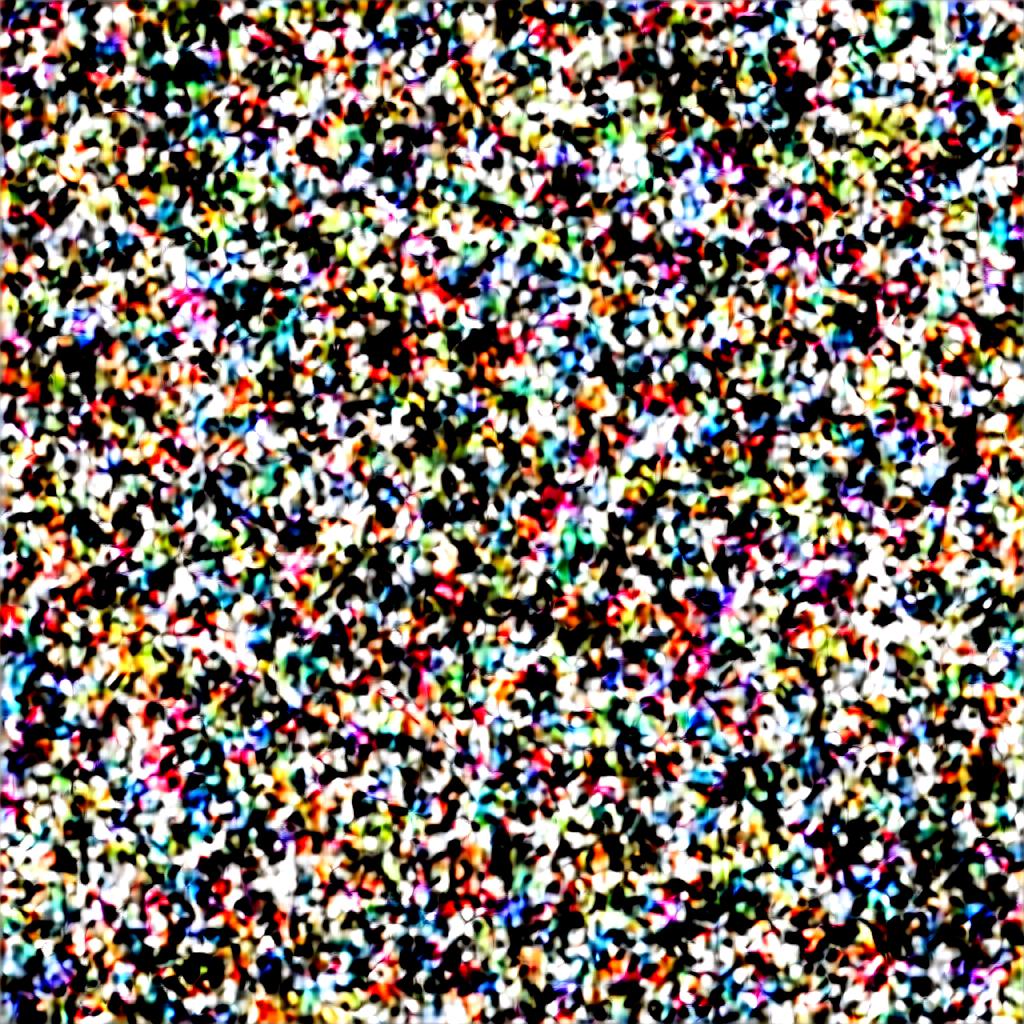} &
        \includegraphics[width=0.07\textwidth]{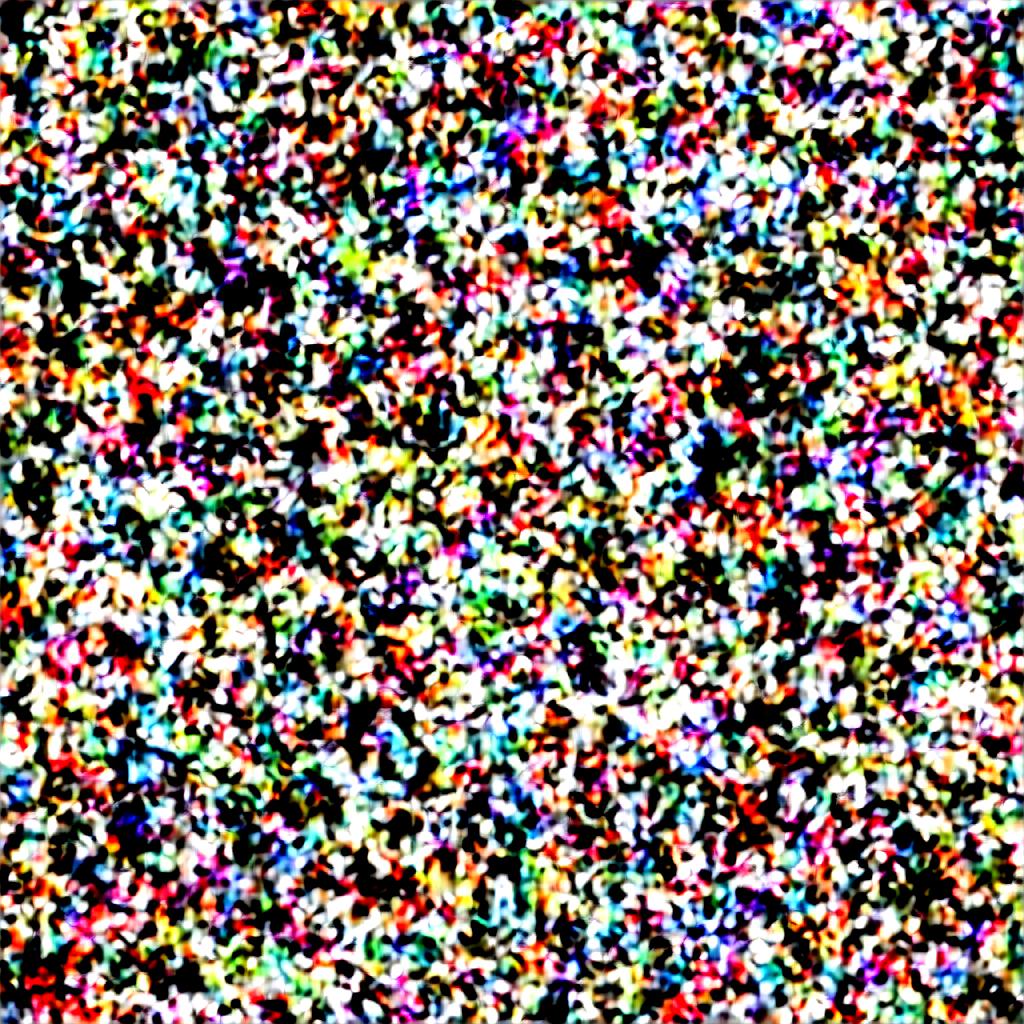}
    \end{tabular}

    \vspace{4pt}
    \captionsetup{font=small}
    \caption{\textbf{TRS perturbation landscape.} Interaction between perturbation length (rows) and masking probability (columns) for SD1.5 and SDXL-Lightning. For both we used the prompt: "A breathtaking view from behind the Cristo Redentor (Christ the Redeemer) statue in Rio de Janeiro, Brazil, with layered mountains stretching into the distance and the sparkling Atlantic Ocean clearly visible below with blue color; warm golden-hour light, atmospheric haze, ultra-detailed, cinematic, wide-angle landscape, beautiful and serene and the city below."}
    \label{fig:trs_comparison_grid}
\end{figure}

\subsection{Warm-Up}
\label{sec:app:ablations:warm-up}

\paragraph{Setup.} 
We investigate the impact of the warm-up phase on performance to determine the optimal allocation of the total compute budget between initialization and the main optimization process. This experiment uses Stable Diffusion 1.5 on the subset of 55 DrawBench prompts, equally distributed across all categories. We evaluate three total evaluation budgets ($N_{\text{total}} \in \{120, 360, 720\}$) with a fixed batch size of $B=24$.

\paragraph{Metrics.} 
To assess the effectiveness of different budget allocations, we utilize ImageReward as the reward function. We report the final mean best rewards achieved across all 55 prompts, tracking how the optimization efficiency changes as the ratio of warm-up to main optimization varies.

\paragraph{Results.} 
As shown in \Cref{fig:ablations_combined}a, TRS achieves peak performance when the warm-up phase accounts for $10\%$ to $20\%$ of the total compute budget across all tested settings. We note that while performance is sensitive to very small warm-up budgets, it degrades only slowly when the warm-up phase is slightly larger than the optimal range. This suggests that the exact choice is not overly critical, provided the initialization period is sufficient; consequently, we set the warm-up fraction to $20\%$ for all main experiments.

\begin{figure}[t]
\centering
\begin{tikzpicture}
\begin{groupplot}[
    group style={
        group size=2 by 1,
        horizontal sep=1.8cm,
    },
    width=5cm,
    height=4cm,
    grid=both,
    grid style={gray!30},
    tick label style={font=\footnotesize},
    label style={font=\footnotesize},
    title style={font=\small\bfseries},
    ymin=1.3,
    ymax=1.8,
    scaled y ticks=false,
    yticklabel={\pgfmathprintnumber[fixed,precision=2]{\tick}},
]

\nextgroupplot[
    title={(a) Warm-up Fraction},
    xlabel={Warmup Fraction (\%)},
    ylabel={Mean Best ImageReward},
    xmin=0, xmax=90,
    xtick={0,20,40,60,80},
    legend to name=sharedlegend,
    legend columns=3,
    legend style={font=\small, draw=none},
]
\addplot[very thick, color=method1color, mark=*, mark size=1.2pt] coordinates {(20, 1.4983) (40, 1.4921) (60, 1.4724) (80, 1.4185)};
\addlegendentry{Budget 120}
\addplot[very thick, color=method2color, mark=*, mark size=1.2pt] coordinates {(7, 1.635) (13, 1.6505) (20, 1.6237) (30, 1.6378) (53, 1.6339) (73, 1.57975)};
\addlegendentry{Budget 360}
\addplot[very thick, color=method3color, mark=*, mark size=1.2pt] coordinates {(3.3, 1.64425) (10, 1.67043) (20, 1.70452) (30, 1.69788) (50, 1.69501) (70, 1.66783)};
\addlegendentry{Budget 720}

\% ---------- First subplot (Image) ----------
\nextgroupplot[
    title={Image (SD1.5 \& IR)},
    xtick={0,1,2,3,4,5,6},
    xticklabels={1,2,4,8,12,16,24}, %
    xmin=-0.5,
    xmax=6.5,
    ymin=1.55,
    ymax=1.68,
    ybar,
    bar width=0.7,
]

\addplot[ultra thick, fill=method2color, draw=method2color]
coordinates {
    (0, 1.58582)
    (1, 1.61497)
    (2, 1.64558)
    (3, 1.6298)
    (4, 1.6377)
    (5, 1.6578)
    (6, 1.6327)
};

\end{groupplot}

\node at ($(group c1r1.south)!0.5!(group c2r1.south) + (0,-1.5cm)$) {\pgfplotslegendfromname{sharedlegend}};

\end{tikzpicture}
\caption{\textbf{Ablation Studies.} (a) Impact of the warm-up phase fraction across different compute budgets. (b) Sensitivity of TRS to the number of regions $k$. Performance remains stable across a wide range of $k$ values.}
\label{fig:ablations_combined}
\end{figure}

\subsection{Number of Regions}
\label{sec:app:ablations:n_regions}

\paragraph{Setup.} 
We ablate the impact of the number of trust regions $k \in \{1, 2, 4, 8, 16, 24\}$ on the optimization performance for text-to-image generation. These experiments use SD1.5 on a 55-prompt subset of DrawBench, utilizing a fixed batch size of $B=24$. The optimization process is partitioned into 3 warm-up iterations followed by 12 trust-region iterations.

\paragraph{Metrics.} 
Consistent with our previous ablations, we align the optimization to ImageReward. We evaluate the final mean best rewards across the prompt subset to identify the optimal range for $k$ and to assess how the number of regions interacts with the total evaluation budget and batch size.

\paragraph{Results.} 
As illustrated in \Cref{fig:ablations_combined}b, using a very small number of regions leads to sub-optimal performance. However, we find that selecting $k$ in a medium range between $1$ and the batch size $B$ consistently yields strong results. These findings suggest that $k$ is not a highly sensitive hyperparameter, as performance remains stable across a wide range of values, indicating that it does not require extensive per-task tuning. Since $k$ is the only hpyerparameter that was changed between the main experiments in Section~\ref{sec:benchmarks}, we conclude that TRS is a highly robust algorithm for noise optimization of flow and diffusion models.

\section{ODE vs. SDE Sampling Dynamics}
\label{sec:app:proteina_discussion}

TRS is a black-box optimization method that relies on local exploration to find better samples. This approach is more effective with ODE sampling because the deterministic paths provide a clear sense of locality. However, since SDE sampling with noise reduction leads to higher designability in modern generative 3D protein models \citep{geffner2025proteina, bose2024sestochastic}, we want to see how our algorithm handles the trade-off between the designability gains of SDEs and the diversity preserved by ODEs. We also include experiments for text-to-Image (T2I) generation to see if these sampling dynamics remain consistent across different model types.

\paragraph{Setup.} 
We compare TRS against standard random search (RS) in both deterministic and stochastic settings. For protein generation, we generate 100 proteins of length 50, similar to Section~\ref{sec:exp:protein_design} optimizing specifically for designability. For T2I generation, we use the same setup as in Section~\ref{sec:exp:text-to-image}, utilizing the DDIM scheduler with both a stochastic setting ($\eta = 1.0$) and a deterministic setting ($\eta = 0.0$). A key difference between these setups is the inference budget: proteina uses 400 steps, while the T2I model uses 50.

\paragraph{Metrics.} 
We evaluate the protein results based on designability, diversity, and novelty and the Rank-1 percentage and show the results in \Cref{tab:app:ode_vs_sde}. With this last metric, we want to show not only into how many clusters the proteins fall in this analysis, but also what the distribution between those clusters is. Rank-1 is the percentage of proteins that fall into the most dominant cluster. 
In more detail \Cref{fig:app:cluster_dist} shows the distribution across the 10 most dominant clusters using rank–size and cumulative mass plots to show how concentrated the generated samples are. 

\paragraph{Results and Discussion.} 
As shown in  \Cref{tab:app:ode_vs_sde}, SDE sampling does lead to higher designability, but it significantly diminishes diversity and novelty. 
When we compare TRS with ODE to random search and TRS with SDE ($\gamma=0.6$), we observe that the designability metrics are on par, while TRS with ODE shows much better diversity and novelty metrics. This becomes especially clear when observing the Rank-1 metric, where we see that SDE-optimized samples, for which $54$ or even $68\%$ of the samples belong to the same cluster.
\Cref{fig:app:cluster_dist} further illustrates the mode collapse of SDE-based sampling in protein design, where we can see that in SDE-based optimization, the designable samples are highly concentrated at the few most dominant clusters. ODE-based optimization instead shows a healthier distribution, creating structures distributed at different clusters more equally.

We also find that most current SDE-based scaling methods are not practical for proteina, either because they lack necessary value estimations \citep{li2024derivative, singhal2025general} or because they are too slow without efficient batch utilization \citep{jain2025diffusion}, especially, when many inference steps are necessary and the reward function is expensive. 

Interestingly, TRS shows a smaller advantage over random search in proteina when using SDEs compared to T2I. This is likely because the high number of inference steps in proteina allows the stochastic noise to eventually override the local search signal provided by the trust region.

\begin{table}[h]
\centering
\caption{\textbf{ODE vs.\ SDE sampling.} Comparison of ODE and SDE variants for random search (RS) and our trust-region search (TRS) in protein design and text-to-image (T2I). ImageReward and designability are the respective alignment rewards and Rank-1, Cluster Divergence, pairwise TM-Score and PDB Novelty are additional metrics to uncover the effects of SDE vs. ODE based alignment with the 400-step proteina model. In T2I we use the 50-step SD1.5 as generative model.}
\renewcommand{\arraystretch}{1}
\setlength{\tabcolsep}{1.67pt}
\small
\begin{tabular}{@{} l c cccccc @{}}
\toprule
& \textbf{T2I} & \multicolumn{5}{c}{\textbf{Protein (SDE with $\gamma=0.6$})} \\
\cmidrule(lr){2-2} \cmidrule(l){3-7}
\textbf{Algorithm} & {ImageReward$\uparrow$} & {Design.$\uparrow$} & {Rank-1 $\downarrow$} & {Clust. Div.$\uparrow$} & {TM Div.$\downarrow$} & {Nov.$\downarrow$} \\
\midrule
RS + ODE         & 1.43          & 0.53          & \underline{37} &    \underline{0.27}          & 0.68          & \textbf{0.85} \\
RS + SDE         & 1.45          & \textbf{0.66} & 68 & 0.19          & 0.73          & 0.94          \\
TRS (Ours) + ODE & \textbf{1.62} & \underline{0.65} & \textbf{18} & \textbf{0.30} & \textbf{0.67} & \underline{0.89} \\
TRS (Ours) + SDE & \underline{1.52} & \textbf{0.66} & 54 & \underline{0.27} & \underline{0.70} & 0.92          \\
\bottomrule
\end{tabular}
\label{tab:app:ode_vs_sde}
\end{table}

\begin{figure}[h]
\centering
\begin{tikzpicture}
\begin{groupplot}[
    group style={
        group size=2 by 1,
        horizontal sep=1.5cm,
    },
    width=6.2cm,
    height=5cm,
    grid=both,
    grid style={gray!20},
    tick label style={font=\small},
    label style={font=\small},
    title style={font=\small\bfseries},
    every axis plot/.append style={line width=1.2pt, mark size=1.8}
]

\nextgroupplot[
    title={Cluster Rank--Size Distribution},
    xlabel={Cluster rank},
    ylabel={Cluster size},
    ymode=log,
    log ticks with fixed point,
    xmin=0.5, xmax=10.5,
    xtick={1,2,3,4,5,6,7,8,9,10},
    ymin=0.8, ymax=100,
    ytick={1,2,5,10,20,50,100},
    legend style={at={(1.1,-0.25)}, anchor=north, legend columns=2, font=\small, draw=none}
]

\addplot[color=method1color, mark=*, xshift=-0.5pt]
    coordinates {(1,37) (2,17) (3,7) (4,5) (5,3) (6,3) (7,3) (8,2) (9,2) (10,2)};
\addlegendentry{RS + ODE}

\addplot[color=method2color, mark=square*, xshift=0.5pt]
    coordinates {(1,68) (2,7) (3,5) (4,4) (5,2) (6,1) (7,1) (8,1) (9,1) (10,1)};
\addlegendentry{RS + SDE}

\addplot[color=method3color, mark=triangle*, xshift=-1pt]
    coordinates {(1,18) (2,16) (3,14) (4,7) (5,6) (6,5) (7,4) (8,4) (9,3) (10,2)};
\addlegendentry{TRS + ODE}

\addplot[color=method4color, mark=diamond*, xshift=1pt]
    coordinates {(1,54) (2,6) (3,5) (4,5) (5,3) (6,2) (7,2) (8,2) (9,2) (10,2)};
\addlegendentry{TRS + SDE}

\nextgroupplot[
    title={Cumulative Cluster Mass},
    xlabel={Fraction of clusters},
    ylabel={Cumulative fraction},
    ymin=0, ymax=1.05,
    xmin=0, xmax=1,
    ytick={0,0.25,0.5,0.75,1.0}
]

\addplot[color=method1color, mark=none] coordinates {(0.037,0.37) (0.074,0.54) (0.111,0.61) (0.222,0.72) (1.0,1.0)};
\addplot[color=method2color, mark=none] coordinates {(0.053,0.68) (0.105,0.75) (0.158,0.80) (0.263,0.86) (1.0,1.0)};
\addplot[color=method3color, mark=none] coordinates {(0.033,0.18) (0.067,0.34) (0.133,0.55) (0.233,0.70) (1.0,1.0)};
\addplot[color=method4color, mark=none] coordinates {(0.037,0.54) (0.074,0.60) (0.148,0.70) (0.259,0.77) (1.0,1.0)};

\end{groupplot}
\end{tikzpicture}
\caption{\textbf{Protein cluster distributions.} The rank--size plot (left) shows that ODE methods maintain higher diversity, whereas SDE methods suffer from mode-collapse. The cumulative mass (right) confirms that TRS + ODE provides the most balanced distribution of samples.}
\label{fig:app:cluster_dist}
\end{figure}

\section{Baselines}
\label{sec:app:baselines}

In this section, we provide more detail about the baselines and explain their hyperparameters. The exact numbers used in the experiments are shown in \Cref{tab:hyperparameters}.

\subsection{Gradient-based guidance}
\label{sec:app:baselines:grad}

\paragraph{Optimal Control Flow (OC-Flow).} OC-Flow~\citep{wang2024training} provides a theoretically grounded, training-free framework for guided flow matching by framing generation as an optimal control problem. It augments pre-trained flow dynamics with a time-dependent control term $\mathbf{u}_t$: 
$\dot{\mathbf{x}}_t = \mathbf{v}_t(\mathbf{x}_t) + \mathbf{u}_t.$
The framework seeks to minimize a cost functional $J(\mathbf{u}) = R(\mathbf{x}_1) + \int_0^1 \frac{1}{2\lambda} \|\mathbf{u}_t\|^2 dt$, comprising a terminal reward loss $R(\mathbf{x}_1)$ and a quadratic running cost that regulates the trajectory's deviation from the prior distribution. Leveraging Pontryagin's Maximum Principle, the control trajectory is optimized iteratively over a fixed number of steps using either SGD or L-BFGS. In each iteration, gradient information is propagated backward via a co-state flow $\boldsymbol{\mu}_t$ to update the control parameters with a step size $\eta$ and weight decay. To ensure trajectory regularity and stable convergence, a weight constraint is enforced on the magnitude of the control term throughout the optimization process.

\subsection{Noise sequence search}
\label{sec:app:baselines:grad_noise_seq}

\textbf{Diffusion Tree Sampling (DTS).} DTS~\citep{jain2025diffusion} frames the inference-time alignment of diffusion models as a tree-structured optimization problem over the denoising sequence. The framework employs a recursive value-based search guided by a soft value function $V(\mathbf{x}_t)$, which is estimated via a soft-Bellman backup with an exploration constant $\lambda$:
\[
V(\mathbf{x}_t) = \frac{1}{\lambda} \log \mathbb{E}_{\mathbf{x}_{t-1} \sim p_\theta(\cdot|\mathbf{x}_t)} \left[ \exp(\lambda V(\mathbf{x}_{t-1})) \right]
\]
At each node, the search proceeds for a fixed number of expansion steps, where the selection of trajectories is governed by a specific exploration type (e.g., UCT). To effectively navigate the continuous branching space of the diffusion process, the algorithm utilizes progressive widening to determine the number of children $k$ for a node $\mathbf{x}_t$ based on its visit count $N(\mathbf{x}_t)$:
\[
k(\mathbf{x}_t) = \lceil C \cdot N(\mathbf{x}_t)^\alpha \rceil
\]
where $C$ is the progressive width constant and $\alpha$ is the progressive width exponent. A stochastic rollout parameter $\rho$ serves as a decision gate: with probability $\rho$, the algorithm performs a full rollout to the terminal state $t=0$ to obtain an exact reward, while with probability $1-\rho$, it continues recursive tree expansion.

\paragraph{Fast Direct.}
Fast Direct ~\citep{tan2025fast} is a trajectory-level optimization method designed for black-box guidance and optimizes the complete noise sequence $\{\boldsymbol{\epsilon}_t\}_{t=1}^T$ simultaneously. The method identifies a pseudo-target $\mathbf{x}^*$ on the manifold via Gaussian Process (GP) regression over previous evaluations and computes a universal direction to update the entire sequence:
\[
    \boldsymbol{\epsilon}_t^{\text{new}} = \text{Norm}(\boldsymbol{\epsilon}_t^{\text{old}} + \alpha(\mathbf{x}^* - \mathbf{x}_0))
\]
where $\alpha$ is the step size. This global refinement allows it to converge to high-reward regions in fewer iterations than per-step filtering methods.

\subsection{Source noise search}
\label{sec:app:baselines:source_noise}

\paragraph{Random Search.}
Random search is the simplest search-based method one can apply, where $N$ source noises $\mathbf{x}_0$ are randomly sampled and evaluated. Thus, it is often referred to as best-of-N in the literature \cite{jain2025diffusion}. This technique can be used both with ODE- and SDE-based samplers.

\paragraph{Zero-Order Search.} 
Zero-order Search~\citep{ma2025inference} can be seen as a special case of our algorithm. Like ours, it iteratively refines the initial latent noise to maximize a target reward. The algorithm begins by sampling an initial set of $B$ Gaussian noise vectors $\{ \mathbf{x}_0^j \}_{j=1}^B$ and selecting the candidate that yields the highest reward as the initial center $\mathbf{x}_0^{\mathrm{c}}$. This is equivalent to using only one warm-up iteration in TRS and setting the number of regions $k=1$.
In each subsequent search iteration, $B$ new candidates are generated by perturbing the current center:
\[
    \mathbf{x}_0^{\text{new}} = \mathbf{x}_0^{\mathrm{c}} + \varepsilon \cdot \boldsymbol{\delta}, \quad \boldsymbol{\delta} \sim \mathcal{N}(\mathbf{0}, \mathbf{I})
\]
where $\varepsilon$ (scalar) controls the radius of the local neighborhood search. This perturbation scheme is similar to using TRS without probability masks and choosing a fixed region length. Similar to ours, this technique works best for ODE samplers.

\paragraph{CMA-ES (diagonal).}
The Covariance Matrix Adaptation Evolution Strategy (CMA-ES)~\citep{hansen2001completely} is a state-of-the-art black-box evolution strategy that we adapt to source noise search, closely related to concurrent work on evolutionary/genetic approaches for noise optimization~\citep{jajal2025inference}. At each generation $g$, a population of $\lambda$ candidate noises is sampled from a multivariate Gaussian centered at the current mean $\mathbf{m}^{(g)}$,
\[
    \mathbf{x}_0^{j} \sim \mathcal{N}\!\left(\mathbf{m}^{(g)}, (\sigma^{(g)})^2\, \mathbf{C}^{(g)}\right), \quad j = 1, \dots, \lambda,
\]
where $\sigma^{(g)}$ is the global step size and $\mathbf{C}^{(g)}$ the covariance matrix. The $\mu$ best candidates by reward are recombined into the new mean, and both $\sigma^{(g)}$ and $\mathbf{C}^{(g)}$ are updated from the evolution path of successful steps. To remain tractable in the high-dimensional noise space, we use the \emph{diagonal} variant, which restricts $\mathbf{C}^{(g)}$ to its diagonal and thus scales linearly with the latent dimensionality while still adapting a per-coordinate step size. As with the other source noise search methods, candidates are renormalized to the noise manifold and the strategy is most effective with ODE samplers.

\section{Metrics}
\label{sec:app:metrics}

In this section, we describe all additional evaluation metrics used in this work that are not employed as reward functions. The reward-based metrics are detailed in \Cref{sec:app:experiment_details}. We consider evaluation criteria for both molecule generation and protein design.

\subsection{Molecule Generation}
\label{sec:app:metrics:molecules}

For the evaluation of 3D molecule generation, we follow the methodology described by \citep{song2023equivariant}, which assesses the physical and chemical plausibility of the generated Cartesian coordinates and atom types.

\paragraph{Stability.}
We evaluate the structural integrity of the generated samples using \textit{Atom Stability} and \textit{Molecule Stability}. Following the convention of Hoogeboom \textit{et al.} \citep{hoogeboom2022equivariant}, chemical bonds are inferred based on the Euclidean distances between atoms using a threshold-based lookup table of covalent radii. An individual atom is considered stable if its inferred bond count matches its expected chemical valency (e.g., 4 for carbon, 1 for hydrogen). \textit{Molecule Stability} is then defined as the percentage of generated molecules for which all constituent atoms are stable. This metric serves as a proxy for the geometric consistency of the generated 3D structures.

\paragraph{Validity, Uniqueness, and Novelty.}
The chemical validity of a molecule is determined by its ability to be successfully parsed into a molecular graph using RDKit \citep{Landrum2010}. We report the \textit{Valid Fraction}, the proportion of samples satisfying fundamental valency constraints. The \textit{Unique Fraction} indicates the percentage of non-redundant molecules among generated compounds. We also report Novelty, defined as the fraction of generated molecules not present in the training set. To provide a stringent quality assessment, we track the Molecule Stability Percentage (MSP) and the Valid and Unique Percentage (VUP). The former assesses the geometric consistency of the 3D coordinates, while the latter ensures the model explores the chemical space without collapsing into a few valid structures.

\subsection{Protein Design}
\label{sec:app:metrics:proteins}

To evaluate the structural quality and variability of the generated protein backbones, we adopt the metric suite introduced by \citep{geffner2025proteina}. These metrics assess two complementary aspects: the uniqueness of the generated structures relative to known proteins (\textit{Novelty}) and the structural variability within the generated set (\textit{Diversity}). Following \citep{geffner2025proteina}, we compute these scores only for designable samples, which means that scRMSD $<$ 2.0 \AA{} (equivalently designability $>$ 0.1353).

\paragraph{Novelty.}
Novelty measures the extent to which the model generates protein folds that differ from those observed in existing experimental and predicted structure databases. For each generated, designable backbone structure, we compute the maximum TM-score against all entries in a reference set using Foldseek \citep{van2024fast}. We then report the average of these maximum TM-scores across the entire sample set. Lower average maximum TM-scores indicate higher structural novelty. We evaluate novelty with respect to the PDB dataset \citep{berman2000protein}.

\paragraph{Diversity.}
We quantify the internal diversity of the generated, designable backbone samples using two complementary metrics:
\begin{enumerate}
    \item \textbf{Average Pairwise TM-score:} We compute the mean pairwise TM-score between all designable backbone samples for each generated protein length. These values are subsequently aggregated to obtain a global average. Since the TM-score measures structural similarity on a scale from 0 to 1, lower values correspond to higher diversity.
    \item \textbf{Cluster Ratio:} We cluster the generated backbones using Foldseek with a TM-score threshold of 0.5. The diversity score is defined as the ratio of the number of unique clusters to the total number of designable samples. Because a more diverse set of samples results in a higher number of clusters, higher ratios indicate greater structural variety within the generated set.
\end{enumerate}

\section{Runtime Comparison}
\label{sec:app:runtime}

We analyze the runtime of the experiments described in Section~\ref{sec:benchmarks} for all considered methods on a single NVIDIA A100-SXM4-40GB. The quantitative comparison of computation time and memory usage is visualized in \Cref{fig:full_comparison}. For these benchmarks, we use the settings of ImageReward for T2I, $R_6$ for molecule generation, and protein designability for sequences of 50 residues.

We observe that DTS\textsuperscript{*} generally exhibits the highest compute time. This arises because the algorithm's sequential nature is difficult to combine with fixed-batch-size calls of the generative model. In our benchmarks, this results in DTS\textsuperscript{*} being approximately $4\times$ slower than TRS, which is naturally parallelizable.

For OC-Flow, which operates with a batch size of 1, the computation time is high for SD1.5 due to the requirement of back-propagating through 50 integration steps. Conversely, for the distilled SDXL Lightning, OC-Flow is faster but incurs substantially higher memory usage. In the molecule generation task, OC-Flow is the slowest method, while its memory footprint in this domain is lower than that of TRS and other sampling-based methods, which leverage a larger batch size of 100.

Regarding Fast Direct, we employ slightly higher NFE budgets because their iterative step-increasing mechanism makes it difficult to match an exact NFE while maintaining a constant batch size. Beyond this, their computation times and memory usage are only marginally higher than our baselines, as the overhead from their Gaussian Process (GP) is minimal. Finally, Random Search, Zero-Order Search, and TRS are the most efficient overall, as their runtime consists almost entirely of the generative model forward passes and reward function evaluations.

\begin{figure}[ht]
\centering
\resizebox{\textwidth}{!}{%
\begin{tikzpicture}[scale=0.9, transform shape]
\begin{groupplot}[
    group style={
        group size=4 by 2,
        horizontal sep=0.7cm,
        vertical sep=1.2cm,
    },
    width=4.1cm,
    height=3.6cm,
    grid style={gray!30},
    tick label style={font=\tiny},
    label style={font=\tiny},
    title style={font=\tiny},
    scaled x ticks=false,
    scaled y ticks=false,
    yticklabel={\pgfmathprintnumber[fixed,precision=0]{\tick}},
]

\nextgroupplot[ybar, bar width=8pt, ymin=0, ymax=1250, xtick={0,1,2,3,4,5}, xticklabels={}, xmin=0, xmax=6, grid=none, ymajorgrids, enlarge x limits={abs=0.4}, forget plot, title={SD1.5 + IR Runtime ($\downarrow$)}, ylabel={Time (s)}]
\addplot[fill=method1color, draw=black!70, line width=0.4pt] coordinates {(0,336)};
\addplot[fill=method2color, draw=black!70, line width=0.4pt] coordinates {(1,315)};
\addplot[fill=method3color, draw=black!70, line width=0.4pt] coordinates {(2,325)};
\addplot[fill=method6color, draw=black!70, line width=0.4pt] coordinates {(3,356)};
\addplot[fill=method5color, draw=black!70, line width=0.4pt] coordinates {(4,443)};
\addplot[fill=method4color, draw=black!70, line width=0.4pt] coordinates {(5,1168)};

\nextgroupplot[ybar, bar width=8pt, ymin=0, ymax=950, xtick={0,1,2,3,4,5}, xticklabels={}, xmin=0, xmax=6, grid=none, ymajorgrids, enlarge x limits={abs=0.4}, forget plot, title={SDXL + IR Runtime ($\downarrow$)}]
\addplot[fill=method1color, draw=black!70, line width=0.4pt] coordinates {(0,260)};
\addplot[fill=method2color, draw=black!70, line width=0.4pt] coordinates {(1,235)};
\addplot[fill=method3color, draw=black!70, line width=0.4pt] coordinates {(2,240)};
\addplot[fill=method6color, draw=black!70, line width=0.4pt] coordinates {(3,280)};
\addplot[fill=method5color, draw=black!70, line width=0.4pt] coordinates {(4,212)};
\addplot[fill=method4color, draw=black!70, line width=0.4pt] coordinates {(5,888)};

\nextgroupplot[ybar, bar width=11pt, ymin=0, ymax=140, xtick={0,1,2,3,4}, xticklabels={}, xmin=0, xmax=3.5, grid=none, ymajorgrids, enlarge x limits={abs=0.4}, forget plot, title={EquiFM + $R_6$ Runtime ($\downarrow$)}]
\addplot[fill=method1color, draw=black!70, line width=0.4pt] coordinates {(0,79.48)};
\addplot[fill=method2color, draw=black!70, line width=0.4pt] coordinates {(1,61.12)};
\addplot[fill=method3color, draw=black!70, line width=0.4pt] coordinates {(2,73.48)};
\addplot[fill=method5color, draw=black!70, line width=0.4pt] coordinates {(3,120.29)};

\nextgroupplot[
    ybar, bar width=11pt, ymin=0, ymax=800, xtick={0,1,2}, xticklabels={}, xmin=-2, xmax=3, grid=none, ymajorgrids, enlarge x limits={abs=0.4}, 
    title={Prot. $n_{\text{res}}=50$ Runtime ($\downarrow$)},
    legend style={at={(-1.5,-1.8)}, anchor=north, legend columns=6, font=\small, draw=none, /tikz/every even column/.style={column sep=5pt}}
]
\addplot[fill=method1color, draw=black!70, line width=0.4pt] coordinates {(1,643.59)}; \addlegendentry{TRS (Ours)}
\addplot[fill=method2color, draw=black!70, line width=0.4pt] coordinates {(2,642.15)}; \addlegendentry{RS}
\addplot[fill=method3color, draw=black!70, line width=0.4pt] coordinates {(3,647.96)}; \addlegendentry{ZO}
\addplot[fill=method6color, draw=black!70, line width=0.4pt] coordinates {(9,9)}; \addlegendentry{FD}
\addplot[fill=method5color, draw=black!70, line width=0.4pt] coordinates {(9,9)}; \addlegendentry{OC-Flow}
\addplot[fill=method4color, draw=black!70, line width=0.4pt] coordinates {(9,9)}; \addlegendentry{DTS*}

\nextgroupplot[ybar, bar width=8pt, ymin=0, ymax=30, xtick={0,1,2,3,4,5}, xticklabels={}, xmin=0, xmax=6, grid=none, ymajorgrids, enlarge x limits={abs=0.4}, forget plot, title={SD1.5 + HPS Mem. ($\downarrow$)}, ylabel={VRAM (GB)}]
\addplot[fill=method1color, draw=black!70, line width=0.4pt] coordinates {(0,15.9627)};
\addplot[fill=method2color, draw=black!70, line width=0.4pt] coordinates {(1,15.9678)};
\addplot[fill=method3color, draw=black!70, line width=0.4pt] coordinates {(2,15.9836)};
\addplot[fill=method6color, draw=black!70, line width=0.4pt] coordinates {(3,16.6655)};
\addplot[fill=method5color, draw=black!70, line width=0.4pt] coordinates {(4,25.4528)};
\addplot[fill=method4color, draw=black!70, line width=0.4pt] coordinates {(5,11.9)};

\nextgroupplot[ybar, bar width=8pt, ymin=0, ymax=35, xtick={0,1,2,3,4,5}, xticklabels={}, xmin=0, xmax=6, grid=none, ymajorgrids, enlarge x limits={abs=0.4}, forget plot, title={SDXL + HPS Mem. ($\downarrow$)}]
\addplot[fill=method1color, draw=black!70, line width=0.4pt] coordinates {(0,16.3748)};
\addplot[fill=method2color, draw=black!70, line width=0.4pt] coordinates {(1,16.3947)};
\addplot[fill=method3color, draw=black!70, line width=0.4pt] coordinates {(2,16.4077)};
\addplot[fill=method6color, draw=black!70, line width=0.4pt] coordinates {(3,17.2165)};
\addplot[fill=method5color, draw=black!70, line width=0.4pt] coordinates {(4,31.5986)};
\addplot[fill=method4color, draw=black!70, line width=0.4pt] coordinates {(5,13.056)};

\nextgroupplot[ybar, bar width=11pt, ymin=0, ymax=12, xtick={0,1,2,3,4}, xticklabels={}, xmin=0, xmax=3.5, grid=none, ymajorgrids, enlarge x limits={abs=0.4}, forget plot, title={EquiFM + $R_3$ Mem. ($\downarrow$)}]
\addplot[fill=method1color, draw=black!70, line width=0.4pt] coordinates {(0,9.1221)};
\addplot[fill=method2color, draw=black!70, line width=0.4pt] coordinates {(1,9.1219)};
\addplot[fill=method3color, draw=black!70, line width=0.4pt] coordinates {(2,9.1220)};
\addplot[fill=method5color, draw=black!70, line width=0.4pt] coordinates {(3,4.7719)};

\nextgroupplot[ybar, bar width=11pt, ymin=0, ymax=12, xtick={0,1,2}, xticklabels={}, xmin=0, xmax=2.5, grid=none, ymajorgrids, enlarge x limits={abs=0.4}, forget plot, title={Prot. $n_{\text{res}}=50$ Mem. ($\downarrow$)}]
\addplot[fill=method1color, draw=black!70, line width=0.4pt] coordinates {(0,8.4279)};
\addplot[fill=method2color, draw=black!70, line width=0.4pt] coordinates {(1,8.4278)};
\addplot[fill=method3color, draw=black!70, line width=0.4pt] coordinates {(2,8.4275)};

\end{groupplot}
\end{tikzpicture}%
}
\caption{\textbf{Resource efficiency benchmarks.} Comparison of runtime (top row) and peak VRAM consumption (bottom row) across methods on a single NVIDIA A100-SXM4-40GB. Arrows ($\downarrow$) indicate that lower values are preferable. We evaluate across three modalities: text-to-image (SD1.5 and SDXL), molecule generation (EquiFM), and protein design. TRS (Ours) demonstrates competitive efficiency across all benchmarks. In contrast, DTS* suffers from significantly higher runtimes due to sequential sampling constraints, and OC-Flow suffers from high memory overhead when scaling to large-scale models like SDXL.}
\label{fig:full_comparison}
\end{figure}

\section{Limitations}
\label{sec:app:limitations}

\begin{figure}[htbp]
    \centering
    \small

    \setlength{\tabcolsep}{0pt} 
    \begin{tabular}{cc @{\hspace{12pt}} cc @{\hspace{12pt}} cc}
        
        \multicolumn{2}{p{0.3\textwidth}}{\centering\scriptsize\itshape Two cats and two dogs sitting on the grass} & 
        \multicolumn{2}{p{0.3\textwidth}}{\centering\scriptsize\itshape A carrot on the left of a broccoli} & 
        \multicolumn{2}{p{0.3\textwidth}}{\centering\scriptsize\itshape A cat on the right of a tennis racket} \\
        \cmidrule(lr{12pt}){1-2} \cmidrule(lr{12pt}){3-4} \cmidrule(l){5-6}
        
        \imgwithreward{0.148\textwidth}{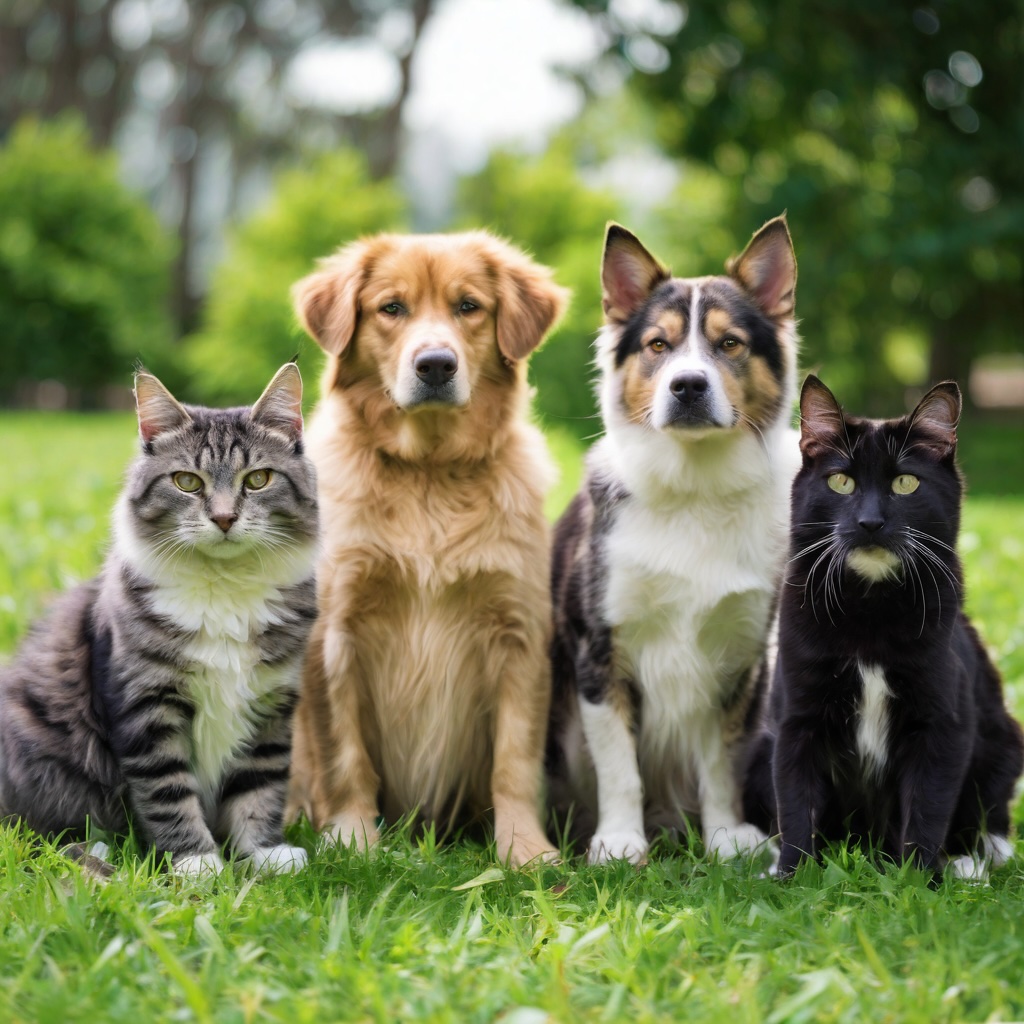}{1.86} &
        \imgwithreward{0.148\textwidth}{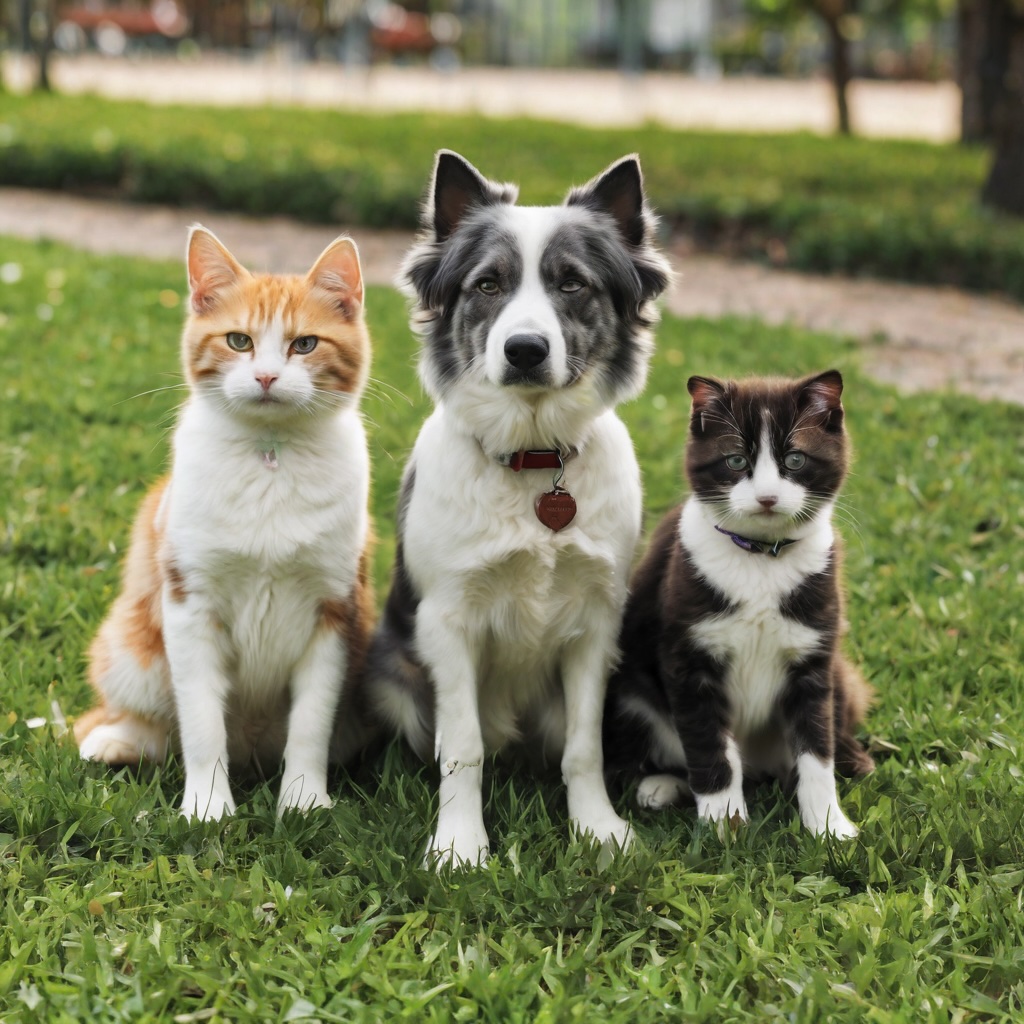}{1.90} &
        
        \imgwithreward{0.148\textwidth}{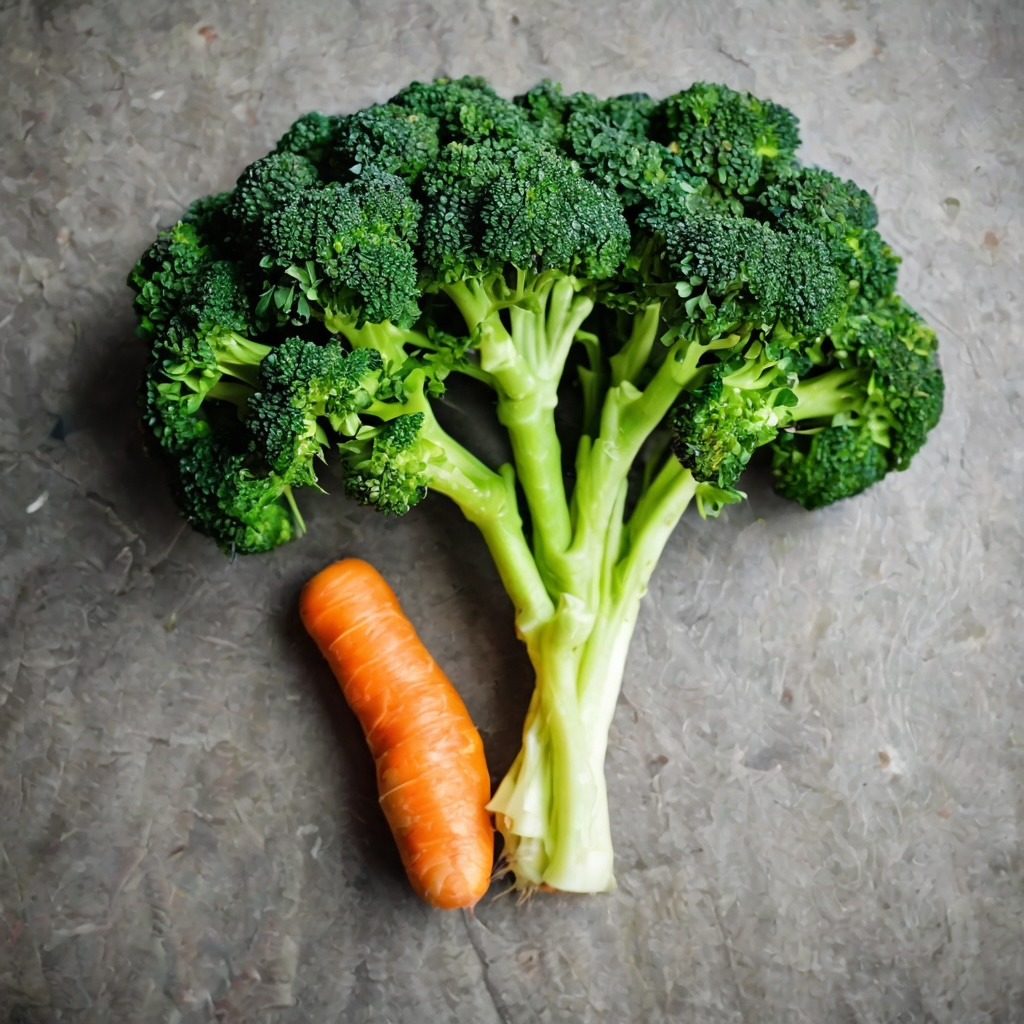}{1.53} &
        \imgwithreward{0.148\textwidth}{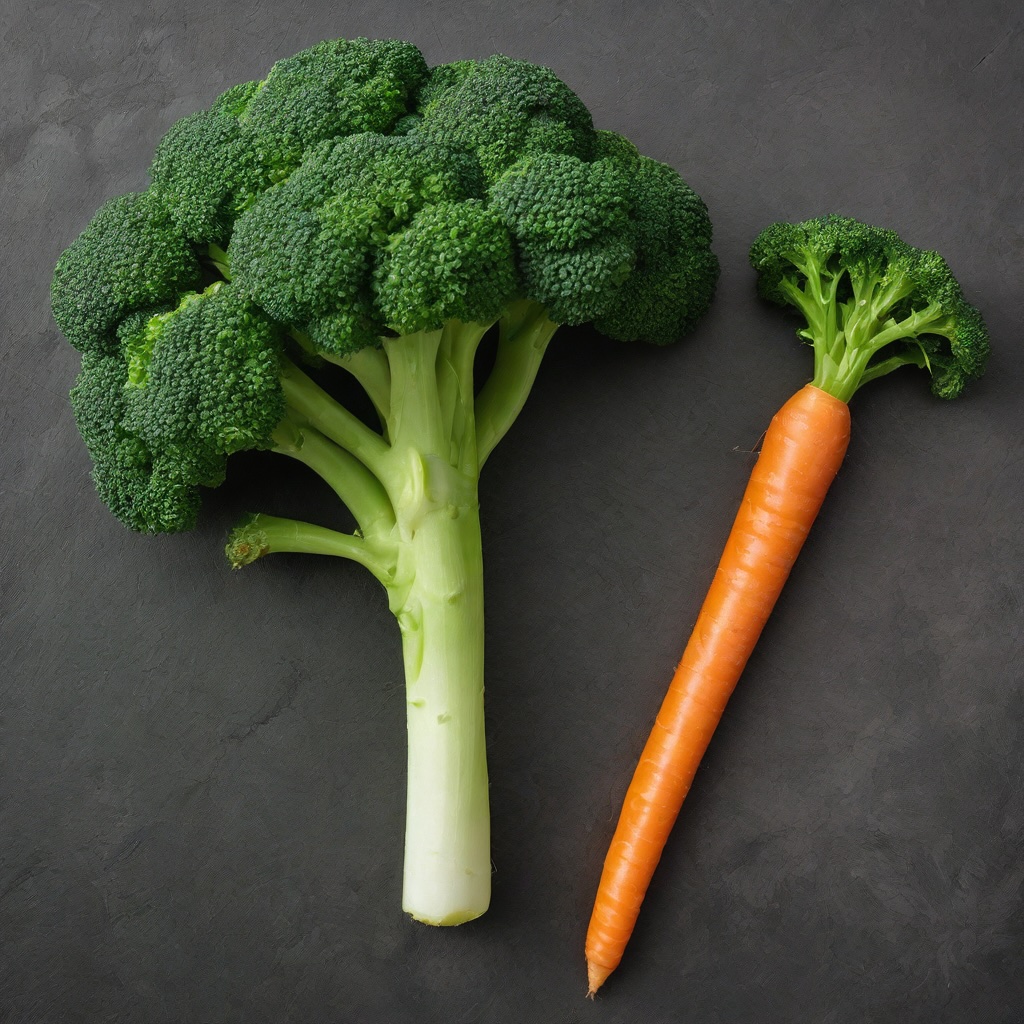}{1.57} &
        
        \imgwithreward{0.148\textwidth}{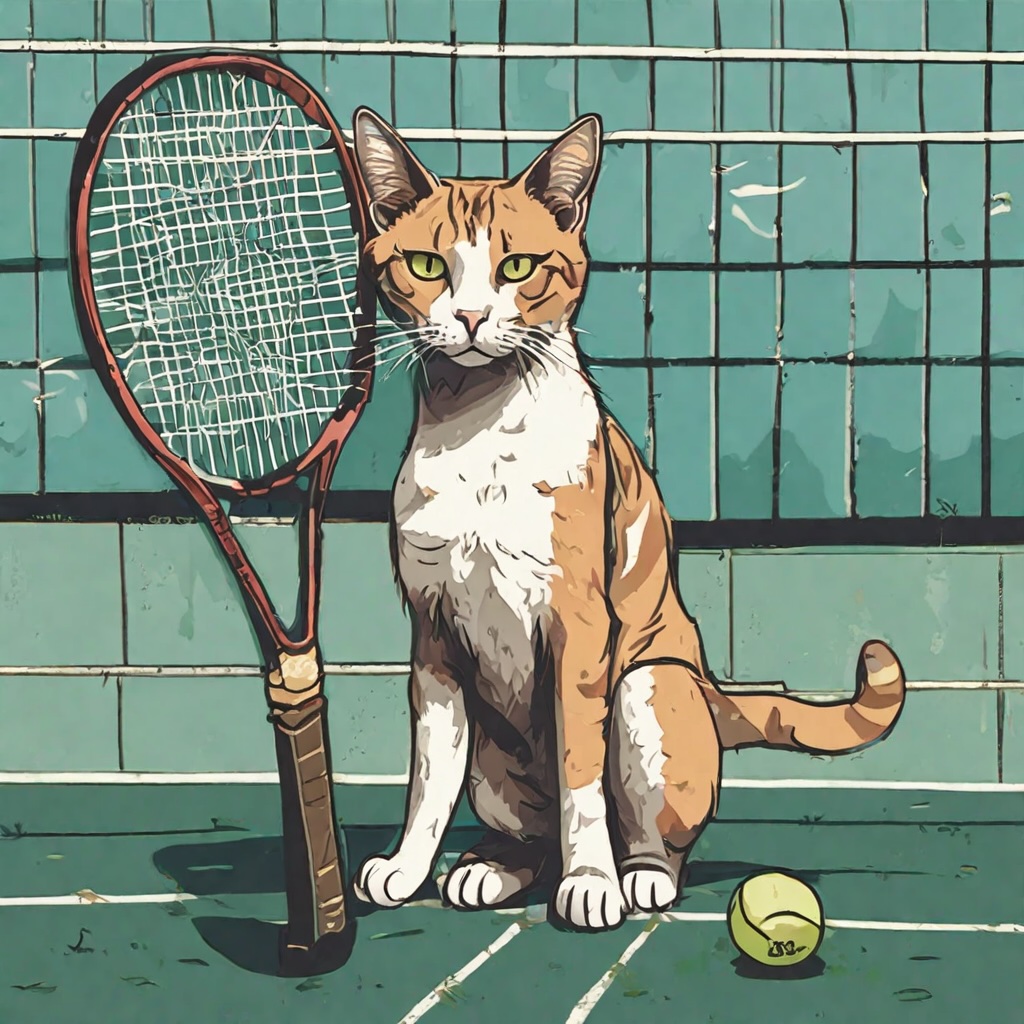}{1.83} &
        \imgwithreward{0.148\textwidth}{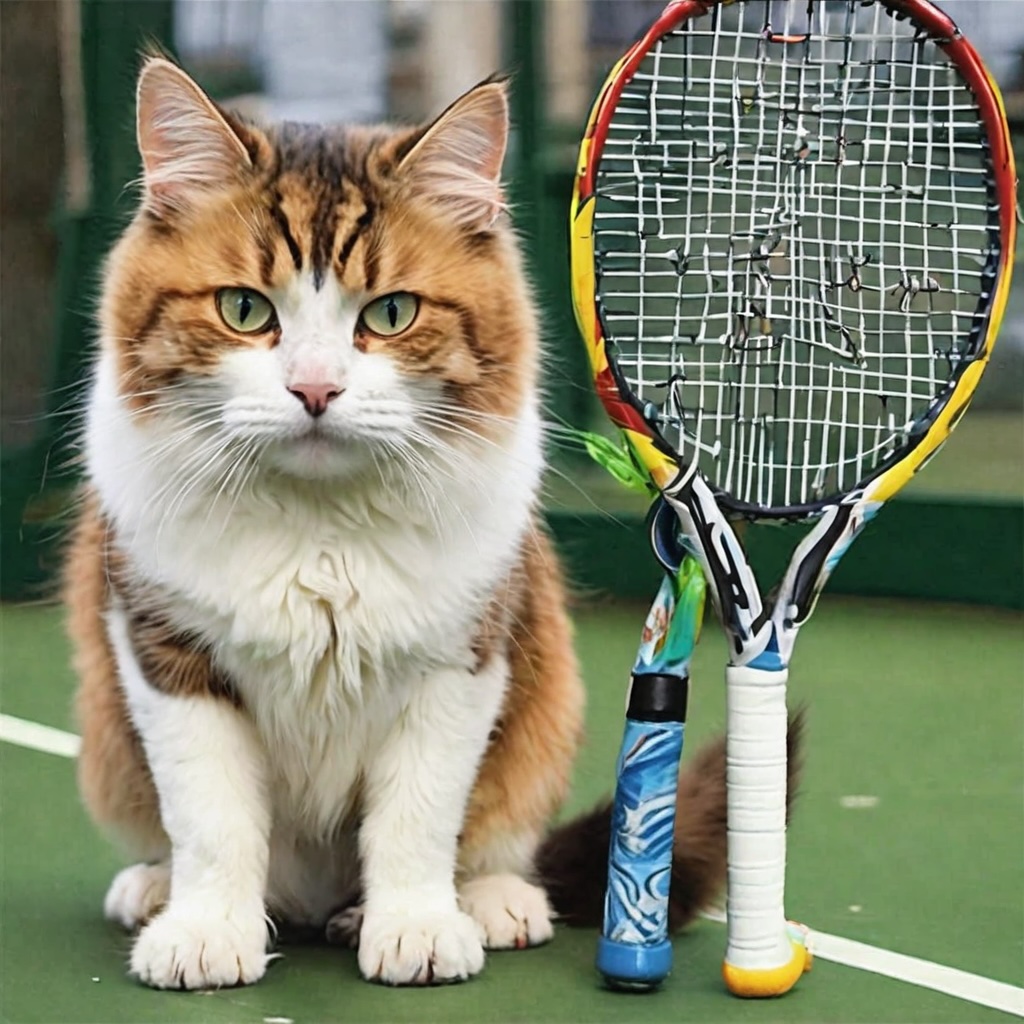}{1.89} \\

        \vspace{6pt} \\ %

        \multicolumn{2}{p{0.3\textwidth}}{\centering\scriptsize\itshape Four cars on the street} & 
        \multicolumn{2}{p{0.3\textwidth}}{\centering\scriptsize\itshape A painting by Grant Wood of an astronaut couple, american gothic style} & 
        \multicolumn{2}{p{0.3\textwidth}}{\centering\scriptsize\itshape A banana on the left of an apple} \\
        \cmidrule(lr{12pt}){1-2} \cmidrule(lr{12pt}){3-4} \cmidrule(l){5-6}

        \imgwithreward{0.148\textwidth}{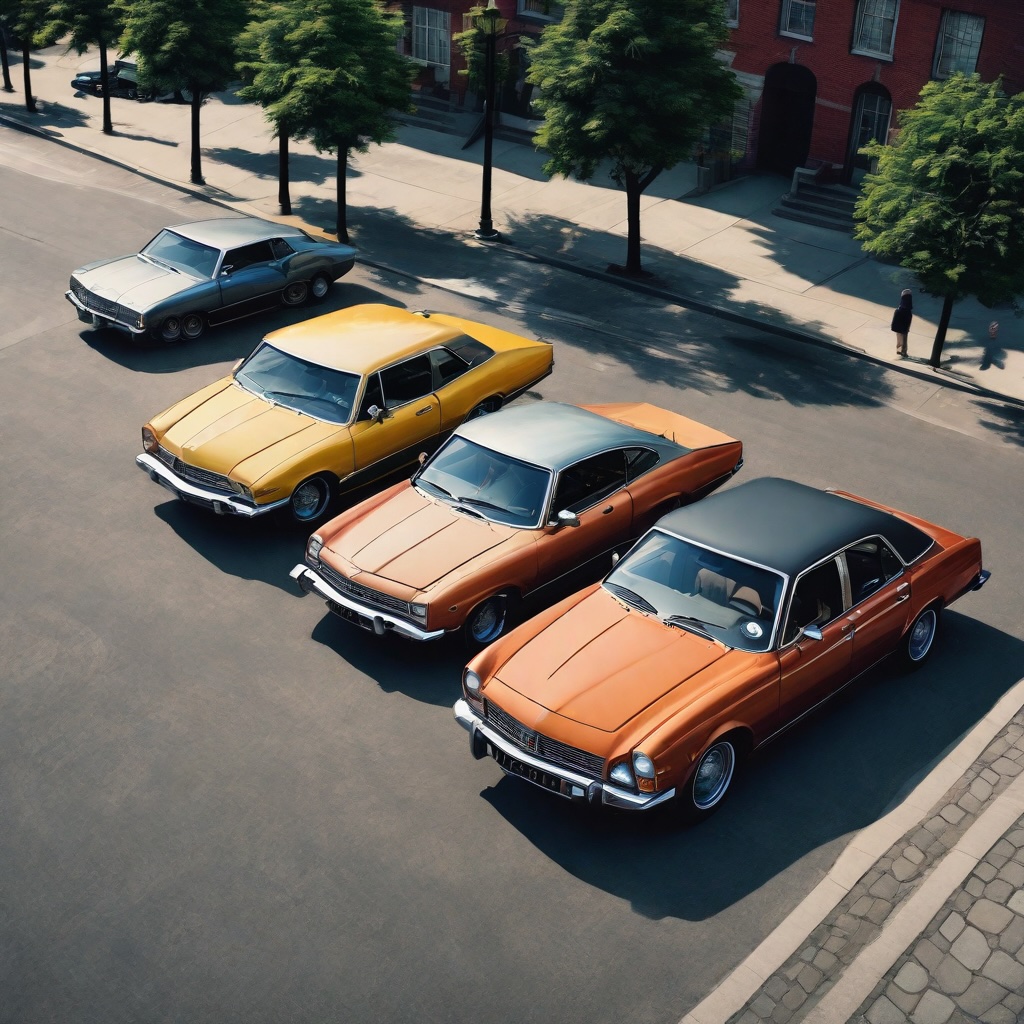}{0.317} &
        \imgwithreward{0.148\textwidth}{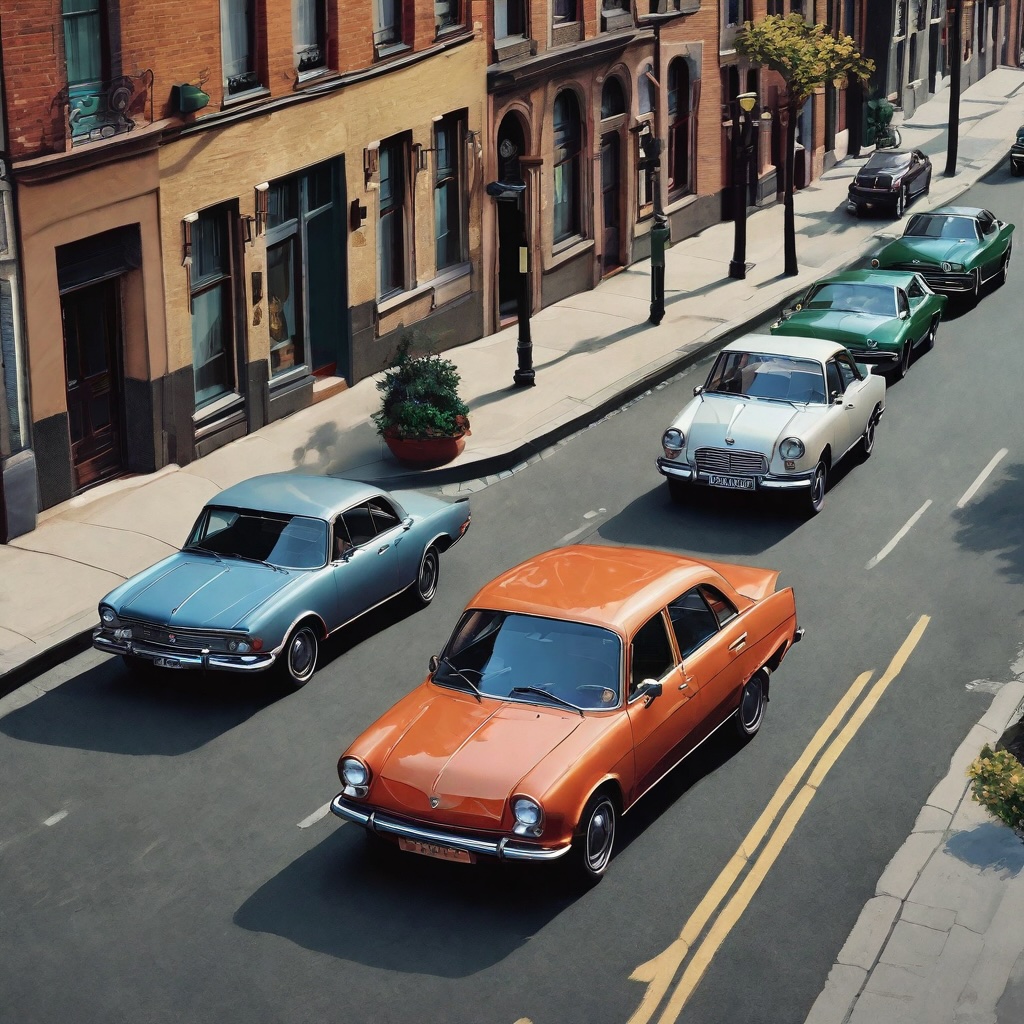}{0.329} &
        
        \imgwithreward{0.148\textwidth}{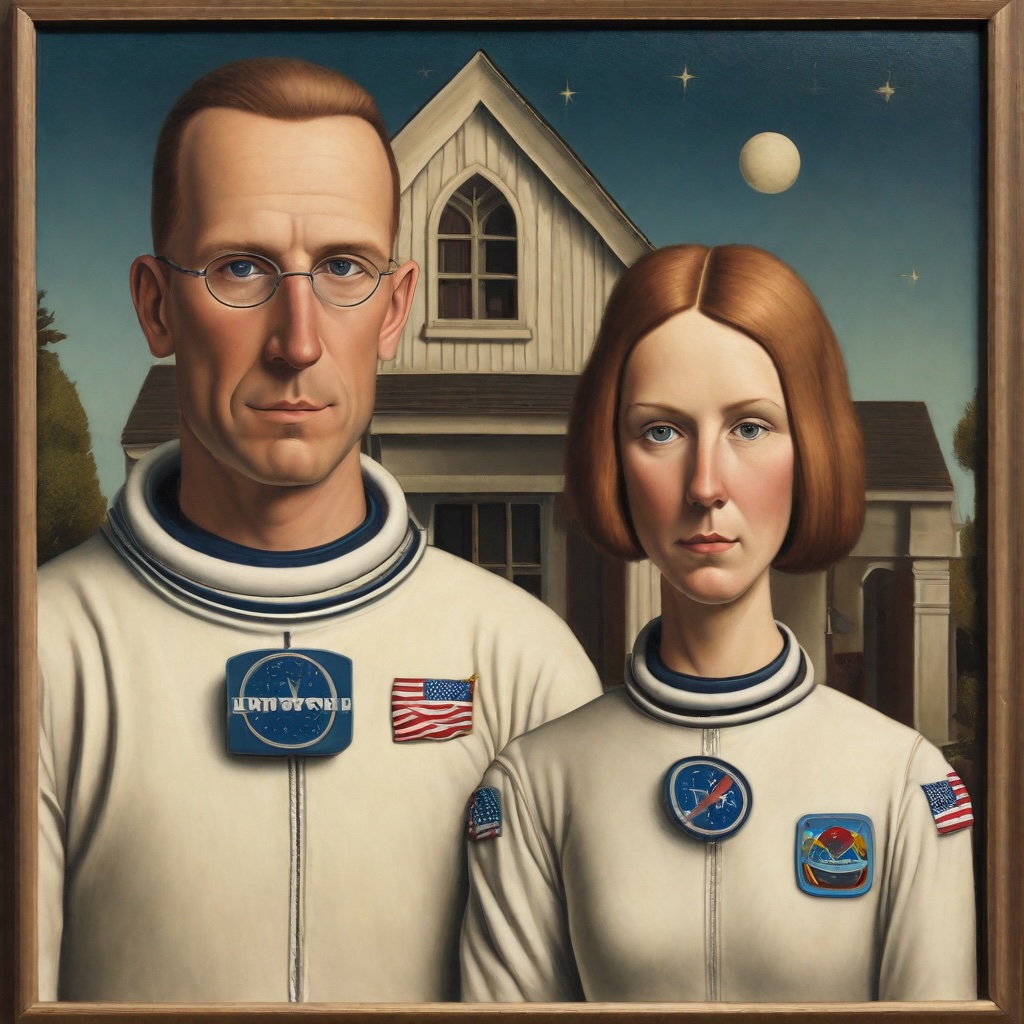}{0.303} &
        \imgwithreward{0.148\textwidth}{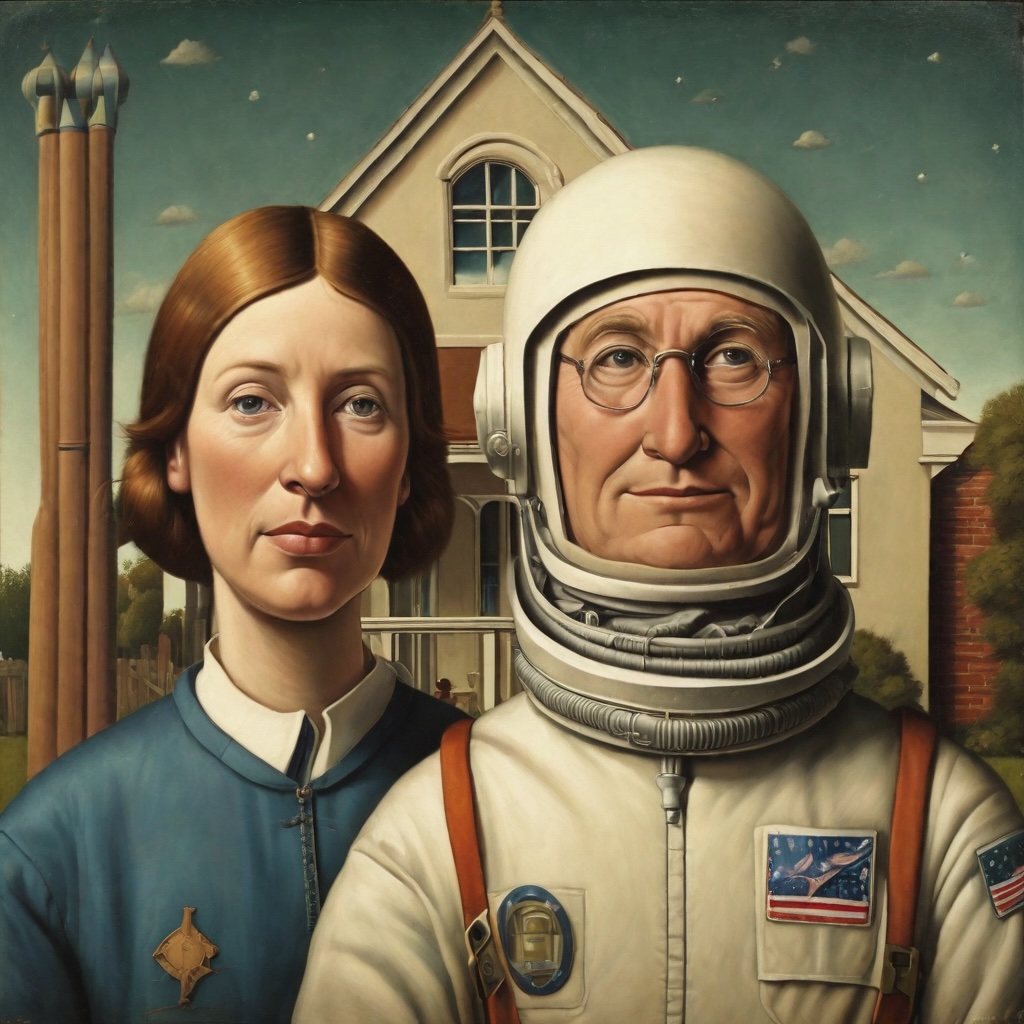}{0.309} &
        
        \imgwithreward{0.148\textwidth}{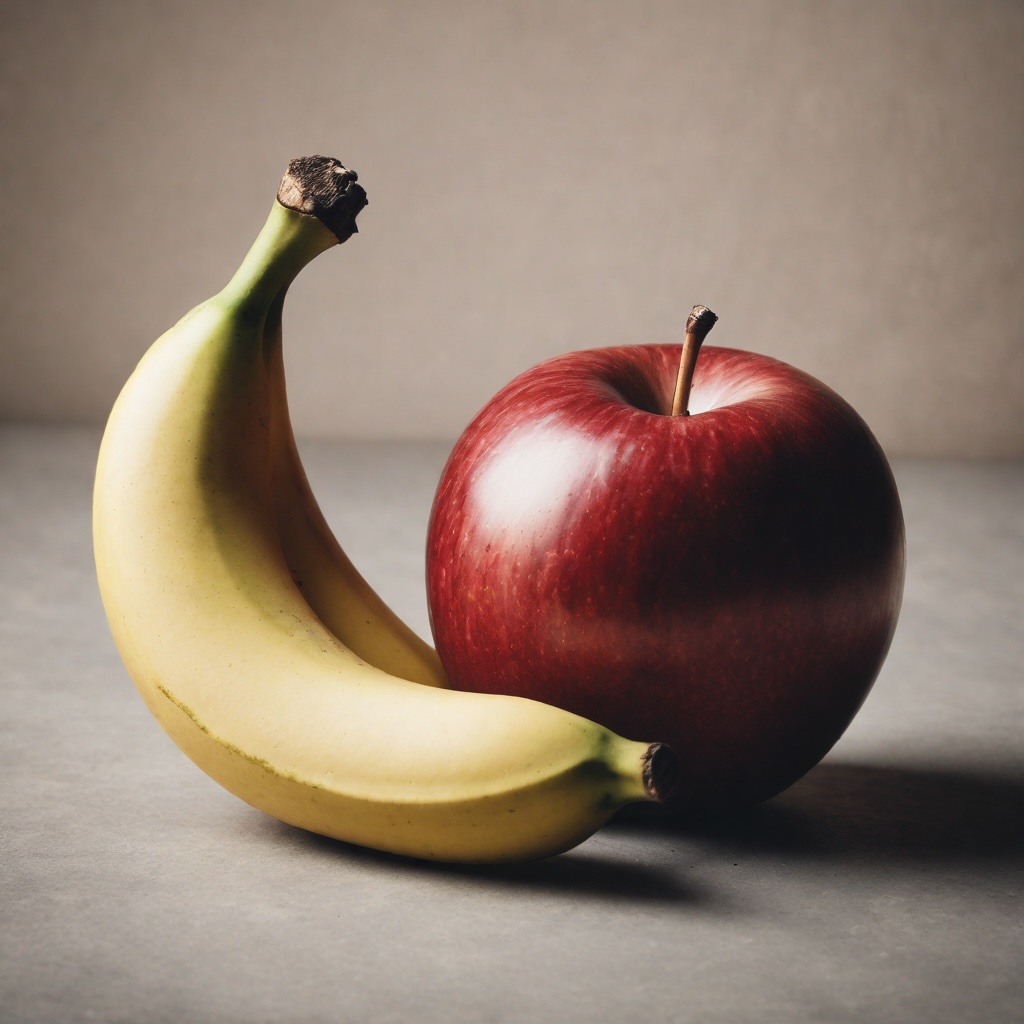}{0.303} &
        \imgwithreward{0.148\textwidth}{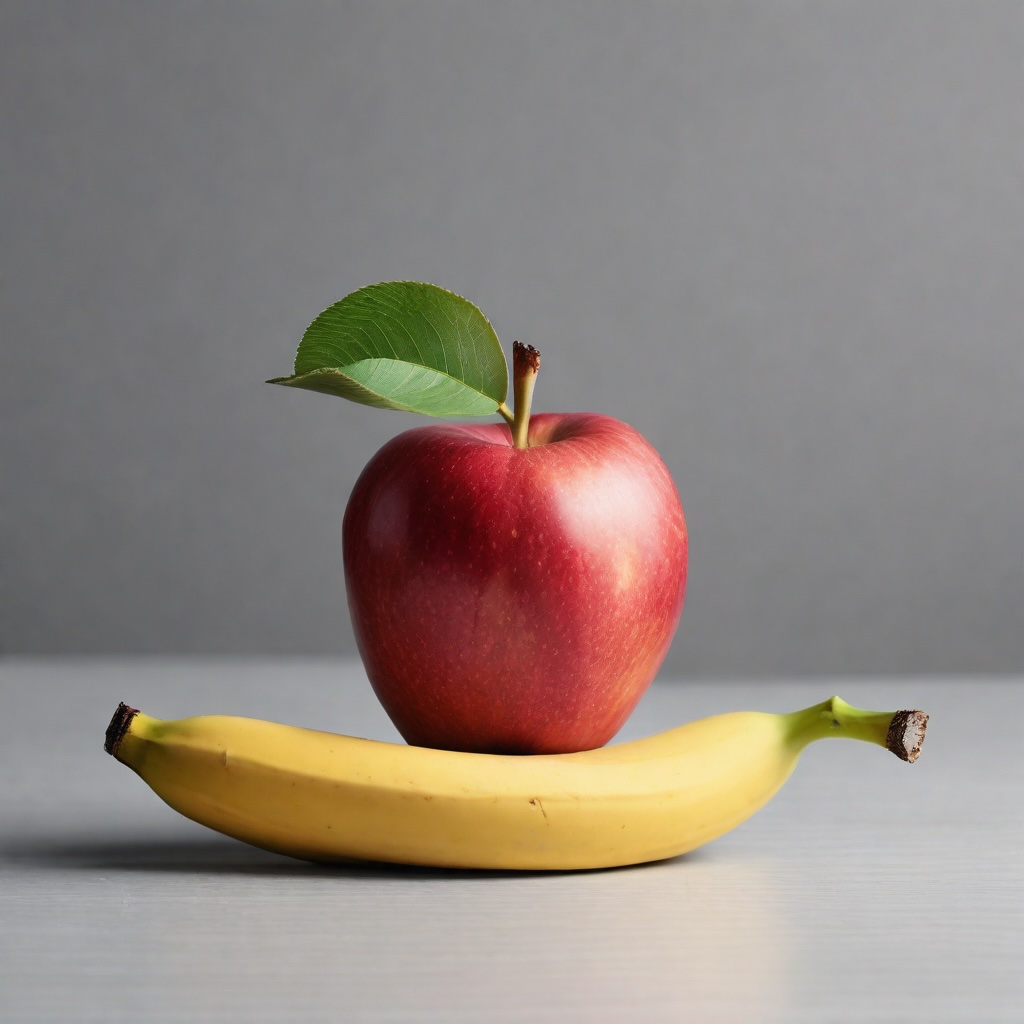}{0.308} \\
        
    \end{tabular}
    \caption{\textbf{Reward misalignment.} Cases where higher reward values are assigned even though the model fails to correctly generate the specific conceptual or spatial requirements of the prompt. All images are generated with SDXL. The upper row shows misalignment to ImageReward and the lower one to HPSv2.}
    \label{fig:reward_failure_cases}
\end{figure}

\paragraph{Reward Functions.} A primary limitation of these methods lies in the reliability of reward functions; when derived from neural network predictors, performance is inherently constrained by these models. In our experiments, reliance on pretrained networks for feedback makes optimization susceptible to model bias and failure modes.  \Cref{fig:reward_failure_cases} illustrates this via TRS-optimized SDXL images that receive high rewards despite failing to capture the core prompt intent. This highlights a fundamental challenge: reward models may score samples highly that exploit shortcuts in the learned signal rather than satisfying the true objective. To address this, recent work has focused on larger, more accurate reward models \citep{wu2025rewarddance}. TRS is well-positioned to integrate with these advances, offering a flexible and scalable framework as the field progresses toward more expressive models.

\paragraph{Diversity.}
While TRS begins with a global exploration phase across multiple promising regions, it ultimately converges to and exploits the most dominant region. Our experimental results demonstrate that this is an effective strategy for identifying a single optimal sample that maximizes the reward objective. However, TRS may be less suited for tasks requiring a diverse ensemble of samples. A direction for future work would be extending the framework to optimize for a diverse set of high-reward samples rather than a single point estimate. For instance, diversity could be explicitly enforced in TRS by incorporating a cosine similarity constraint on the center noise vectors of the trust regions.

\section{Optimized samples}
\label{sec:app:optimized_samples_examples}

\begin{figure}[t]
\centering
\setlength{\tabcolsep}{2pt}
\renewcommand{\arraystretch}{1.05} 
\setlength{\extrarowheight}{3pt}   

\definecolor{highlight}{rgb}{0.94, 0.96, 1.0}

\resizebox{\textwidth}{!}{%
\begin{tabular}{ >{\centering\arraybackslash}m{0.06\textwidth}
                 *{6}{>{\centering\arraybackslash}m{0.14\textwidth}} }



& \parbox[c][2.0cm][c]{0.14\textwidth}{\centering\arraybackslash\tiny\itshape A panda making latte art.}
& \parbox[c][2.0cm][c]{0.14\textwidth}{\centering\arraybackslash\tiny\itshape 35mm macro shot a kitten licking a baby duck, studio lighting.}
& \parbox[c][2.0cm][c]{0.14\textwidth}{\centering\arraybackslash\tiny\itshape Lego Arnold Schwarzenegger.}
& \parbox[c][2.0cm][c]{0.14\textwidth}{\centering\arraybackslash\tiny\itshape An old photograph of a 1920s airship shaped like a pig, floating over a wheat field.}
& \parbox[c][2.0cm][c]{0.14\textwidth}{\centering\arraybackslash\tiny\itshape A triangular purple flower pot. A purple flower pot in the shape of a triangle.}
& \parbox[c][2.0cm][c]{0.14\textwidth}{\centering\arraybackslash\tiny\itshape An ancient Egyptian painting depicting an argument over whose turn it is to take out the trash.}
\\[6pt] 
\noalign{\vspace{0pt}}

\rotatebox{90}{\shortstack{\scriptsize SD1.5}} &
\includegraphics[width=0.14\textwidth]{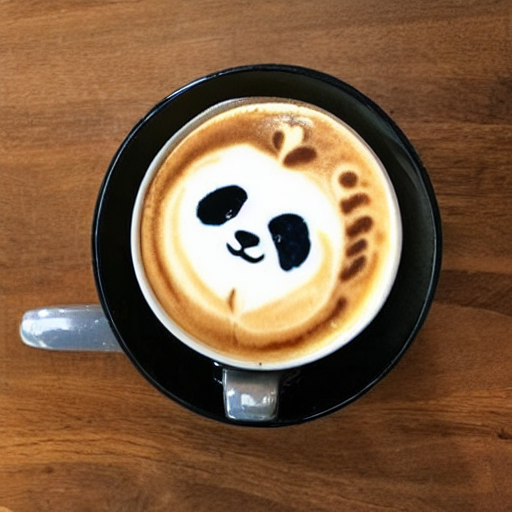} &
\includegraphics[width=0.14\textwidth]{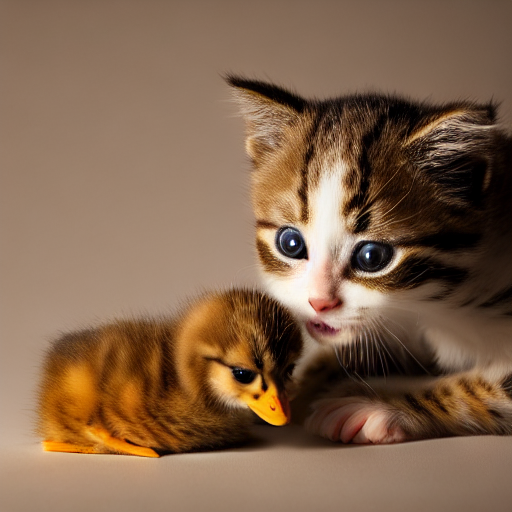} &
\includegraphics[width=0.14\textwidth]{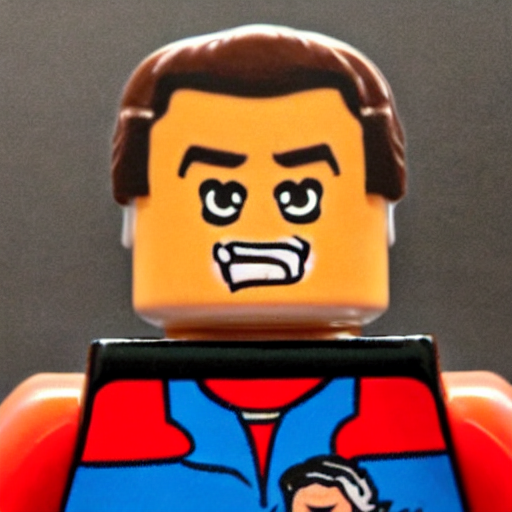} &
\includegraphics[width=0.14\textwidth]{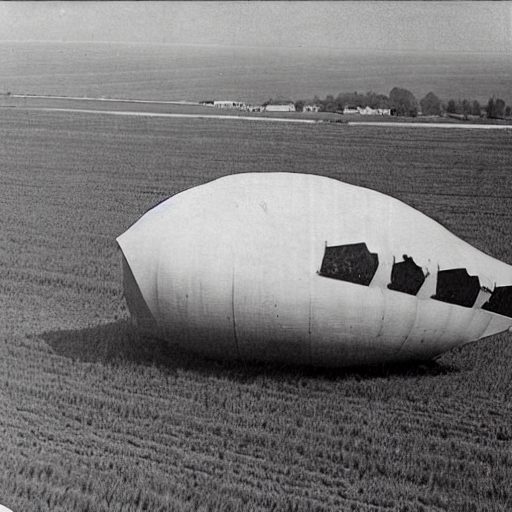} &
\includegraphics[width=0.14\textwidth]{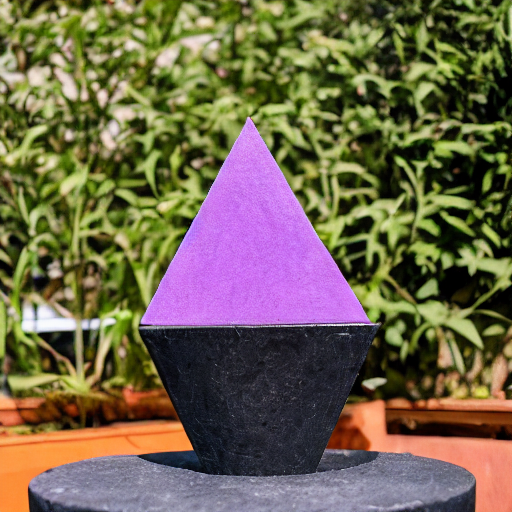} &
\includegraphics[width=0.14\textwidth]{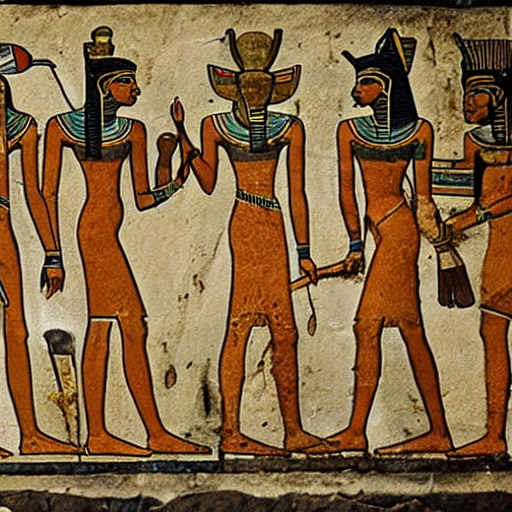}
\\

\rotatebox{90}{\shortstack{\scriptsize FLUX.1-dev}} &
\includegraphics[width=0.14\textwidth]{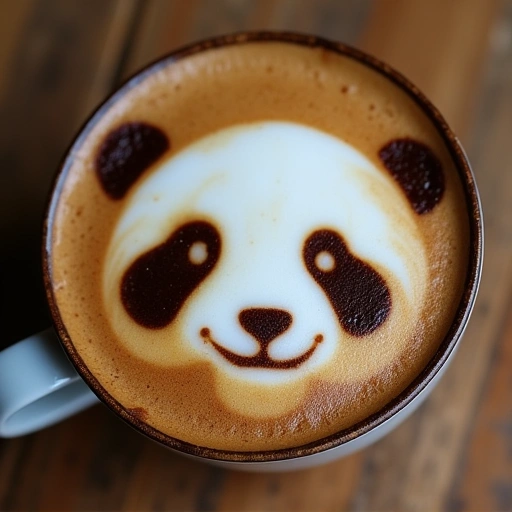} &
\includegraphics[width=0.14\textwidth]{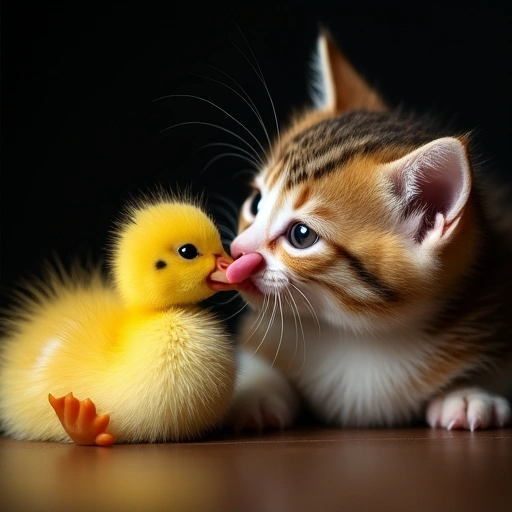} &
\includegraphics[width=0.14\textwidth]{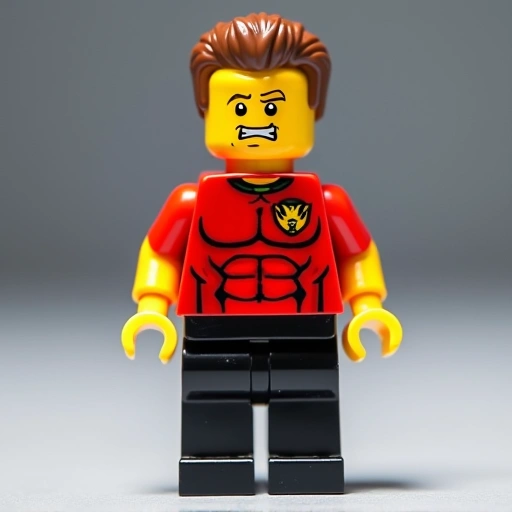} &
\includegraphics[width=0.14\textwidth]{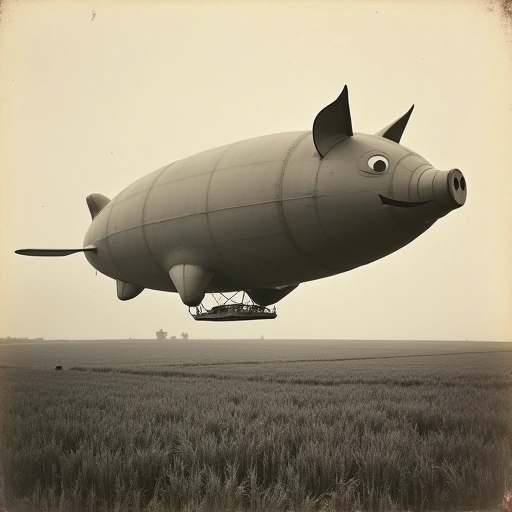} &
\includegraphics[width=0.14\textwidth]{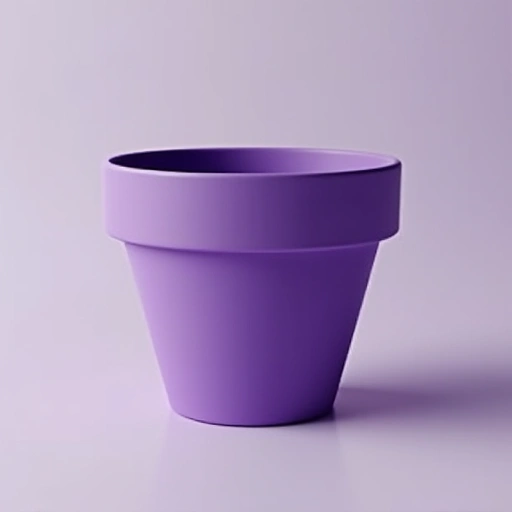} &
\includegraphics[width=0.14\textwidth]{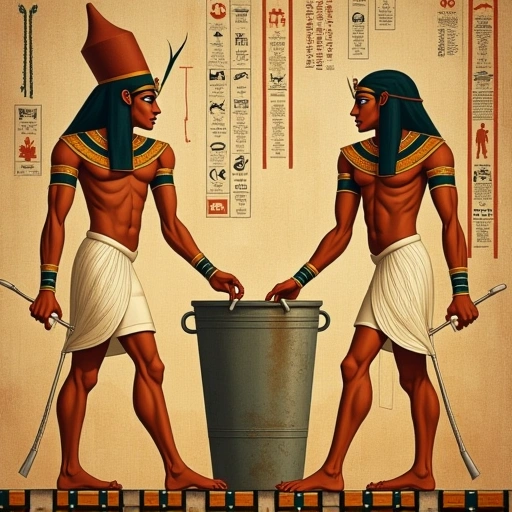}
\\

\rowcolor{highlight}[0pt][0pt]
\rotatebox{90}{\shortstack{\scriptsize SD1.5 + TRS\\ \scriptsize (Ours)}} &
\includegraphics[width=0.14\textwidth]{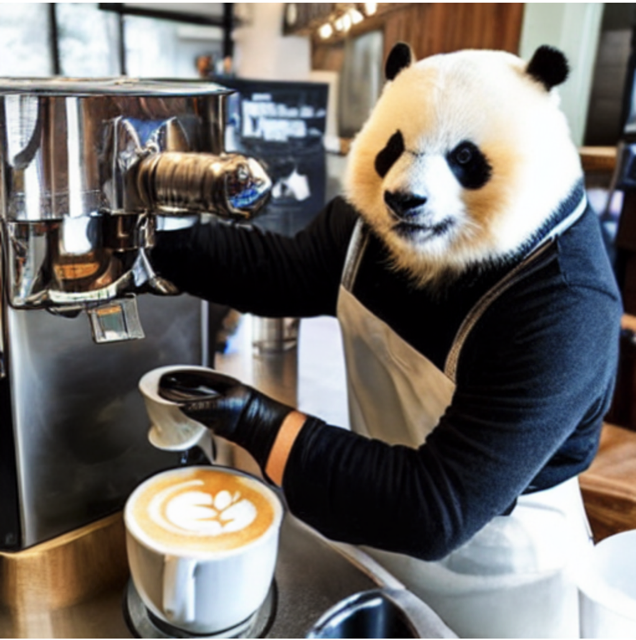} &
\includegraphics[width=0.14\textwidth]{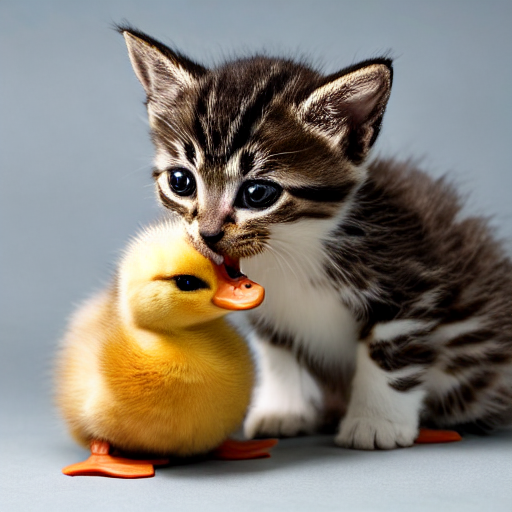} &
\includegraphics[width=0.14\textwidth]{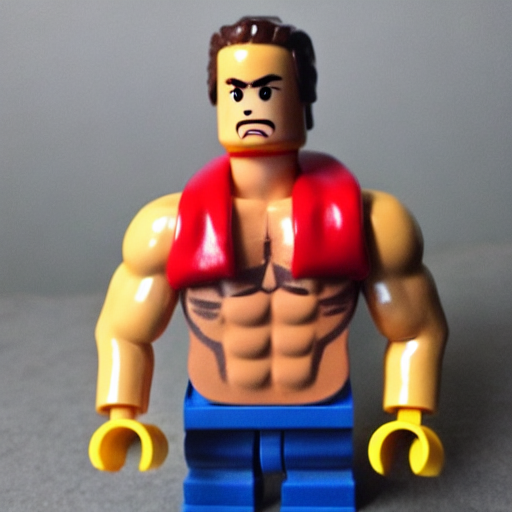} &
\includegraphics[width=0.14\textwidth]{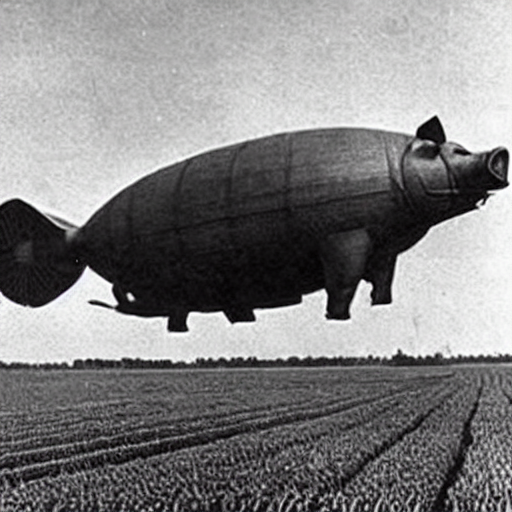} &
\includegraphics[width=0.14\textwidth]{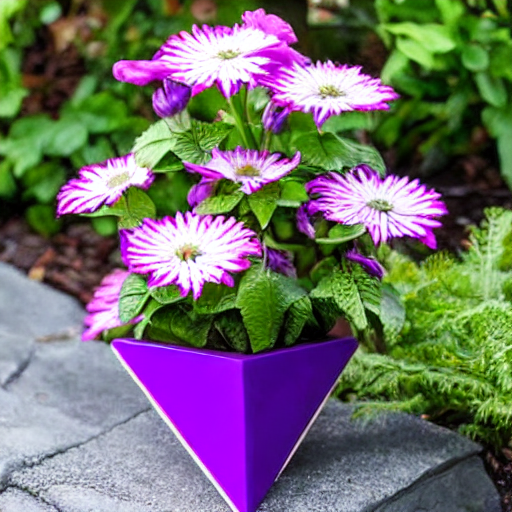} &
\includegraphics[width=0.14\textwidth]{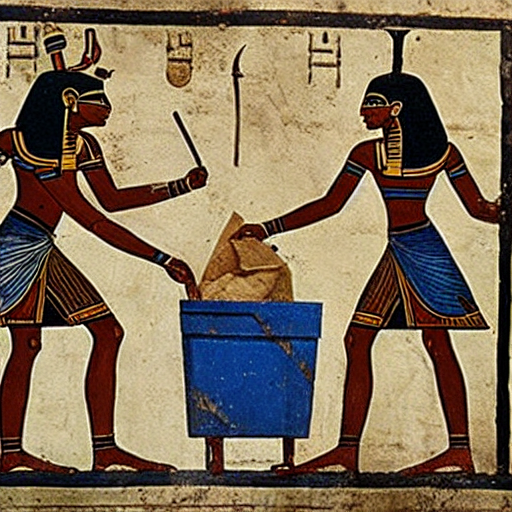}
\\

\end{tabular}%
}
\captionsetup{font=small}
\caption{\textbf{Inference-time scaling can be better than picking a stronger model.} Comparing vanilla Stable Diffusion 1.5, vanilla FLUX.1-dev and our method after $1500$ reward function evaluations, with prompts from DrawBench and ImageReward as the reward function. Investing additional compute into finding the best source noise for a smaller, more efficient model can match or surpass a much larger and more expensive state-of-the-art model.}
\label{fig:second_teaser}
\end{figure}

\Cref{fig:second_teaser} illustrates the broader potential of source noise search as an alternative to scaling up the generative backbone. We compare vanilla Stable Diffusion 1.5, vanilla FLUX.1-dev, and SD1.5 with our method, showing that spending additional compute on finding good source noise for a smaller, more efficient model can match or surpass a much larger and more expensive state-of-the-art model.

\begin{figure}[htbp]
    \centering
    \definecolor{highlight}{rgb}{0.94, 0.96, 1.0}
    
    \setlength{\tabcolsep}{0pt} 
    \setlength{\aboverulesep}{2pt} 
    \setlength{\belowrulesep}{2pt}

    \setlength{\hsep}{4pt} 
    \setlength{\vsep}{28pt}
    
    \renewcommand{\arraystretch}{0} 

    \begin{tabular}{
        >{\raggedright\arraybackslash\tiny\itshape}m{1.5cm} 
        @{\hspace{\hsep}} >{\columncolor{highlight}}c 
        @{\hspace{\hsep}}c@{\hspace{\hsep}}c@{\hspace{\hsep}}c@{\hspace{\hsep}}c@{\hspace{\hsep}}c 
    }
        \toprule
        \rule{0pt}{3ex} \textbf{\scriptsize Prompt} & \textbf{\scriptsize TRS (Ours)} & \textbf{\scriptsize RS} & \textbf{\scriptsize ZO} & \textbf{\scriptsize DTS*} & \textbf{\scriptsize FD} & \textbf{\scriptsize OC-Flow} \\[1.5ex]
        \midrule
        
        A keyboard made of water, the water is made of light, the light is turned off. & 
        $\vcenter{\hbox{\includegraphics[width=0.136\textwidth]{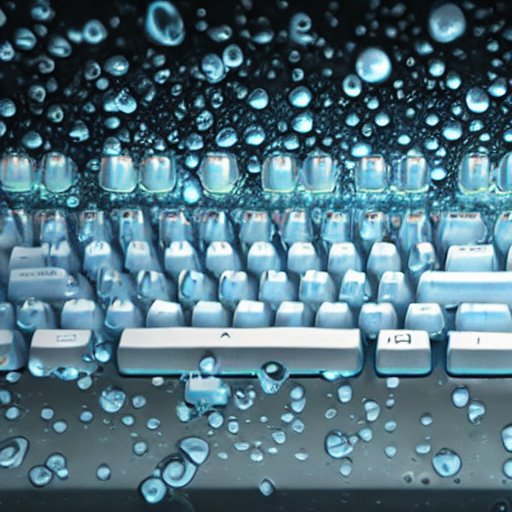}}}$ & 
        $\vcenter{\hbox{\includegraphics[width=0.136\textwidth]{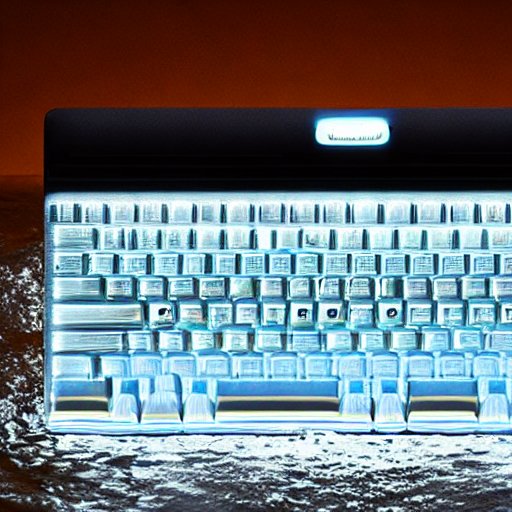}}}$ & 
        $\vcenter{\hbox{\includegraphics[width=0.136\textwidth]{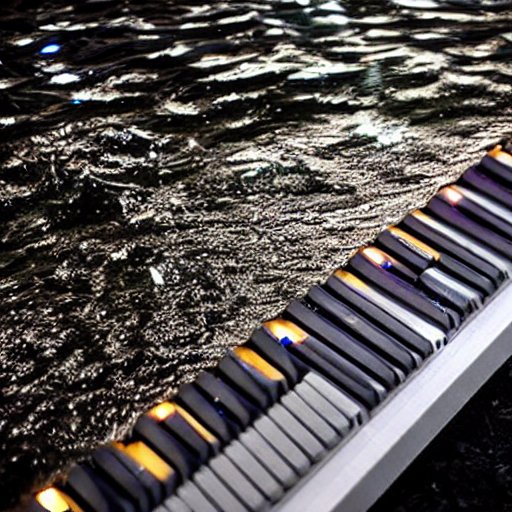}}}$ & 
        $\vcenter{\hbox{\includegraphics[width=0.136\textwidth]{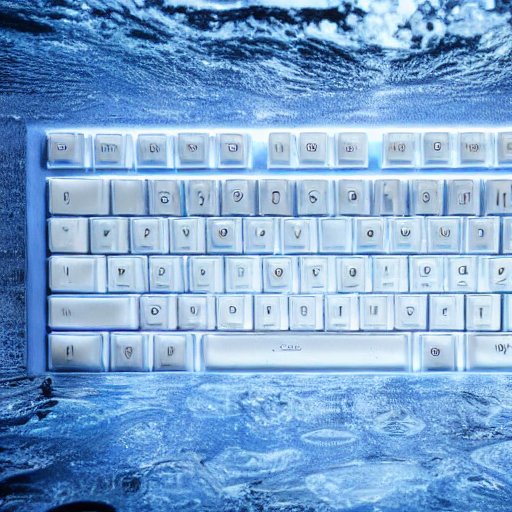}}}$ & 
        $\vcenter{\hbox{\includegraphics[width=0.136\textwidth]{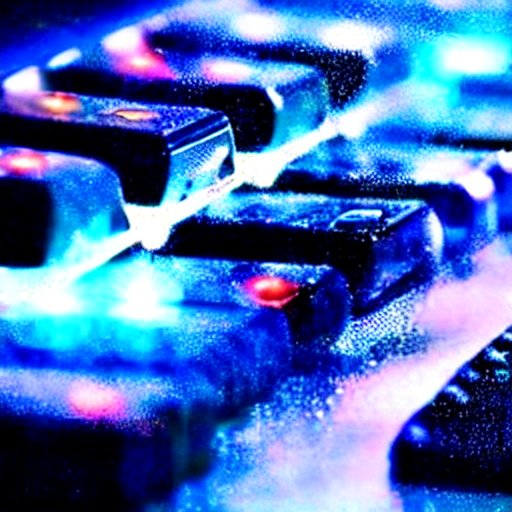}}}$ & 
        $\vcenter{\hbox{\includegraphics[width=0.136\textwidth]{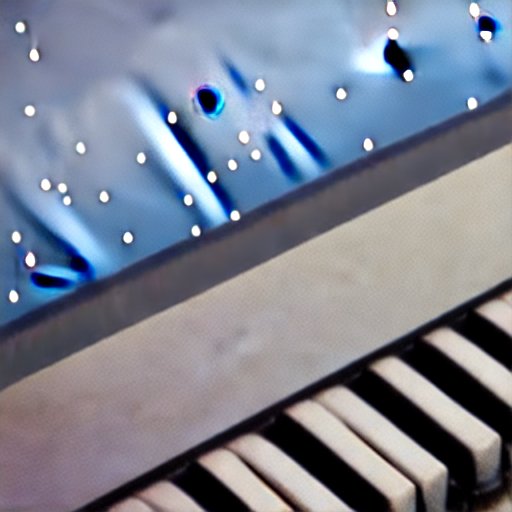}}}$ \\[\vsep]

        A red colored dog. & 
        $\vcenter{\hbox{\includegraphics[width=0.136\textwidth]{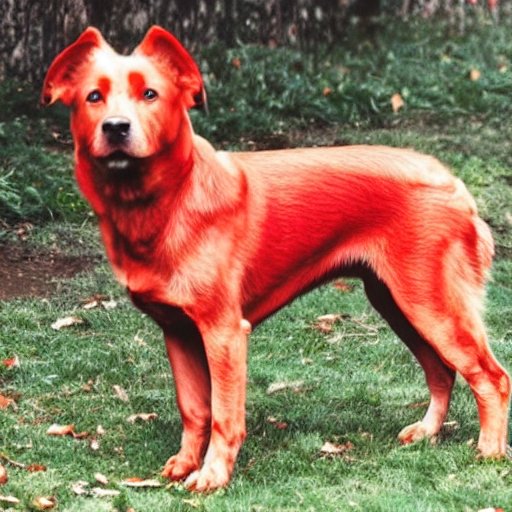}}}$ & 
        $\vcenter{\hbox{\includegraphics[width=0.136\textwidth]{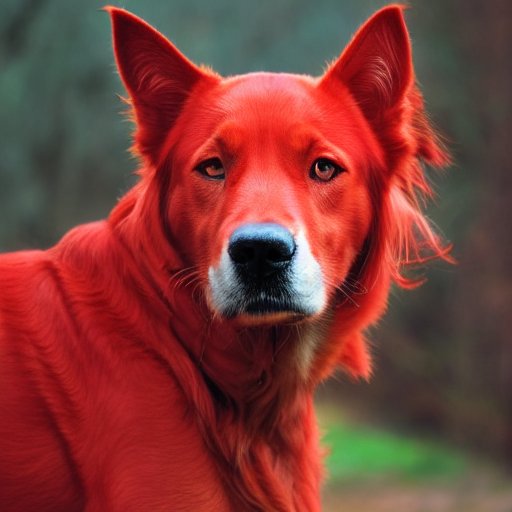}}}$ & 
        $\vcenter{\hbox{\includegraphics[width=0.136\textwidth]{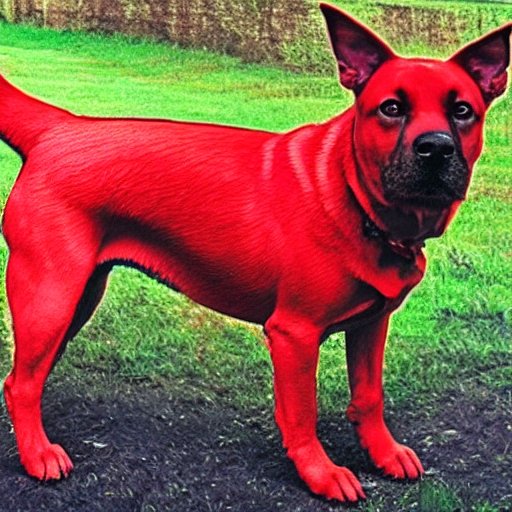}}}$ & 
        $\vcenter{\hbox{\includegraphics[width=0.136\textwidth]{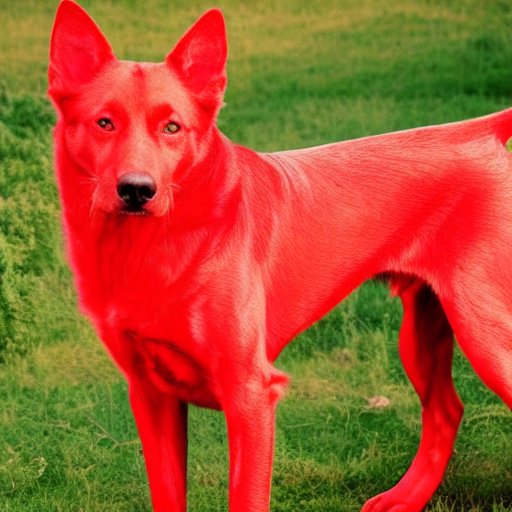}}}$ & 
        $\vcenter{\hbox{\includegraphics[width=0.136\textwidth]{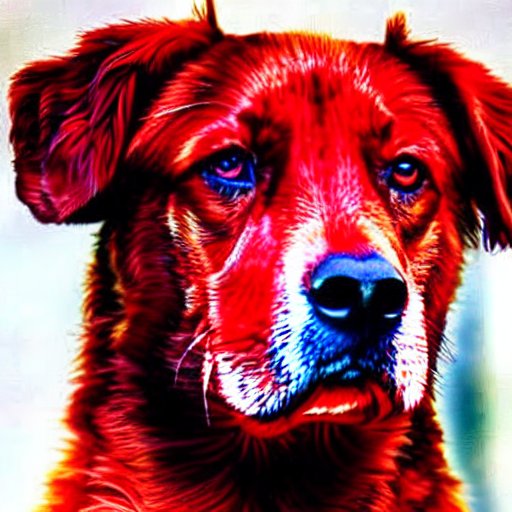}}}$ & 
        $\vcenter{\hbox{\includegraphics[width=0.136\textwidth]{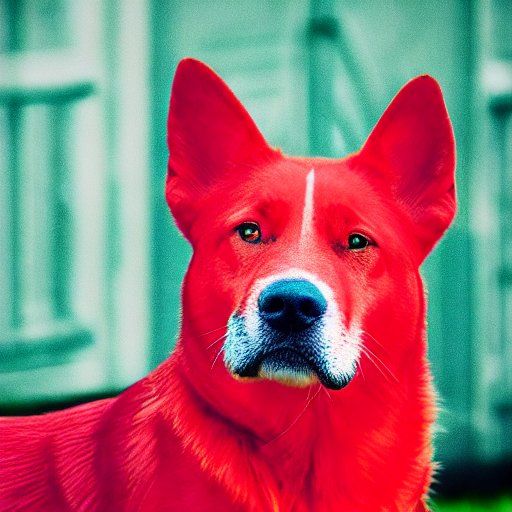}}}$ \\[\vsep]

        Hyper-realistic photo of an abandoned industrial site during a storm. & 
        $\vcenter{\hbox{\includegraphics[width=0.136\textwidth]{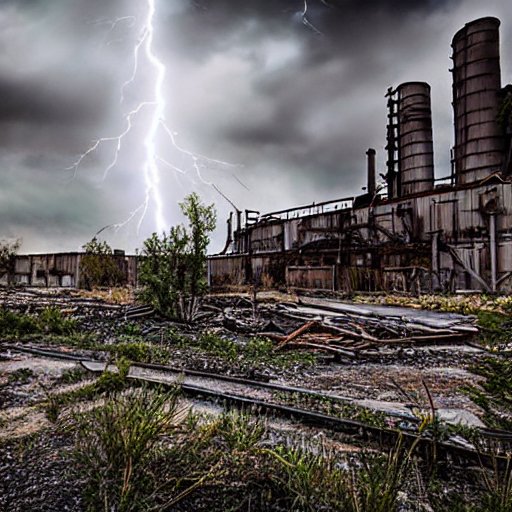}}}$ & 
        $\vcenter{\hbox{\includegraphics[width=0.136\textwidth]{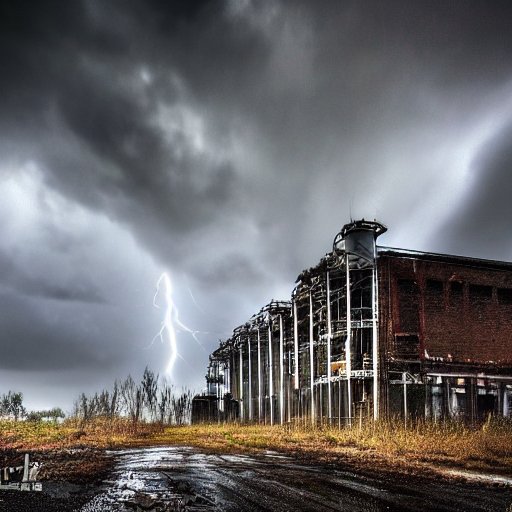}}}$ & 
        $\vcenter{\hbox{\includegraphics[width=0.136\textwidth]{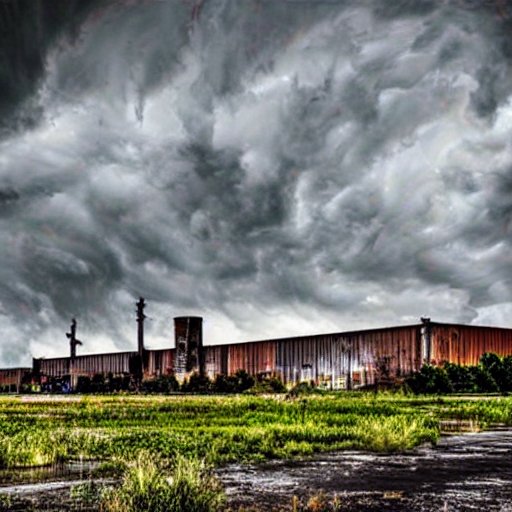}}}$ & 
        $\vcenter{\hbox{\includegraphics[width=0.136\textwidth]{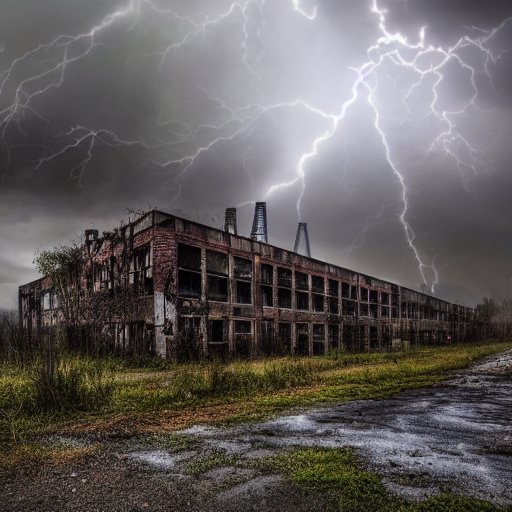}}}$ & 
        $\vcenter{\hbox{\includegraphics[width=0.136\textwidth]{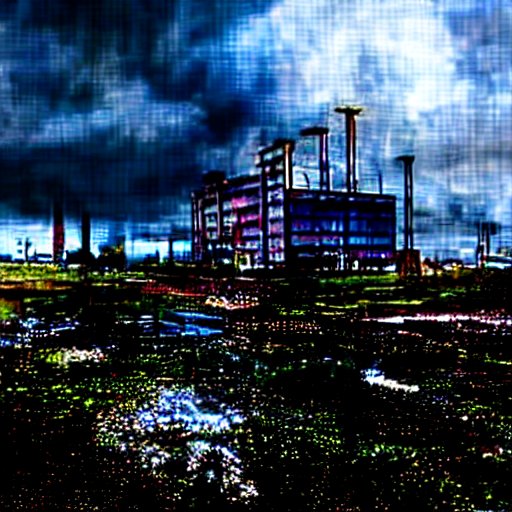}}}$ & 
        $\vcenter{\hbox{\includegraphics[width=0.136\textwidth]{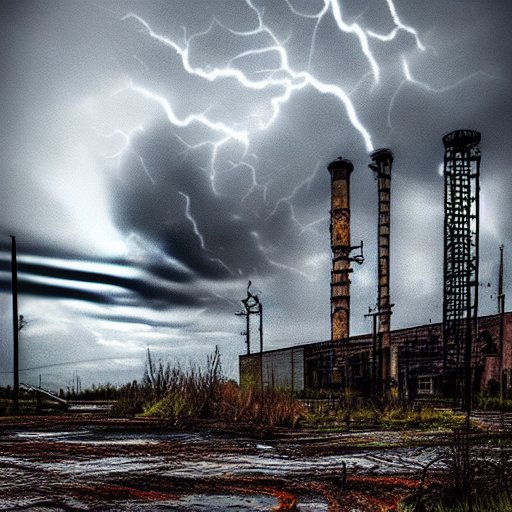}}}$ \\[\vsep]

        A cat on the left of a dog. & 
        $\vcenter{\hbox{\includegraphics[width=0.136\textwidth]{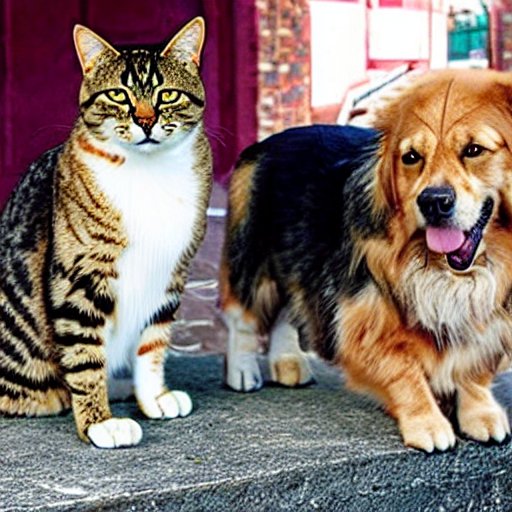}}}$ & 
        $\vcenter{\hbox{\includegraphics[width=0.136\textwidth]{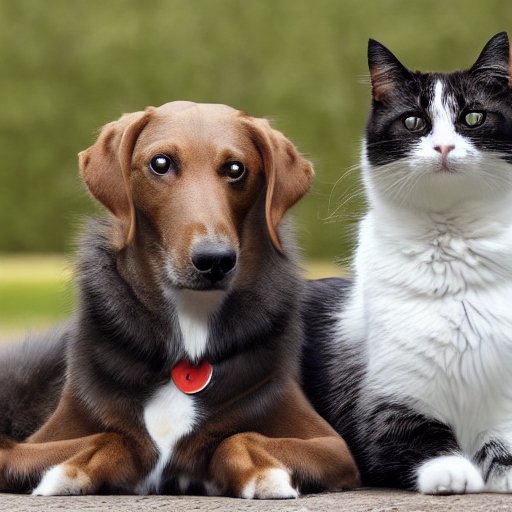}}}$ & 
        $\vcenter{\hbox{\includegraphics[width=0.136\textwidth]{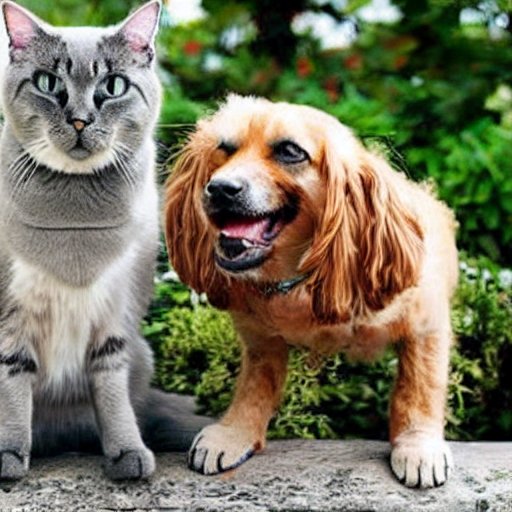}}}$ & 
        $\vcenter{\hbox{\includegraphics[width=0.136\textwidth]{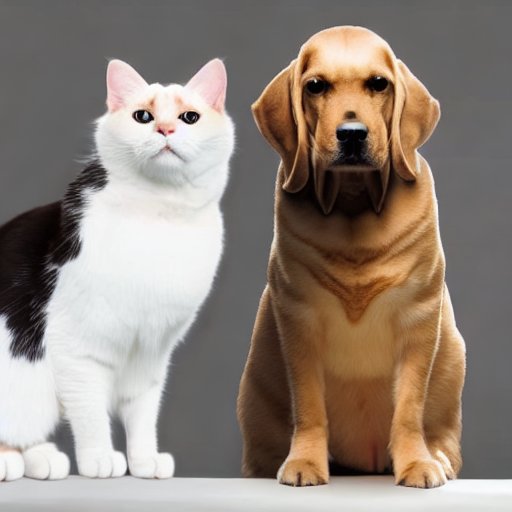}}}$ & 
        $\vcenter{\hbox{\includegraphics[width=0.136\textwidth]{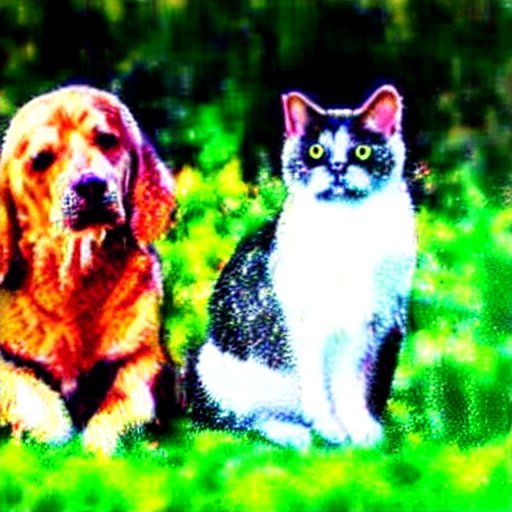}}}$ & 
        $\vcenter{\hbox{\includegraphics[width=0.136\textwidth]{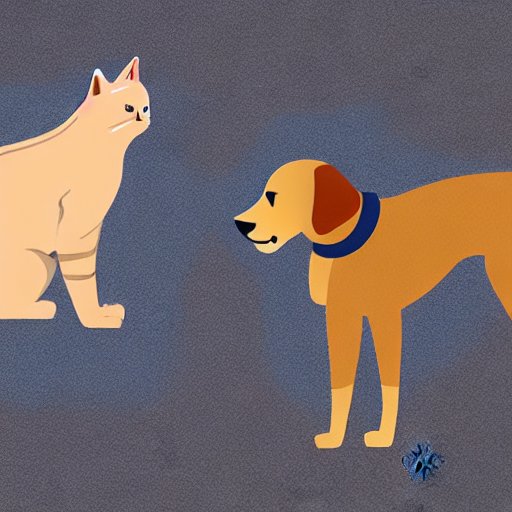}}}$ \\[\vsep]

        A giraffe underneath a microwave. & 
        $\vcenter{\hbox{\includegraphics[width=0.136\textwidth]{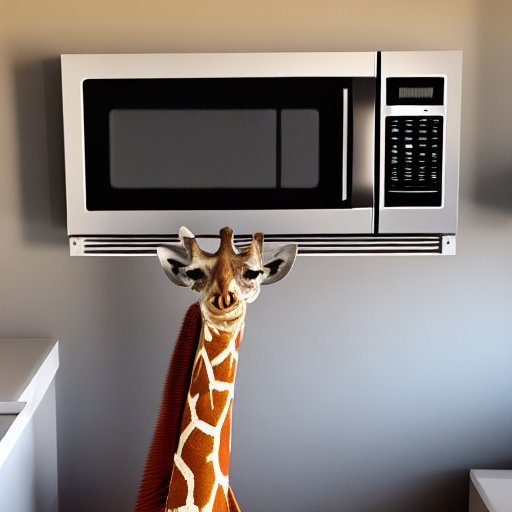}}}$ & 
        $\vcenter{\hbox{\includegraphics[width=0.136\textwidth]{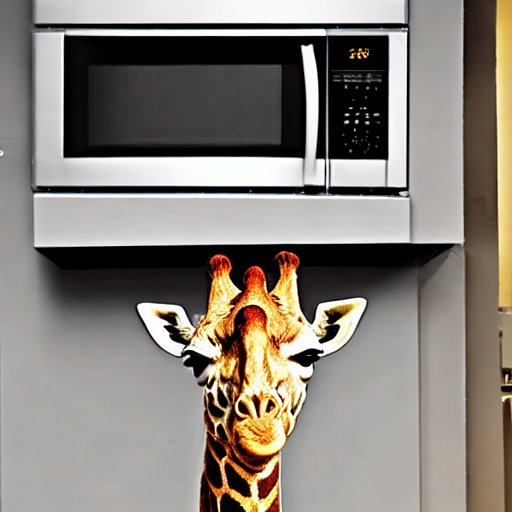}}}$ & 
        $\vcenter{\hbox{\includegraphics[width=0.136\textwidth]{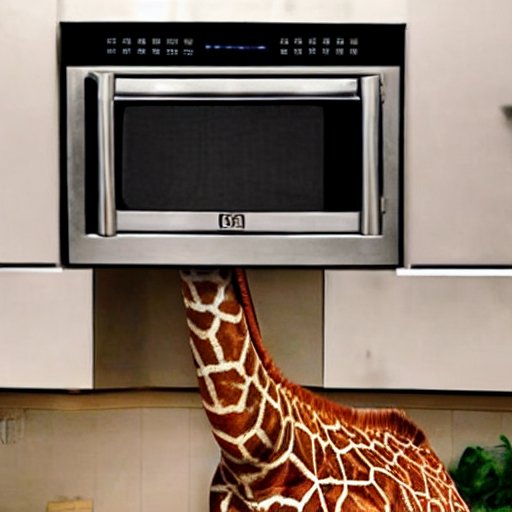}}}$ & 
        $\vcenter{\hbox{\includegraphics[width=0.136\textwidth]{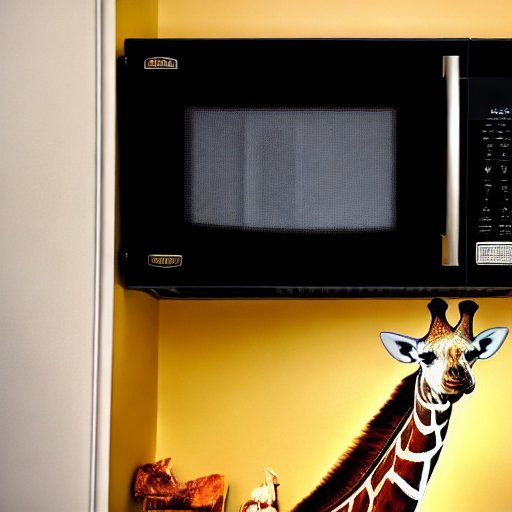}}}$ & 
        $\vcenter{\hbox{\includegraphics[width=0.136\textwidth]{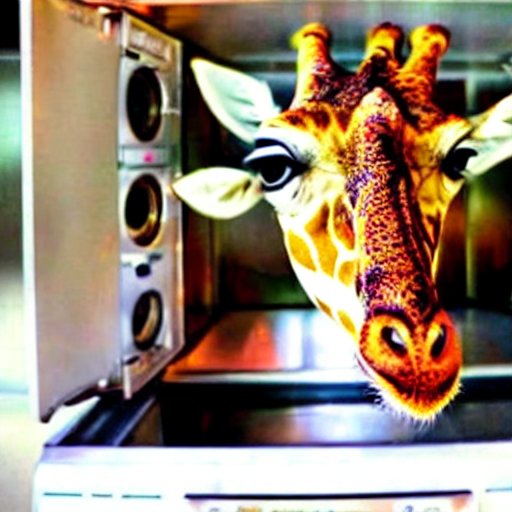}}}$ & 
        $\vcenter{\hbox{\includegraphics[width=0.136\textwidth]{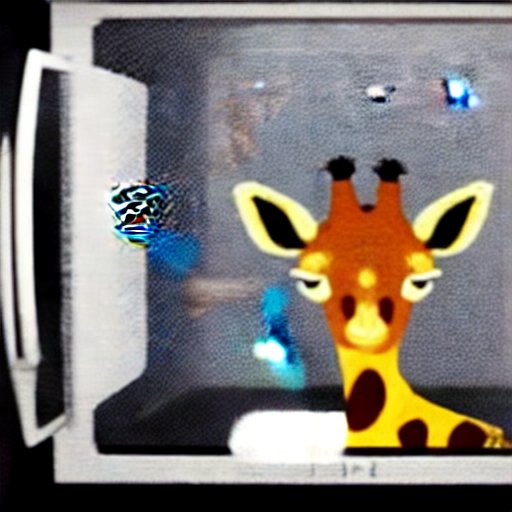}}}$ \\[\vsep]

        A baby fennec sneezing onto a strawberry, detailed, macro, studio light, droplets, backlit ears. & 
        $\vcenter{\hbox{\includegraphics[width=0.136\textwidth]{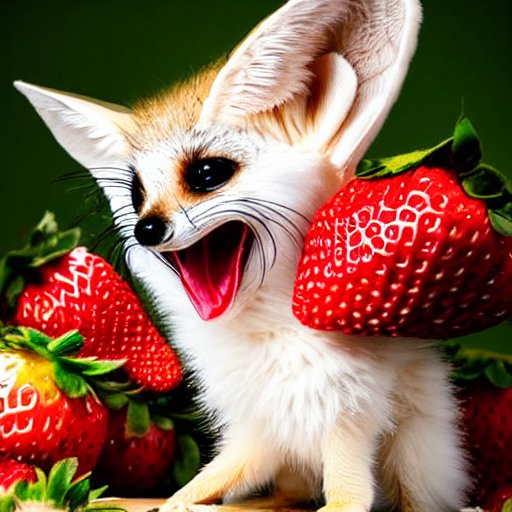}}}$ & 
        $\vcenter{\hbox{\includegraphics[width=0.136\textwidth]{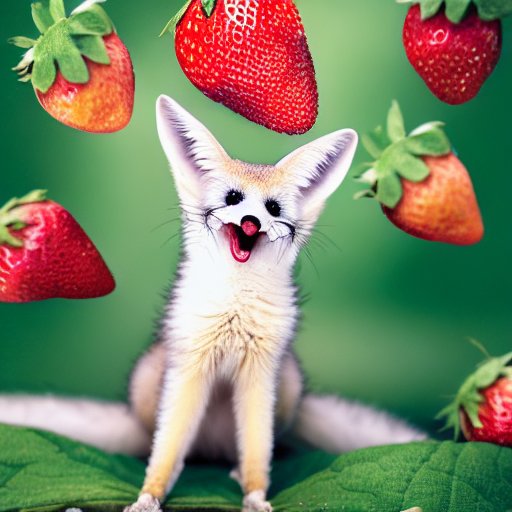}}}$ & 
        $\vcenter{\hbox{\includegraphics[width=0.136\textwidth]{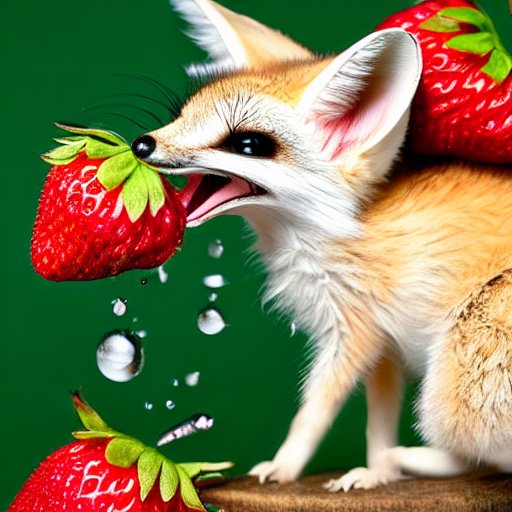}}}$ & 
        $\vcenter{\hbox{\includegraphics[width=0.136\textwidth]{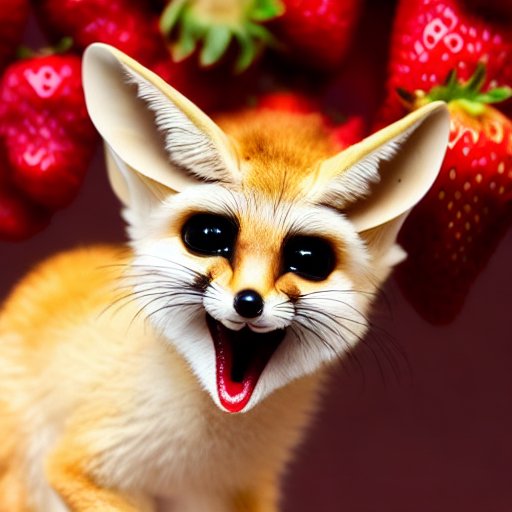}}}$ & 
        $\vcenter{\hbox{\includegraphics[width=0.136\textwidth]{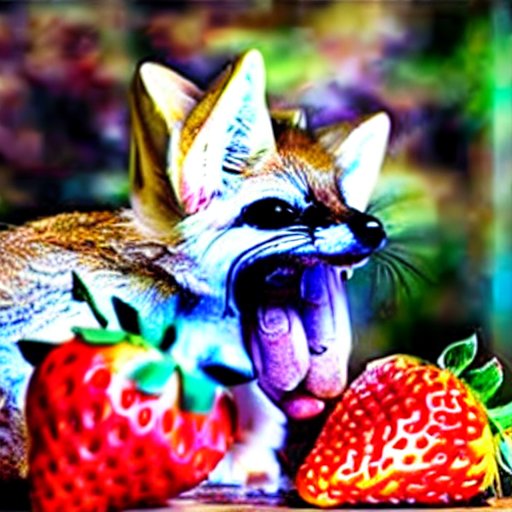}}}$ & 
        $\vcenter{\hbox{\includegraphics[width=0.136\textwidth]{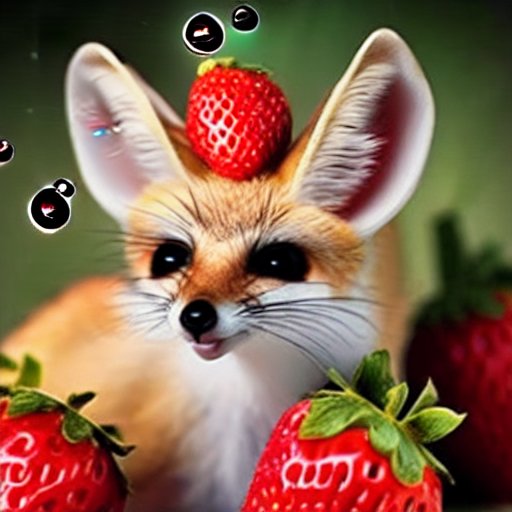}}}$ \\[\vsep]

        One car on the street. & 
        $\vcenter{\hbox{\includegraphics[width=0.136\textwidth]{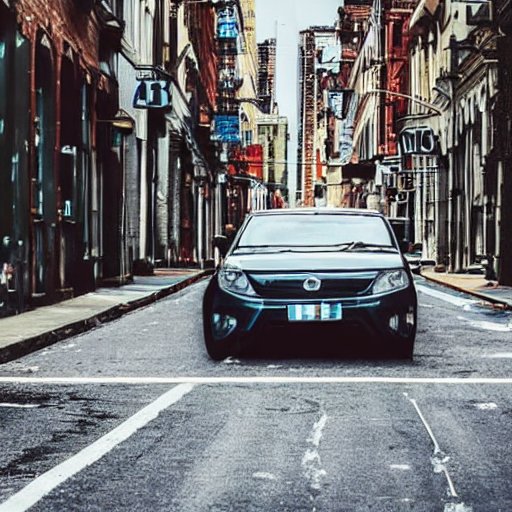}}}$ & 
        $\vcenter{\hbox{\includegraphics[width=0.136\textwidth]{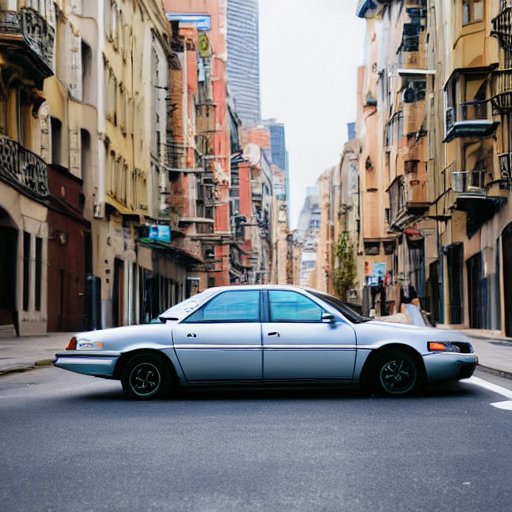}}}$ & 
        $\vcenter{\hbox{\includegraphics[width=0.136\textwidth]{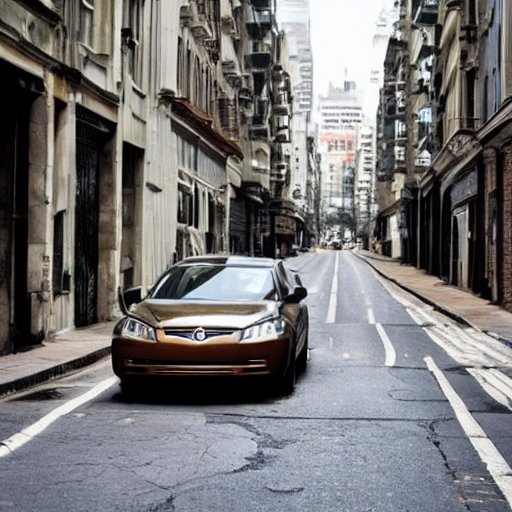}}}$ & 
        $\vcenter{\hbox{\includegraphics[width=0.136\textwidth]{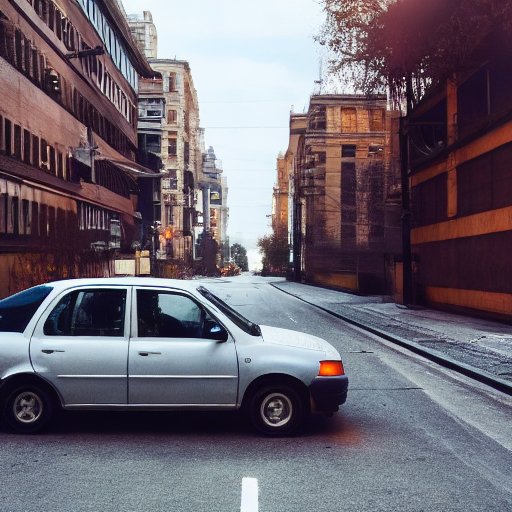}}}$ & 
        $\vcenter{\hbox{\includegraphics[width=0.136\textwidth]{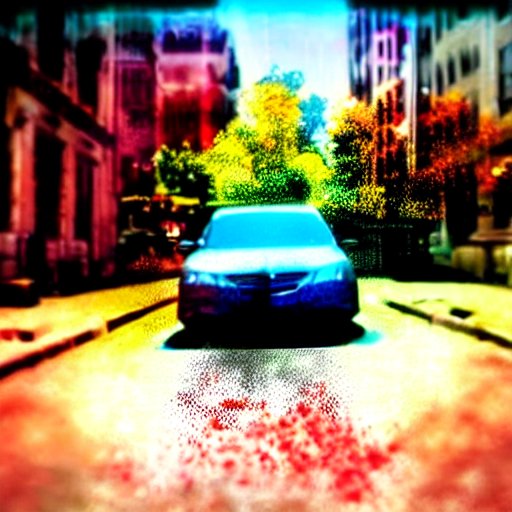}}}$ & 
        $\vcenter{\hbox{\includegraphics[width=0.136\textwidth]{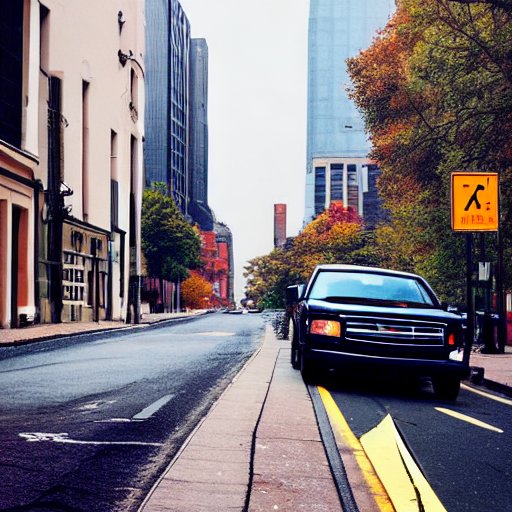}}}$ \\[\vsep]

        A pink colored car. & 
        $\vcenter{\hbox{\includegraphics[width=0.136\textwidth]{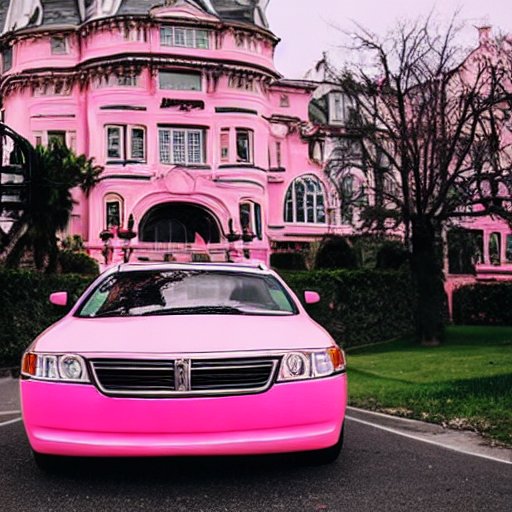}}}$ & 
        $\vcenter{\hbox{\includegraphics[width=0.136\textwidth]{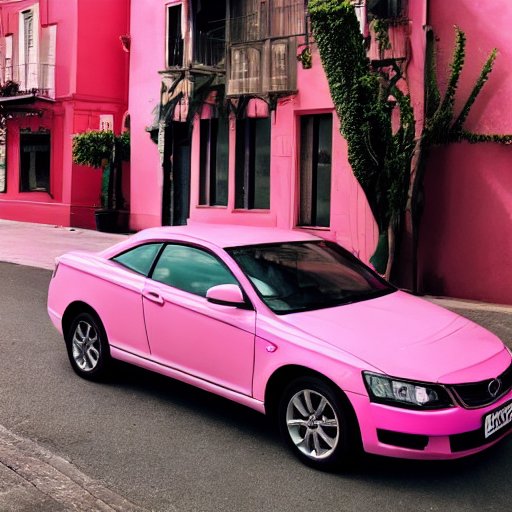}}}$ & 
        $\vcenter{\hbox{\includegraphics[width=0.136\textwidth]{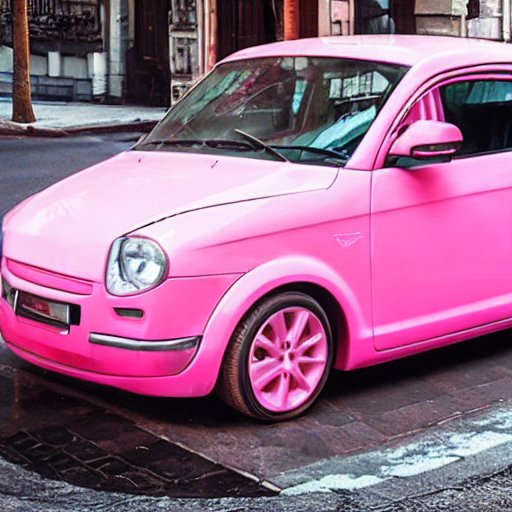}}}$ & 
        $\vcenter{\hbox{\includegraphics[width=0.136\textwidth]{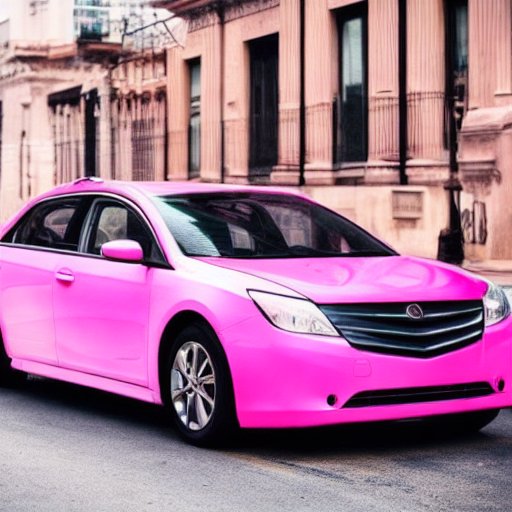}}}$ & 
        $\vcenter{\hbox{\includegraphics[width=0.136\textwidth]{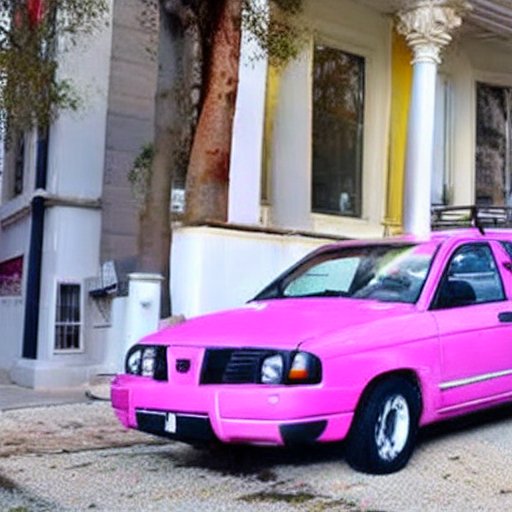}}}$ & 
        $\vcenter{\hbox{\includegraphics[width=0.136\textwidth]{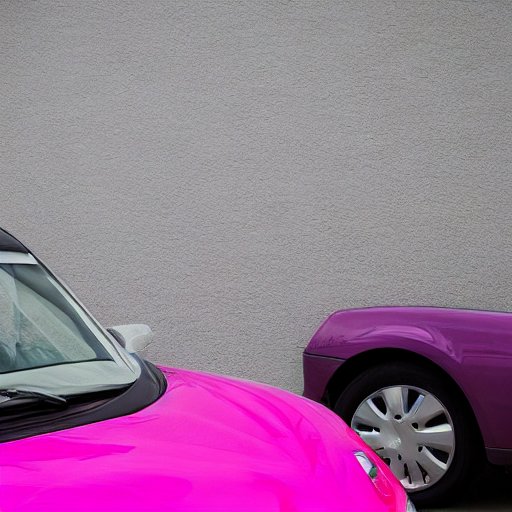}}}$ \\[\vsep]

        An umbrella on top of a spoon. & 
        $\vcenter{\hbox{\includegraphics[width=0.136\textwidth]{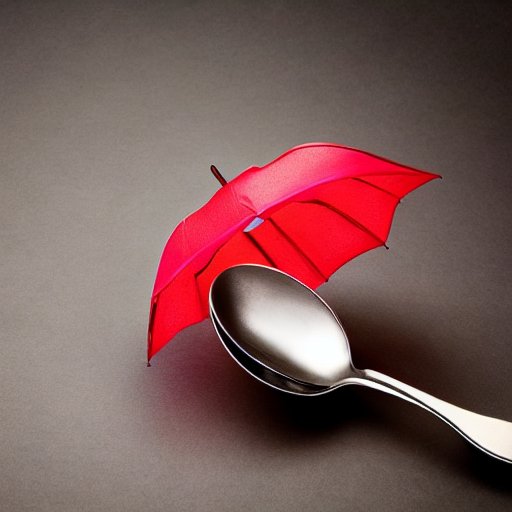}}}$ & 
        $\vcenter{\hbox{\includegraphics[width=0.136\textwidth]{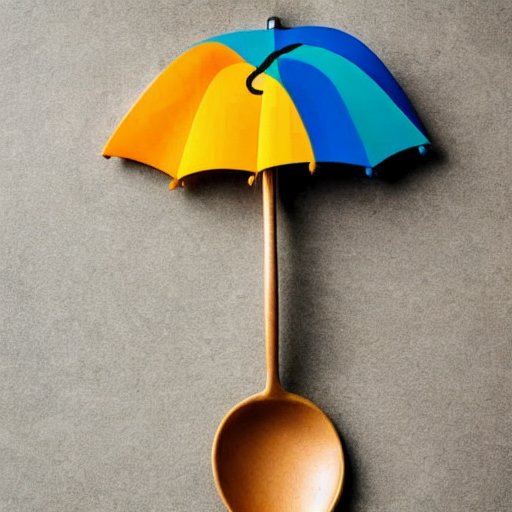}}}$ & 
        $\vcenter{\hbox{\includegraphics[width=0.136\textwidth]{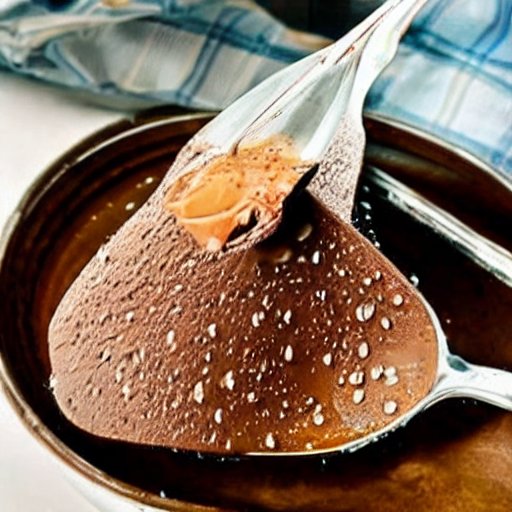}}}$ & 
        $\vcenter{\hbox{\includegraphics[width=0.136\textwidth]{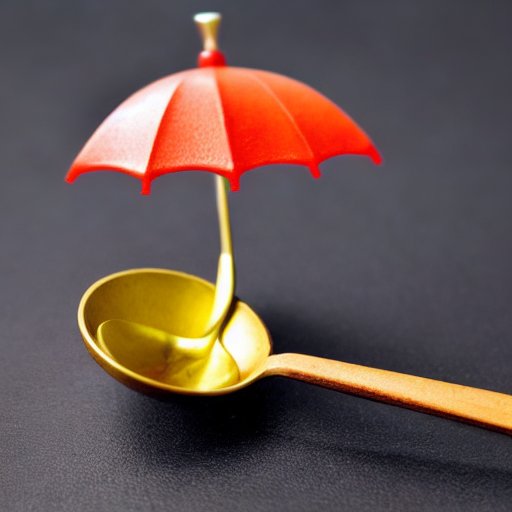}}}$ & 
        $\vcenter{\hbox{\includegraphics[width=0.136\textwidth]{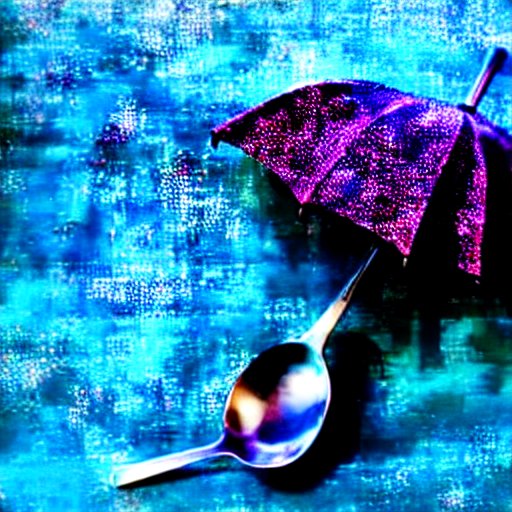}}}$ & 
        $\vcenter{\hbox{\includegraphics[width=0.136\textwidth]{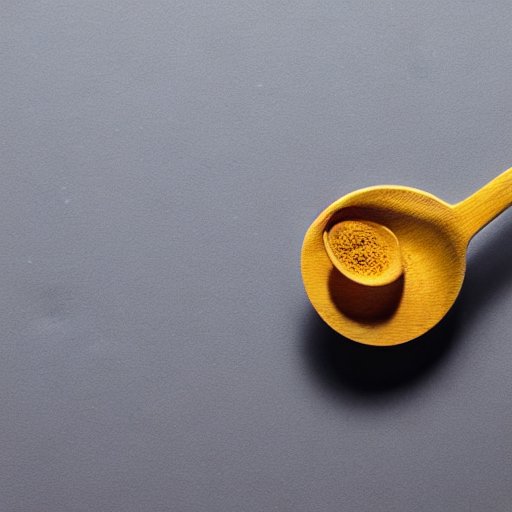}}}$ \\[\vsep]

        A single clock is sitting on a table. & 
        $\vcenter{\hbox{\includegraphics[width=0.136\textwidth]{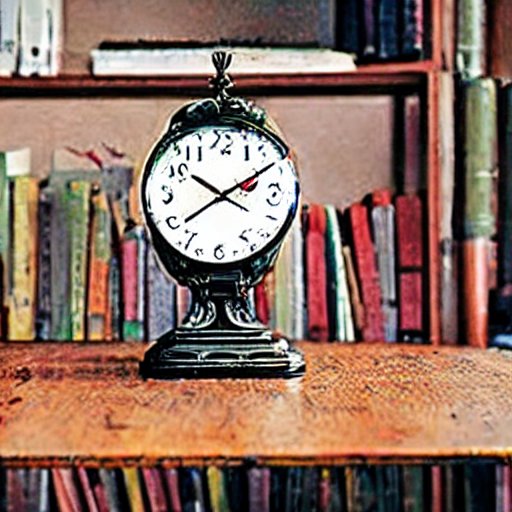}}}$ & 
        $\vcenter{\hbox{\includegraphics[width=0.136\textwidth]{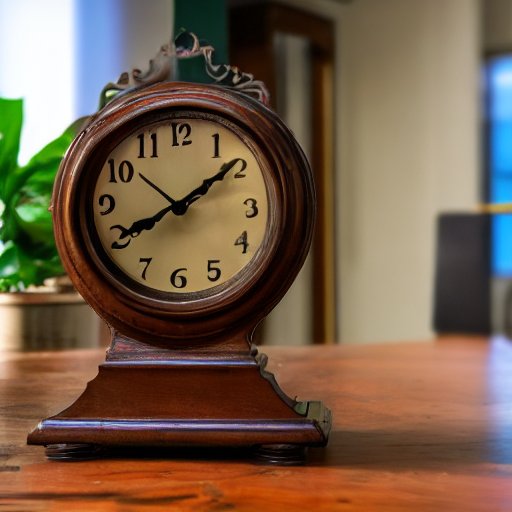}}}$ & 
        $\vcenter{\hbox{\includegraphics[width=0.136\textwidth]{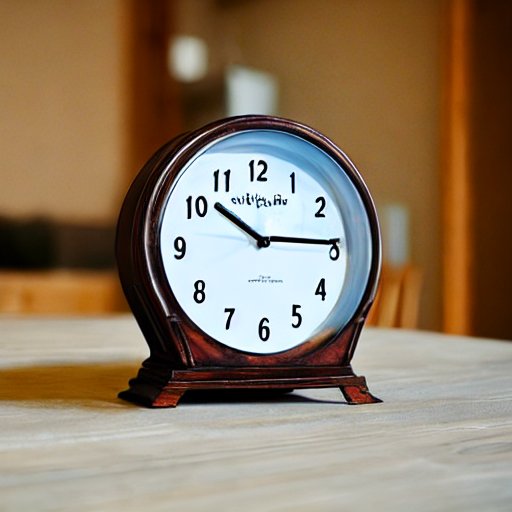}}}$ & 
        $\vcenter{\hbox{\includegraphics[width=0.136\textwidth]{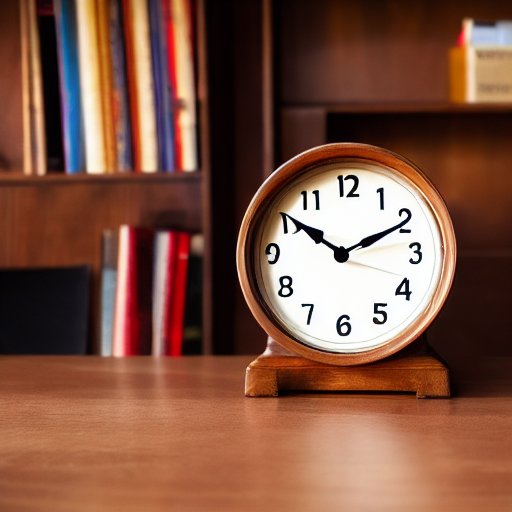}}}$ & 
        $\vcenter{\hbox{\includegraphics[width=0.136\textwidth]{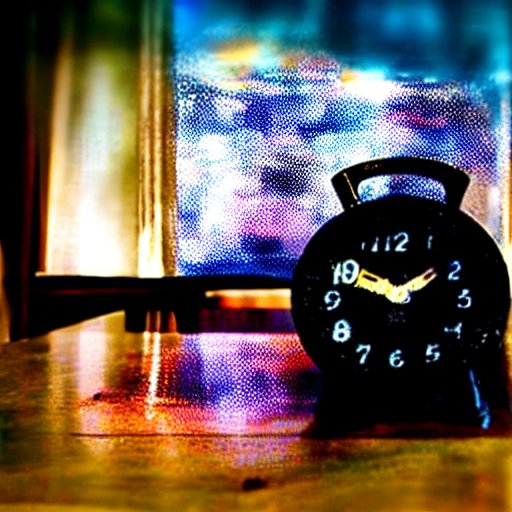}}}$ & 
        $\vcenter{\hbox{\includegraphics[width=0.136\textwidth]{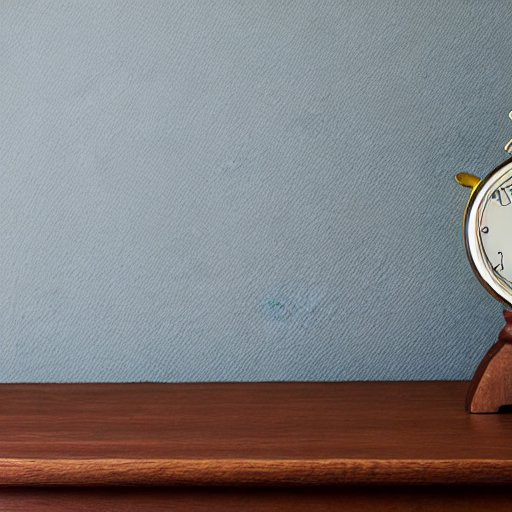}}}$ \\
        
        \bottomrule
    \end{tabular}
    \raggedright
    \caption{\textbf{Qualitative comparison (SD1.5 + ImageReward).} 10 randomly chosen prompts for different alignment methods using Stable Diffusion v1.5 and ImageReward.}
    \label{fig:app:sd_reward}
\end{figure}

\begin{figure}[htbp]
    \centering
    \definecolor{highlight}{rgb}{0.94, 0.96, 1.0}
    
    \setlength{\tabcolsep}{0pt} 
    \setlength{\aboverulesep}{2pt} 
    \setlength{\belowrulesep}{2pt}

    \setlength{\hsep}{4pt} 
    \setlength{\vsep}{28pt}
    
    \renewcommand{\arraystretch}{0} 

    \begin{tabular}{
        >{\raggedright\arraybackslash\tiny\itshape}m{1.5cm} 
        @{\hspace{\hsep}} >{\columncolor{highlight}}c 
        @{\hspace{\hsep}}c@{\hspace{\hsep}}c@{\hspace{\hsep}}c@{\hspace{\hsep}}c@{\hspace{\hsep}}c 
    }
        \toprule
        \rule{0pt}{3ex} \textbf{\scriptsize Prompt} & \textbf{\scriptsize TRS (Ours)} & \textbf{\scriptsize RS} & \textbf{\scriptsize ZO} & \textbf{\scriptsize DTS*} & \textbf{\scriptsize FD} & \textbf{\scriptsize OC-Flow} \\[1.5ex]
        \midrule
        
        A keyboard made of water, the water is made of light, the light is turned off. & 
        $\vcenter{\hbox{\includegraphics[width=0.136\textwidth]{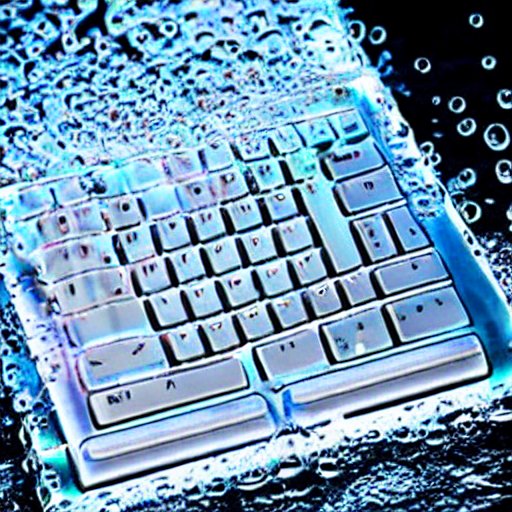}}}$ & 
        $\vcenter{\hbox{\includegraphics[width=0.136\textwidth]{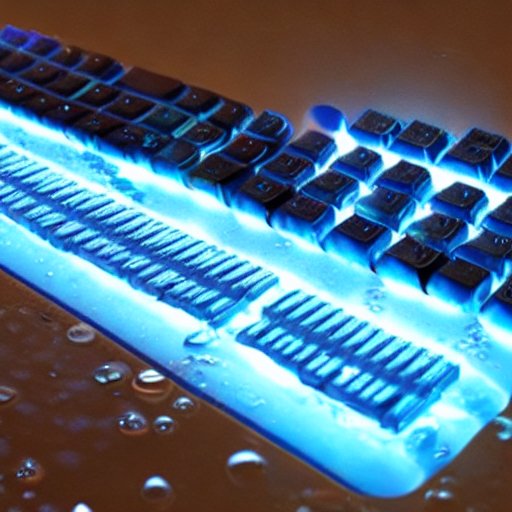}}}$ & 
        $\vcenter{\hbox{\includegraphics[width=0.136\textwidth]{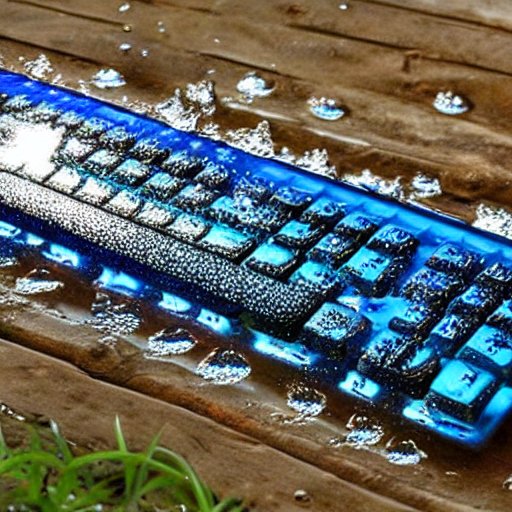}}}$ & 
        $\vcenter{\hbox{\includegraphics[width=0.136\textwidth]{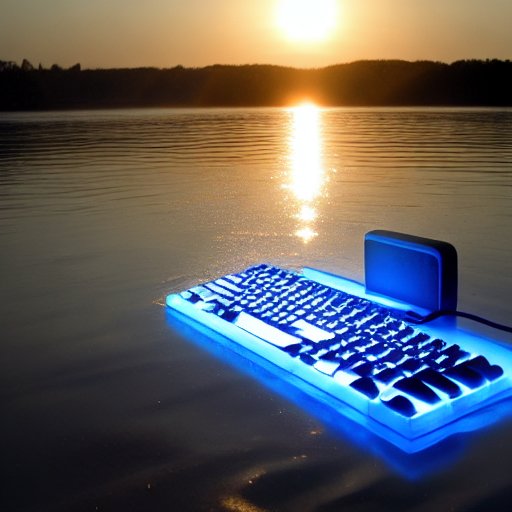}}}$ & 
        $\vcenter{\hbox{\includegraphics[width=0.136\textwidth]{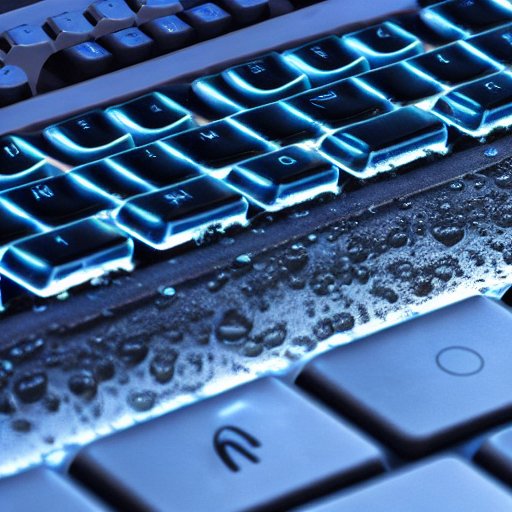}}}$ & 
        $\vcenter{\hbox{\includegraphics[width=0.136\textwidth]{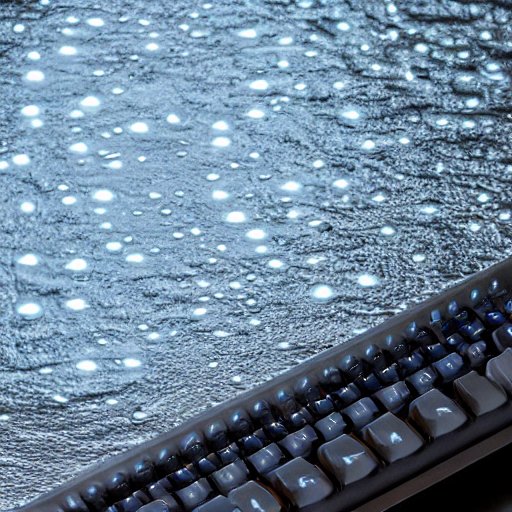}}}$ \\[\vsep]

        A red colored dog. & 
        $\vcenter{\hbox{\includegraphics[width=0.136\textwidth]{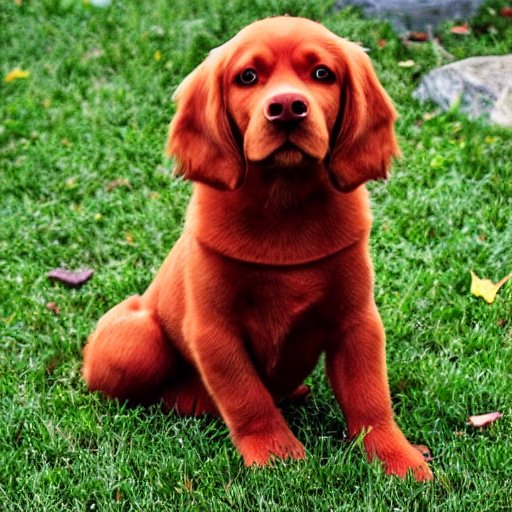}}}$ & 
        $\vcenter{\hbox{\includegraphics[width=0.136\textwidth]{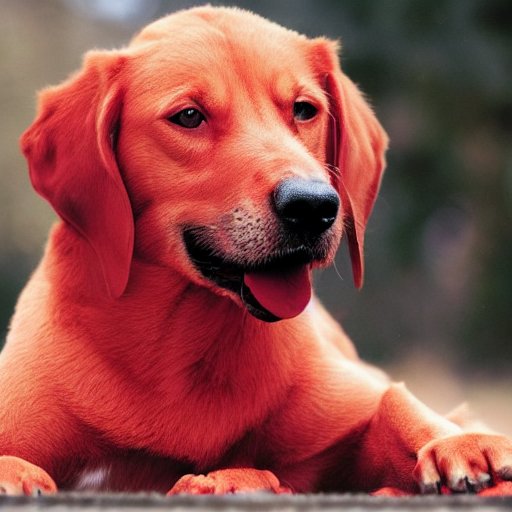}}}$ & 
        $\vcenter{\hbox{\includegraphics[width=0.136\textwidth]{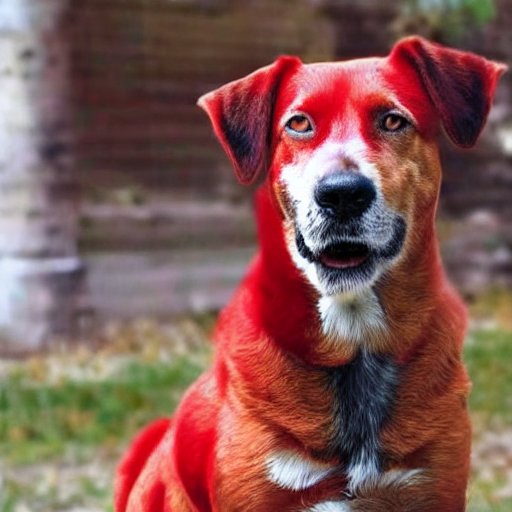}}}$ & 
        $\vcenter{\hbox{\includegraphics[width=0.136\textwidth]{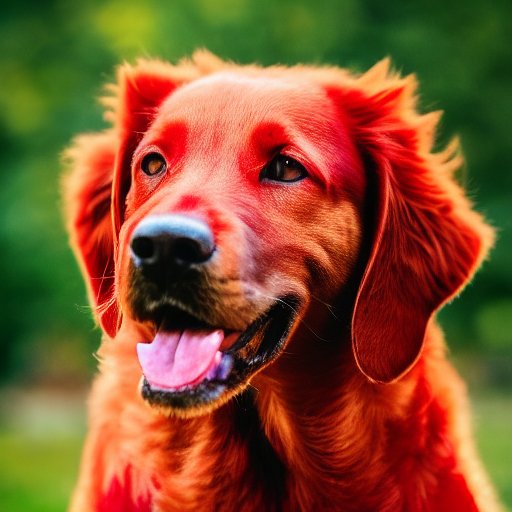}}}$ & 
        $\vcenter{\hbox{\includegraphics[width=0.136\textwidth]{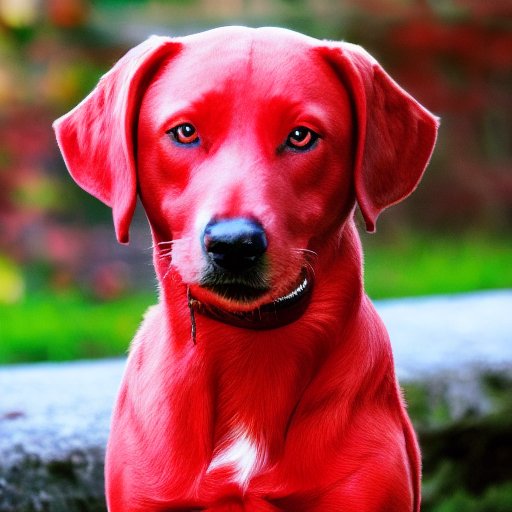}}}$ & 
        $\vcenter{\hbox{\includegraphics[width=0.136\textwidth]{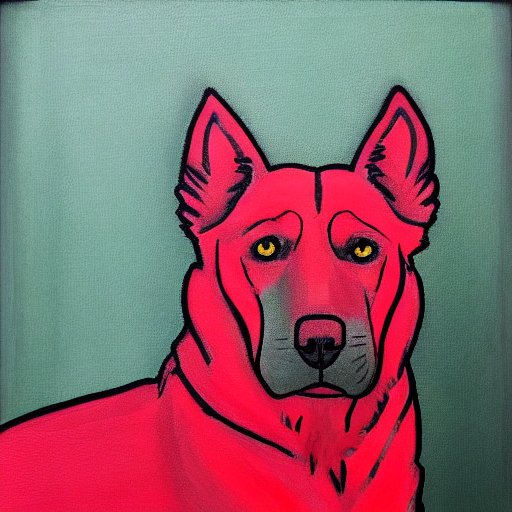}}}$ \\[\vsep]

        Hyper-realistic photo of an abandoned industrial site during a storm. & 
        $\vcenter{\hbox{\includegraphics[width=0.136\textwidth]{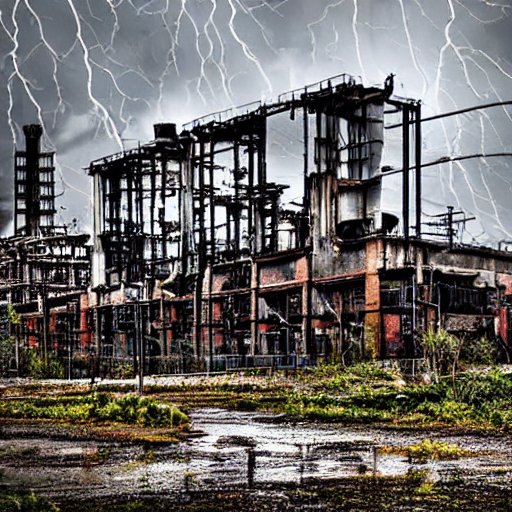}}}$ & 
        $\vcenter{\hbox{\includegraphics[width=0.136\textwidth]{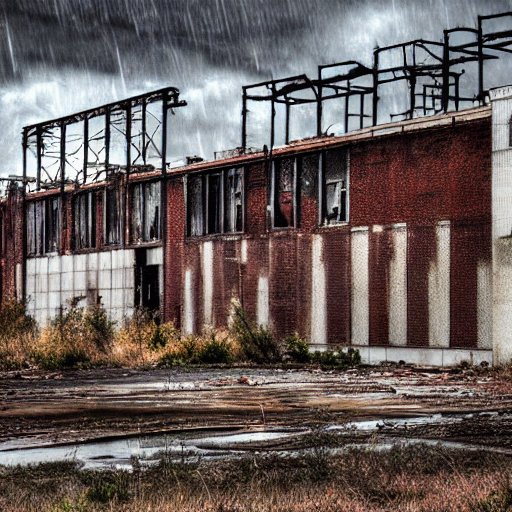}}}$ & 
        $\vcenter{\hbox{\includegraphics[width=0.136\textwidth]{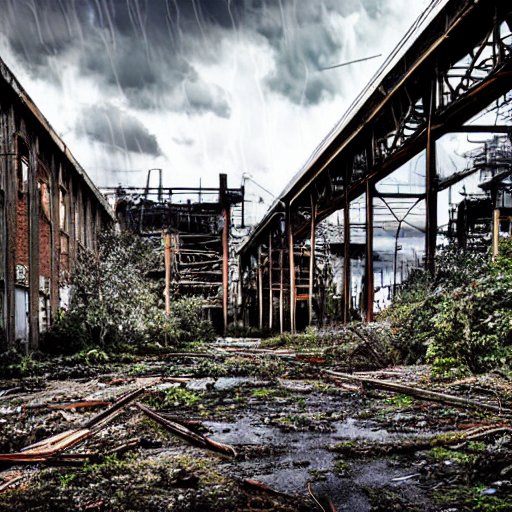}}}$ & 
        $\vcenter{\hbox{\includegraphics[width=0.136\textwidth]{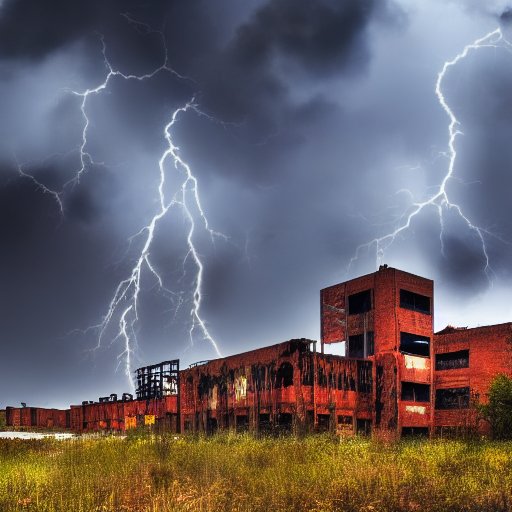}}}$ & 
        $\vcenter{\hbox{\includegraphics[width=0.136\textwidth]{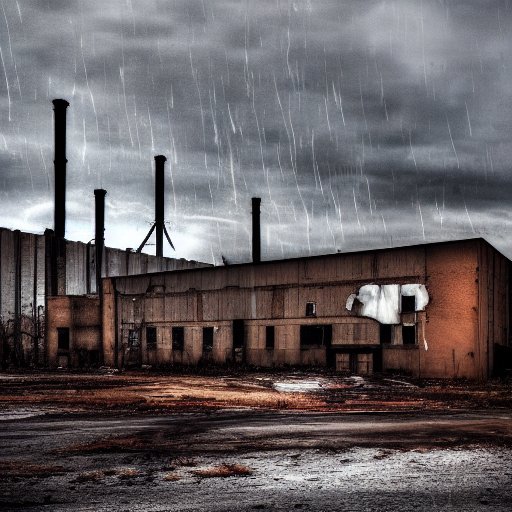}}}$ & 
        $\vcenter{\hbox{\includegraphics[width=0.136\textwidth]{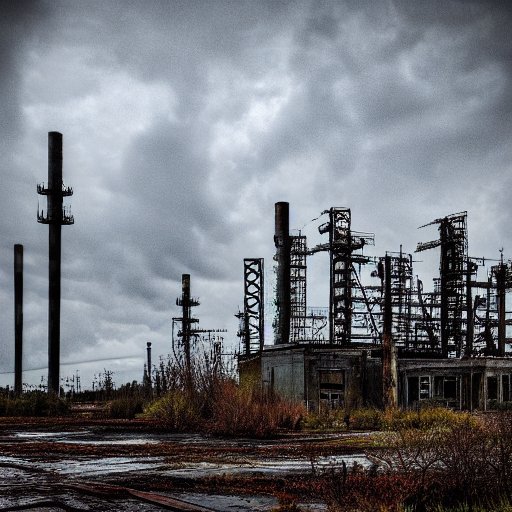}}}$ \\[\vsep]

        A cat on the left of a dog. & 
        $\vcenter{\hbox{\includegraphics[width=0.136\textwidth]{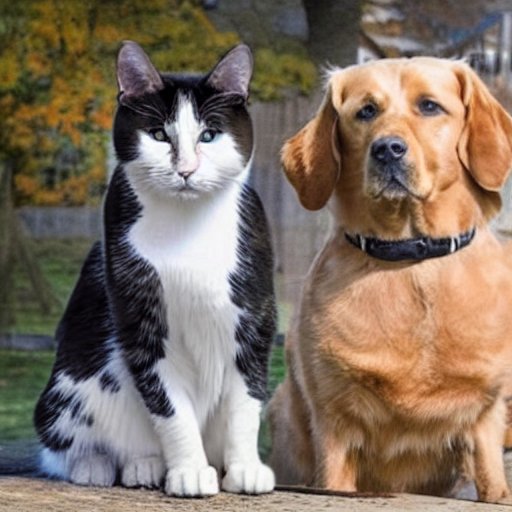}}}$ & 
        $\vcenter{\hbox{\includegraphics[width=0.136\textwidth]{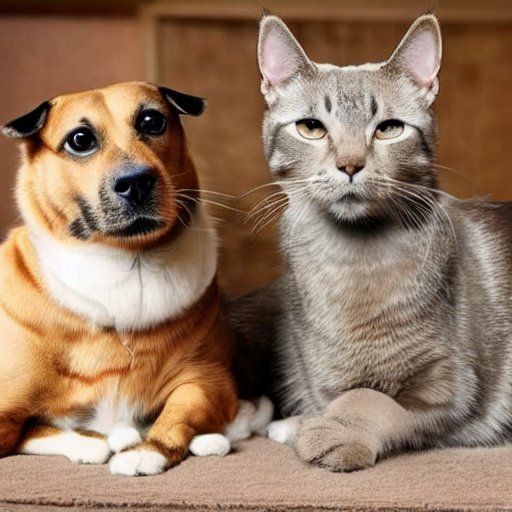}}}$ & 
        $\vcenter{\hbox{\includegraphics[width=0.136\textwidth]{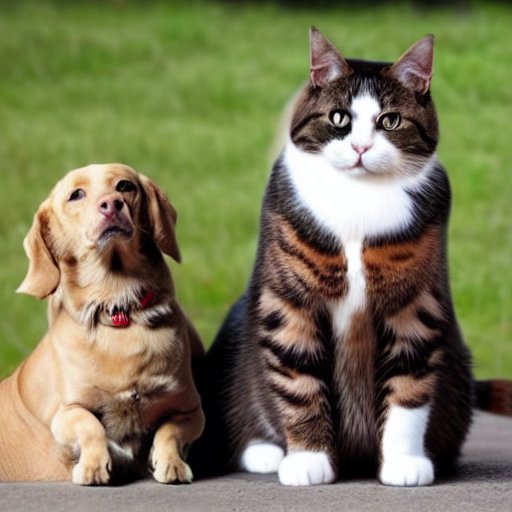}}}$ & 
        $\vcenter{\hbox{\includegraphics[width=0.136\textwidth]{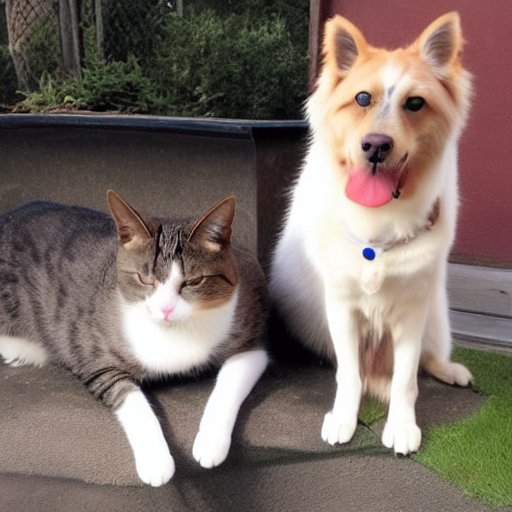}}}$ & 
        $\vcenter{\hbox{\includegraphics[width=0.136\textwidth]{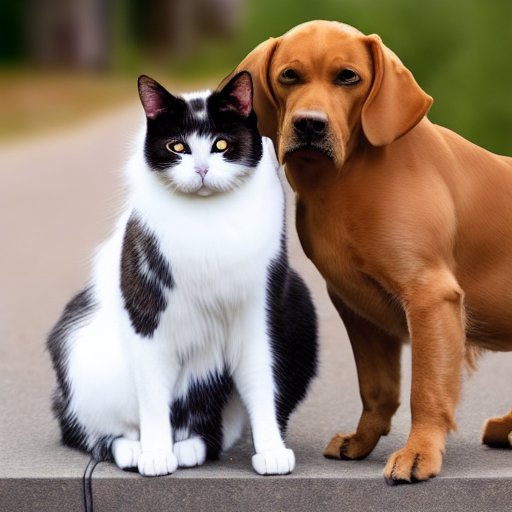}}}$ & 
        $\vcenter{\hbox{\includegraphics[width=0.136\textwidth]{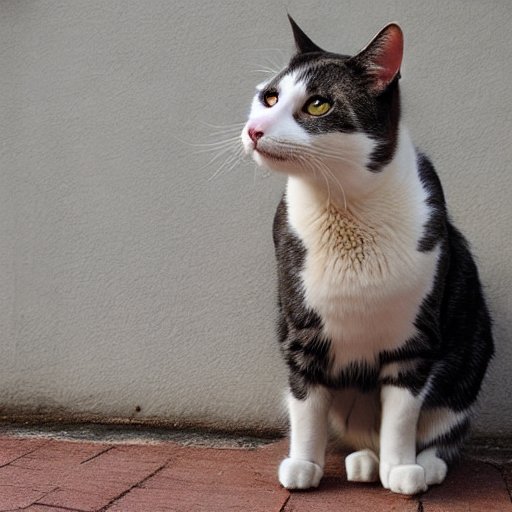}}}$ \\[\vsep]

        A giraffe underneath a microwave. & 
        $\vcenter{\hbox{\includegraphics[width=0.136\textwidth]{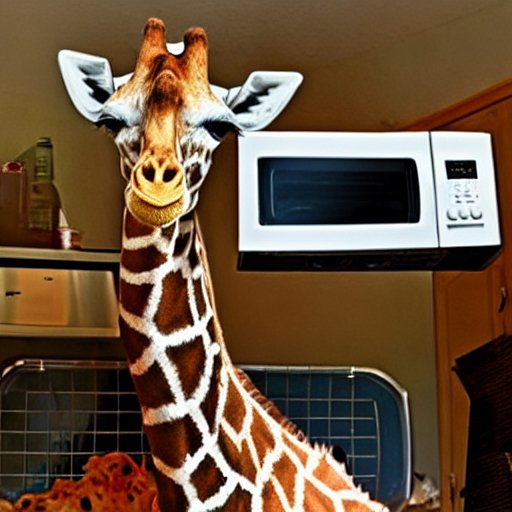}}}$ & 
        $\vcenter{\hbox{\includegraphics[width=0.136\textwidth]{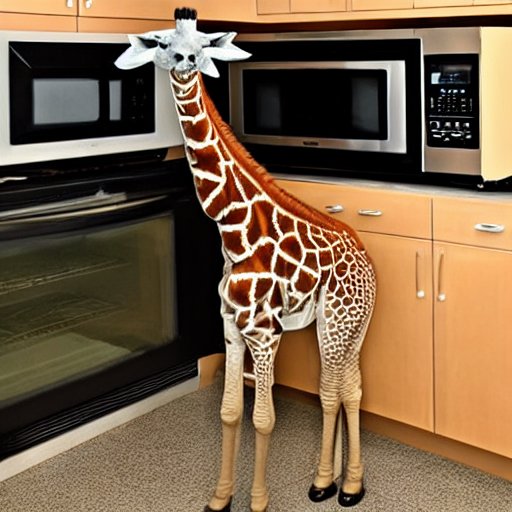}}}$ & 
        $\vcenter{\hbox{\includegraphics[width=0.136\textwidth]{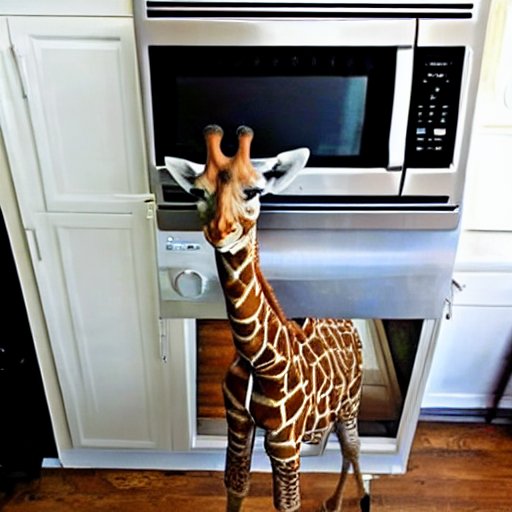}}}$ & 
        $\vcenter{\hbox{\includegraphics[width=0.136\textwidth]{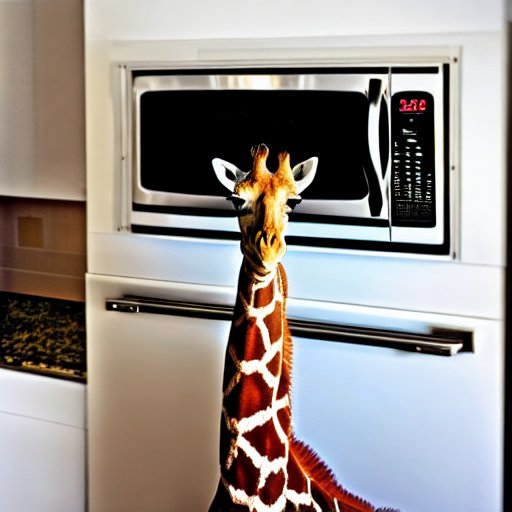}}}$ & 
        $\vcenter{\hbox{\includegraphics[width=0.136\textwidth]{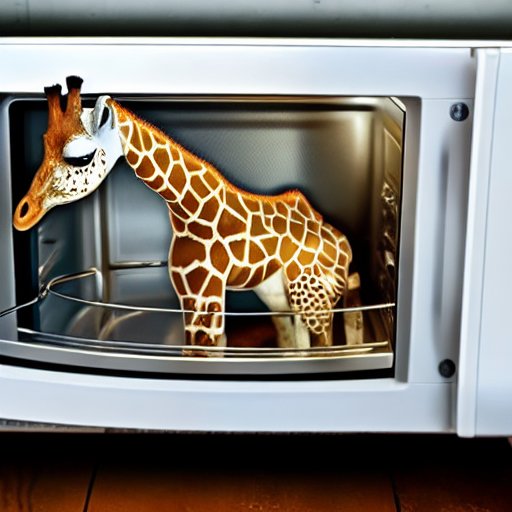}}}$ & 
        $\vcenter{\hbox{\includegraphics[width=0.136\textwidth]{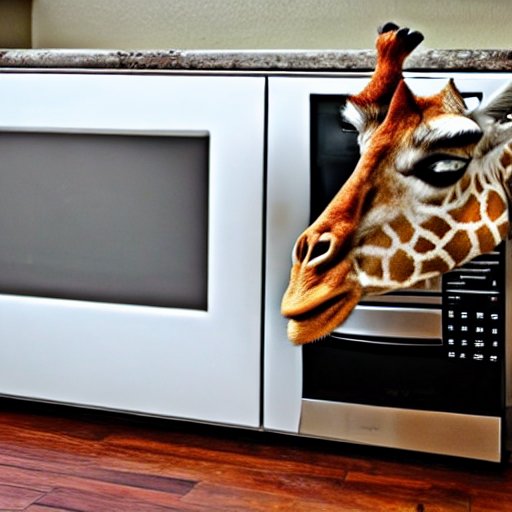}}}$ \\[\vsep]

        A baby fennec sneezing onto a strawberry, detailed, macro, studio light, droplets, backlit ears. & 
        $\vcenter{\hbox{\includegraphics[width=0.136\textwidth]{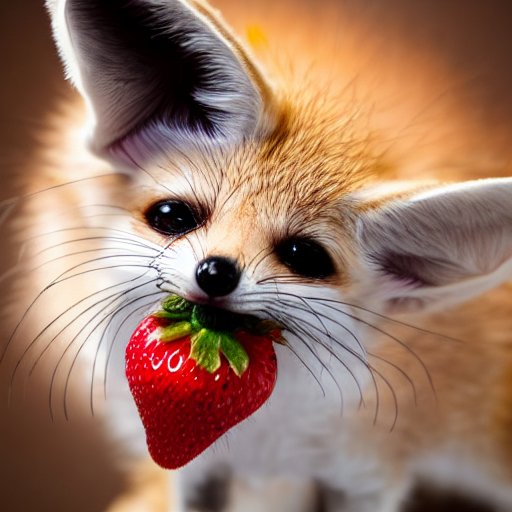}}}$ & 
        $\vcenter{\hbox{\includegraphics[width=0.136\textwidth]{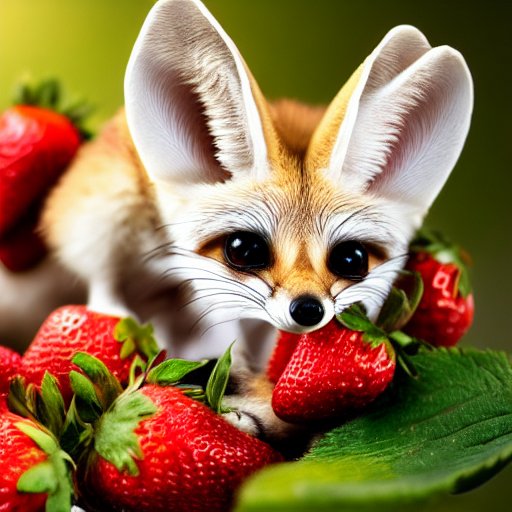}}}$ & 
        $\vcenter{\hbox{\includegraphics[width=0.136\textwidth]{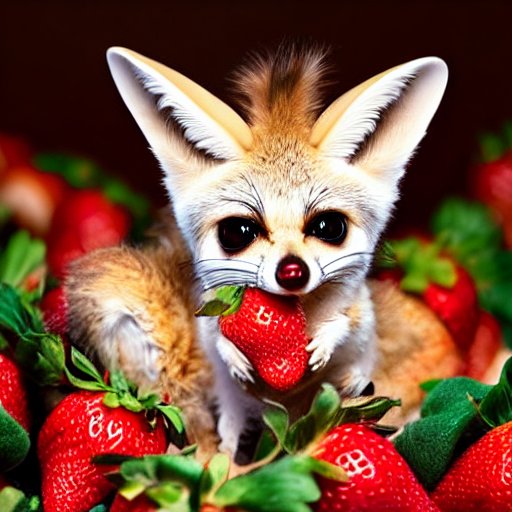}}}$ & 
        $\vcenter{\hbox{\includegraphics[width=0.136\textwidth]{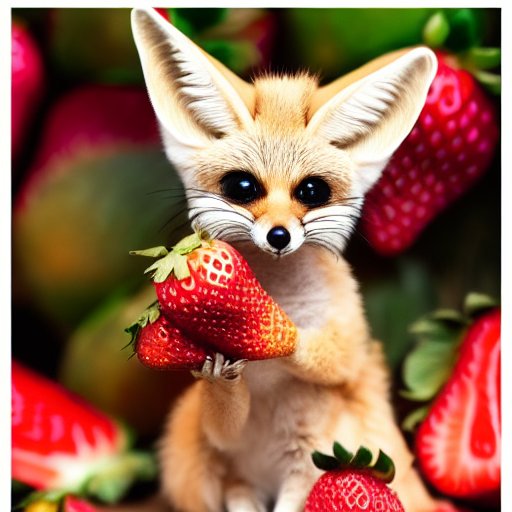}}}$ & 
        $\vcenter{\hbox{\includegraphics[width=0.136\textwidth]{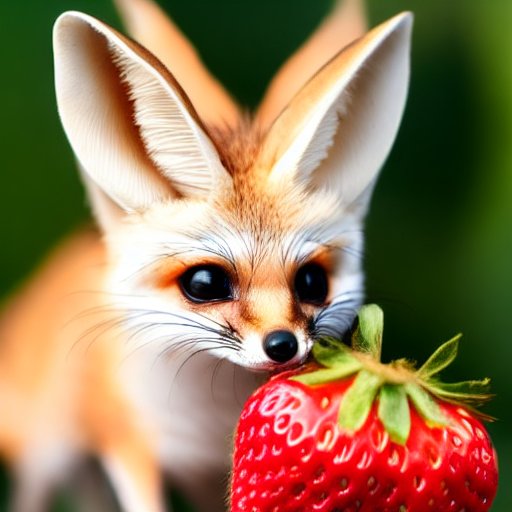}}}$ & 
        $\vcenter{\hbox{\includegraphics[width=0.136\textwidth]{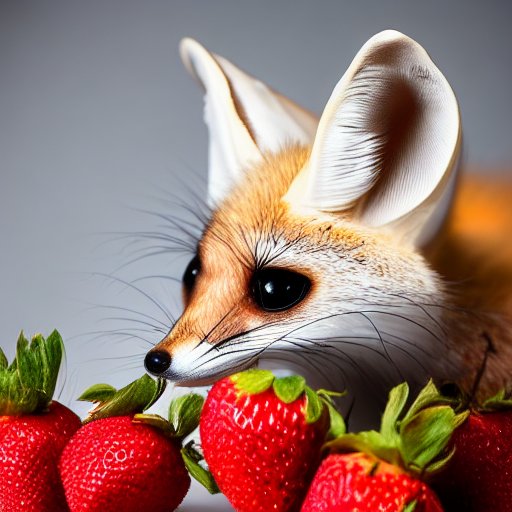}}}$ \\[\vsep]

        One car on the street. & 
        $\vcenter{\hbox{\includegraphics[width=0.136\textwidth]{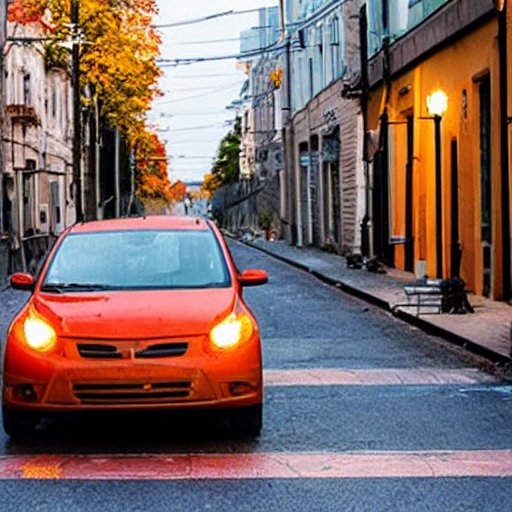}}}$ & 
        $\vcenter{\hbox{\includegraphics[width=0.136\textwidth]{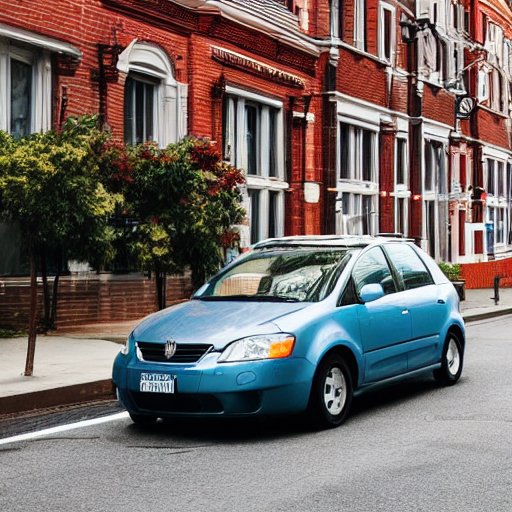}}}$ & 
        $\vcenter{\hbox{\includegraphics[width=0.136\textwidth]{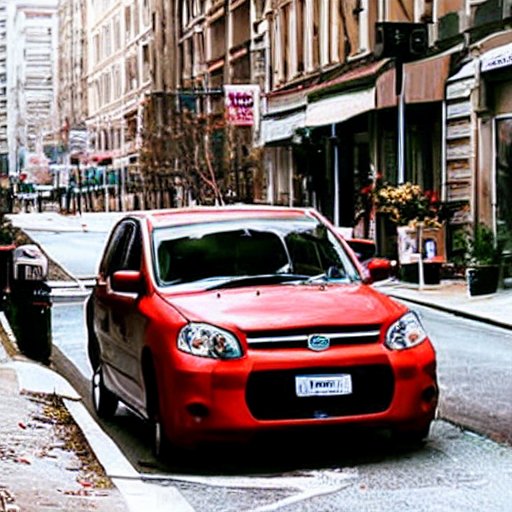}}}$ & 
        $\vcenter{\hbox{\includegraphics[width=0.136\textwidth]{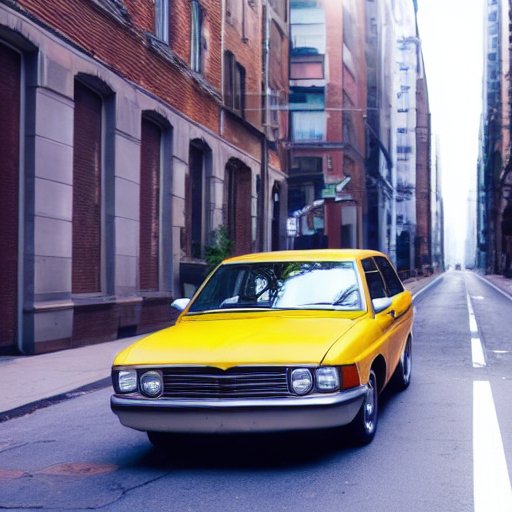}}}$ & 
        $\vcenter{\hbox{\includegraphics[width=0.136\textwidth]{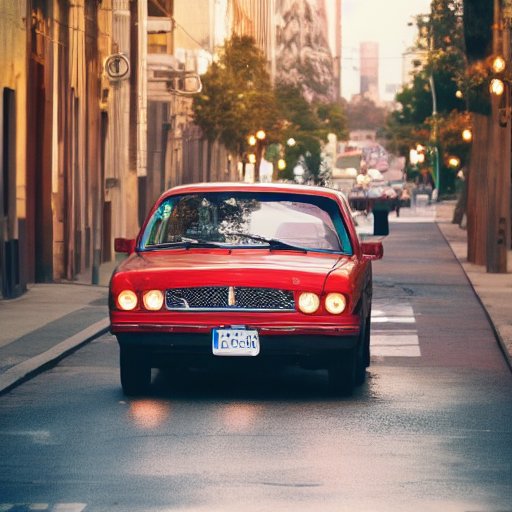}}}$ & 
        $\vcenter{\hbox{\includegraphics[width=0.136\textwidth]{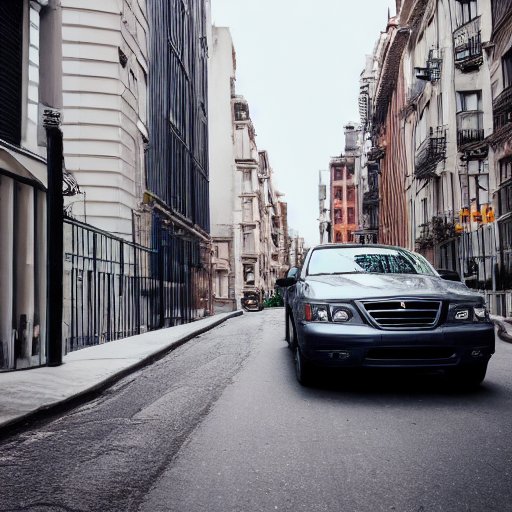}}}$ \\[\vsep]

        A pink colored car. & 
        $\vcenter{\hbox{\includegraphics[width=0.136\textwidth]{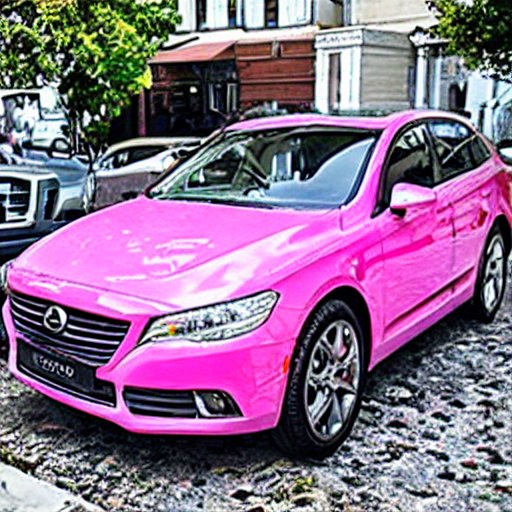}}}$ & 
        $\vcenter{\hbox{\includegraphics[width=0.136\textwidth]{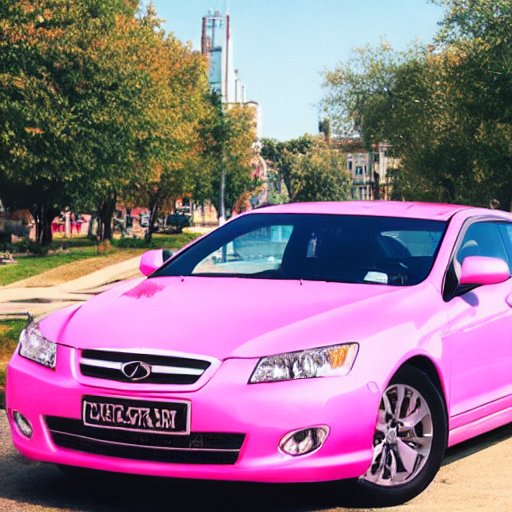}}}$ & 
        $\vcenter{\hbox{\includegraphics[width=0.136\textwidth]{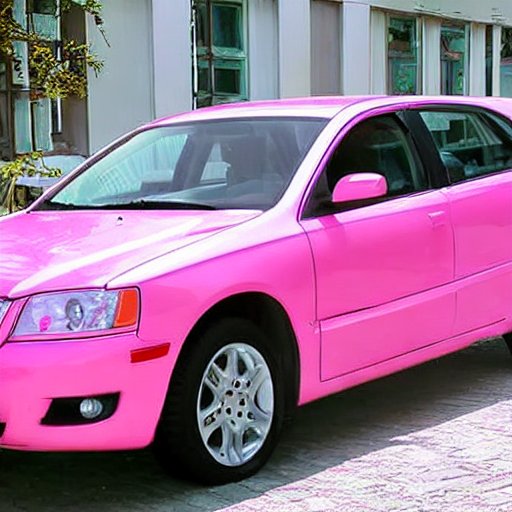}}}$ & 
        $\vcenter{\hbox{\includegraphics[width=0.136\textwidth]{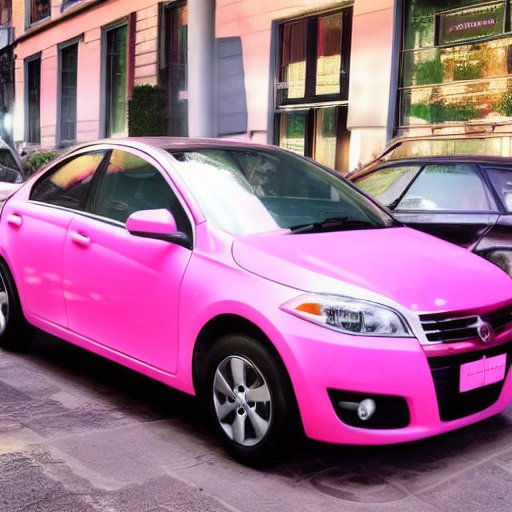}}}$ & 
        $\vcenter{\hbox{\includegraphics[width=0.136\textwidth]{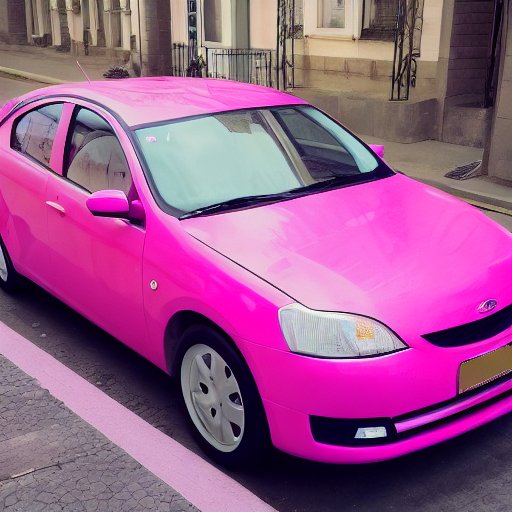}}}$ & 
        $\vcenter{\hbox{\includegraphics[width=0.136\textwidth]{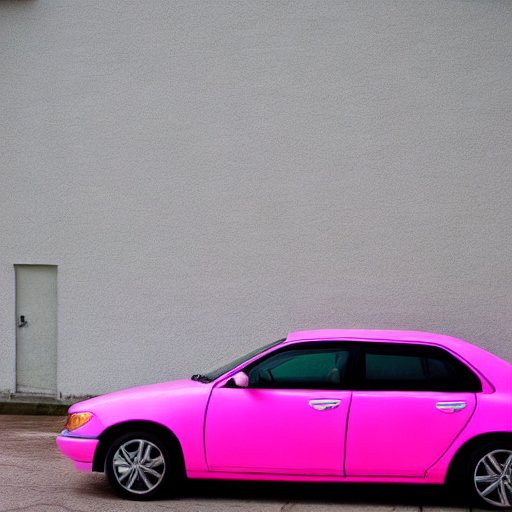}}}$ \\[\vsep]

        An umbrella on top of a spoon. & 
        $\vcenter{\hbox{\includegraphics[width=0.136\textwidth]{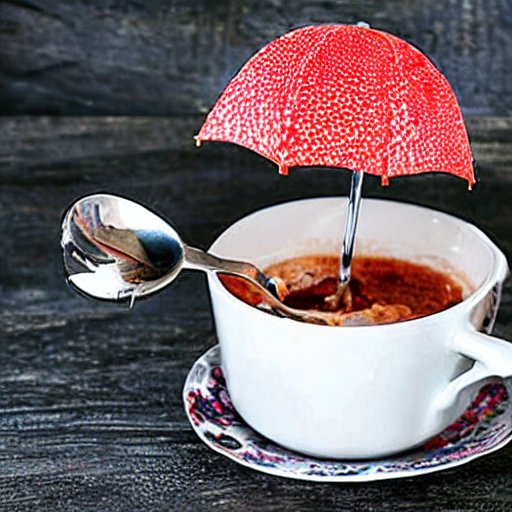}}}$ & 
        $\vcenter{\hbox{\includegraphics[width=0.136\textwidth]{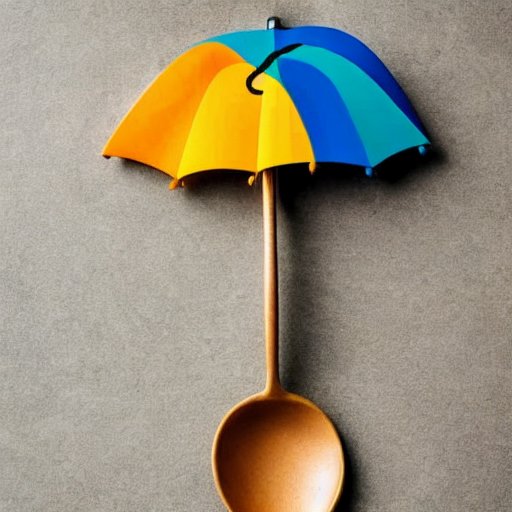}}}$ & 
        $\vcenter{\hbox{\includegraphics[width=0.136\textwidth]{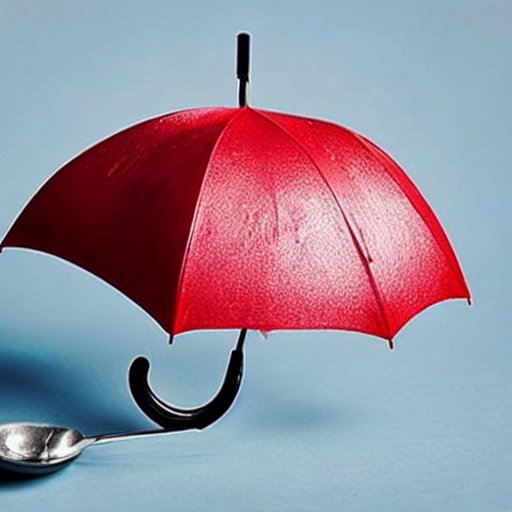}}}$ & 
        $\vcenter{\hbox{\includegraphics[width=0.136\textwidth]{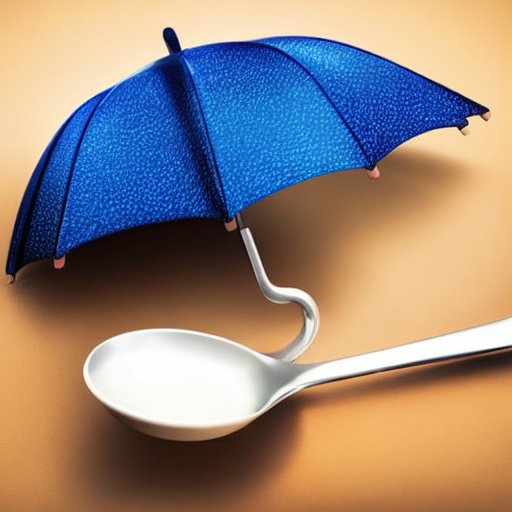}}}$ & 
        $\vcenter{\hbox{\includegraphics[width=0.136\textwidth]{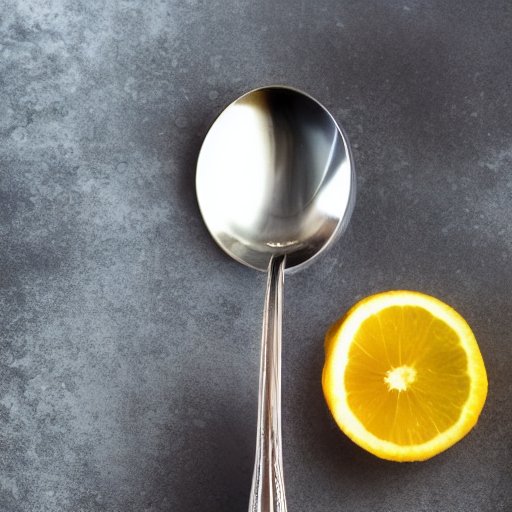}}}$ & 
        $\vcenter{\hbox{\includegraphics[width=0.136\textwidth]{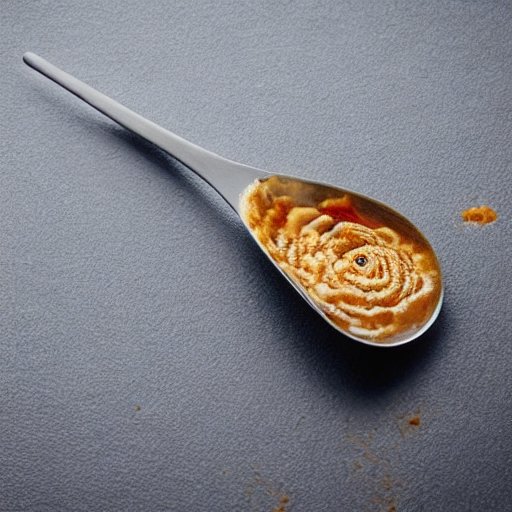}}}$ \\[\vsep]

        A single clock is sitting on a table. & 
        $\vcenter{\hbox{\includegraphics[width=0.136\textwidth]{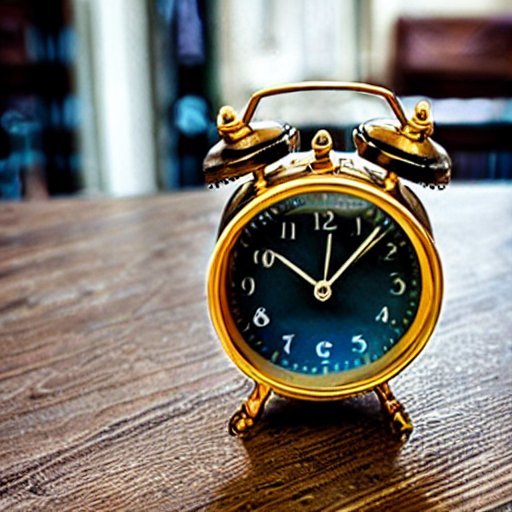}}}$ & 
        $\vcenter{\hbox{\includegraphics[width=0.136\textwidth]{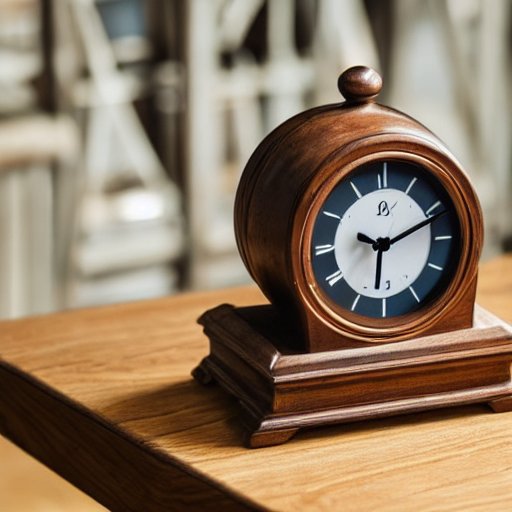}}}$ & 
        $\vcenter{\hbox{\includegraphics[width=0.136\textwidth]{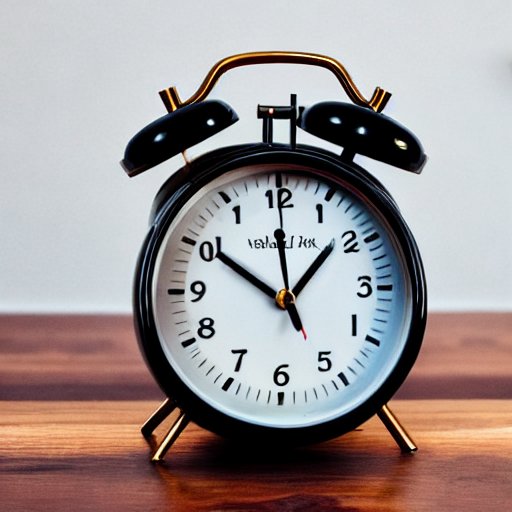}}}$ & 
        $\vcenter{\hbox{\includegraphics[width=0.136\textwidth]{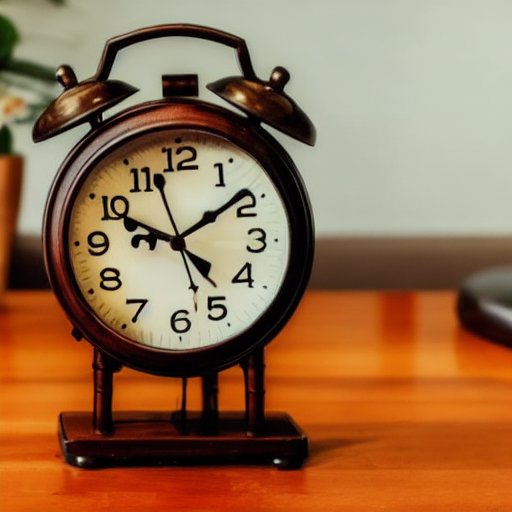}}}$ & 
        $\vcenter{\hbox{\includegraphics[width=0.136\textwidth]{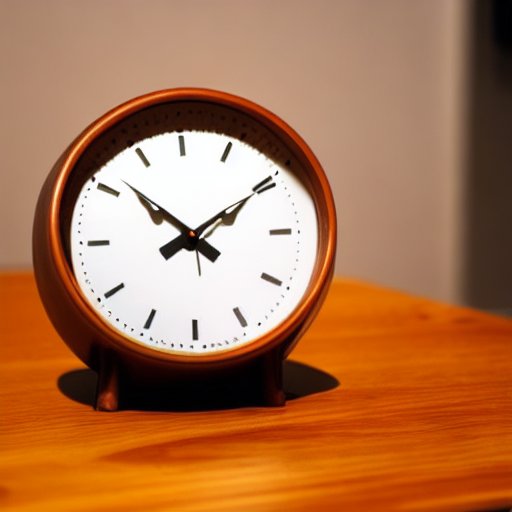}}}$ & 
        $\vcenter{\hbox{\includegraphics[width=0.136\textwidth]{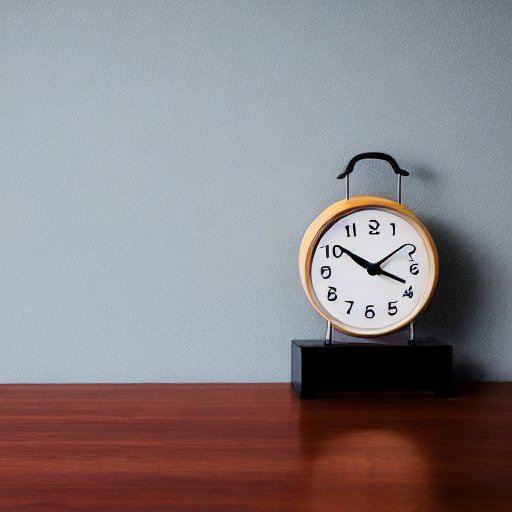}}}$ \\
        
        \bottomrule
    \end{tabular}
    \raggedright
    \caption{\textbf{Qualitative comparison (SD1.5 + HPSv2).} 10 randomly chosen prompts for different alignment methods using Stable Diffusion v1.5 and HPSv2.}
    \label{fig:app:sd_hps}
\end{figure}

\begin{figure}[htbp]
    \centering
    \definecolor{highlight}{rgb}{0.94, 0.96, 1.0}
    
    \setlength{\tabcolsep}{0pt} 
    \setlength{\aboverulesep}{2pt} 
    \setlength{\belowrulesep}{2pt}

    \setlength{\hsep}{4pt} 
    \setlength{\vsep}{28pt}
    
    \renewcommand{\arraystretch}{0} 

    \begin{tabular}{
        >{\raggedright\arraybackslash\tiny\itshape}m{1.5cm} 
        @{\hspace{\hsep}} >{\columncolor{highlight}}c 
        @{\hspace{\hsep}}c@{\hspace{\hsep}}c@{\hspace{\hsep}}c@{\hspace{\hsep}}c@{\hspace{\hsep}}c 
    }
        \toprule
        \rule{0pt}{3ex} \textbf{\scriptsize Prompt} & \textbf{\scriptsize TRS (Ours)} & \textbf{\scriptsize RS} & \textbf{\scriptsize ZO} & \textbf{\scriptsize DTS*} & \textbf{\scriptsize FD} & \textbf{\scriptsize OC-Flow} \\[1.5ex]
        \midrule
        
        A keyboard made of water, the water is made of light, the light is turned off. & 
        $\vcenter{\hbox{\includegraphics[width=0.136\textwidth]{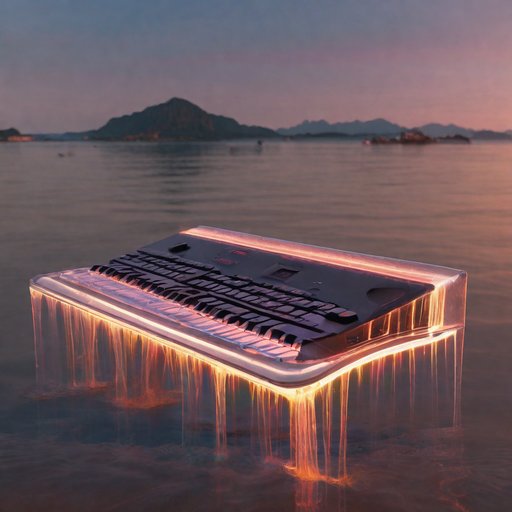}}}$ & 
        $\vcenter{\hbox{\includegraphics[width=0.136\textwidth]{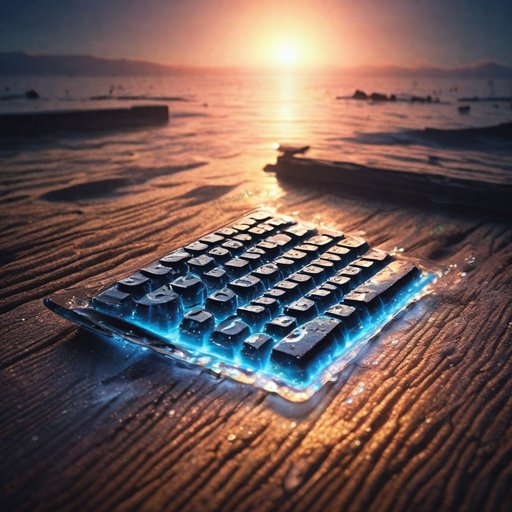}}}$ & 
        $\vcenter{\hbox{\includegraphics[width=0.136\textwidth]{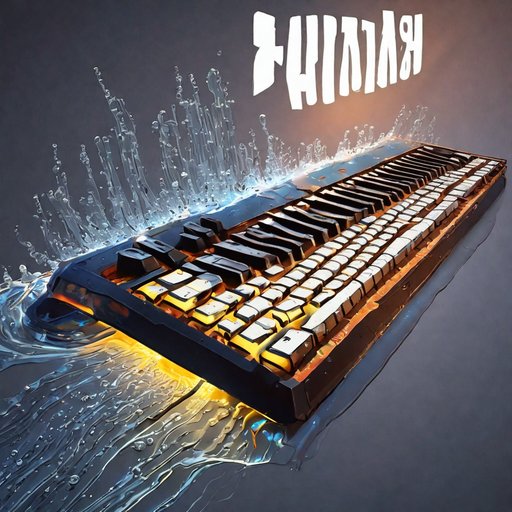}}}$ & 
        $\vcenter{\hbox{\includegraphics[width=0.136\textwidth]{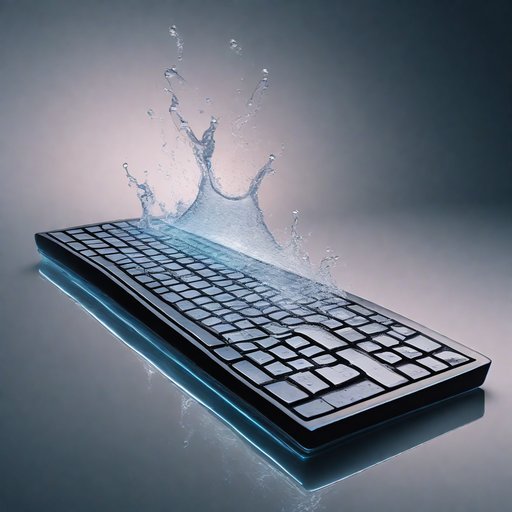}}}$ & 
        $\vcenter{\hbox{\includegraphics[width=0.136\textwidth]{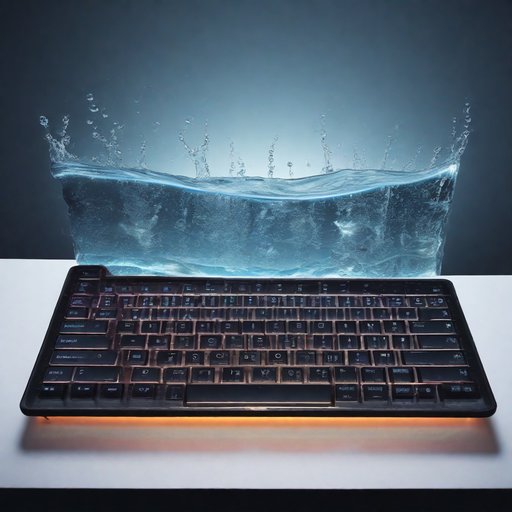}}}$ & 
        $\vcenter{\hbox{\includegraphics[width=0.136\textwidth]{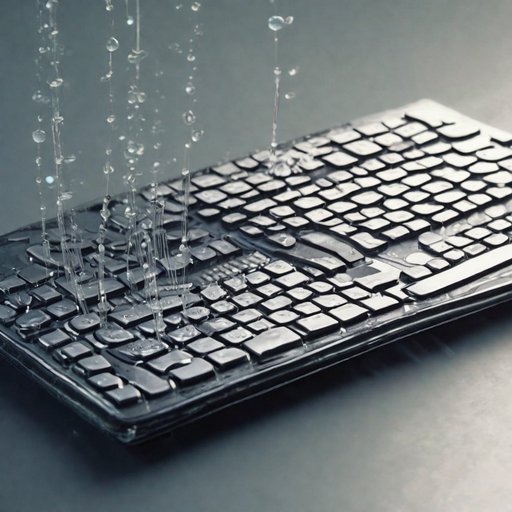}}}$ \\[\vsep]

        A red colored dog. & 
        $\vcenter{\hbox{\includegraphics[width=0.136\textwidth]{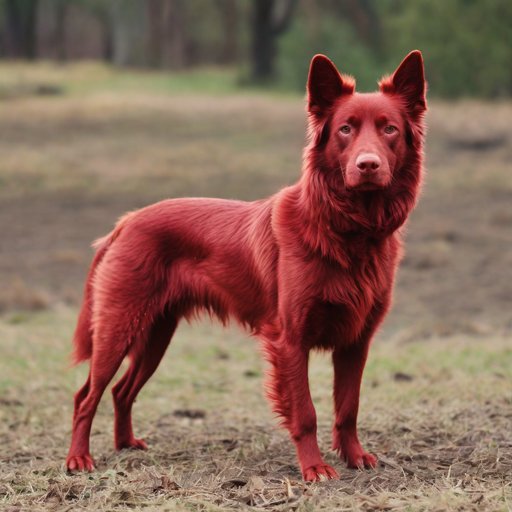}}}$ & 
        $\vcenter{\hbox{\includegraphics[width=0.136\textwidth]{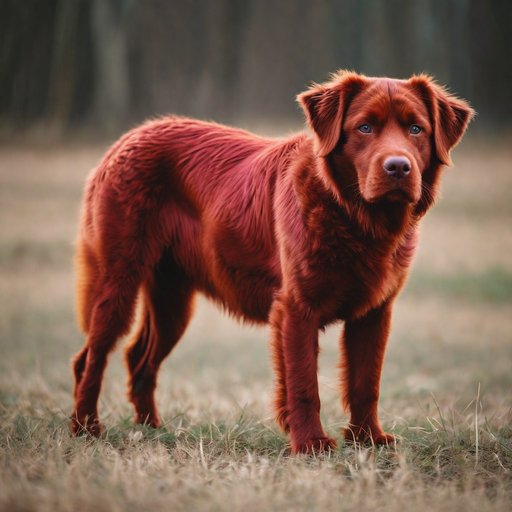}}}$ & 
        $\vcenter{\hbox{\includegraphics[width=0.136\textwidth]{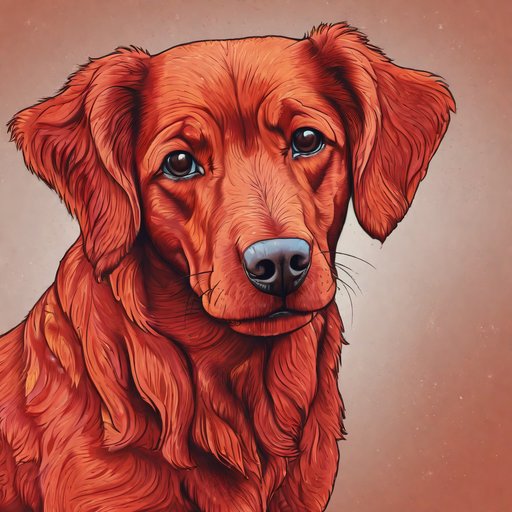}}}$ & 
        $\vcenter{\hbox{\includegraphics[width=0.136\textwidth]{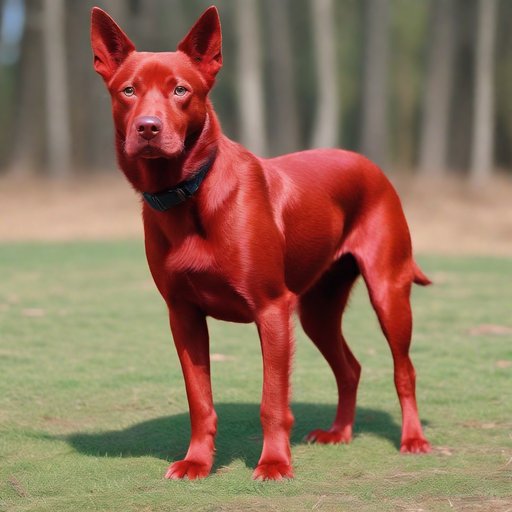}}}$ & 
        $\vcenter{\hbox{\includegraphics[width=0.136\textwidth]{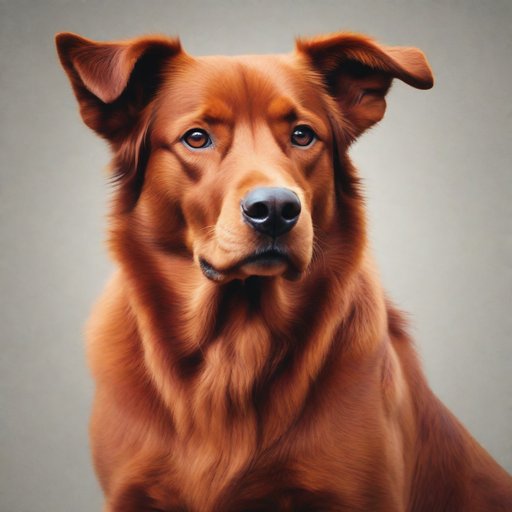}}}$ & 
        $\vcenter{\hbox{\includegraphics[width=0.136\textwidth]{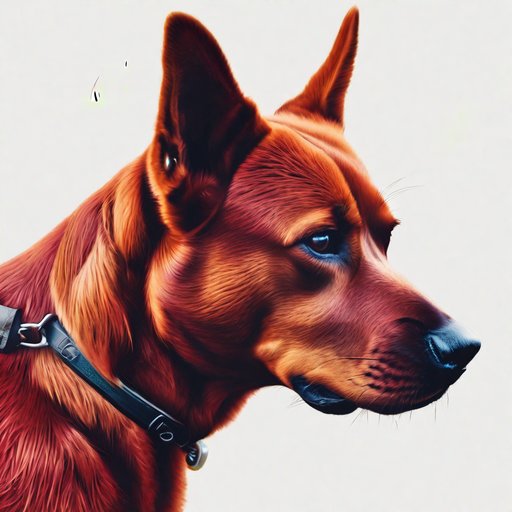}}}$ \\[\vsep]

        Hyper-realistic photo of an abandoned industrial site during a storm. & 
        $\vcenter{\hbox{\includegraphics[width=0.136\textwidth]{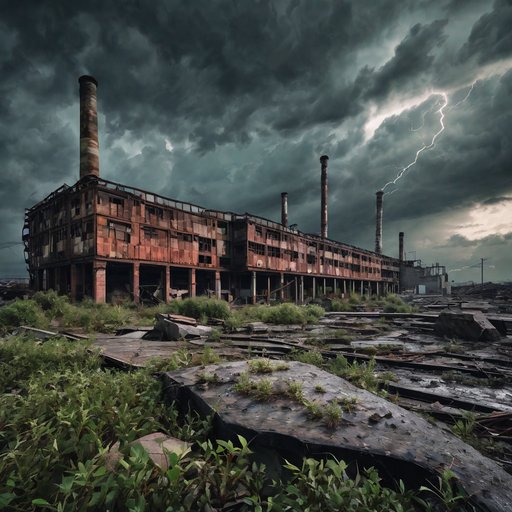}}}$ & 
        $\vcenter{\hbox{\includegraphics[width=0.136\textwidth]{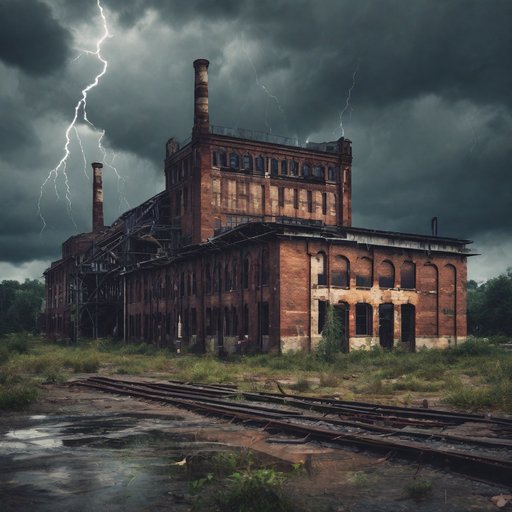}}}$ & 
        $\vcenter{\hbox{\includegraphics[width=0.136\textwidth]{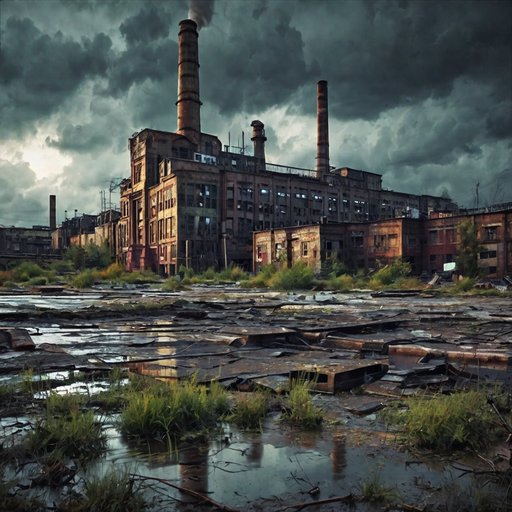}}}$ & 
        $\vcenter{\hbox{\includegraphics[width=0.136\textwidth]{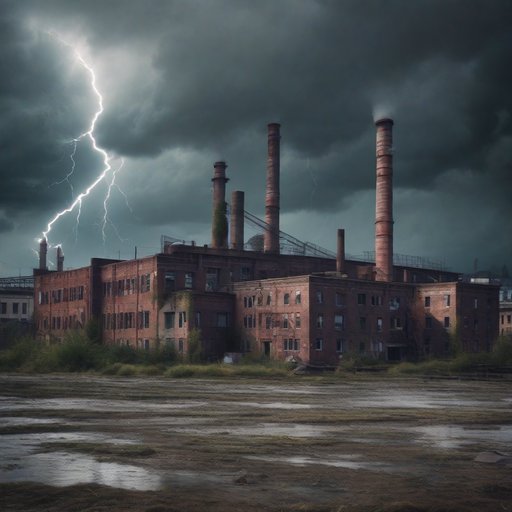}}}$ & 
        $\vcenter{\hbox{\includegraphics[width=0.136\textwidth]{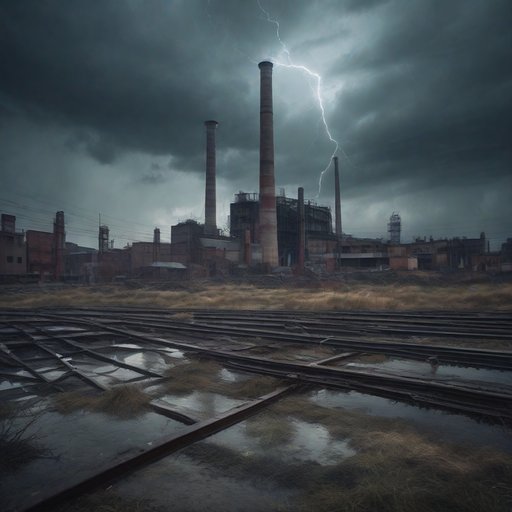}}}$ & 
        $\vcenter{\hbox{\includegraphics[width=0.136\textwidth]{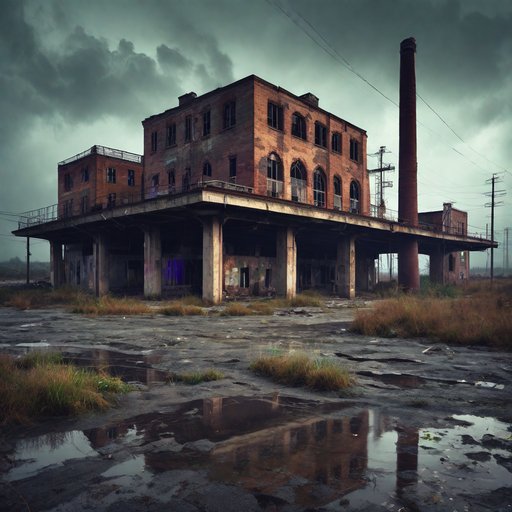}}}$ \\[\vsep]

        A cat on the left of a dog. & 
        $\vcenter{\hbox{\includegraphics[width=0.136\textwidth]{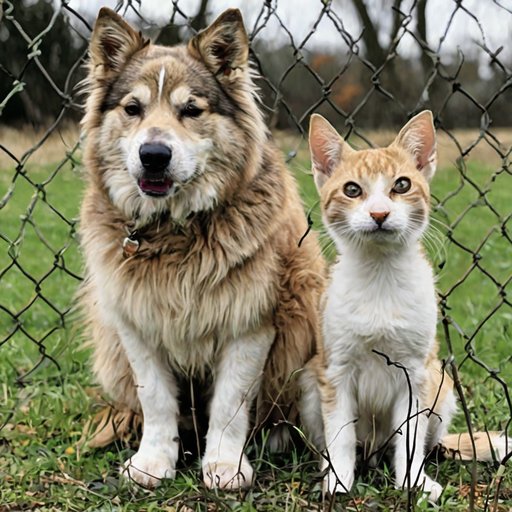}}}$ & 
        $\vcenter{\hbox{\includegraphics[width=0.136\textwidth]{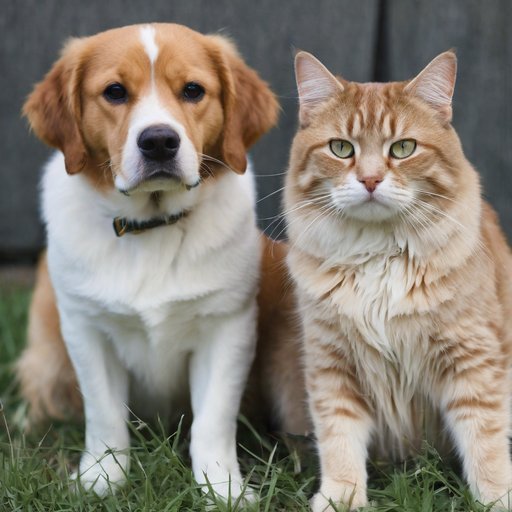}}}$ & 
        $\vcenter{\hbox{\includegraphics[width=0.136\textwidth]{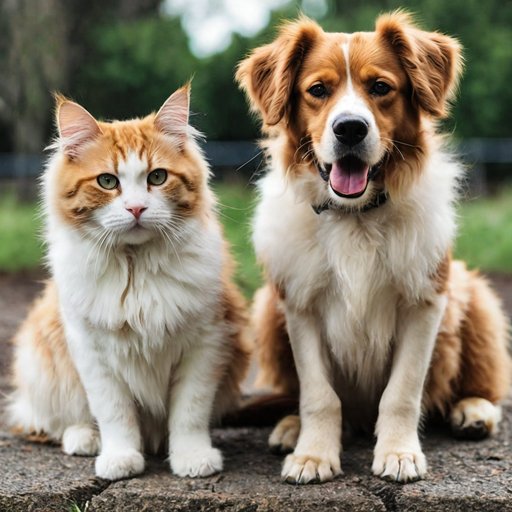}}}$ & 
        $\vcenter{\hbox{\includegraphics[width=0.136\textwidth]{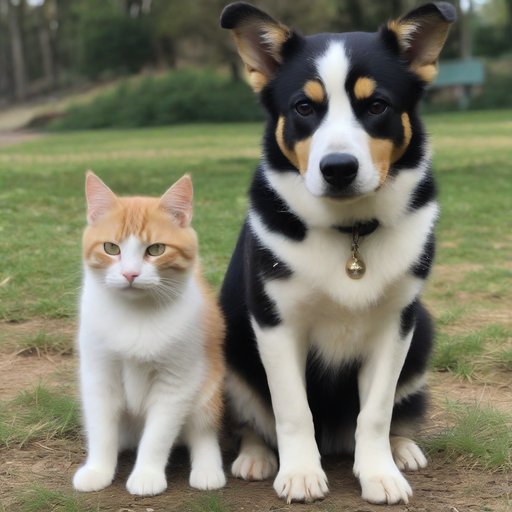}}}$ & 
        $\vcenter{\hbox{\includegraphics[width=0.136\textwidth]{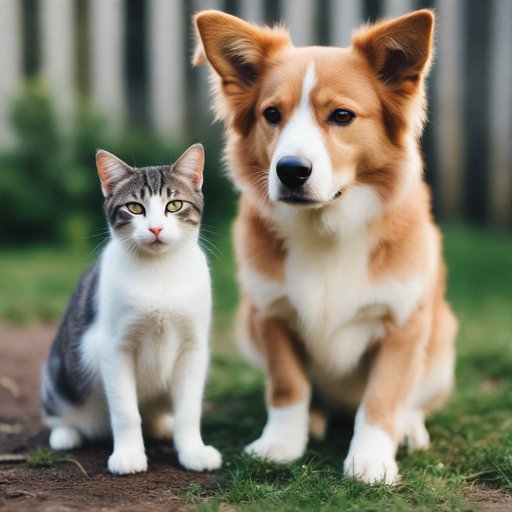}}}$ & 
        $\vcenter{\hbox{\includegraphics[width=0.136\textwidth]{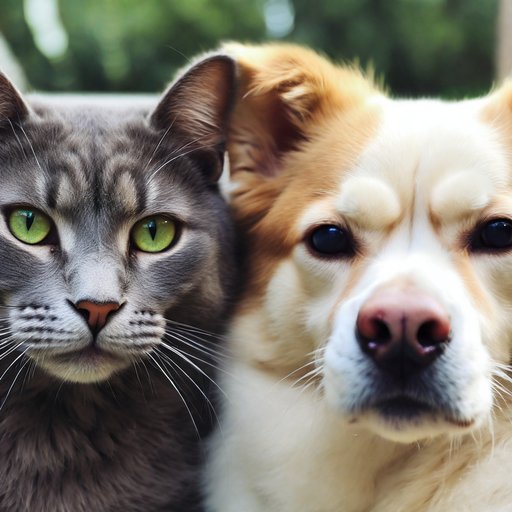}}}$ \\[\vsep]

        A giraffe underneath a microwave. & 
        $\vcenter{\hbox{\includegraphics[width=0.136\textwidth]{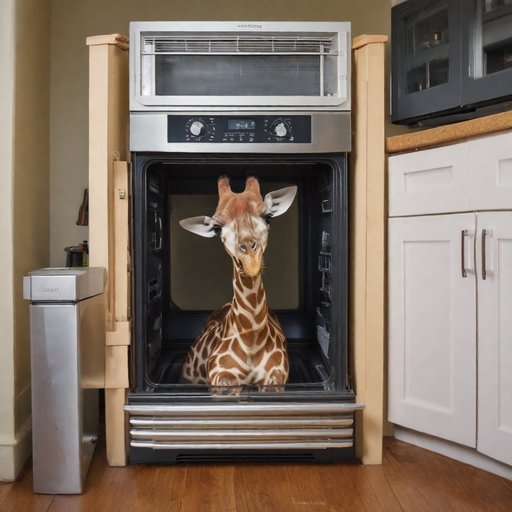}}}$ & 
        $\vcenter{\hbox{\includegraphics[width=0.136\textwidth]{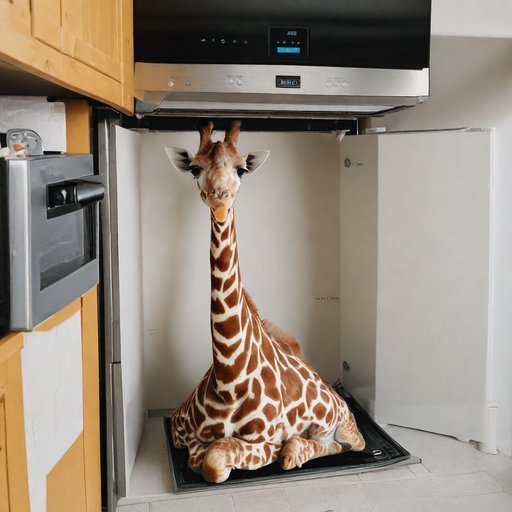}}}$ & 
        $\vcenter{\hbox{\includegraphics[width=0.136\textwidth]{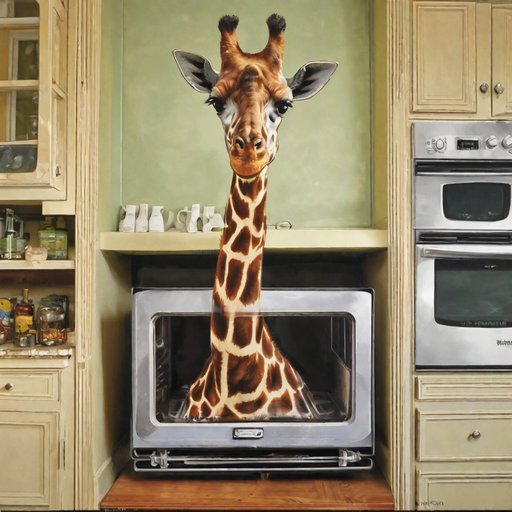}}}$ & 
        $\vcenter{\hbox{\includegraphics[width=0.136\textwidth]{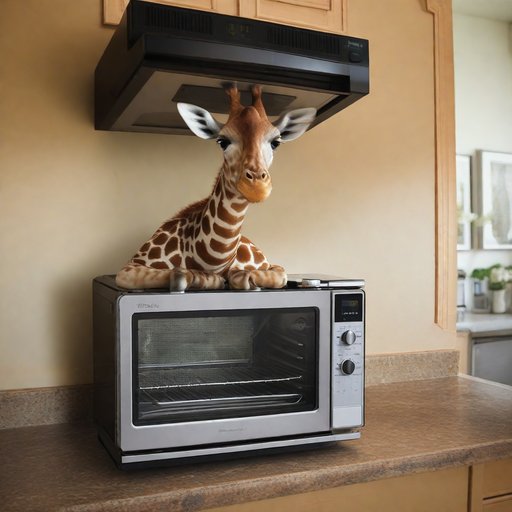}}}$ & 
        $\vcenter{\hbox{\includegraphics[width=0.136\textwidth]{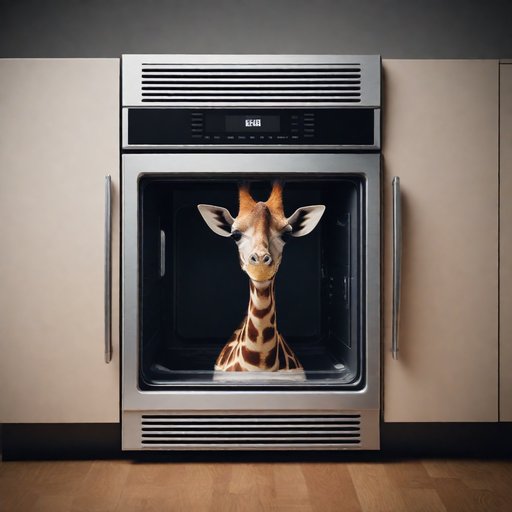}}}$ & 
        $\vcenter{\hbox{\includegraphics[width=0.136\textwidth]{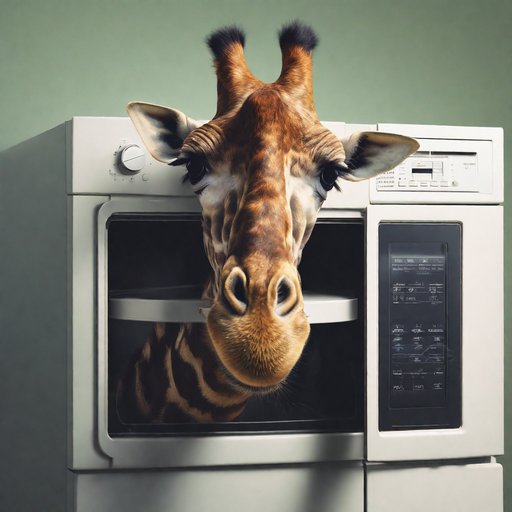}}}$ \\[\vsep]

        A baby fennec sneezing onto a strawberry, detailed, macro, studio light, droplets, backlit ears. & 
        $\vcenter{\hbox{\includegraphics[width=0.136\textwidth]{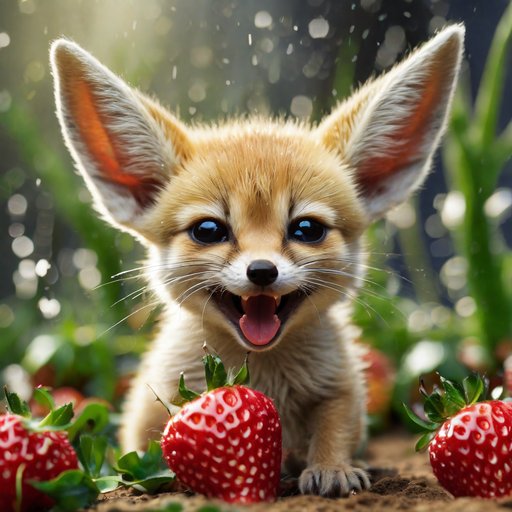}}}$ & 
        $\vcenter{\hbox{\includegraphics[width=0.136\textwidth]{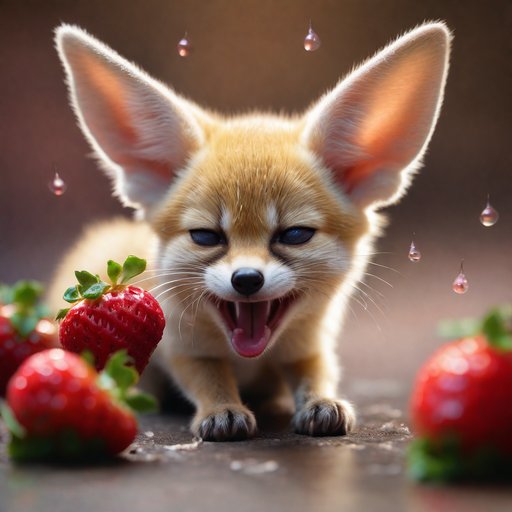}}}$ & 
        $\vcenter{\hbox{\includegraphics[width=0.136\textwidth]{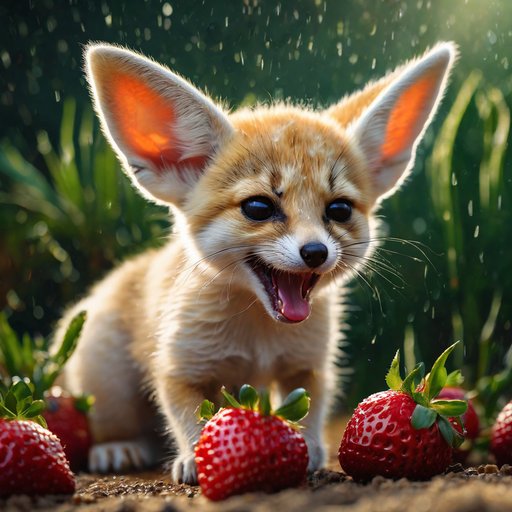}}}$ & 
        $\vcenter{\hbox{\includegraphics[width=0.136\textwidth]{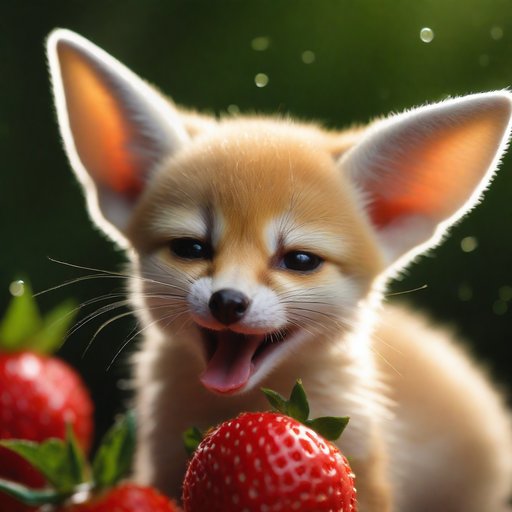}}}$ & 
        $\vcenter{\hbox{\includegraphics[width=0.136\textwidth]{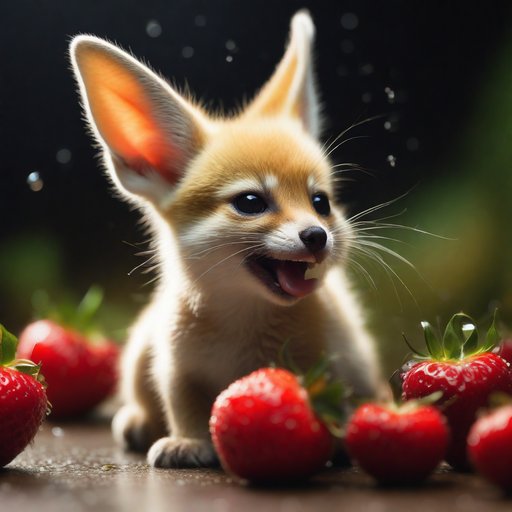}}}$ & 
        $\vcenter{\hbox{\includegraphics[width=0.136\textwidth]{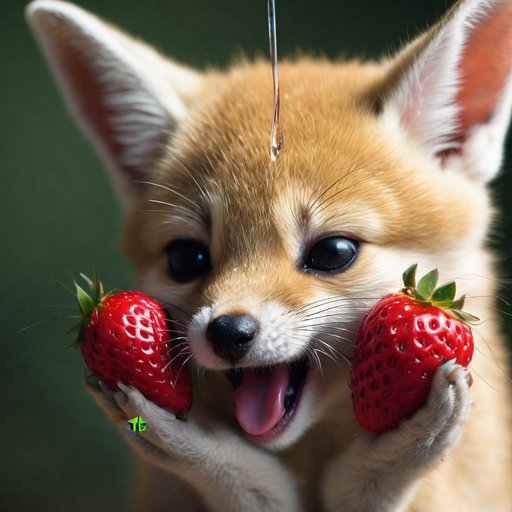}}}$ \\[\vsep]

        One car on the street. & 
        $\vcenter{\hbox{\includegraphics[width=0.136\textwidth]{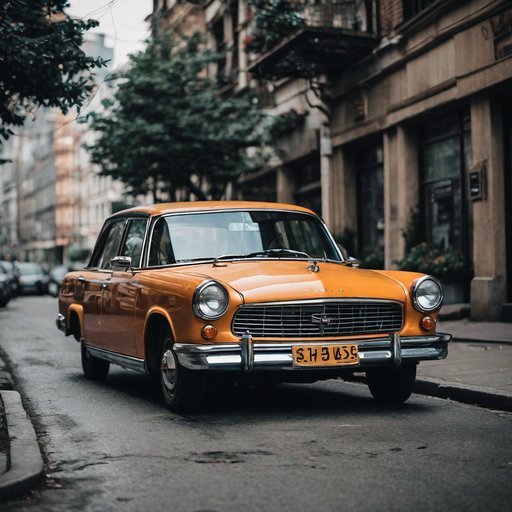}}}$ & 
        $\vcenter{\hbox{\includegraphics[width=0.136\textwidth]{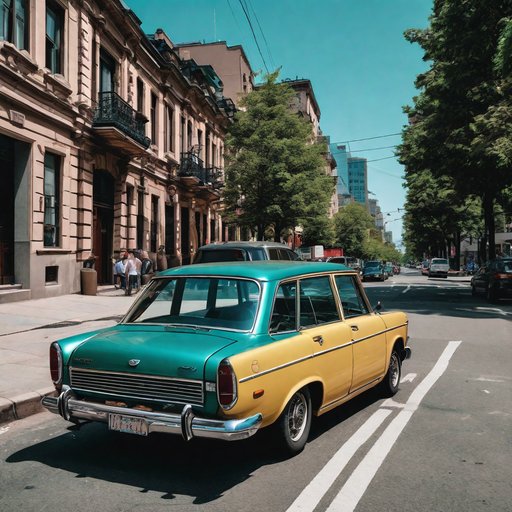}}}$ & 
        $\vcenter{\hbox{\includegraphics[width=0.136\textwidth]{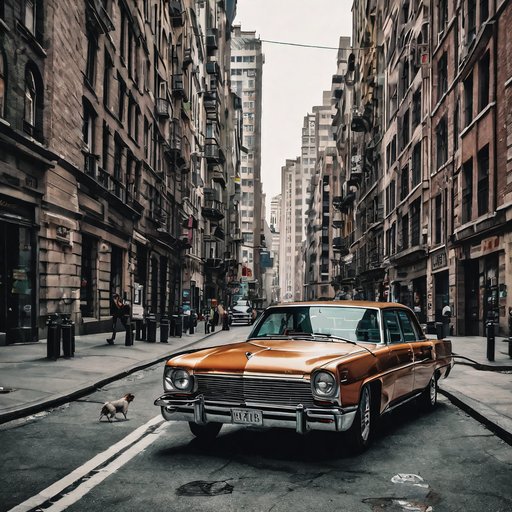}}}$ & 
        $\vcenter{\hbox{\includegraphics[width=0.136\textwidth]{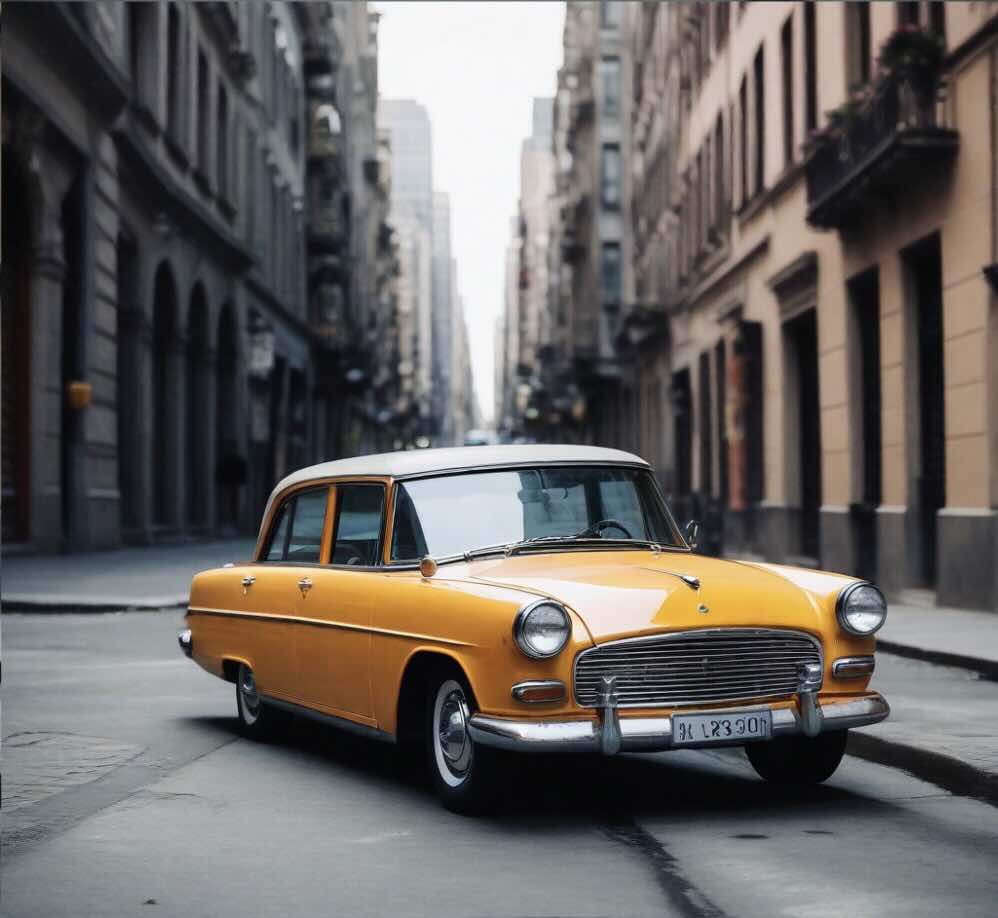}}}$ & 
        $\vcenter{\hbox{\includegraphics[width=0.136\textwidth]{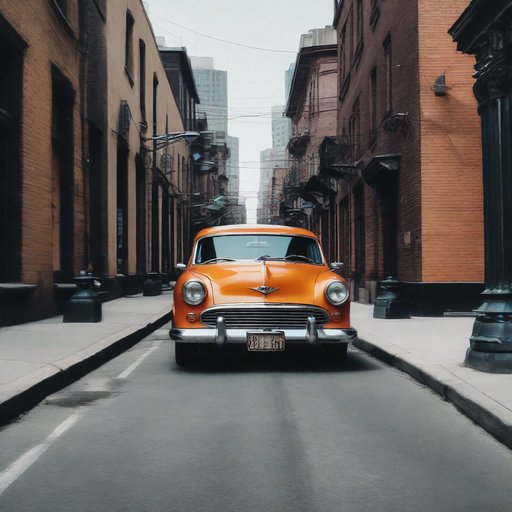}}}$ & 
        $\vcenter{\hbox{\includegraphics[width=0.136\textwidth]{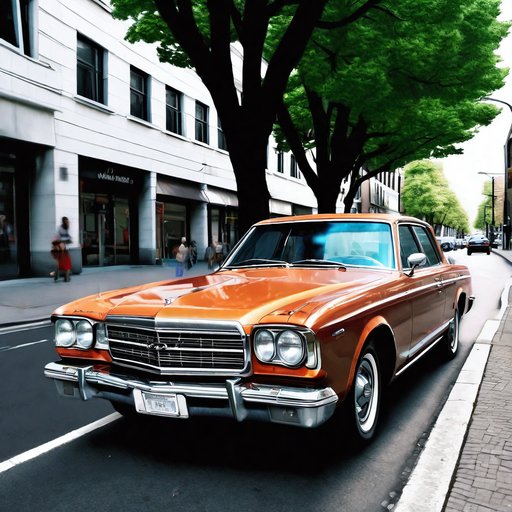}}}$ \\[\vsep]

        A pink colored car. & 
        $\vcenter{\hbox{\includegraphics[width=0.136\textwidth]{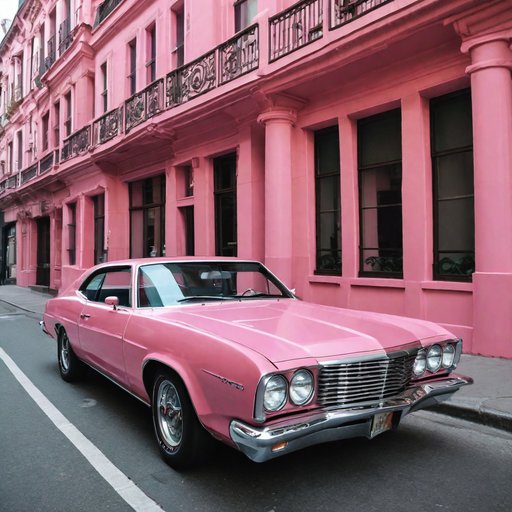}}}$ & 
        $\vcenter{\hbox{\includegraphics[width=0.136\textwidth]{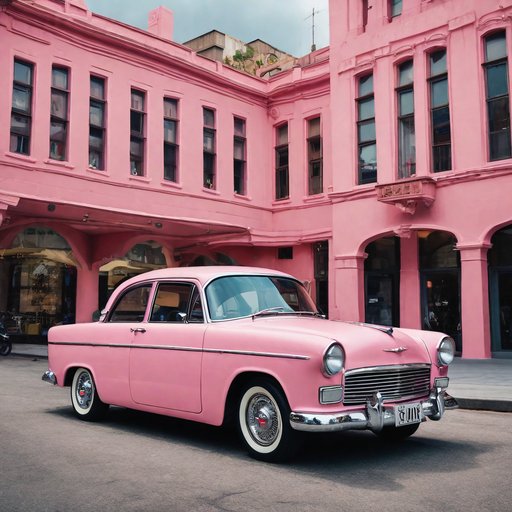}}}$ & 
        $\vcenter{\hbox{\includegraphics[width=0.136\textwidth]{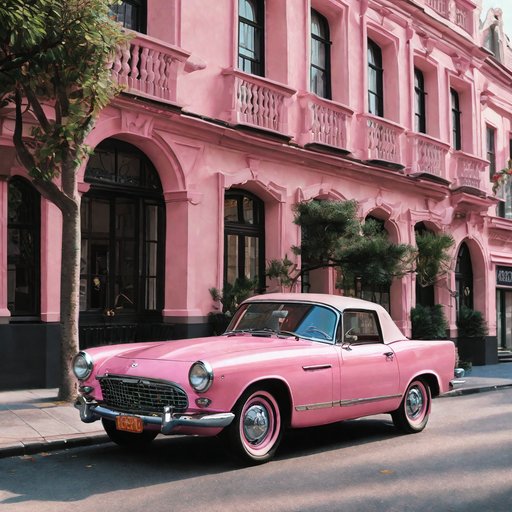}}}$ & 
        $\vcenter{\hbox{\includegraphics[width=0.136\textwidth]{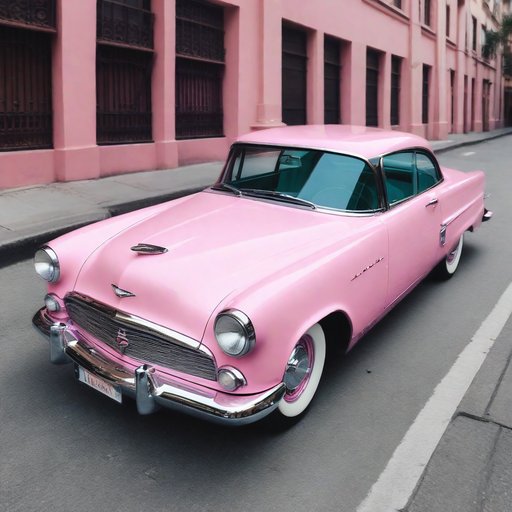}}}$ & 
        $\vcenter{\hbox{\includegraphics[width=0.136\textwidth]{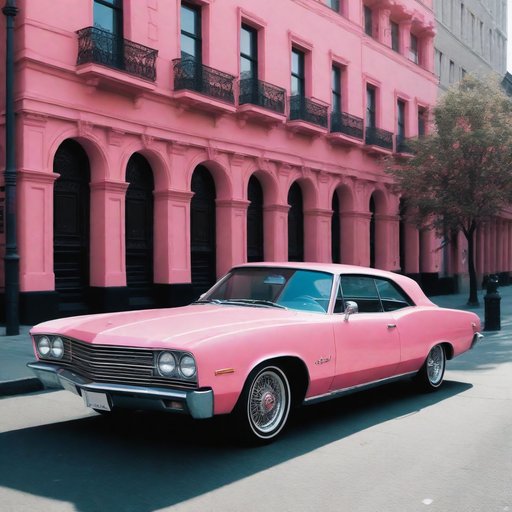}}}$ & 
        $\vcenter{\hbox{\includegraphics[width=0.136\textwidth]{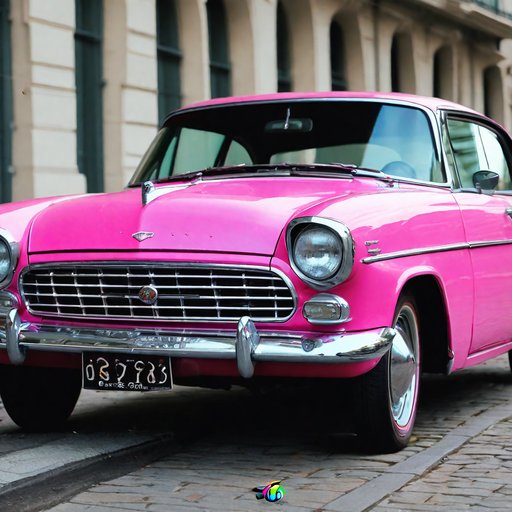}}}$ \\[\vsep]

        An umbrella on top of a spoon. & 
        $\vcenter{\hbox{\includegraphics[width=0.136\textwidth]{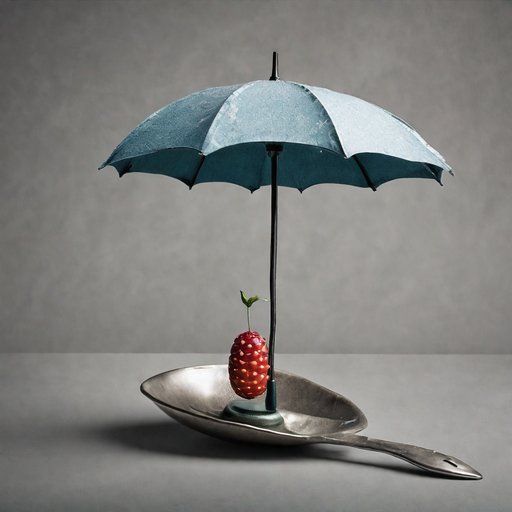}}}$ & 
        $\vcenter{\hbox{\includegraphics[width=0.136\textwidth]{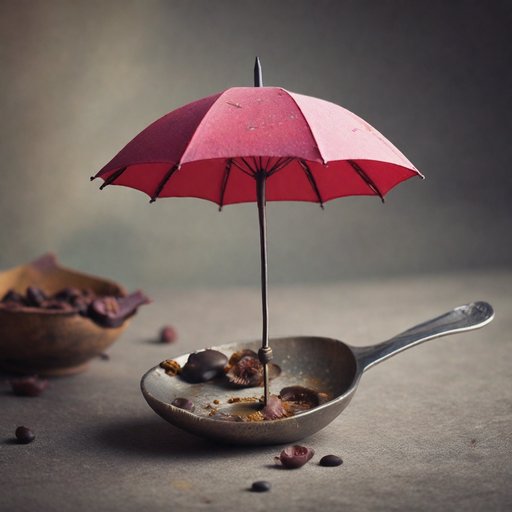}}}$ & 
        $\vcenter{\hbox{\includegraphics[width=0.136\textwidth]{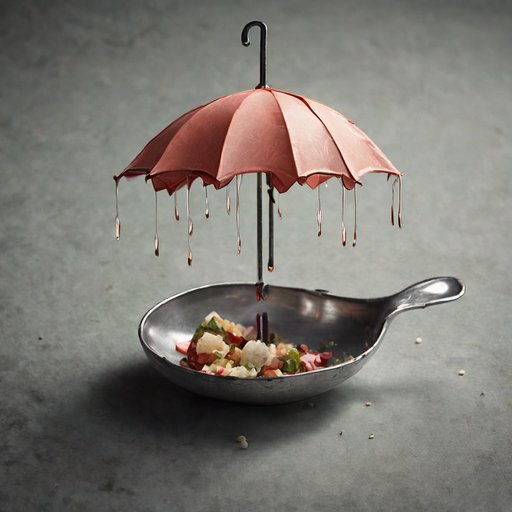}}}$ & 
        $\vcenter{\hbox{\includegraphics[width=0.136\textwidth]{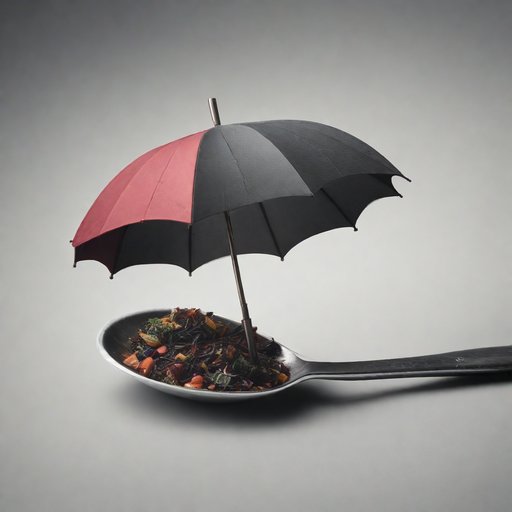}}}$ & 
        $\vcenter{\hbox{\includegraphics[width=0.136\textwidth]{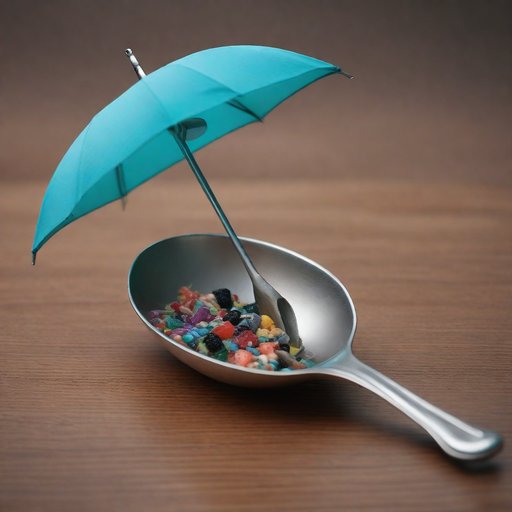}}}$ & 
        $\vcenter{\hbox{\includegraphics[width=0.136\textwidth]{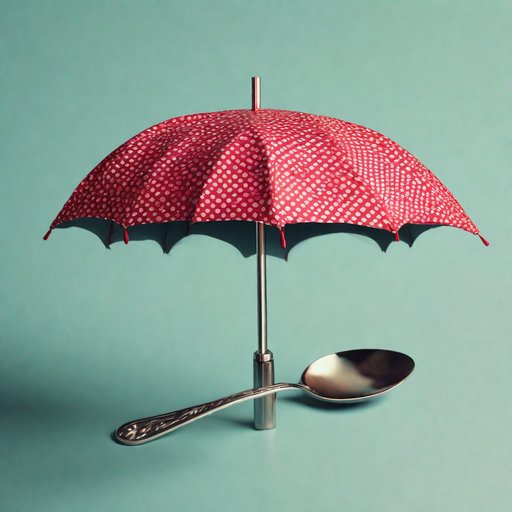}}}$ \\[\vsep]

        A single clock is sitting on a table. & 
        $\vcenter{\hbox{\includegraphics[width=0.136\textwidth]{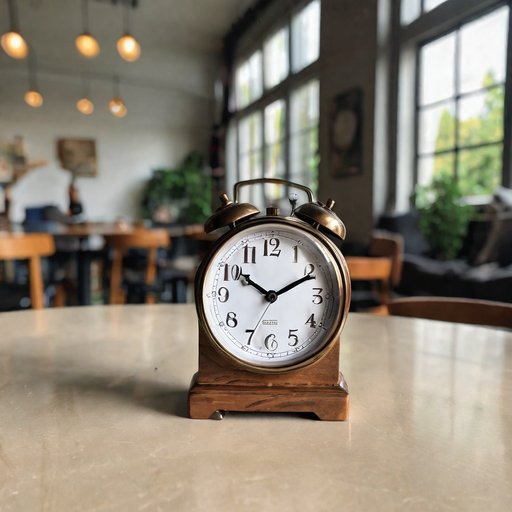}}}$ & 
        $\vcenter{\hbox{\includegraphics[width=0.136\textwidth]{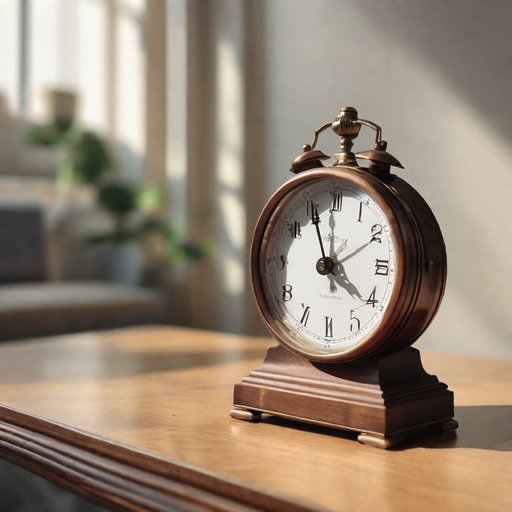}}}$ & 
        $\vcenter{\hbox{\includegraphics[width=0.136\textwidth]{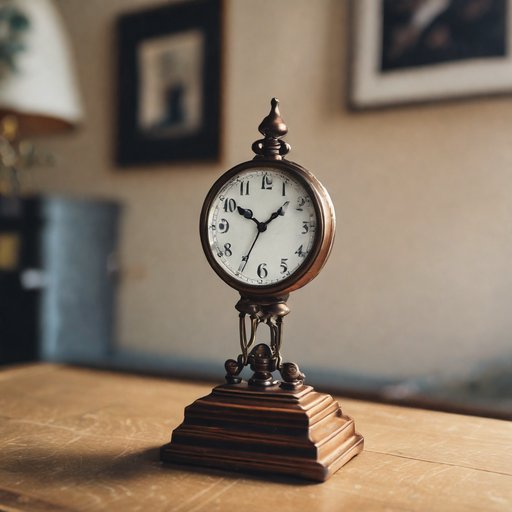}}}$ & 
        $\vcenter{\hbox{\includegraphics[width=0.136\textwidth]{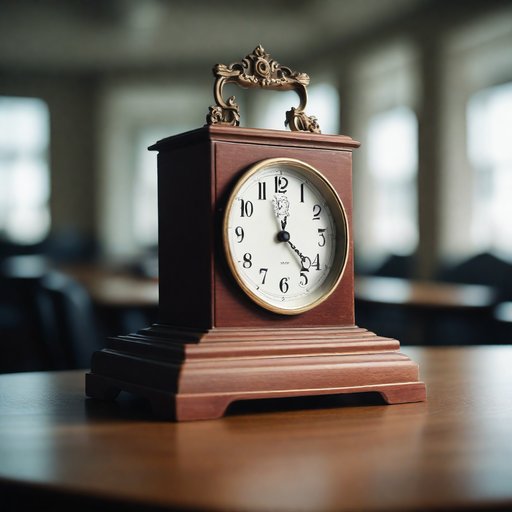}}}$ & 
        $\vcenter{\hbox{\includegraphics[width=0.136\textwidth]{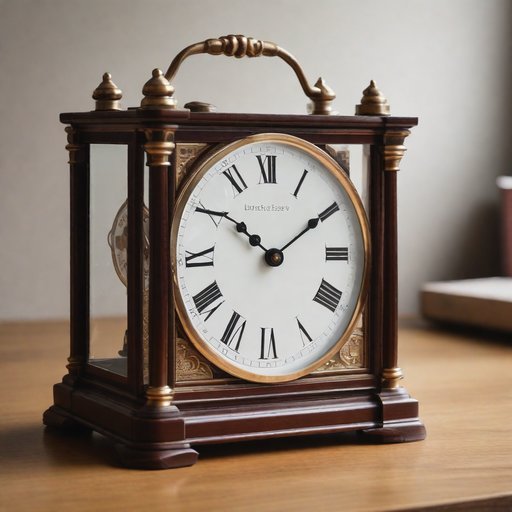}}}$ & 
        $\vcenter{\hbox{\includegraphics[width=0.136\textwidth]{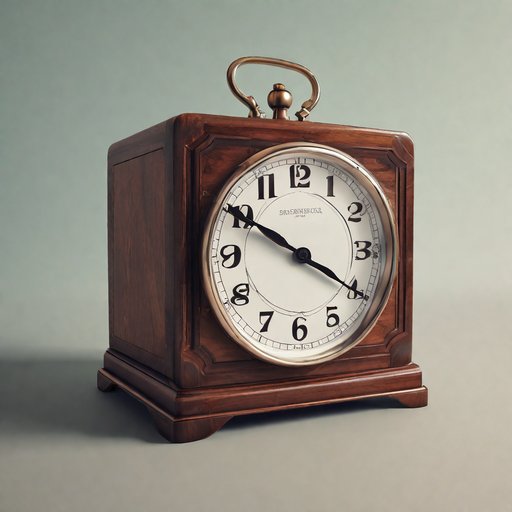}}}$ \\
        
        \bottomrule
    \end{tabular}
    \raggedright
    \caption{\textbf{Qualitative comparison (SDXL-Lightning + ImageReward).} 10 randomly chosen prompts for different alignment methods using SDXL-Lightning and ImageReward.}
    \label{fig:app:sdxl_reward}
\end{figure}

\begin{figure}[htbp]
    \centering
    \definecolor{highlight}{rgb}{0.94, 0.96, 1.0}
    
    \setlength{\tabcolsep}{0pt} 
    \setlength{\aboverulesep}{2pt} 
    \setlength{\belowrulesep}{2pt}

    \setlength{\hsep}{4pt} 
    \setlength{\vsep}{28pt}
    
    \renewcommand{\arraystretch}{0} 

    \begin{tabular}{
        >{\raggedright\arraybackslash\tiny\itshape}m{1.5cm} 
        @{\hspace{\hsep}} >{\columncolor{highlight}}c 
        @{\hspace{\hsep}}c@{\hspace{\hsep}}c@{\hspace{\hsep}}c@{\hspace{\hsep}}c@{\hspace{\hsep}}c 
    }
        \toprule
        \rule{0pt}{3ex} \textbf{\scriptsize Prompt} & \textbf{\scriptsize TRS (Ours)} & \textbf{\scriptsize RS} & \textbf{\scriptsize ZO} & \textbf{\scriptsize DTS*} & \textbf{\scriptsize FD} & \textbf{\scriptsize OC-Flow} \\[1.5ex]
        \midrule
        
        A keyboard made of water, the water is made of light, the light is turned off. & 
        $\vcenter{\hbox{\includegraphics[width=0.136\textwidth]{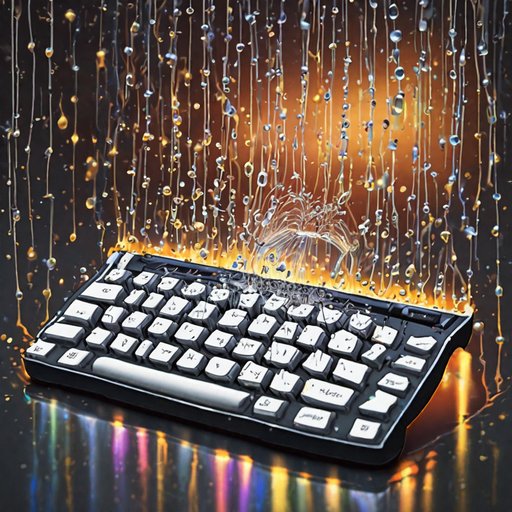}}}$ & 
        $\vcenter{\hbox{\includegraphics[width=0.136\textwidth]{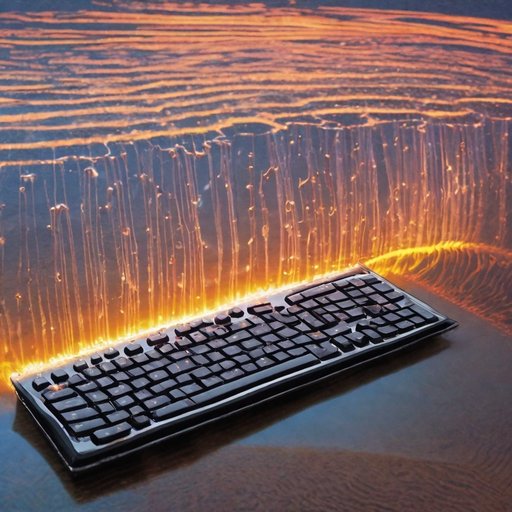}}}$ & 
        $\vcenter{\hbox{\includegraphics[width=0.136\textwidth]{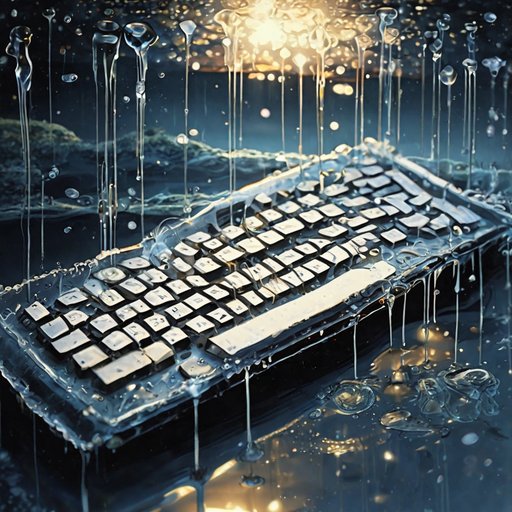}}}$ & 
        $\vcenter{\hbox{\includegraphics[width=0.136\textwidth]{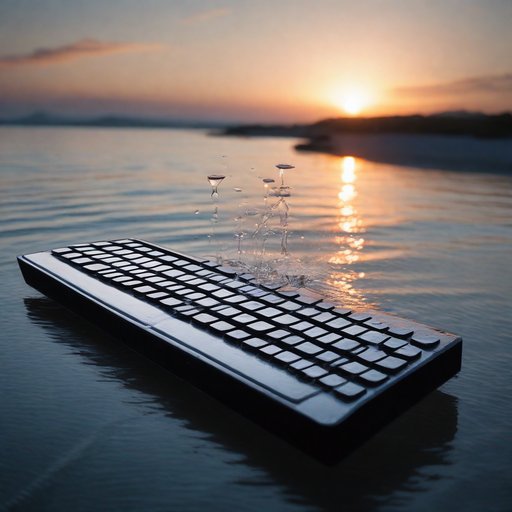}}}$ & 
        $\vcenter{\hbox{\includegraphics[width=0.136\textwidth]{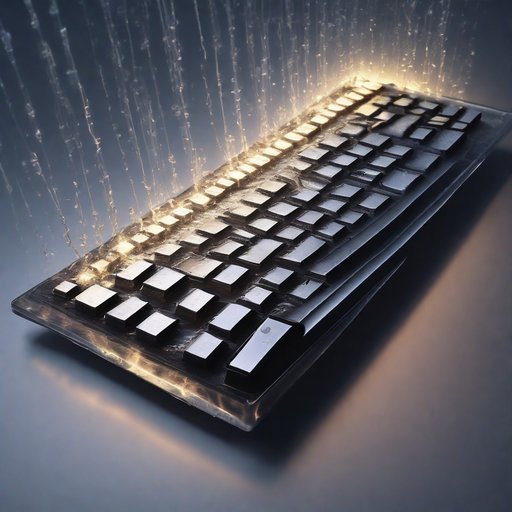}}}$ & 
        $\vcenter{\hbox{\includegraphics[width=0.136\textwidth]{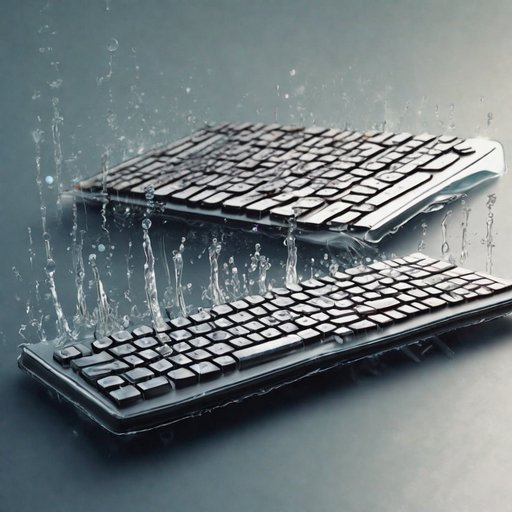}}}$ \\[\vsep]

        A red colored dog. & 
        $\vcenter{\hbox{\includegraphics[width=0.136\textwidth]{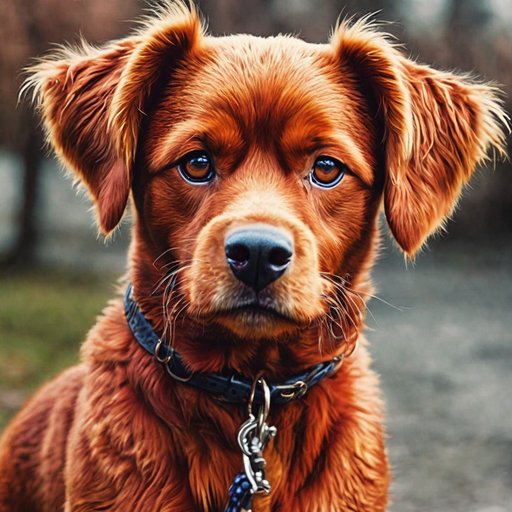}}}$ & 
        $\vcenter{\hbox{\includegraphics[width=0.136\textwidth]{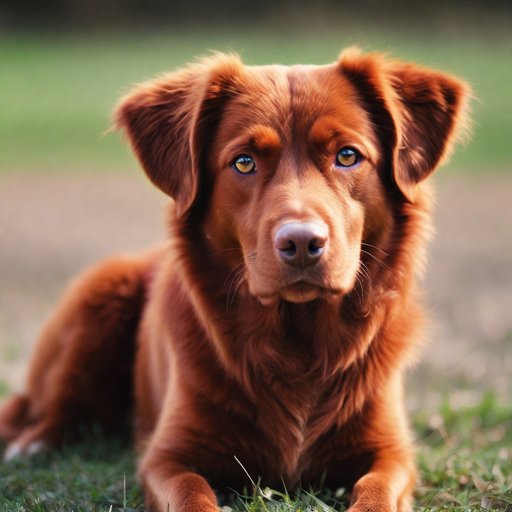}}}$ & 
        $\vcenter{\hbox{\includegraphics[width=0.136\textwidth]{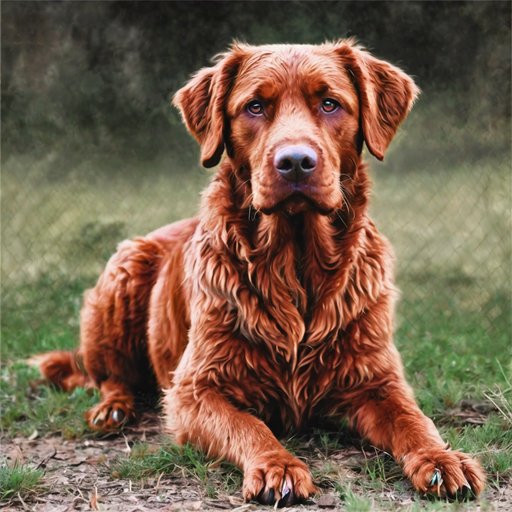}}}$ & 
        $\vcenter{\hbox{\includegraphics[width=0.136\textwidth]{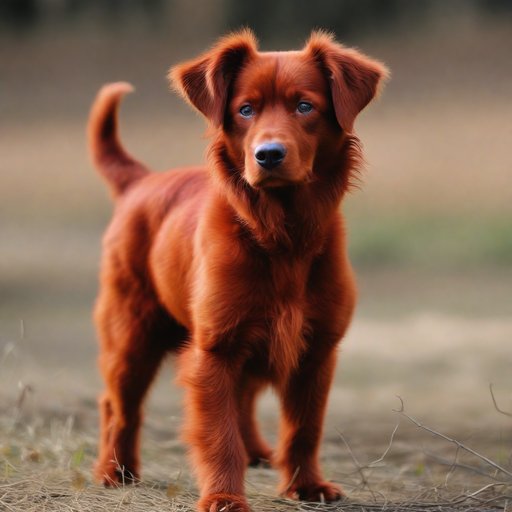}}}$ & 
        $\vcenter{\hbox{\includegraphics[width=0.136\textwidth]{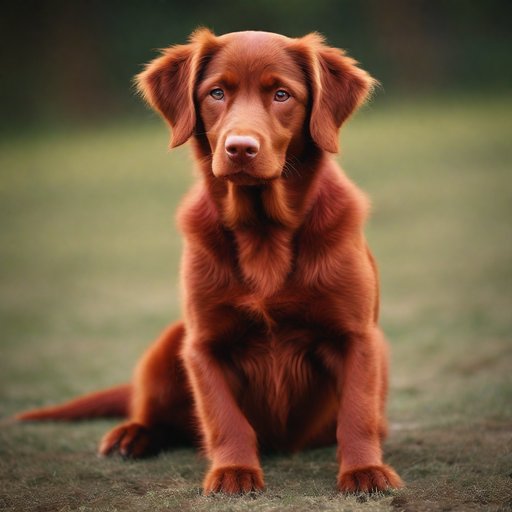}}}$ & 
        $\vcenter{\hbox{\includegraphics[width=0.136\textwidth]{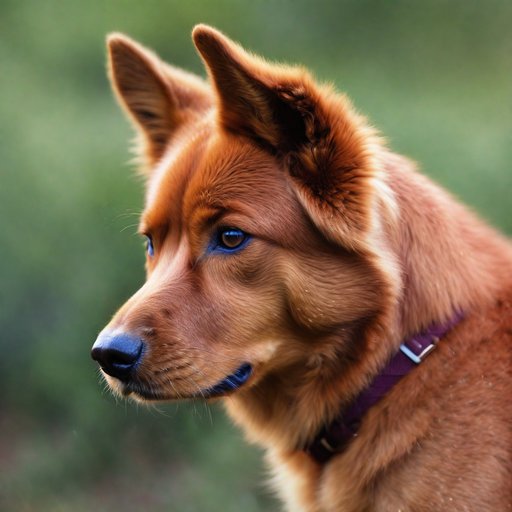}}}$ \\[\vsep]

        Hyper-realistic photo of an abandoned industrial site during a storm. & 
        $\vcenter{\hbox{\includegraphics[width=0.136\textwidth]{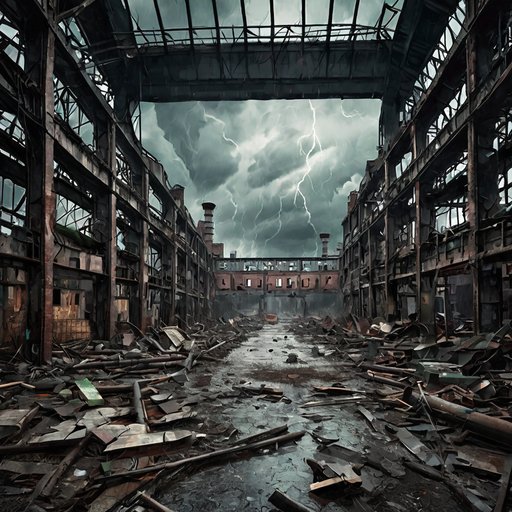}}}$ & 
        $\vcenter{\hbox{\includegraphics[width=0.136\textwidth]{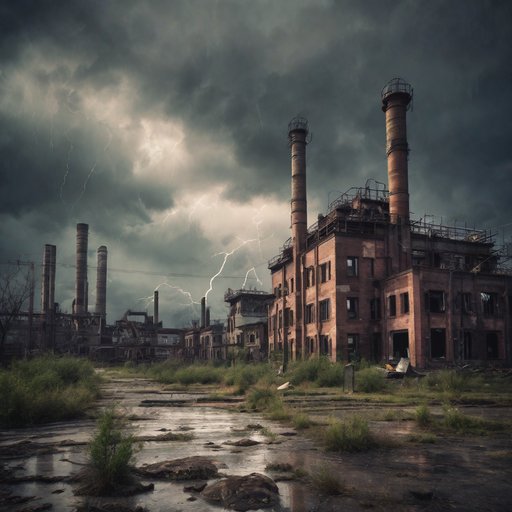}}}$ & 
        $\vcenter{\hbox{\includegraphics[width=0.136\textwidth]{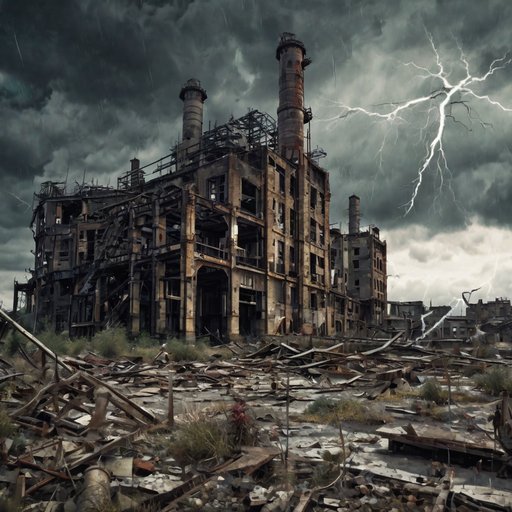}}}$ & 
        $\vcenter{\hbox{\includegraphics[width=0.136\textwidth]{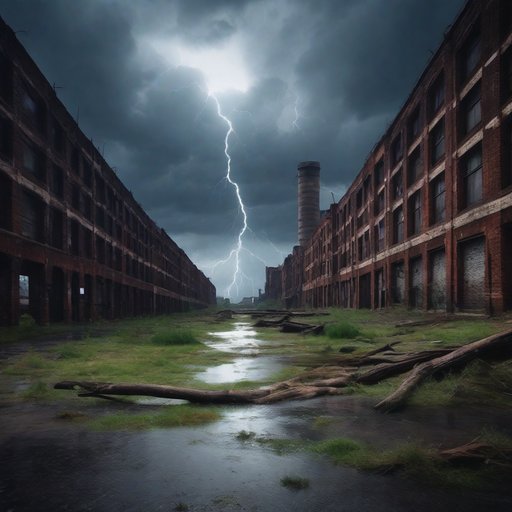}}}$ & 
        $\vcenter{\hbox{\includegraphics[width=0.136\textwidth]{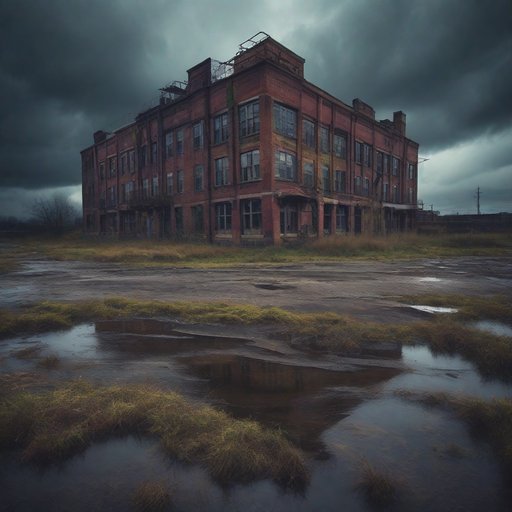}}}$ & 
        $\vcenter{\hbox{\includegraphics[width=0.136\textwidth]{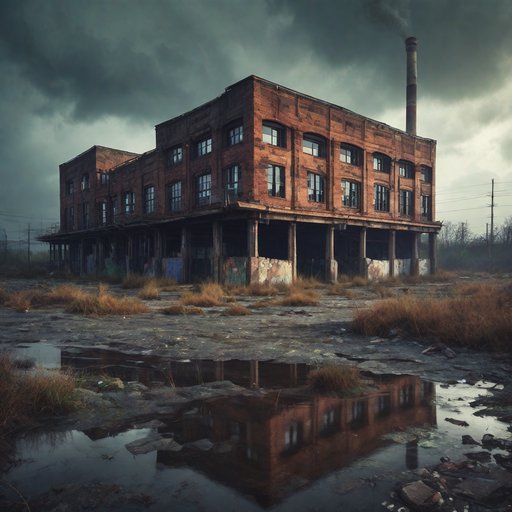}}}$ \\[\vsep]

        A cat on the left of a dog. & 
        $\vcenter{\hbox{\includegraphics[width=0.136\textwidth]{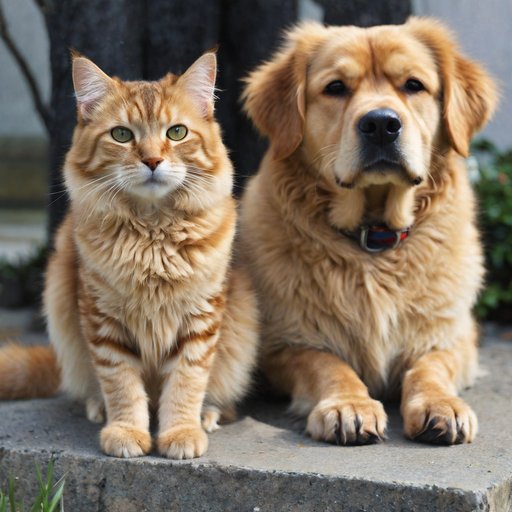}}}$ & 
        $\vcenter{\hbox{\includegraphics[width=0.136\textwidth]{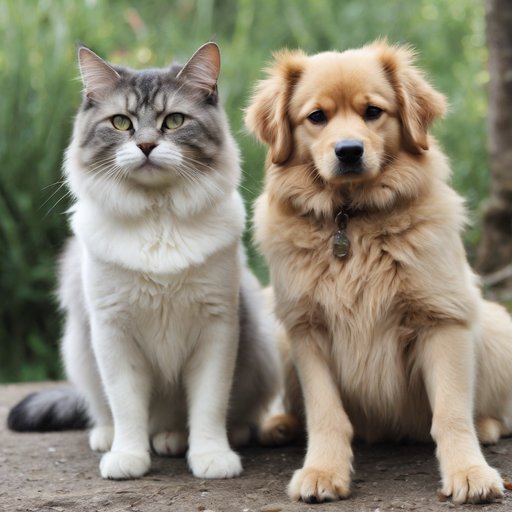}}}$ & 
        $\vcenter{\hbox{\includegraphics[width=0.136\textwidth]{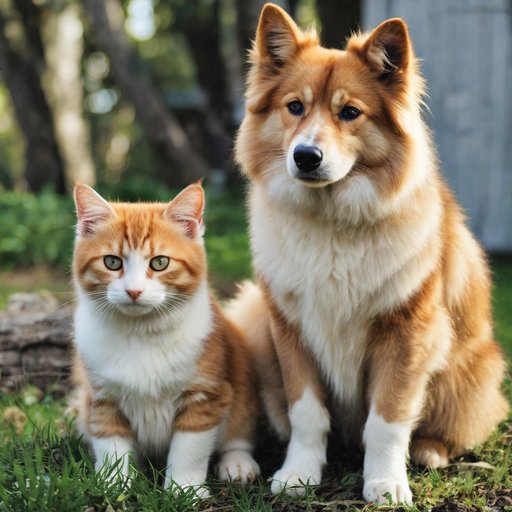}}}$ & 
        $\vcenter{\hbox{\includegraphics[width=0.136\textwidth]{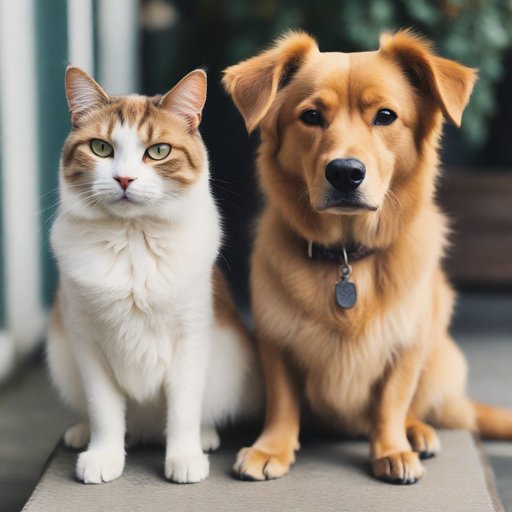}}}$ & 
        $\vcenter{\hbox{\includegraphics[width=0.136\textwidth]{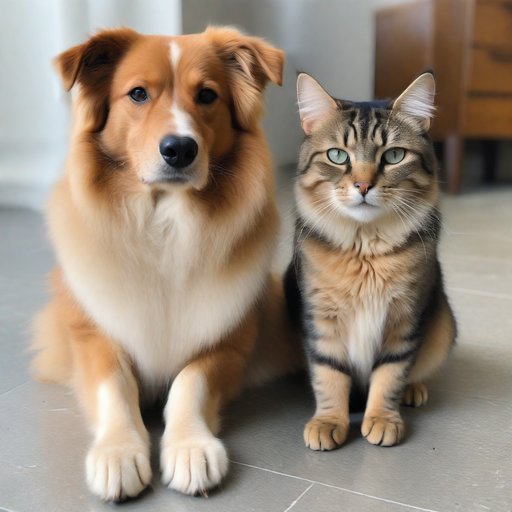}}}$ & 
        $\vcenter{\hbox{\includegraphics[width=0.136\textwidth]{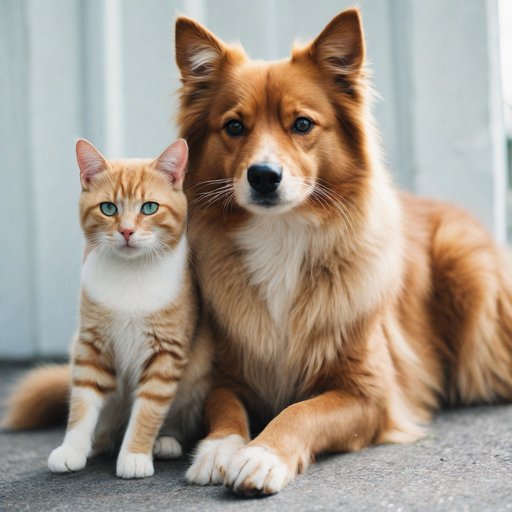}}}$ \\[\vsep]

        A giraffe underneath a microwave. & 
        $\vcenter{\hbox{\includegraphics[width=0.136\textwidth]{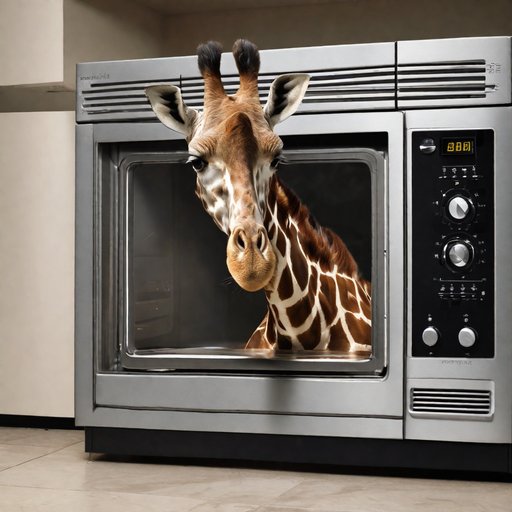}}}$ & 
        $\vcenter{\hbox{\includegraphics[width=0.136\textwidth]{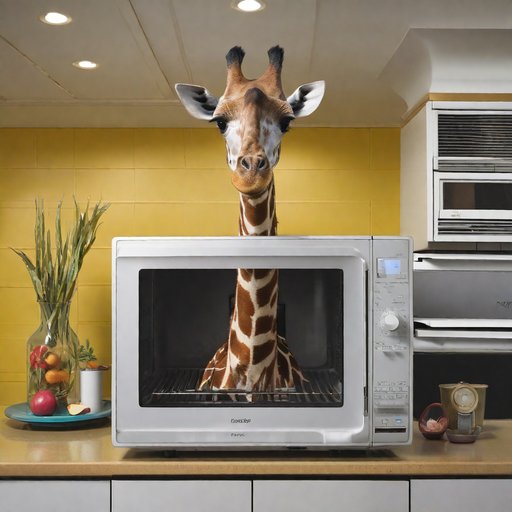}}}$ & 
        $\vcenter{\hbox{\includegraphics[width=0.136\textwidth]{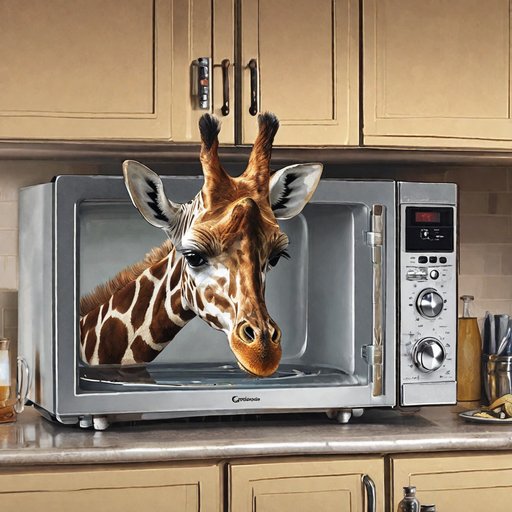}}}$ & 
        $\vcenter{\hbox{\includegraphics[width=0.136\textwidth]{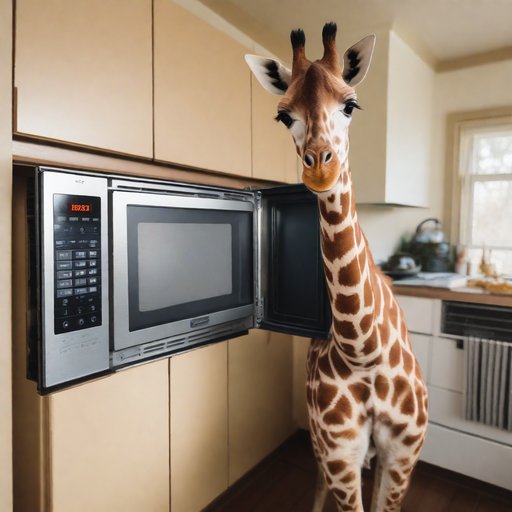}}}$ & 
        $\vcenter{\hbox{\includegraphics[width=0.136\textwidth]{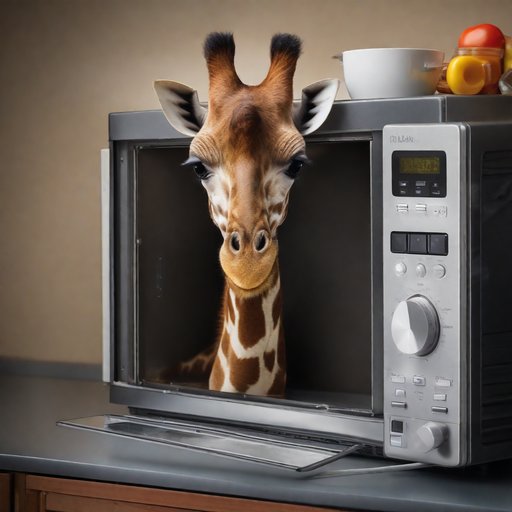}}}$ & 
        $\vcenter{\hbox{\includegraphics[width=0.136\textwidth]{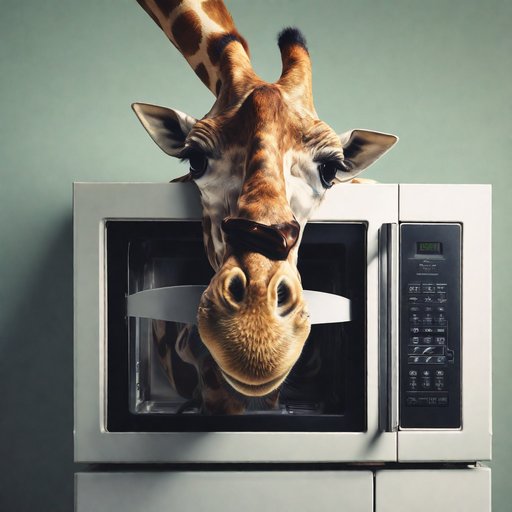}}}$ \\[\vsep]

        A baby fennec sneezing onto a strawberry, detailed, macro, studio light, droplets, backlit ears. & 
        $\vcenter{\hbox{\includegraphics[width=0.136\textwidth]{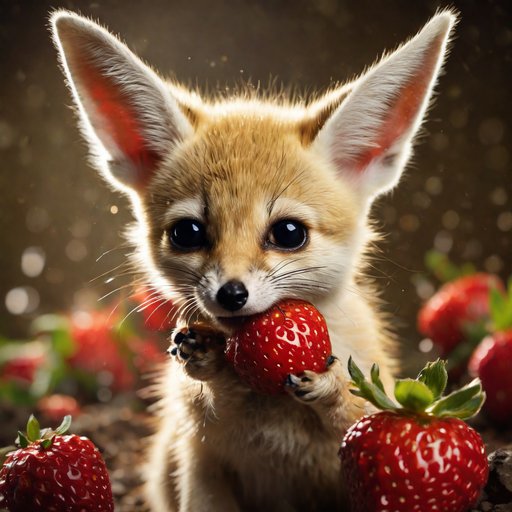}}}$ & 
        $\vcenter{\hbox{\includegraphics[width=0.136\textwidth]{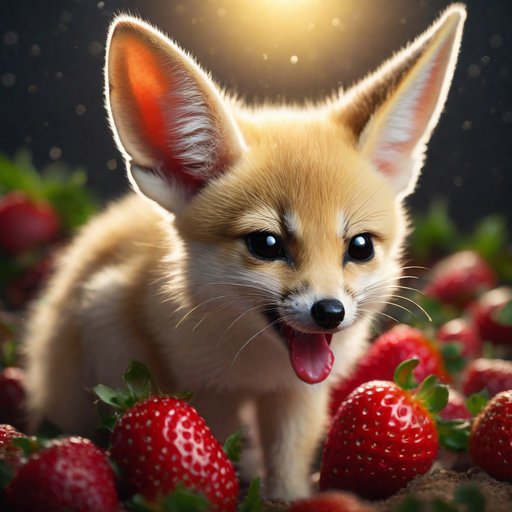}}}$ & 
        $\vcenter{\hbox{\includegraphics[width=0.136\textwidth]{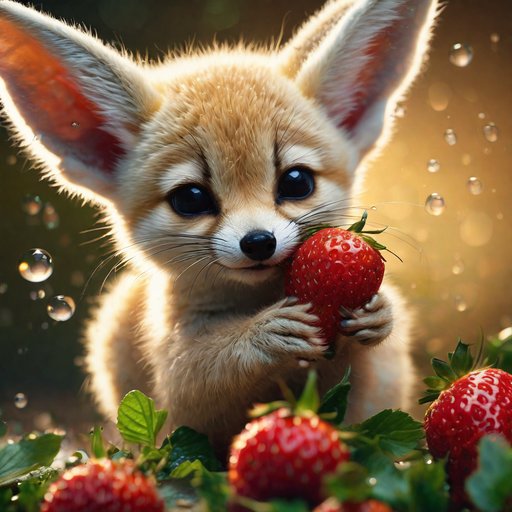}}}$ & 
        $\vcenter{\hbox{\includegraphics[width=0.136\textwidth]{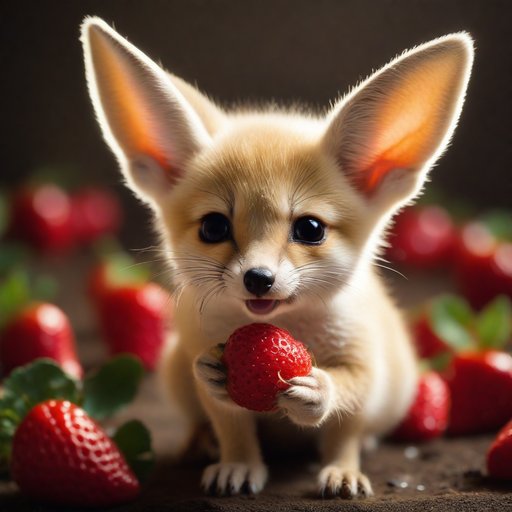}}}$ & 
        $\vcenter{\hbox{\includegraphics[width=0.136\textwidth]{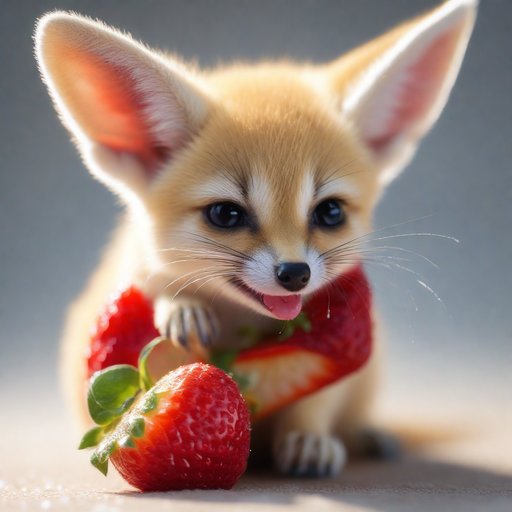}}}$ & 
        $\vcenter{\hbox{\includegraphics[width=0.136\textwidth]{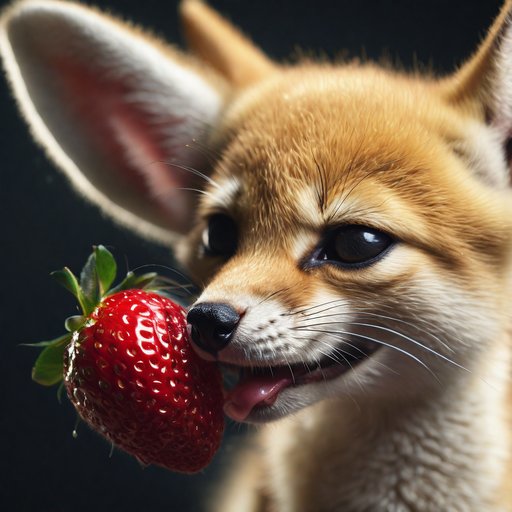}}}$ \\[\vsep]

        One car on the street. & 
        $\vcenter{\hbox{\includegraphics[width=0.136\textwidth]{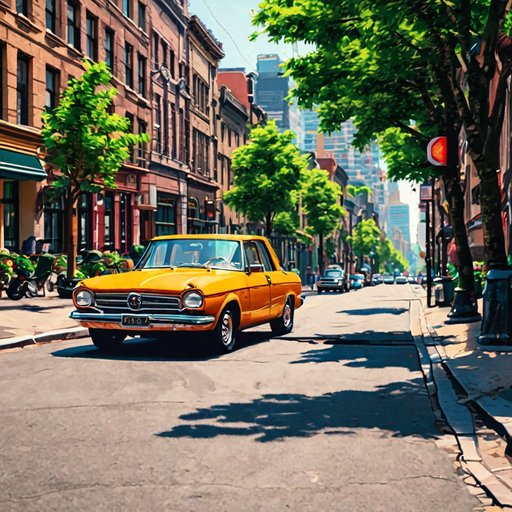}}}$ & 
        $\vcenter{\hbox{\includegraphics[width=0.136\textwidth]{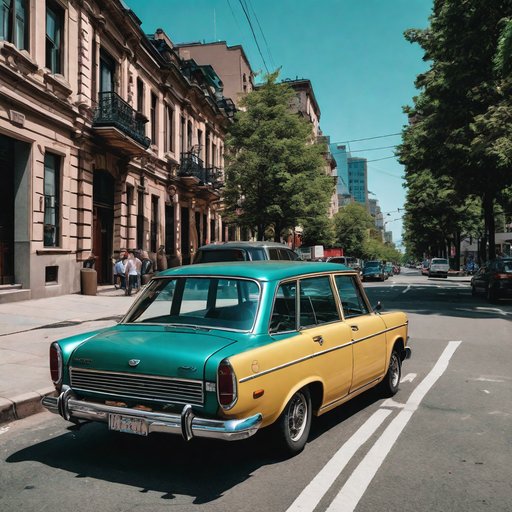}}}$ & 
        $\vcenter{\hbox{\includegraphics[width=0.136\textwidth]{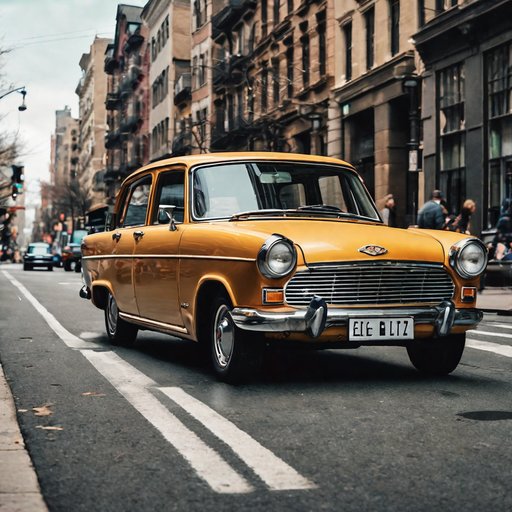}}}$ & 
        $\vcenter{\hbox{\includegraphics[width=0.136\textwidth]{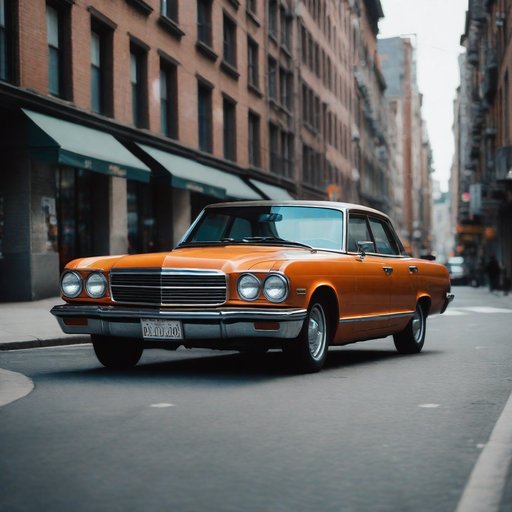}}}$ & 
        $\vcenter{\hbox{\includegraphics[width=0.136\textwidth]{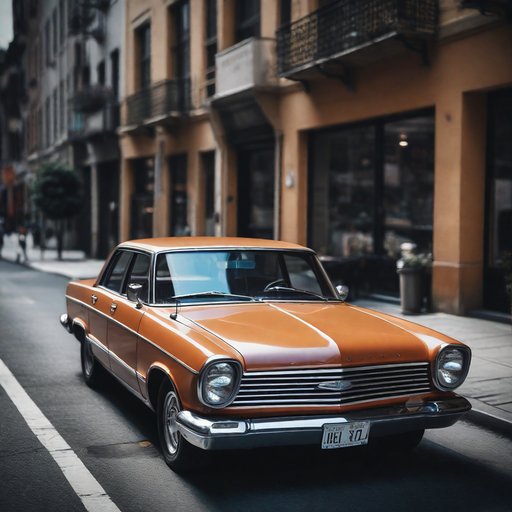}}}$ & 
        $\vcenter{\hbox{\includegraphics[width=0.136\textwidth]{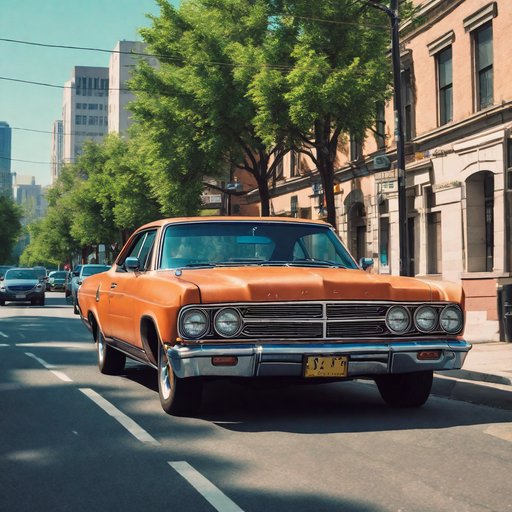}}}$ \\[\vsep]

        A pink colored car. & 
        $\vcenter{\hbox{\includegraphics[width=0.136\textwidth]{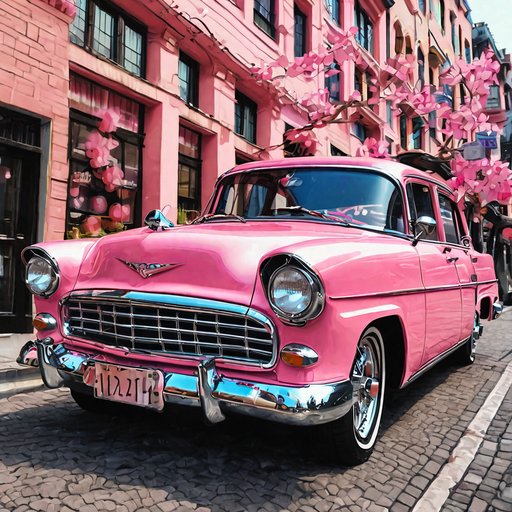}}}$ & 
        $\vcenter{\hbox{\includegraphics[width=0.136\textwidth]{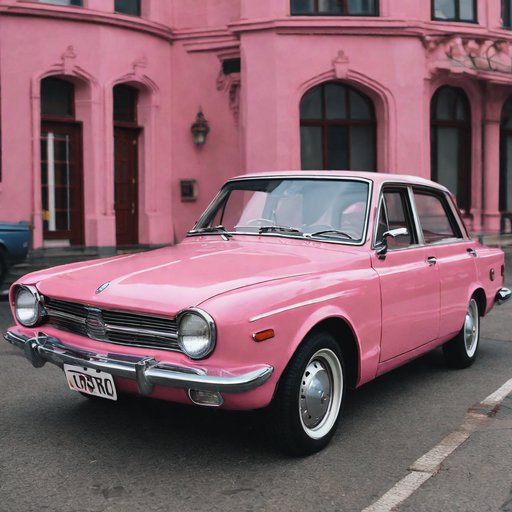}}}$ & 
        $\vcenter{\hbox{\includegraphics[width=0.136\textwidth]{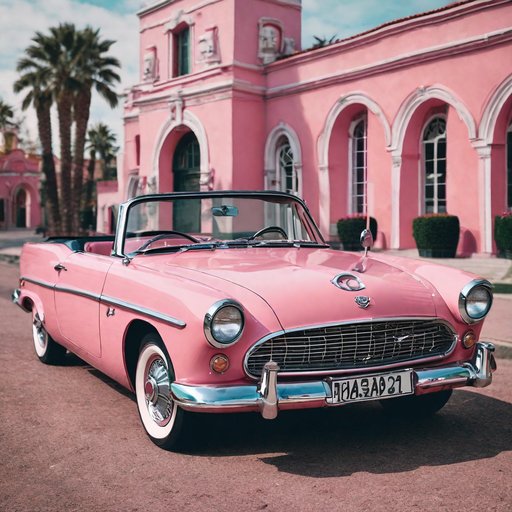}}}$ & 
        $\vcenter{\hbox{\includegraphics[width=0.136\textwidth]{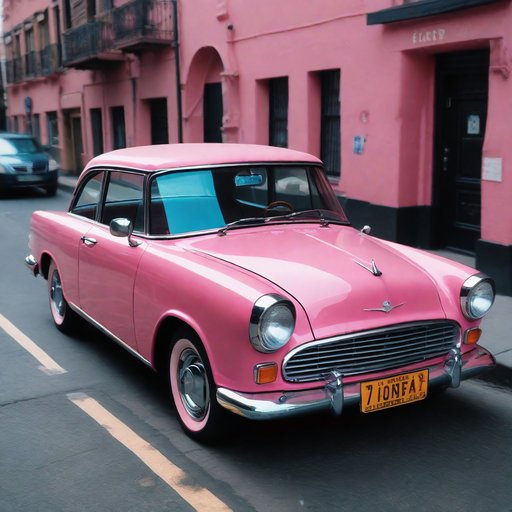}}}$ & 
        $\vcenter{\hbox{\includegraphics[width=0.136\textwidth]{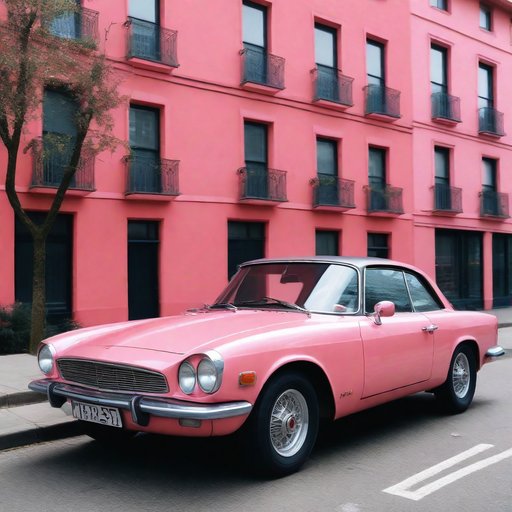}}}$ & 
        $\vcenter{\hbox{\includegraphics[width=0.136\textwidth]{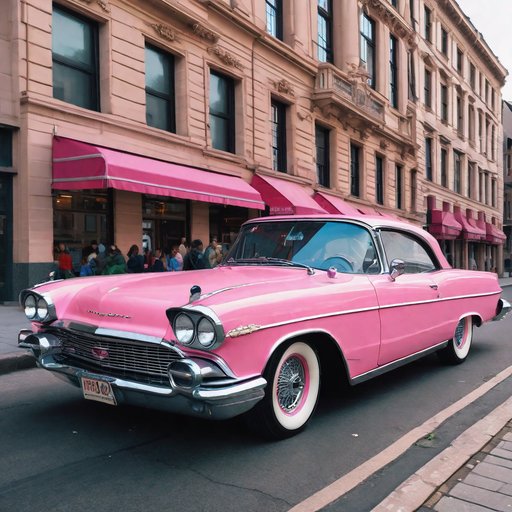}}}$ \\[\vsep]

        An umbrella on top of a spoon. & 
        $\vcenter{\hbox{\includegraphics[width=0.136\textwidth]{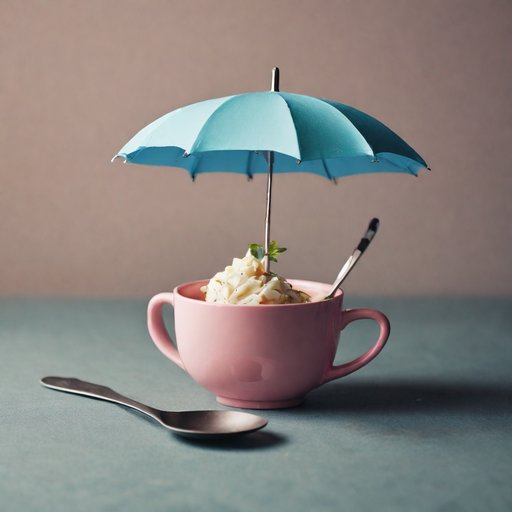}}}$ & 
        $\vcenter{\hbox{\includegraphics[width=0.136\textwidth]{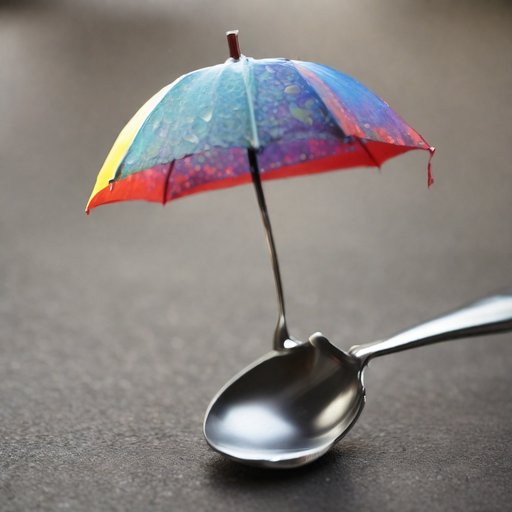}}}$ & 
        $\vcenter{\hbox{\includegraphics[width=0.136\textwidth]{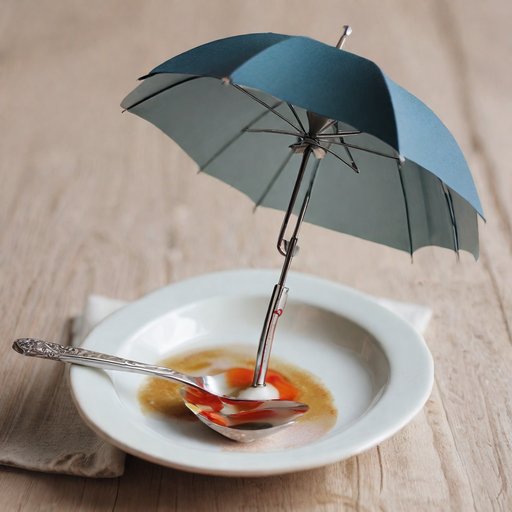}}}$ & 
        $\vcenter{\hbox{\includegraphics[width=0.136\textwidth]{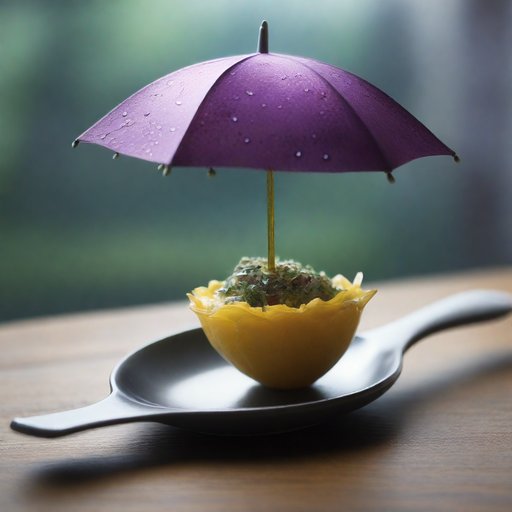}}}$ & 
        $\vcenter{\hbox{\includegraphics[width=0.136\textwidth]{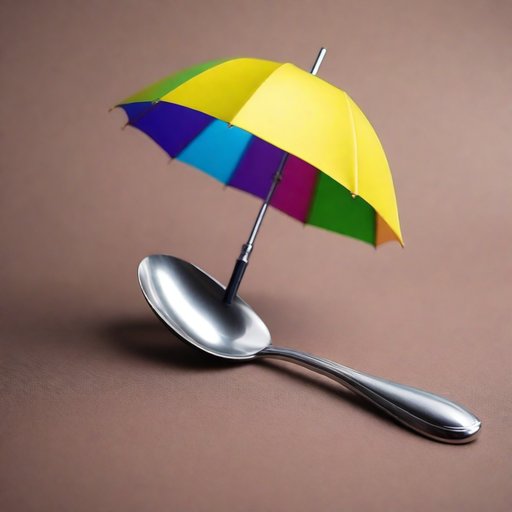}}}$ & 
        $\vcenter{\hbox{\includegraphics[width=0.136\textwidth]{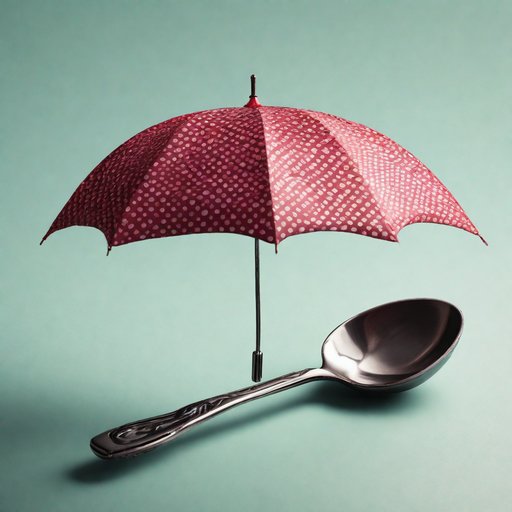}}}$ \\[\vsep]

        A single clock is sitting on a table. & 
        $\vcenter{\hbox{\includegraphics[width=0.136\textwidth]{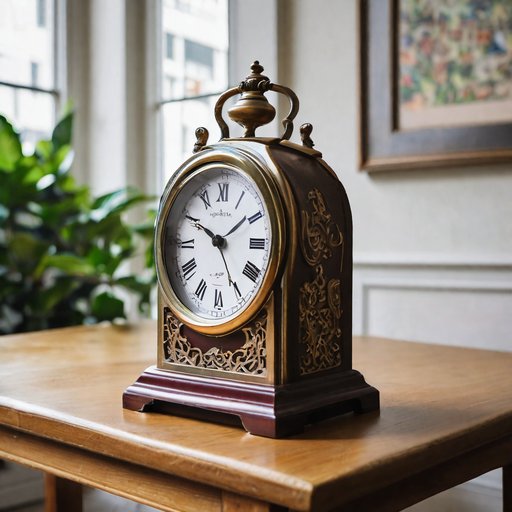}}}$ & 
        $\vcenter{\hbox{\includegraphics[width=0.136\textwidth]{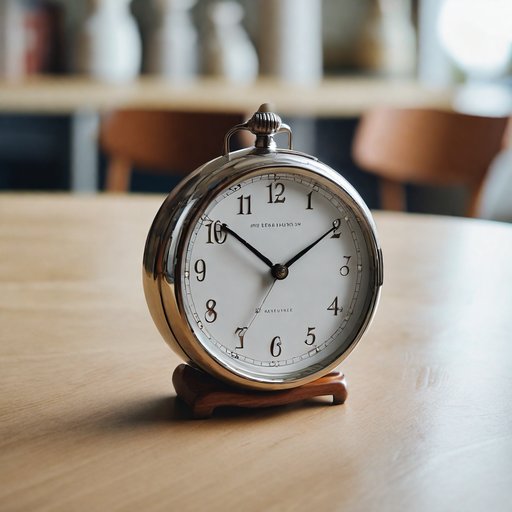}}}$ & 
        $\vcenter{\hbox{\includegraphics[width=0.136\textwidth]{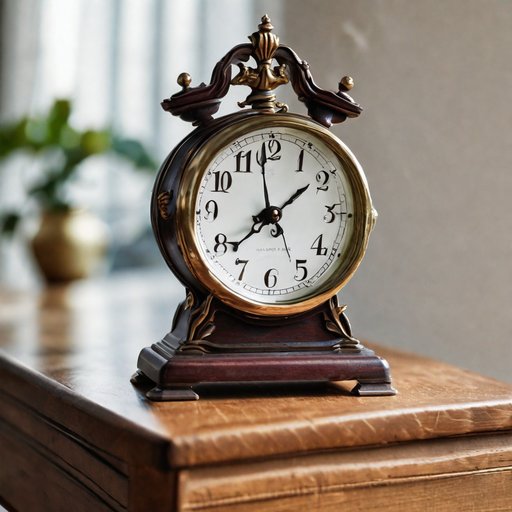}}}$ & 
        $\vcenter{\hbox{\includegraphics[width=0.136\textwidth]{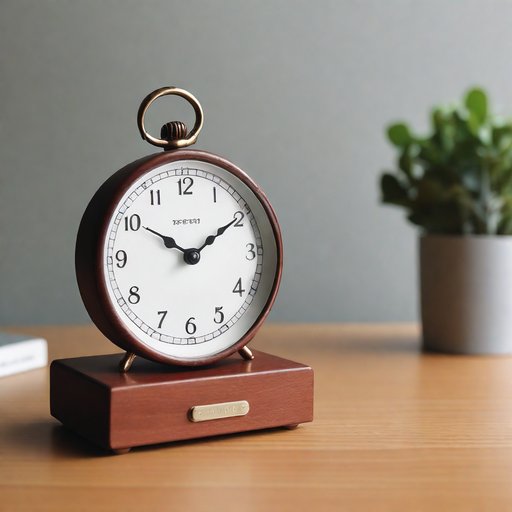}}}$ & 
        $\vcenter{\hbox{\includegraphics[width=0.136\textwidth]{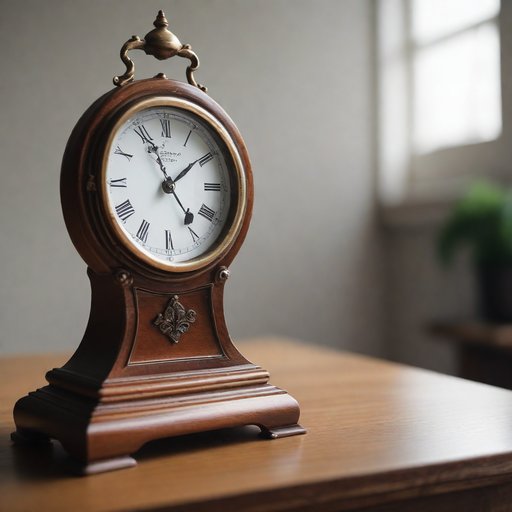}}}$ & 
        $\vcenter{\hbox{\includegraphics[width=0.136\textwidth]{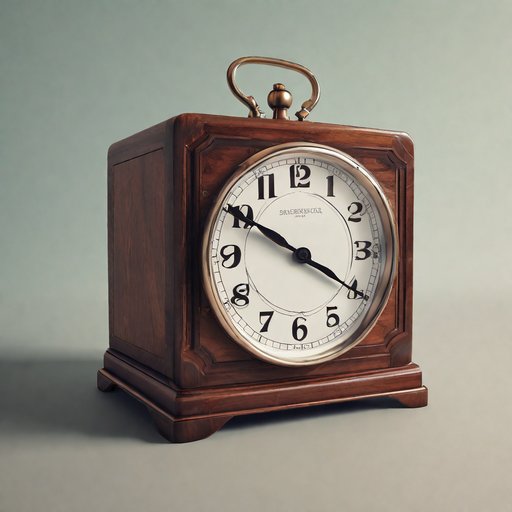}}}$ \\
        
        \bottomrule
    \end{tabular}
    \raggedright
    \caption{\textbf{Qualitative comparison (SDXL-Lightning + HPSv2).} 10 randomly chosen prompts for different alignment methods using SDXL-Lightning and HPSv2.}
    \label{fig:app:sdxl_hps}
\end{figure}

We randomly sample 10 prompts out of DrawBench and show the qualitative results of the six algorithms we compare in Section~\ref{sec:exp:text-to-image} including our TRS. The images are the best images per prompt from the experiment in Section~\ref{sec:exp:text-to-image}. We provide in total four figures, including \Cref{fig:app:sd_reward} for SD1.5 and ImageReward, \Cref{fig:app:sd_hps} for SD1.5 and HPSv2, \Cref{fig:app:sdxl_reward} for SDXL-Lightning and ImageReward and \Cref{fig:app:sdxl_hps} for SDXL-lightning optimized for HPSv2.


\end{document}